\documentclass{article}

     \usepackage[preprint]{neurips_2020}

\usepackage[utf8]{inputenc} % allow utf-8 input
\usepackage[T1]{fontenc}    % use 8-bit T1 fonts
\usepackage{hyperref}       % hyperlinks
\usepackage{url}            % simple URL typesetting
\usepackage{booktabs}       % professional-quality tables
\usepackage{amsfonts}       % blackboard math symbols
\usepackage{nicefrac}       % compact symbols for 1/2, etc.
\usepackage{microtype}      % microtypography

\usepackage{bbm}
\usepackage{xcolor,bm,amsmath,graphicx,comment,bigints}
\usepackage{algorithm,algorithmic,subcaption}
\usepackage{hanging}
\graphicspath{{images-gsvd/},{images_gsvd_mnist/}}

\title{Using Wavelets and Spectral Methods to Study Patterns in Image-Classification Datasets}

\usepackage{wrapfig}

\author{%
Roozbeh Yousefzadeh\\
  Yale University\\
  New Haven, CT \\
  \texttt{roozbeh.yousefzadeh@yale.edu} \\
 \And
Furong Huang \\
  Department of Computer Science \\
  University of Maryland, College Park, MD \\
   \texttt{furongh@cs.umd.edu} \\
}

\begin{document}

\maketitle

\begin{abstract}
%\ry{This is revised.}
Deep learning models extract, before a final classification layer, features or patterns which are key for their unprecedented advantageous performance. However, the process of complex nonlinear feature extraction is not well understood, a major reason why interpretation, adversarial robustness, and generalization of deep neural nets are all open research problems. 
In this paper, we use wavelet transformation and spectral methods to analyze the contents of image classification datasets, extract specific patterns from the datasets and find the associations between patterns and classes. 
We show that each image can be written as the summation of a finite number of rank-1 patterns in the wavelet space, providing a low rank approximation that captures the structures and patterns essential for learning.
Regarding the studies on memorization vs learning, our results clearly reveal disassociation of patterns from classes, when images are randomly labeled.
Our method can be used as a pattern recognition approach to understand and interpret learnability of these datasets. It may also be used for gaining insights about the features and patterns that deep classifiers learn from the datasets.
\end{abstract}

\section{Introduction}

%\ry{I made small changes in the next two paragraphs.}
Deep learning image classifiers are complex and nonlinear mathematical functions that extract features from images and learn to classify them based on those features. Yet, this mathematical process is not well understood, a major reason why interpretation, adversarial robustness, and generalization of them are all open research problems. By bringing computational methods from the image processing literature, in tandem with spectral methods from the numerical analysis literature, here, we develop computational methods to study the contents of image-classification datasets with the aim to understand the fine level patterns and features of images with respect to classification.

Deep learning functions are formed during training, by the contents of their training sets. Understanding the learning process and generalization, for some data types, may start with direct analysis of the data itself \citep{bahri2020statistical}. However, studying the contents of image-classification datasets is not straightforward \citep{strang2019linear}, because of the typical difficulties in working with visual signals.

In practice and in research, learning of images is left to the models. From the adversarial point of view, image-classification models are vulnerable and this vulnerability seems inevitable \citep{shafahi2018adversarial} and even at odds with achieving good accuracy \citep{tsipras2018robustness,ilyas2019adversarial}. From the generalization point of view, models are highly over-parameterized and they can fit the contents of their training sets, even when training images are labeled randomly and even if the contents of images get replaced with random noise \citep{zhang2016understanding}. Recent studies have pursued this dilemma by making the distinction between the concepts of learning vs memorization \citep{belkin2018understand,belkin2019does}.

This lack of understanding about the models can partly be explained by our lack of knowledge about fine-grained details of image-classification datasets. Recently, a new line of research has focused on the contents of these datasets, studied the procedures used to gather and label the images, and raised questions about the learning, statistical bias and possible over-fitting of deep classifiers \citep{recht2019imagenet,yadav2019cold,recht2018cifar,engstrom2020identifying}. %\ry{These are new citations.} 
Still, there is a need for specialized computational procedures to analyze these datasets and explain what can make one image associated with one class and not with other classes. What is the essence of each image when it comes to classification?% What is there about these labeled images that can be associated to specific classes?

From the optimization perspective, there are infinite number of global minimizers for training loss of image classification models, many of which would perform very badly on testing data \citep{arora2019fine,neyshabur2019towards}. Studies that focus on training, aim to find the minimizer that performs well on the testing set. But, the way to confirm that one has chosen a good minimizer of training loss is to measure its accuracy on a testing or validation set, and there is not an independent procedure to investigate whether there are specific associations between patterns and classes. Clearly, we would not expect a randomly labeled training set to have such association and that is why when one achieves zero training loss on such data, we consider it \textit{memorization} and not \textit{learning}.

Of course, a human can look at randomly labeled images and confirm the randomness of labels. But, can we devise computational procedures to evaluate that, without using a testing set? We would like such computational procedure to provide us with information such as:

\begin{hangparas}{.4cm}{1}
    1. There are no specific patterns in a dataset, if content of images are random noise.
    
    2. There are patterns in a dataset but not associated with classes, if images are randomly labeled.
    
    3. There are patterns in a dataset and each class is associated with certain patterns, if images are legitimately labeled and there are patterns in the dataset exclusive to individual classes.
\end{hangparas}

%\subsection{Our approach}

We develop a method to analyze image-classification datasets and provide the information above. Our approach can be considered a pattern recognition method which can analyze the datasets and provide {\bf insights about their learnability}. We show that each image can be written as the summation of a finite number of {\bf rank-1 patterns in the wavelet space}; and the main distinguishable patterns in each image can be reconstructed using a relatively small number of those patterns, providing a {\bf low-rank approximation to each image}. We extract the patterns by tensor decomposition of datasets in the wavelet space and then transform the patterns back to the pixel space. We see that when datasets are randomly labeled, some patterns may emerge but they cannot be associated with specific classes.%, {\bf providing a procedure to distinguish between memorization vs learning, by just investigating the training sets}.% Our method can be used for unsupervised learning and pattern recognition in image-classification datasets. It may also be used to gain insights about the features and patterns that models learn.

%In Section~\ref{sec:deeplearning_literature}, we review the related studies in the context of deep learning. 
%In Section~\ref{sec:wavelets}, we explain the wavelets as a tool to decompose images and extract useful information from them. In Section~\ref{sec:spectral}, we explain spectral methods for analyzing the patterns present in the datasets and identifying which patterns distinguish each class from the others. In Section~\ref{sec:procedure}, we formalize our method and present an algorithm. In Section~\ref{sec:results}, we present our results, and finally, Section~\ref{sec:conclusion} is our conclusions and directions for future research.

\section{Using wavelets to extract features from images} \label{sec:wavelets}

\paragraph{Background.} Wavelets are a class of functions and one of the most capable tools to systematically process images and extract features and patterns from them.
%that are successfully used to extract features and useful information from different kinds of data, especially images and signals. In image processing research, wavelets are widely used to perform different types of analysis on images. Wavelets are widely used in practice, too, e.g., in image compression.
The difficulty of working with images and many signals arises from the spatial complexity of patterns and structures in them. What makes an image to represent a dog, and not an automobile, cannot be explained by one or even a few pixels, rather, it may be explained by the specific patterns that appear in various regions of an image, and how these different regions are arranged.%Image processing and signal processing both have a rich literature and wavelets are one of the most capable tools to systematically process images and extract useful information from them.

\paragraph{Connection of wavelets with deep learning.} In recent years, we have seen the outstanding performance of deep learning models as an image processing tool that can analyze images and learn to classify them. The features learned by these models are sometimes referred to as deep features \citep{noh2017large,romero2015unsupervised,chen2016deep,effland2019image}. Some studies use deep nets to learn specific features in images, e.g., facial attributes \citep{hand2018learning,thom2020facial}. However, the process of complex nonlinear feature extraction by the models is not well understood. \textbf{This is the gap we want to bridge: to bring tools from the image processing literature and use them in tandem with spectral methods to analyze the contents of image classification datasets.}

We note that wavelets have a similar computational nature as the deep learning models. Wavelets decompose an image by convolving a wavelet basis with the image. Deep learning models also rely on convolutional layers that convolve images with stencils/filters. So, our approach of transforming images with wavelets, and then studying their patterns coincides with the approach used by the classification models. Recently, \cite{yousefzadeh2020using} proposed a method that clusters images based on their wavelet coefficients in order to analyze the similarities in image classification datasets, and showed that wavelets identify similar images, the same way that a pre-trained ResNet model does \citep{birodkar2019semantic}. Also, \cite{bruna2015super} have used wavelets to initialize the filters of convolutional neural nets.

\paragraph{Properties and notation.} There are older forms of data transformation, for example, the Fourier transform which has a long history and widespread applications. In fact, wavelets were developed building on the scientific knowledge of Fourier transform in the context of image and signal processing. For example, \cite{daubechies1990wavelet} showed that wavelets perform better than windowed Fourier transform on visual signals, because wavelets handle the frequencies in a nonlinear way. The family of Daubechies wavelets \citep{daubechies1992ten} are one of the most successful types of wavelet transformation, and we use them in this paper.

The orthogonality of Daubechies wavelets is particularly useful for feature extraction, because orthogonality in this setting implies the filters are independent and each filter is measuring a specific feature in the image signals. To process images with wavelets, we use the function
\begin{equation}
    [\omega,\beta] = wavedec(x,\Omega,N),
\end{equation}
which takes as input an image $x$, a wavelet basis $\Omega$, and level number $N$. It returns a vector of real numbers $\omega$, representing the wavelet coefficients obtained from convolving $x$ with $\Omega$, and a book keeping matrix $\beta$ containing the dimensions of wavelet coefficients by level. This operation is reversible, therefore, given $\omega$, $\beta$, and $\Omega$, we can return to pixel space and reconstruct the image $x$, which we denote its operation by
\begin{equation}
    x = waverec(\omega,\beta,\Omega).
\end{equation}
For a given $N$, $\beta$ will be constant for all images of the same size.\footnote{Since we use a uniform $N$ for all images in a dataset, we can consider $\beta$ to be constant, too, and exclude it from function arguments.}
%\fh{needs a citation and a simple description with the most brief math.}\ry{I addressed it.}
%Once we transform images to the wavelet space, we can arrange them in a tensor and decompose that tensor with spectral methods to extract the features/patterns and find their association with classes.%, as explained in the next section.

\section{Using spectral decomposition to understand the patterns in image datasets} \label{sec:spectral}

\paragraph{Background.} In general spectral methods are widely used in supervised and unsupervised learning, for example, for manifold learning \citep{belkin2006manifold}, dictionary learning \citep{huang2015convolutional}, and biomedical data \citep{alter2000singular,alter2003generalized,perros2019temporal}. In deep learning, spectral methods are used for compressing the models \citep{arora2018stronger,li2020understanding}, and also for their training \citep{sedghi2018the}. These studies, however, concern the trainable parameters, not the data.

In fact, spectral methods are rarely used on the contents of image classification datasets. An early use of spectral decomposition for image classification is by \cite{savas2007handwritten}. Their method performs higher order singular value decomposition \citep{de2000multilinear} on pixels of MNIST and uses the decomposition of training set to classify images in the testing set. They apply their method directly to pixels, achieving about 94\% testing accuracy.\footnote{We achieve more than 97\% accuracy on MNIST, in the wavelet space, only by measuring the distance of each testing image to the convex hull of digits in training set, and using the class with shortest distance as the predicted class. This clearly demonstrates the benefit of analyzing images in the wavelet space.}
Other examples of using tensor decomposition directly work on image pixels, too. For example, \cite{fang2017cp} proposed a tensor based compression method for on-ground spectral imaging. \cite{ali2016character} used tensor decomposition for character recognition. Applying tensor decomposition directly on image pixels might be able to extract some usable patterns, if images contain simple patterns like word characters. However, as we explained earlier, extracting sophisticated patterns from images requires more advanced tools, such as wavelets.

%After transforming the contents of datasets with wavelets, we form them into a rank-3 tensor $\mathcal{D}$ and 

%In the rest of this section, we briefly explain two spectral decomposition methods for tensors. We later use them to extract and analyze the patterns in datasets, after images are transformed into the wavelet space. In such analysis, we will form rank-3 tensors, where one dimension represents the classes in datasets, another dimension represents images in each class, and the last dimension represents the wavelet coefficients. It is possible to arrange the data in higher rank tensors which can be the subject of future research.

%\subsection{Higher Order GSVD.} 
\paragraph{Higher Order GSVD.} Higher Order Generalized Singular Value Decomposition (HO-GSVD) is viewed under the broad category of Tucker tensor decomposition methods \citep{kolda2009tensor}. Originally developed by \cite{van1976generalizing}, it was advanced to the higher order case \citep{omberg2007tensor,ponnapalli2011higher}, where $n$ number of (full column rank) matrices, $\mathcal{D}_i, i \in \{1,2, \dots , n\}$, can be decomposed as

\begin{equation}
    \mathcal{D}_i = U_i \Sigma_i V, \; \forall i \in \{ 1,2, \dots, n\},
\end{equation}

where, $U_i$'s, are composed of normalized left basis vectors, the singular values are positive scalars organized in diagonal elements of $\Sigma_i$, and the normalized orthogonal right basis $V$ is common among all the $n$ decompositions. It is proved that the resulting decomposition extends all of the mathematical properties of the GSVD,\footnote{This extension is advantageous both in computation and interpretation of results.} %Most recently, \cite{alter2018advanced} extended this method to tensors of higher rank.} 
except for orthogonality of the columns of left basis vectors \citep{ponnapalli2011higher,alter2018advanced}.
%There are stable numerical methods to compute this decomposition \citep{paige1981towards}, assuming that the individual matrices are not defective. Moreover, the decomposition is unique and comes with certain theoretical results that can be used for its interpretation. 

The vectors in $V$ can generally be viewed as the patterns present in all $\mathcal{D}_i$'s. The singular values are the key to understand which pattern is specific to each class and which patterns are shared among classes. For example, if the $j$th singular value of $\Sigma_i$ is significantly larger (dominant) than the $j$th singular values for all other classes, it means that the $j$th vector in $V$ is specific to the class $i$. We expand further on this kind of interpretation in our numerical results.
Additionally, we can use this decomposition to write each $\mathcal{D}_i$ as the summation of $m$ rank-1 matrices.
\begin{equation} \label{eq:sum_rank1}
    \mathcal{D}_i = \sum_{j=1}^{m} \sigma_j U_i(:,j) V_i(j,:)
\end{equation}

%Using this, we decompose each image as the summation of $m$ rank-1 patterns in the wavelet space.% This way, we can break down each image into many simple components and identify their essential patterns. 

\paragraph{Other methods.}
Using other forms of tensor decomposition may be useful, too. In particular Canonical Polyadic Decomposition can be useful for certain aspects of our analysis, because it breaks the data into rank-1 components (patterns), and the number of components corresponds to the rank of tensor. %\citep{kolda2009tensor}. %This way we can use this as a dimensionality reduction method to extract a compact set of patterns from image-classification datasets. 
Recently, \cite{hong2020generalized} presented an algorithm for generalized canonical polyadic (GCP) low-rank tensor decomposition, which we plan to explore in future research.

%\ry{I have not included any results for this. We have ran out of space. Shall }
%\paragraph{Canonical Polyadic Decomposition.} 
%This form of decomposition, approximates a tensor as sum of rank-1 tensors \citep{kolda2009tensor}. This can be particularly useful for certain aspects of our analysis, because it breaks the data into rank-1 components (patterns), and the number of components corresponds to the rank of tensor. This way we can use this as a dimensionality reduction method to extract a compact set of patterns from image-classification datasets. Most recently, \cite{hong2020generalized} presented an algorithm for generalized canonical polyadic (GCP) low-rank tensor decomposition that allows other loss functions besides squared error. We use this method in our numerical results.

\paragraph{Choosing a subset of influential wavelet coefficients. } \label{sec:rrqr}
The HO-GSVD has stable and efficient numerical algorithms for its computation and there are well-studied theoretical properties about it. But in order to benefit from such algorithms and mathematical properties, the individual matrices for each class should have full column rank.
Some datasets may naturally satisfy such property, as we see for CIFAR-10 dataset. But, this is not satisfied for some other datasets, for example MNIST and Omniglot. In order to choose a subset of most influential wavelet coefficients, satisfying the full column rank requirement, we use rank-revealing QR factorization (RR-QR) \citep{chan1987rank}. This method orders the columns of a matrix based on their rank influence. Hence, we can easily choose a subset of most influential coefficients satisfying the full column rank. This approach of using rank-revealing QR factorization to choose a subset of wavelet coefficients is previously suggested by \cite{yousefzadeh2019interpreting}.\footnote{The computational complexity of RR-QR for a matrix with $n$ rows and $m$ columns is $\mathcal{O}(nm^2)$ which is inexpensive for datasets like CIFAR-10.}

%\section{Our Algorithm} \label{sec:algorithm}

\section{Formal procedure} \label{sec:procedure}

%\subsection{Formal procedure}

%\fh{I think we have two options: (1) We could move the current description in the pseudo code here, and add the more precise math description to the pseudo code. (2) Or the opposite: Maintain the current description in the pseudo code, put more precise description in the section.I often go with option 1. It is debatable which one is better. But we will need to have a more precise description.}\ry{I went with option 1 and added the math notation to the algorithm.}
\paragraph{Our approach.}
The main goal, here, is to extract features and patterns from image-classification dataset, $\mathcal{P}$, and find the association of patterns to classes. Our algorithm first transforms the images into a wavelet space, organizes them into a tensor $\mathcal{D}$, analyzes the tensor and removes its redundancies and rank deficiencies, performs spectral decomposition on $\mathcal{D}$, analyzes the decomposition, and finally, reconstructs the patterns back in the pixel space, and provides low-rank approximation for images.

\paragraph{Our algorithm.}
We now formalize our procedure in Algorithm~\ref{alg:main}. Line~1 counts the number of classes, $n_c$. Line~2 through~7 decomposes all images using wavelet basis $\Omega$ and organize them in a rank-3 tensor, where each $\mathcal{D}_i$ represents each class, rows of each $\mathcal{D}_i$ represent the $n_i$ images in that class and its columns represent the $m$ wavelet coefficients of images. We discuss the choice of wavelet basis for our numerical experiments in Appendix~\ref{sec:appx_data}. Note that the slices of $\mathcal{D}$ have different number of images and HO-GSVD can decompose such tensor.

%\ry{Probably, I can move lines 16 through 30 of the algorithm into a separate algorithm and put it in the appendix, to free up space.}

\begin{algorithm}[ht]
\caption{Wavelets Spectral Decomposition for Pattern Association (WSDPA): Algorithm for analyzing the patterns in image-classification datasets using wavelets and spectral decomposition%\fh{a name for the algorithm: maybe something like Wavelets Spectral Decomposition for Pattern Extraction (WSDPE)}\ry{Done!}
}
\label{alg:main}
\textbf{Inputs}: Dataset $\mathcal{P}$ (in pixel space), wavelet basis $\Omega$, $\tau$, tensor decomposition method $\mathcal{M}$\\
%\textbf{Parameter}:  $k$, $\kappa$\\ comparison function $\mathcal{G}$,
\textbf{Outputs}: Reconstructed patterns in pixel space $\nu_k$ and their association to each class, Decomposition of $\mathcal{D}$
%\fh{It would be nice to exhibit the outputs precisely} \ry{I have tried to make it precise.}
\begin{algorithmic}[1] %[1] enables line numbers
\STATE Count total number of classes in $\mathcal{P}$ as $c$
\FOR{$i=1$ to $c$}
    \STATE $n_i = $ total number of images in class $i$
    \FOR{$j=1$ to $n_i$}
        \STATE $\mathcal{D}(j,:,i) = wavedec(P\{i\}\{j\},\Omega)$
    \ENDFOR
\ENDFOR
%\fh{We should use notations and equations to be precise here. In the explanation of the pseudo-code in the main body, we could use this description.}
\STATE $\alpha = mean(\mathcal{D},``all")$, $m = $ number of wavelet coefficients per image
%Compute the average of all wavelet coefficients for all images, as $\alpha$ \fh{We should use math here as well, the explanation could go the the main body}
\STATE $S_{n,c} = \mathcal{D}_{n,m,c} \times_{m} \mathbbm{1}_{m,1}$
\STATE $\forall{(i \in \{1,\dots,c \},j \in \{1,\dots,n_i \})} \rightarrow \mathcal{D}(j,:,i) = \mathcal{D}(j,:,i) \odot (\alpha/S(j,i))$
% \FOR{$i=1$ to $n_c$}
%     \FOR{$j=1$ to $n_i$}
%         \STATE $\mathcal{D}(j,:,i) = \mathcal{D}(j,:,i) \odot (\alpha/S(j,i))$
%     \ENDFOR
% \ENDFOR
%\STATE Scale the wavelet coefficients for each image so that the sum of wavelet coefficients becomes equal to $\alpha$ for all images\fh{We should use math here as well}
\IF{$m \leq$ total number of images in $\mathcal{P}$}
    \STATE Stack all $\mathcal{D}(:,:,i),\forall{i \in \{1,\dots,c\}}$ into a single matrix $A$ and perform rank-revealing QR factorization to obtain $PA = Q R$
    \FOR{$i=1$ to $n_c$}
        \STATE $\hat{\gamma}_i = \min \big( n_i , \arg \max\limits_{\gamma} \kappa \big(P(1:\gamma,:) \mathcal{D}_i \big) \leq \tau \big)$
    \ENDFOR
    \STATE $m = \min(\hat{\gamma}_i,i \in \{1,\dots,c \})$
    \STATE $\forall (i \in \{1,\dots,c \}) \rightarrow \mathcal{D}_i = P(1:m,:) \; \mathcal{D}_i$
    % \FOR {$i=1$ to $n_c$}
    % \STATE $\mathcal{D}_i = P(1:m,:) \; \mathcal{D}_i$
    % \ENDFOR
\ELSE
    \STATE Warning: ``Samples are fewer than features."
\ENDIF
\STATE Perform spectral decomposition on rank-3 tensor $\mathcal{D}$ using method $\mathcal{M}$\\
$\rightarrow$ e.g., using HO-GSVD: $\mathcal{D} = U_i \Sigma_i V, \forall{i \in \{1,\dots,c\}}$
\STATE Analyze the patterns in decomposition, reconstruct them, and find their association to classes \\
$\rightarrow$ $\nu_k = waverec(V(:,k),\Omega)$ to reconstruct the patterns in pixel space\\
$\rightarrow$ Find association of $\nu_k$'s to each class based on singular values
\STATE \textbf{return} the decomposition and patterns
\end{algorithmic}
\end{algorithm}

Line~8 computes the total average of all the wavelet coefficients in $\mathcal{D}$ and defines $m$ as the number of wavelet coefficients per image. Line~9 sums the values of $\mathcal{D}$ over the dimension of wavelet coefficients, to obtain $S$, and line~10 scales the wavelet coefficients for each image, so that the sum of wavelet coefficients becomes equal for all images. %This makes the darkness/brightness of images uniform across the dataset, and therefore less influential in the patterns we are going to extract with tensor decomposition. In this operation Hadamard product is used, denoted by $\odot$.
Lines~11 through~20 are about choosing the most influential wavelet coefficients. Such feature selection requires the total number of images in the dataset to be more than the number of wavelet coefficients per image. This is naturally satisfied in some image classification datasets, but in cases where it is not satisfied, one can reduce the resolution of images, before feeding them to Algorithm~\ref{alg:main}. Further discussion is provided in Appendix~\ref{sec:appx_algorithm}.

Line~12 performs RR-QR algorithm on all the data to sort the wavelet coefficients based on their importance (linear independency). Lines~13 through~15 loop over slices of the tensor and find the maximum number of wavelet coefficients that can be used while satisfying the conditions discussed in Section~\ref{sec:rrqr}. $\kappa(.)$ computes the 2-norm condition number. Line~16, chooses~$m$ as the number of coefficients that satisfy the requirements for all slices of $\mathcal{D}$. Line~17 keeps the $m$ most influential wavelet coefficients and discards the rest of coefficients for all images. In choosing the~$m$, Algorithm~\ref{alg:main} uses a parameter $\tau$ which is an upper limit on the condition number of slices of the tensor corresponding to each class. %We note that one can define a specific $m$ and bypass Lines~11 through~20 of the algorithm, but the chosen $m$ should satisfy the requirements discussed in Section~\ref{sec:rrqr}.
%\fh{This should be in the main body} \ry{Sure. I moved some of the practical notes to appendix~\ref{sec:appx_algorithm}.}
%If the full column rank condition is not satisfied, Line~29 returns a warning. 
Finally, Lines~21 and~22 perform the spectral decomposition on $\mathcal{D}$ and analyzes the patterns in the decomposition. This process involves reconstructing the patterns in the pixel space. Based on the spectral method used, we can find the association of patterns to classes, by identifying which patterns are almost exclusively contributing to specific classes.% When $\mathcal{M}$ is chosen to be HO-GSVD, this association can be easily performed by analyzing the singular values for each class. 

\section{Numerical Experiments} \label{sec:results}
%\fh{In experiments, I generally think we could have the messages more explicit, such as using titles or bold sentences. Current writing use bold sentences as description of what we do, but does not provide the message of what we are showing, what are the key contributions.} \ry{I tried to address this.}

Here, we study the patterns in CIFAR-10 \citep{krizhevsky2009learning}. Our results on MNIST \citep{lecun1998gradient} are presented in Appendix~\ref{sec:appx_mnist}. 
%Our result on 30 alphabets of Omniglot is presented in Appendix~\ref{sec:appx_omniglot} \fh{section deleted}.\ry{I can include Omniglot, but the paper is getting too long.}
%In our experiments, {\bf we extract and analyze the patterns (rank-1 components of images in the wavelet space)} and identify which ones are discriminative for each class. We then derive {\bf low rank approximation to each image and show that the distinctive features in images can be captured by a relatively small number of those patterns}. We also explore the effect of random labels on our results. We know that deep learning models can achieve perfect accuracy on this training set, even when all the images are labeled randomly, and one would not be able to detect the randomness of labels just by training a model. We show that our method reveals whether there is {\bf learnable classification information} present in the training set, useful in practice, and also useful for studying the concept of {\bf memorization vs learning}.
%\paragraph{CIFAR-10.}
We first consider two classes of Cat and Dog in the CIFAR-10, because they are two of the most similar classes and make up most of the mistakes by the state of art model \citep{kolesnikov2019large}.
%\fh{This paragraph could be a summary of the ``messages'' (observations and conclusions and what we verified using the following experiments) in the following experiments. It doesn't have to be a recap of the experiment settings.}\ry{I changed it completely.}
To decompose images, we use the Daubechies-2 wavelet basis. In Appendix~\ref{sec:appx_data}, we explain the reason for this choice. The number of wavelet coefficients we use is $m = 3,000$ in this example, which corresponds to using $\tau = 7\times 10^4$ in Algorithm~\ref{alg:main}.

%We then choose a subset of unique wavelet coefficients for all images, by doing rank-revealing QR factorization on the matrix that contains wavelet coefficients for all images of Cats and Dogs. The number of wavelet coefficients we choose is $m = 3,000$ in this example, which corresponds to using $\tau = 7\times 10^4$ in Algorithm~\ref{alg:main}. \fh{I feel much of these descriptions here could be merged with the algorithmic description. This would shorten the paper a lot.}

%The algorithm then divides images based on their class and forms them into a rank-3 tensor. $\mathcal{D}_1$ contains the training images of Cats and $\mathcal{D}_2$ contains training images of Dogs, as shown in Figure~\ref{fig:cifar10_cd_matrix}. Each class in the training set of CIFAR-10 has 5,000 images, so each $\mathcal{D}_i$ has 5,000 rows and $m$ columns. We then scale all the rows in both $\mathcal{D}_i$'s such that the sum of elements in all rows become equal.
%\fh{I feel much of these descriptions here could be merged with the algorithmic description. This would shorten the paper a lot.}\ry{Sure.}

\begin{figure}[H]
  \centering
   \includegraphics[width=0.6\linewidth]{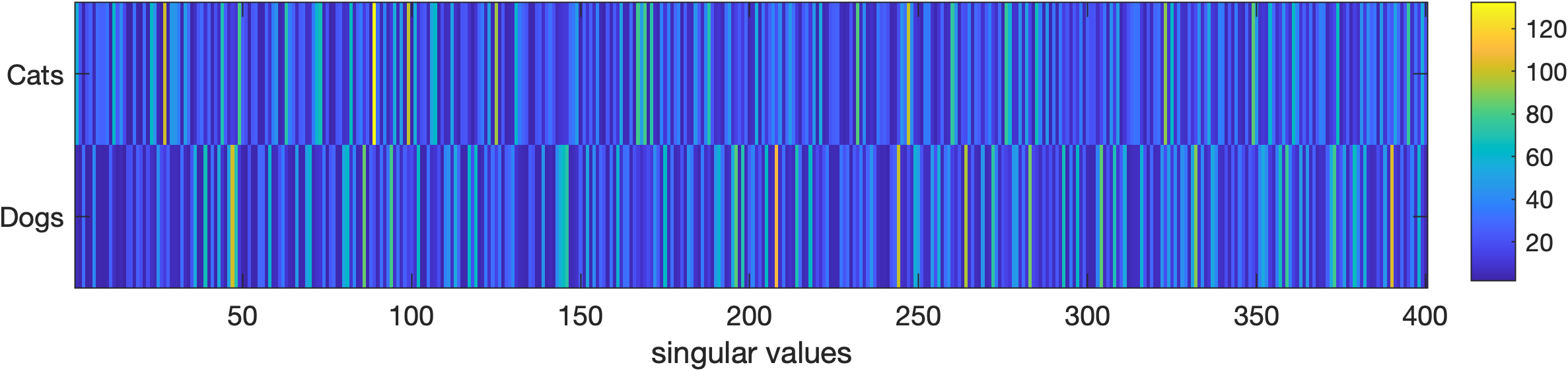}
  \caption{The first 400 singular values of $\Sigma_1$ and $\Sigma_2$. Larger singular value for one class signifies the importance of corresponding right basis for that class. Clearly, there are distinct separations between some of the singular values.}
  \label{fig:cifar_cd_singulars}
\end{figure}

{\bf Singular values reveal patterns specific to each class.} %\fh{this is a good paragraph.}\ry{I am now starting with this paragraph, because it has a clearer message.}
Each $D_i$ has its own set of singular values, $\Sigma_i$. Comparison of singular values in $\Sigma_1$ and $\Sigma_2$ reveals which patterns (i.e., vectors in $V$) are influential for cats and which patterns are influential for dogs. Figure~\ref{fig:cifar_cd_singulars} shows the first 400 singular values for cats and dogs. The ratio of singular values and also the angular distance between them can be insightful in this analysis, as suggested by \cite{bradley2019gsvd}. When two singular values are equal for a particular basis, it means that the basis is equally common for them and not discriminative. But, when a singular value for one class is much larger compared to other classes, it means that the corresponding basis in $V$ is specific to that class. In Figure~\ref{fig:cifar_cd_singulars}, we can see many such cases.

\begin{figure}[H]
  \centering
  \begin{subfigure}[b]{0.49\textwidth}
  \centering
  \fbox{
   \includegraphics[width=0.15\linewidth]{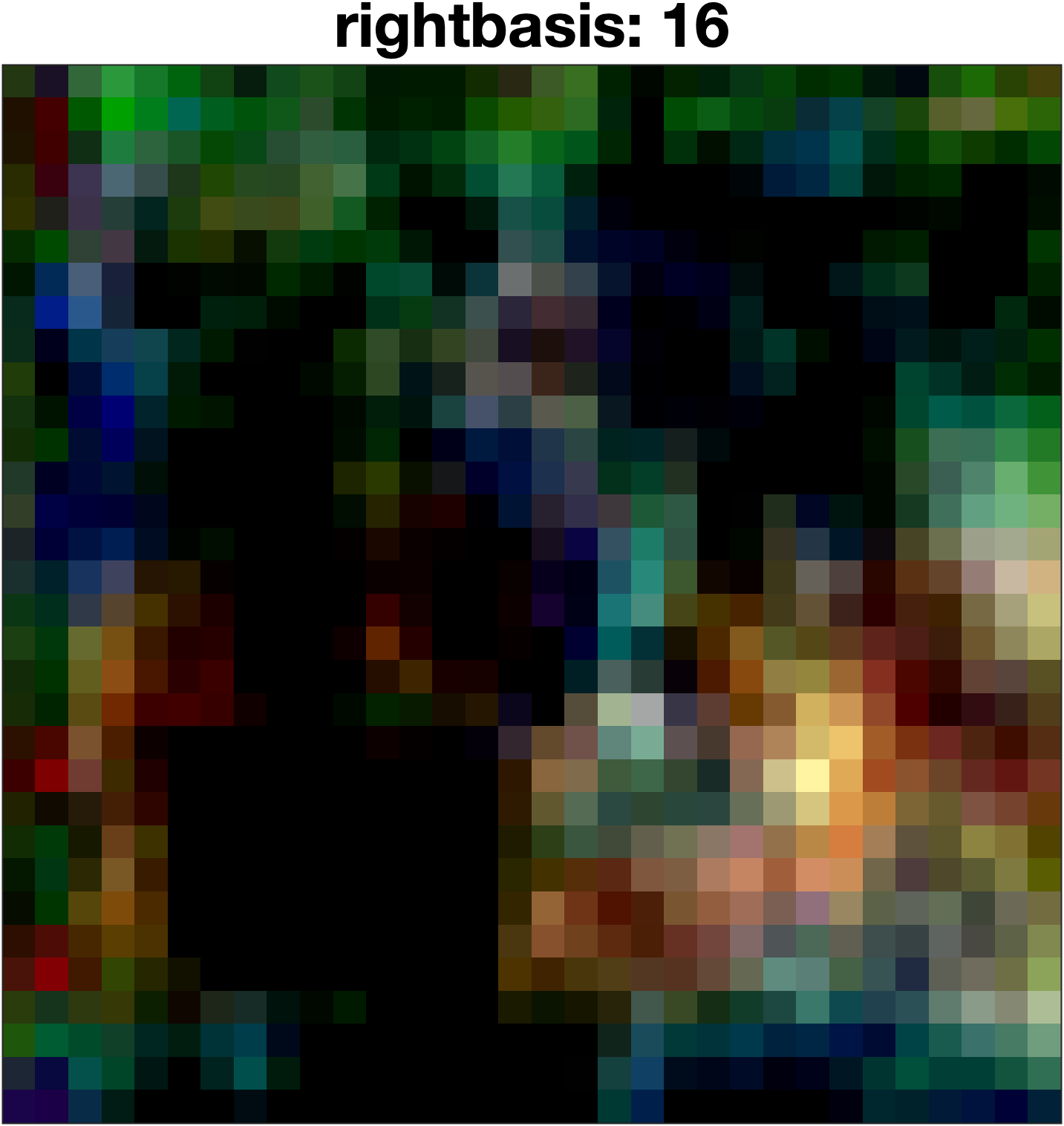}
   \includegraphics[width=0.15\linewidth]{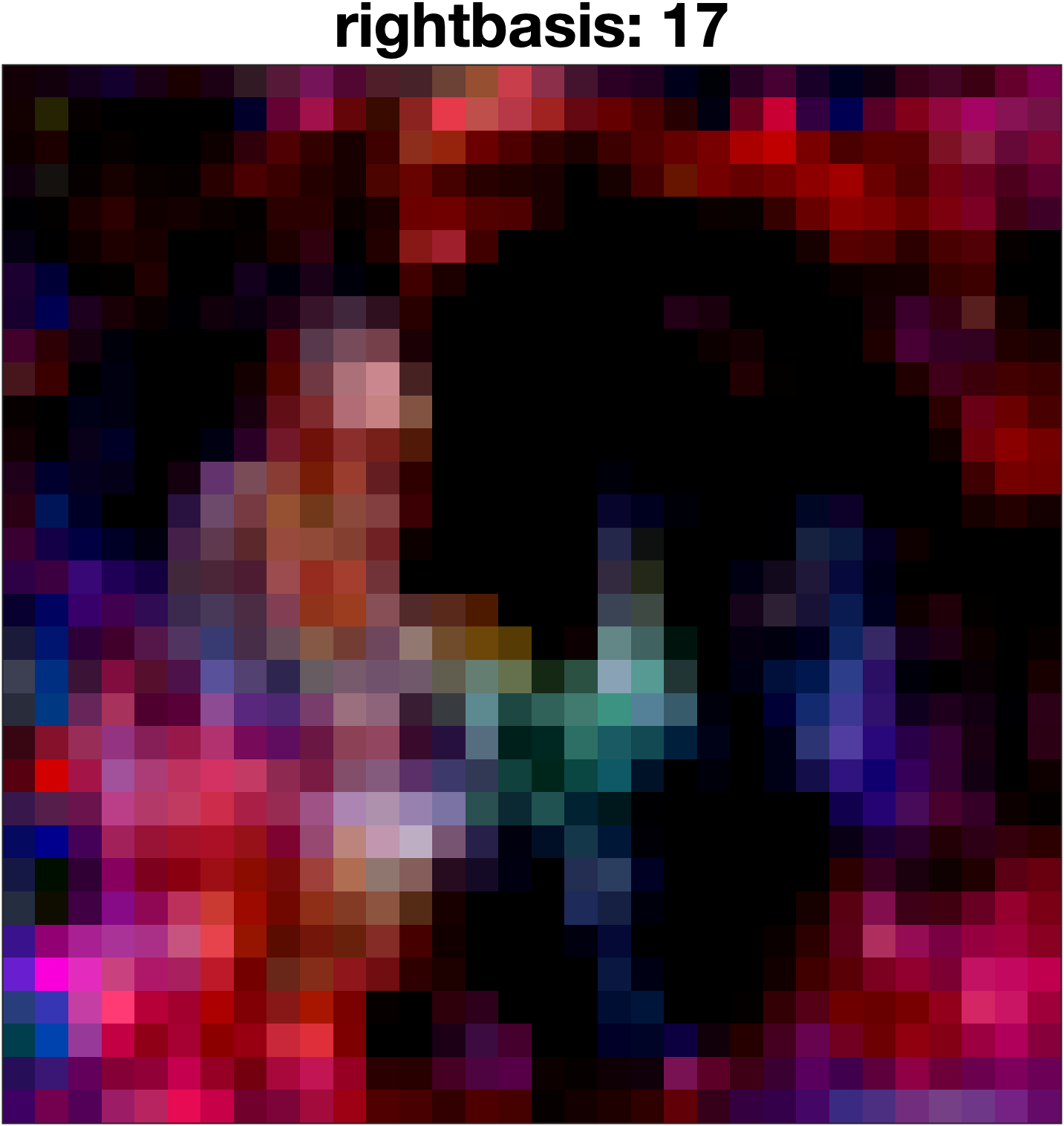}
   \includegraphics[width=0.15\linewidth]{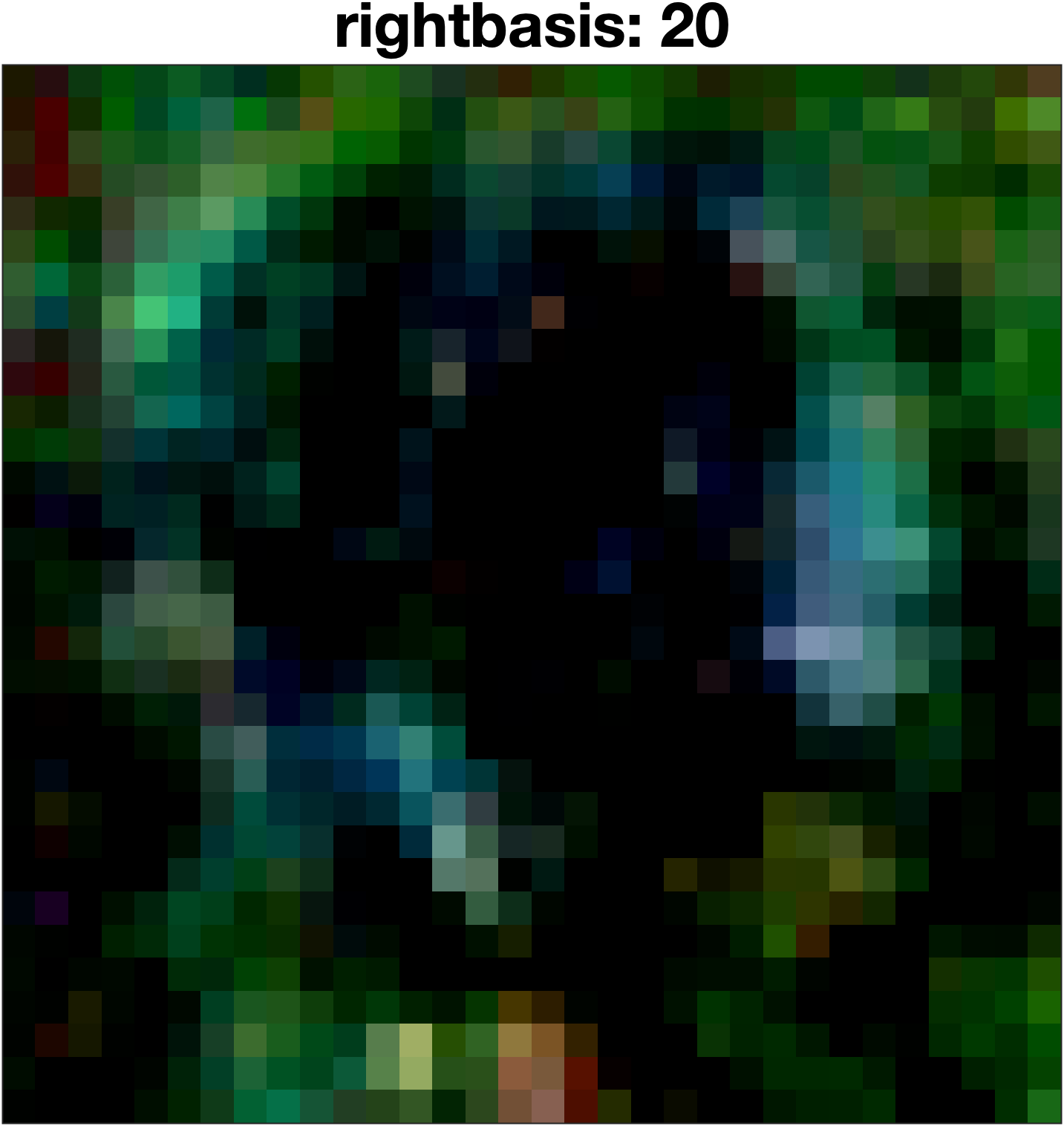}
   }
  \caption{Dogs.}
  \label{fig:rightbasis_dog}
  \end{subfigure}
  \hspace{-2cm}
  \begin{subfigure}[b]{0.49\textwidth}
  \centering
  \fbox{
   \includegraphics[width=0.15\linewidth]{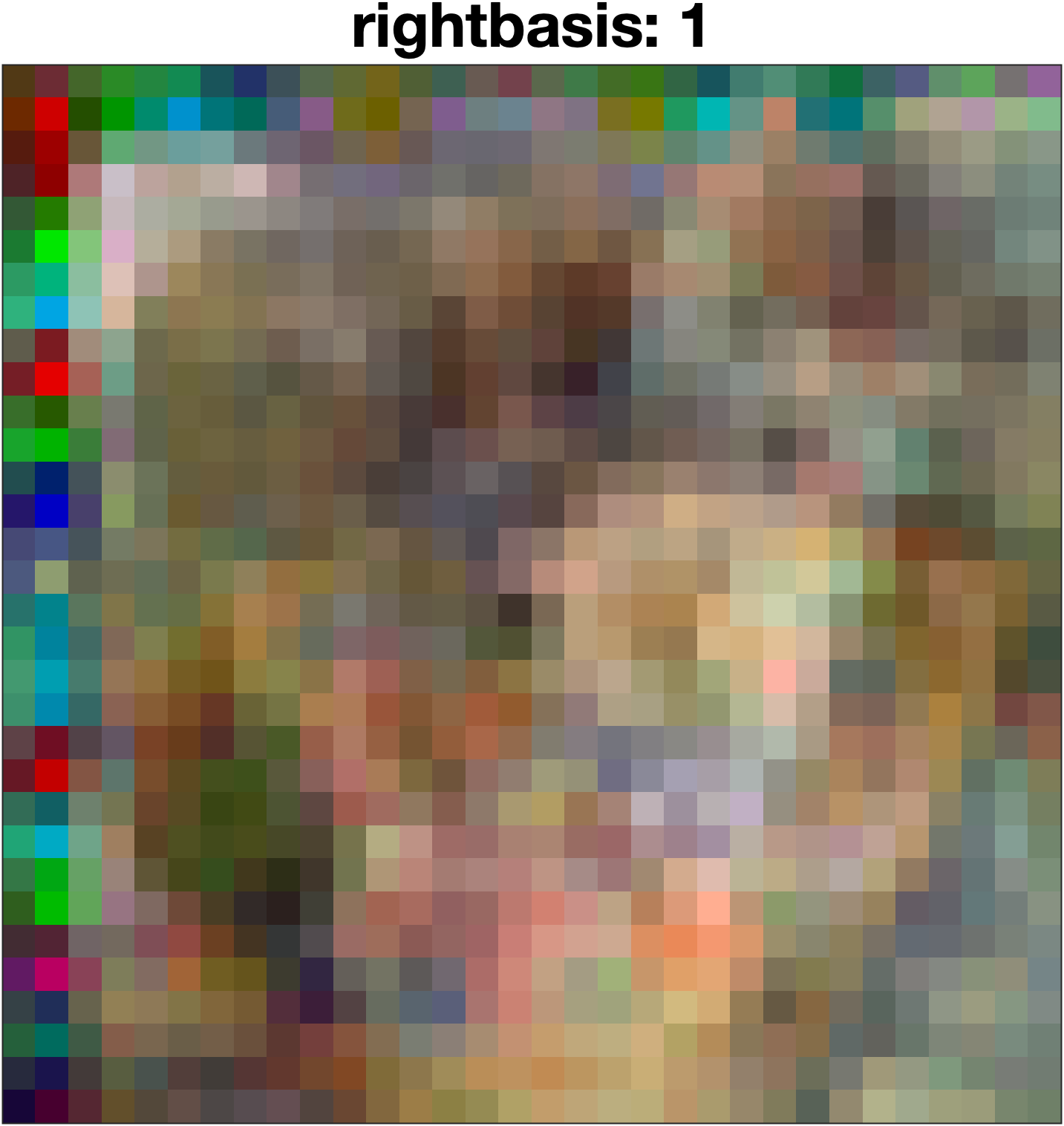}
   \includegraphics[width=0.15\linewidth]{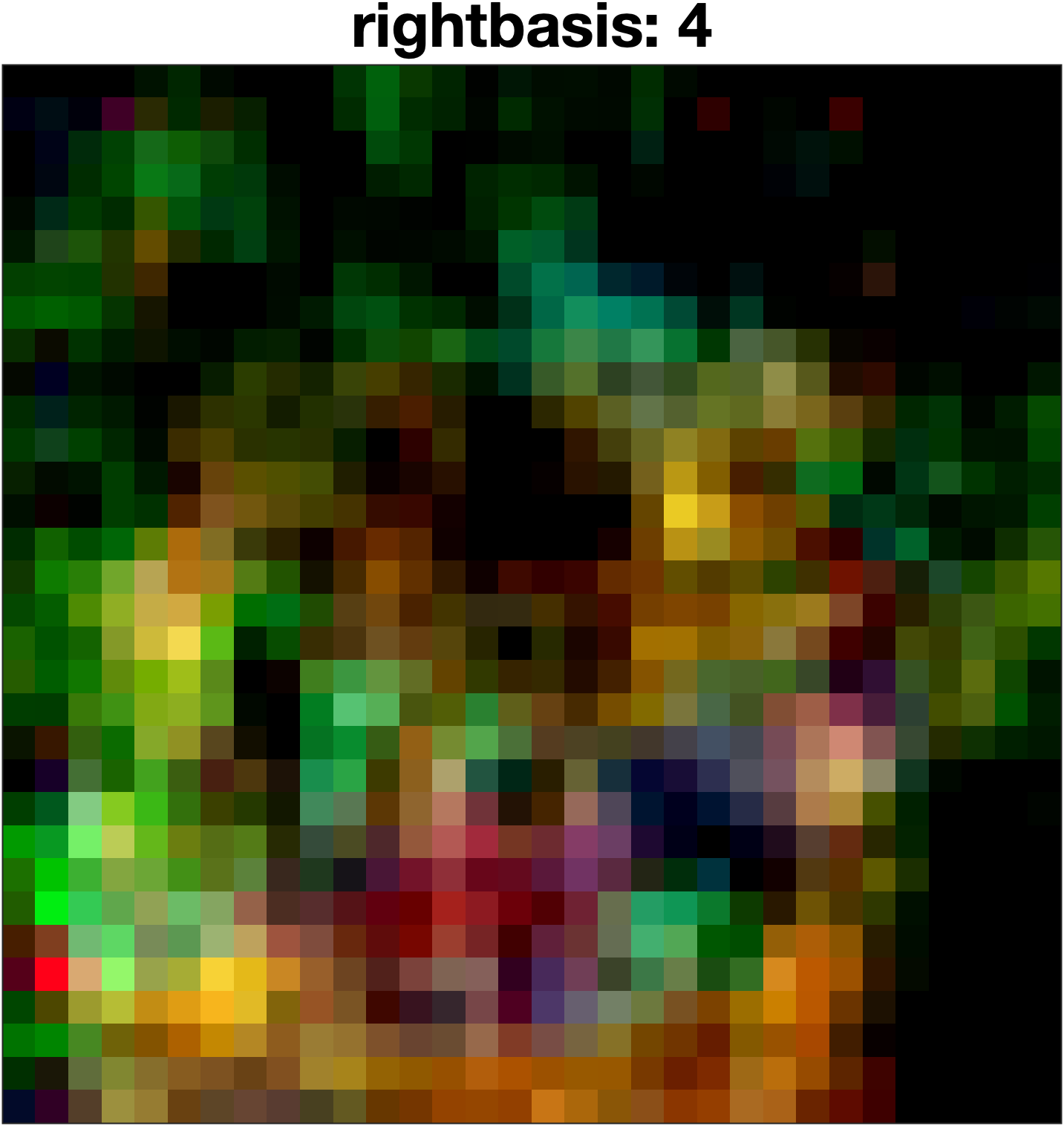}
   \includegraphics[width=0.15\linewidth]{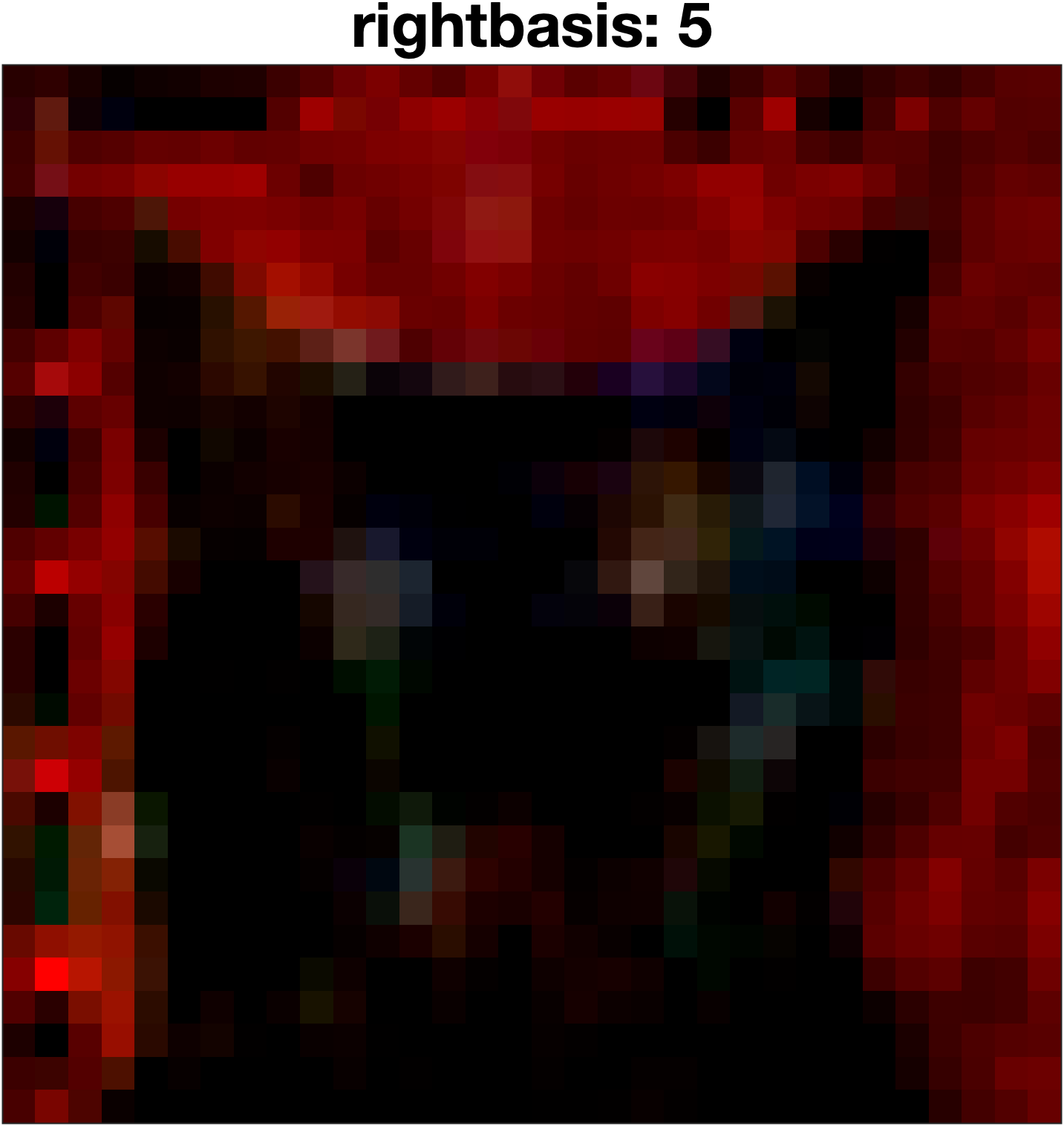}
   }
  \caption{Cats.}
  \label{fig:rightbasis_cat}
  \end{subfigure}
  \caption{Images reconstructed in the pixel space from the most dominant right basis columns.}
  \label{fig:rightbasis}
\end{figure}

{\bf Reconstructing the patterns back in the pixel space.}
Each column in $V$ corresponds to a pattern and we can reconstruct the patterns back in the pixel space, using the same wavelet basis. Figure~\ref{fig:rightbasis_dog} shows the four most dominant bases for Dogs, in the pixel space. By dominant, we mean these bases have large singular values for the Dog class, but small singular values for Cats. Similarly, Figure~\ref{fig:rightbasis_cat} shows the patterns dominant for cats, back in the pixel space.

%\fh{I also don't know how to interpret this result, since in Figure~\ref{fig_rightbasis_cat}, we couldn't really verify that these are indeed basis for dogs.} \ry{I hope it is clearer now with the rearrangements.}

\begin{figure}[h]
  \centering
   \includegraphics[width=0.5\linewidth]{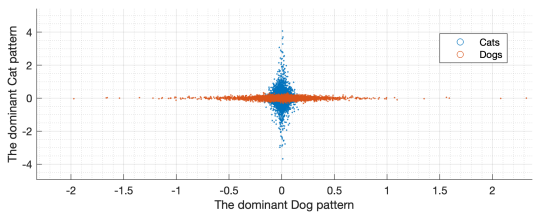}
  \caption{Contribution of most dominant dog pattern vs the most dominant cat pattern to all images in the dataset. There is a clear separation between classes, but there is also a considerable overlap near the origin (i.e., some images do not get considerable contributions from these two patterns.}
  \label{fig:cifar_cd_pattern_coeff}
\end{figure}

{\bf Understanding the contribution of each pattern to each image.}
Each of the patterns present in $V$ contribute to each image in the dataset with some coefficient, determined in the left basis. In fact, we can write each image as the summation of the patterns in the pixel space, using Equation~\ref{eq:sum_rank1}. %We see that the dominant patterns for Dogs do not contribute significantly to any of the cat images, while they contribute with large coefficients to many images of dogs. 
Figure~\ref{fig:cifar_cd_pattern_coeff} shows the coefficients of contribution for all images in this data, for the most dominant dog pattern vs the most dominant cat pattern. While a clear separation between the two classes is visible, we can also see a considerable overlap of points near the origin. This overlap corresponds to images that are not getting a noticeable contribution from these two specific patterns, and such images belong to both classes. To summarize, the most dominant dog pattern never contributes significantly to any of the cat images, it contributes significantly to some of the dog images, and it also does not contribute to many of the images in the dog class. Those dog images get their contributions from other dog patterns in $V$.

\begin{figure}[H]
  \centering
   \includegraphics[width=0.1\linewidth]{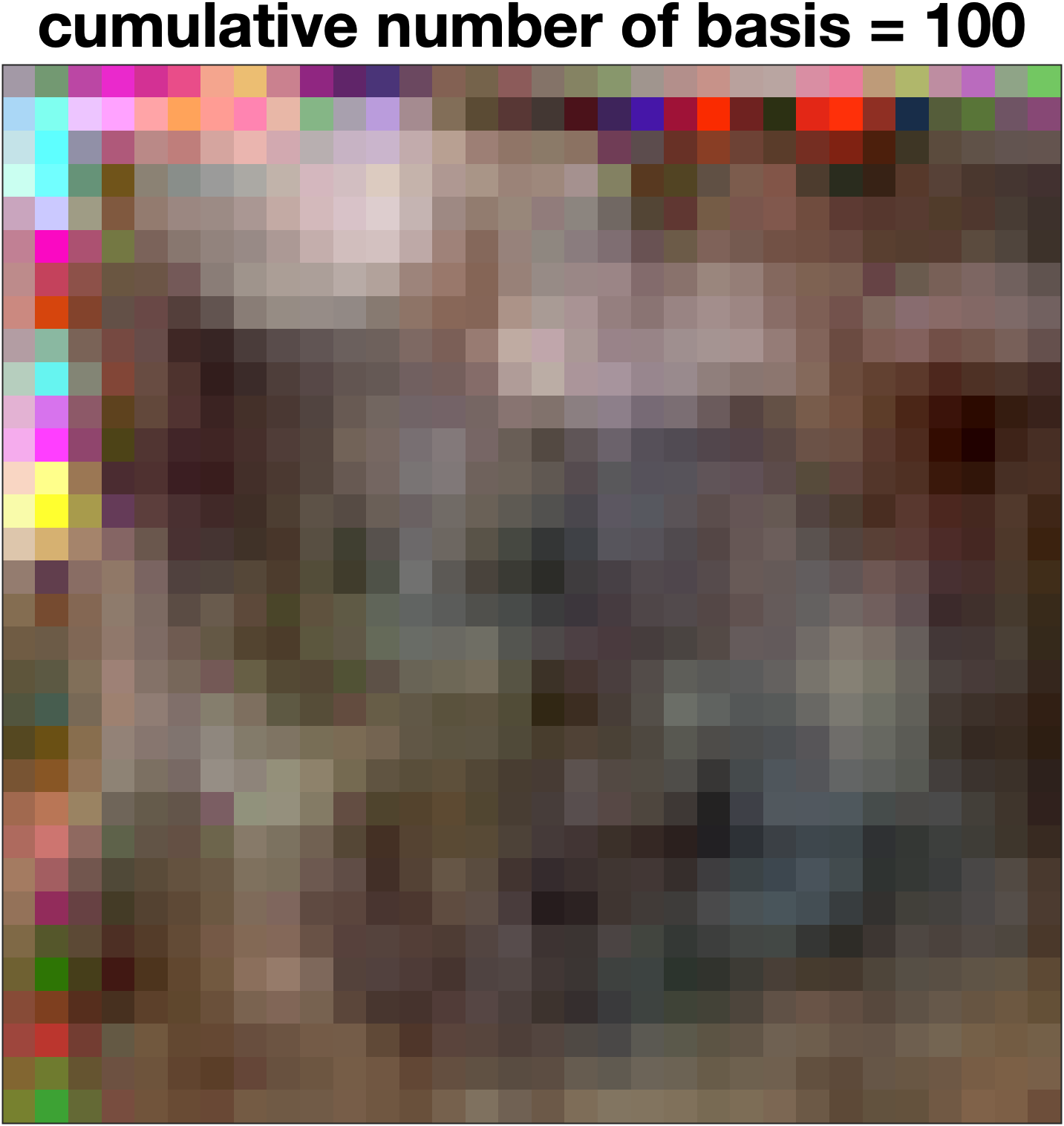}
   \includegraphics[width=0.1\linewidth]{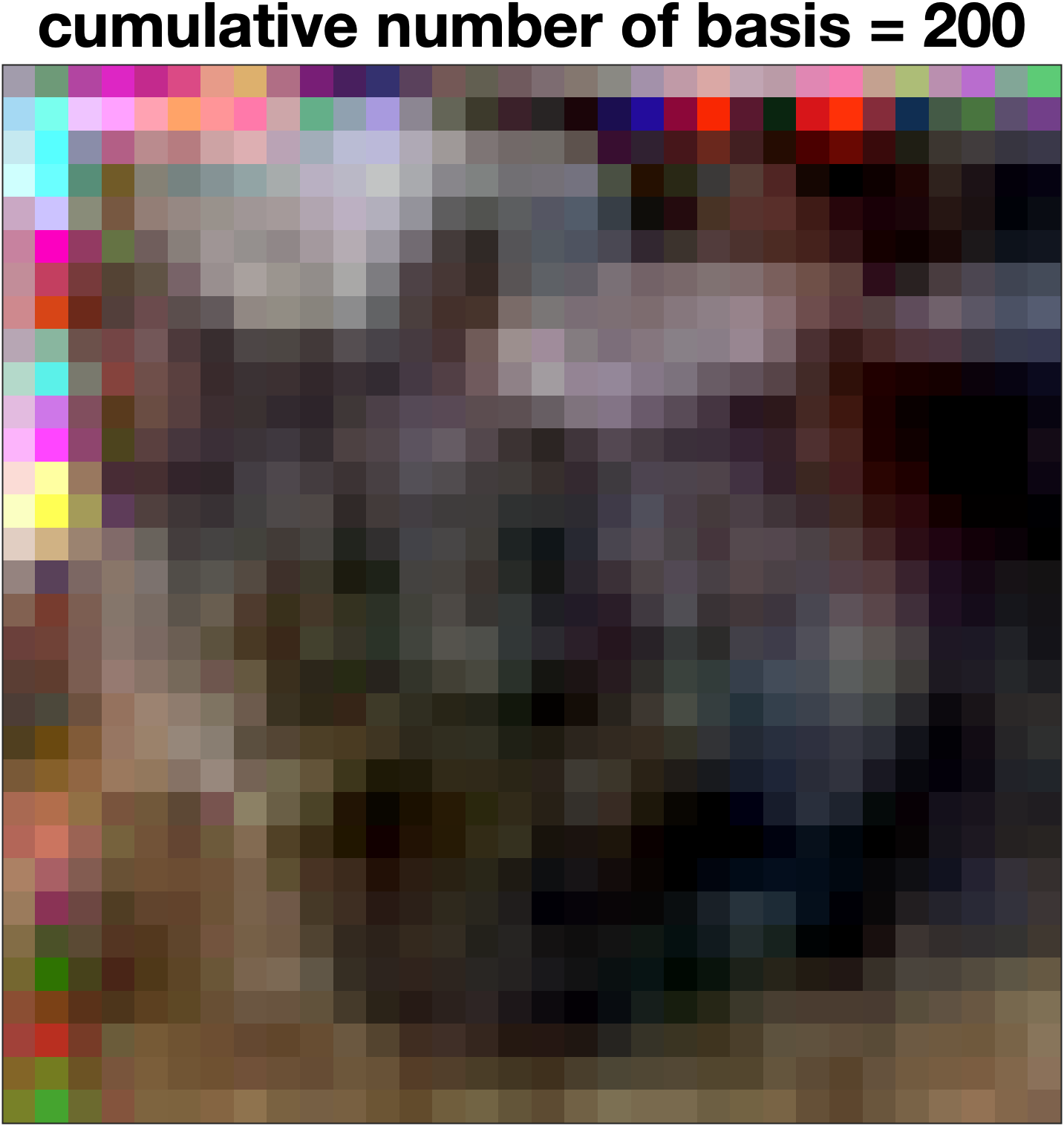}
   \includegraphics[width=0.1\linewidth]{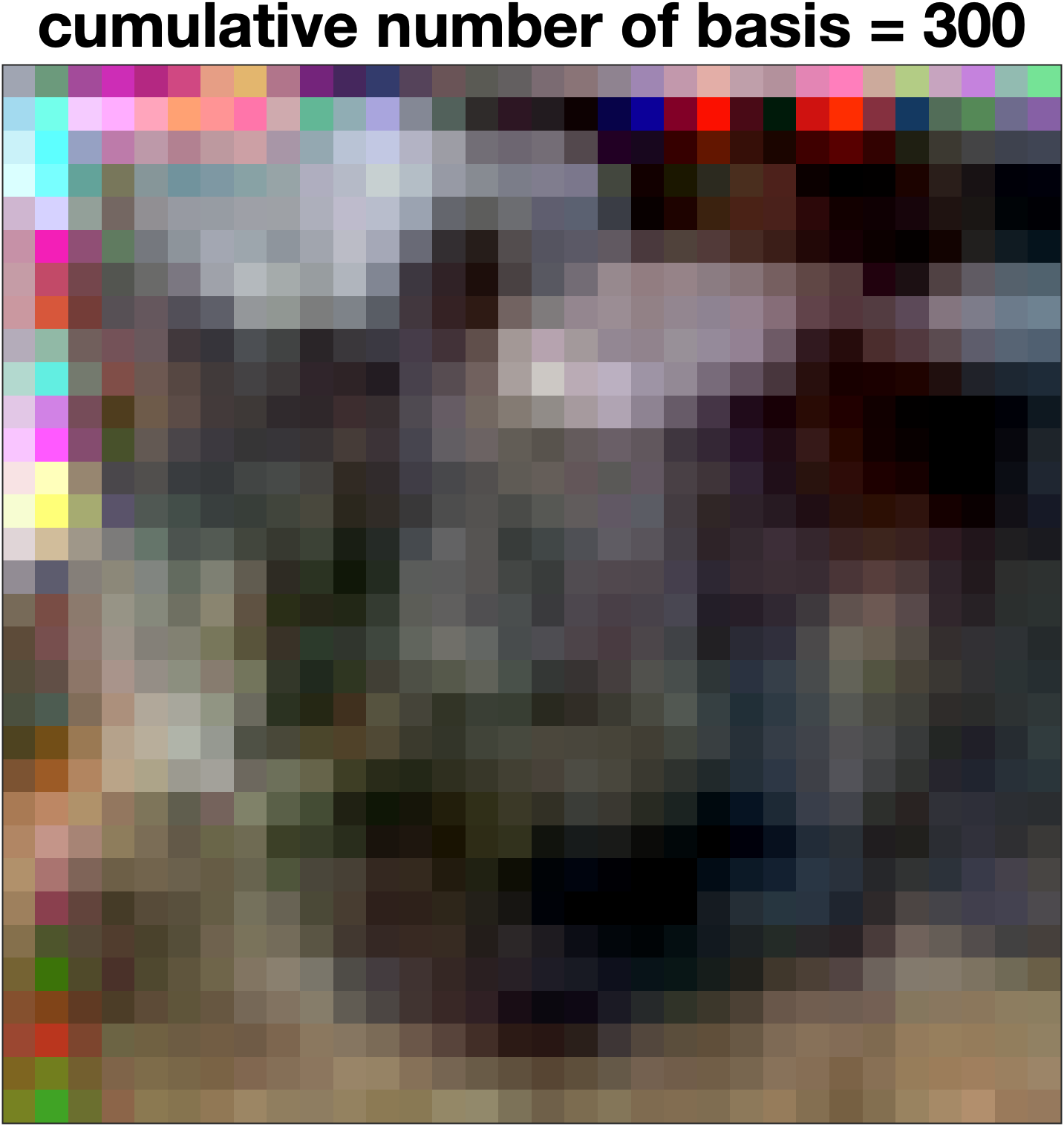}
   \includegraphics[width=0.1\linewidth]{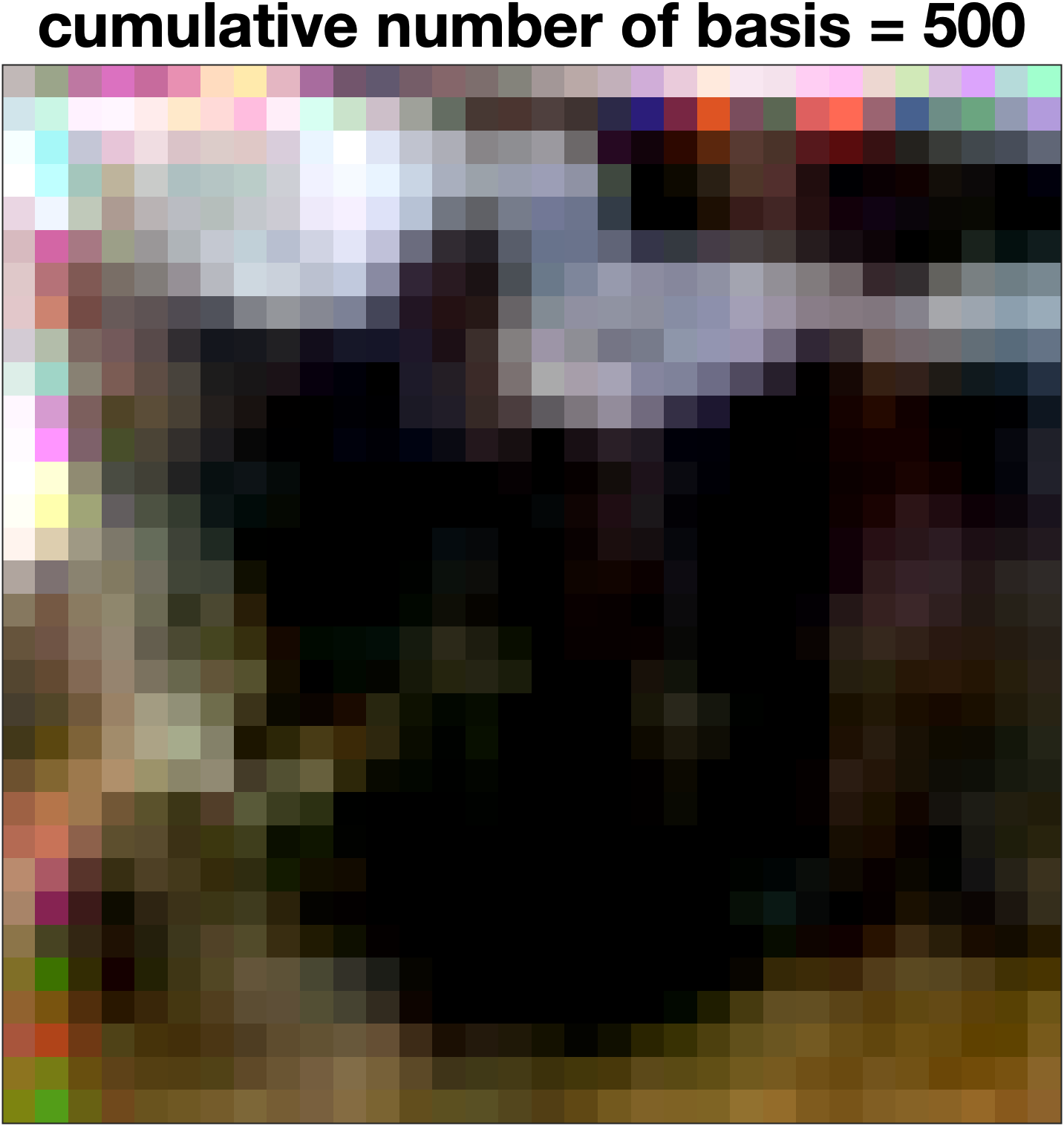}
   \includegraphics[width=0.1\linewidth]{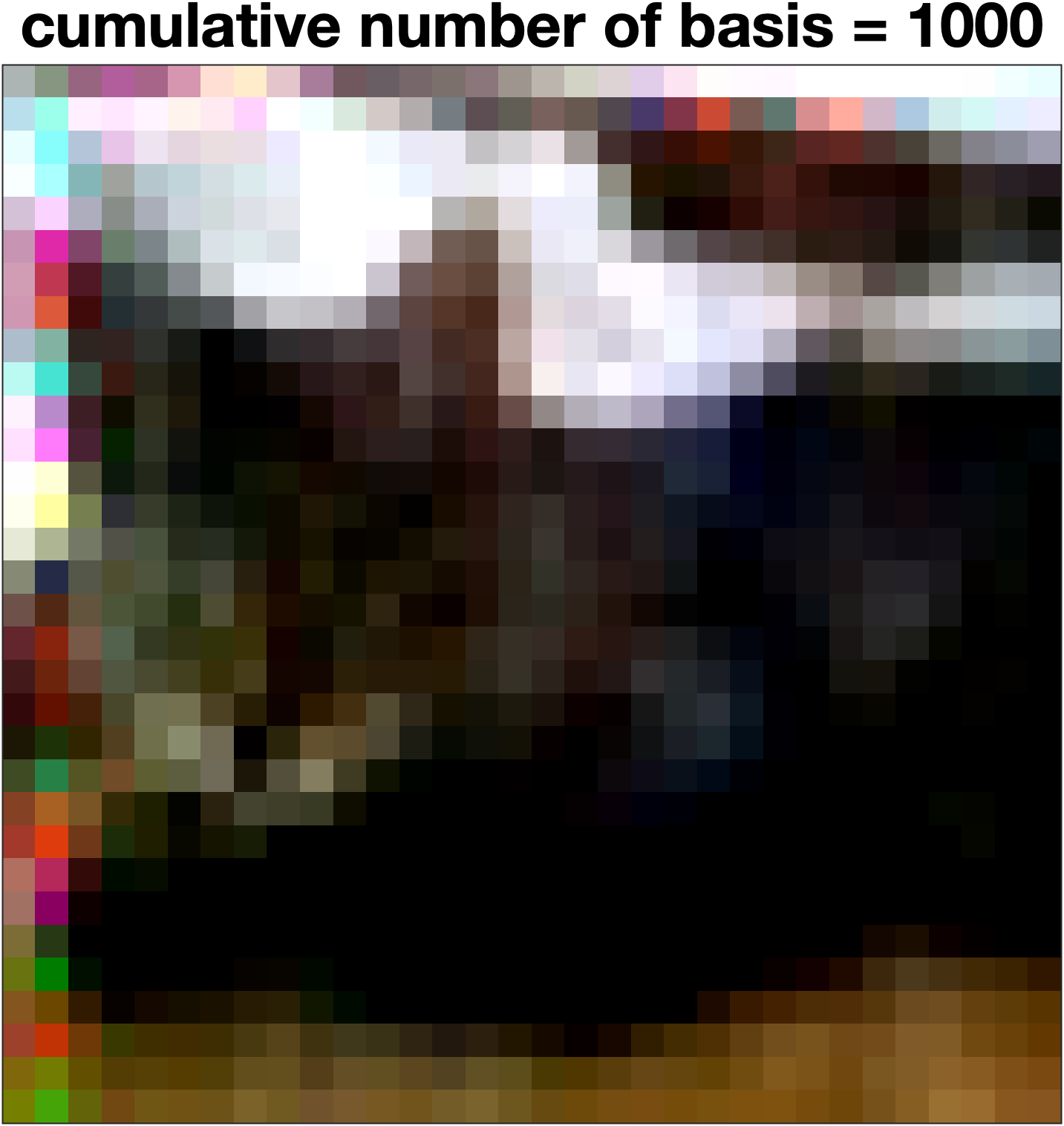}
   \includegraphics[width=0.1\linewidth]{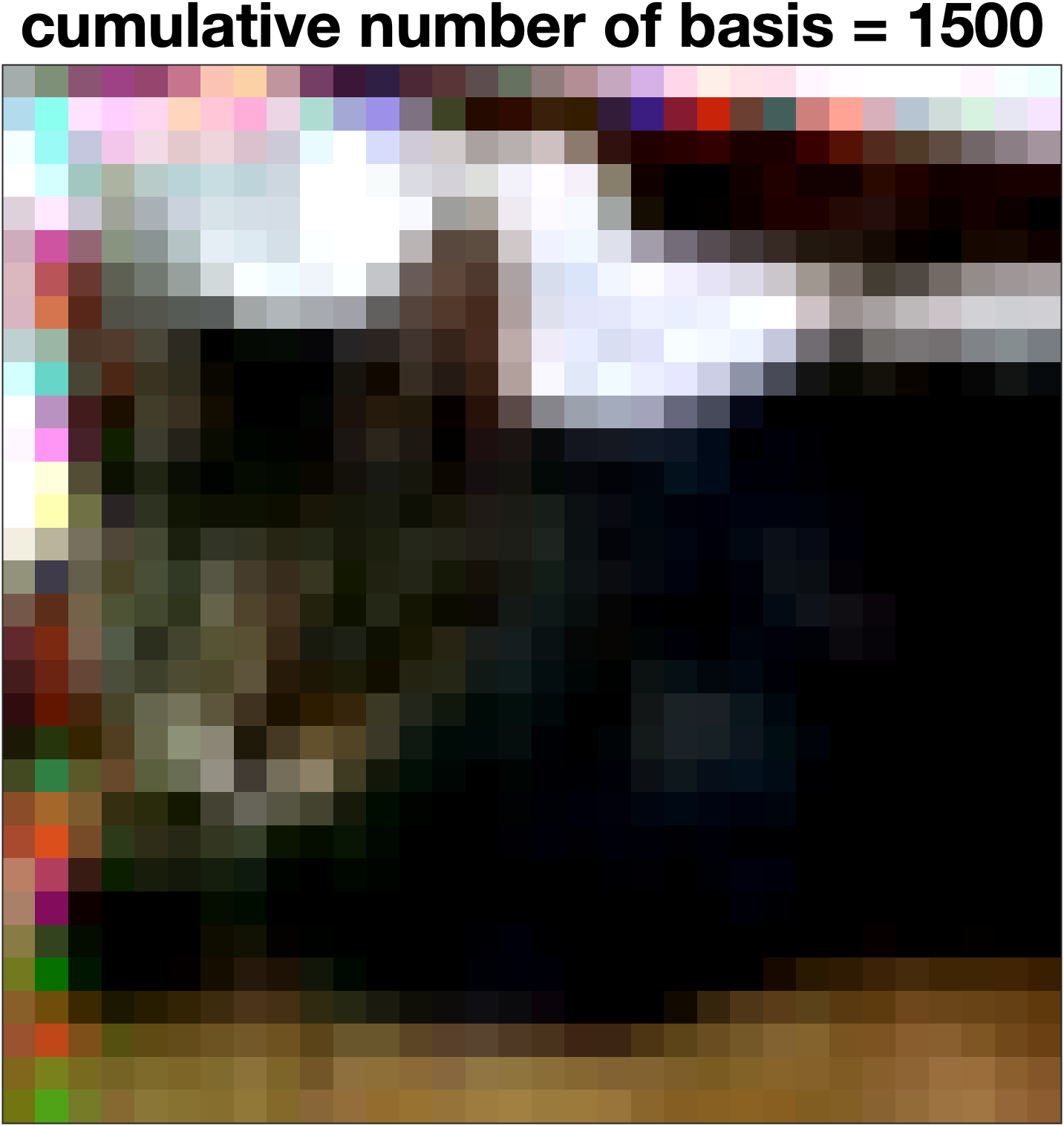}
   \includegraphics[width=0.1\linewidth]{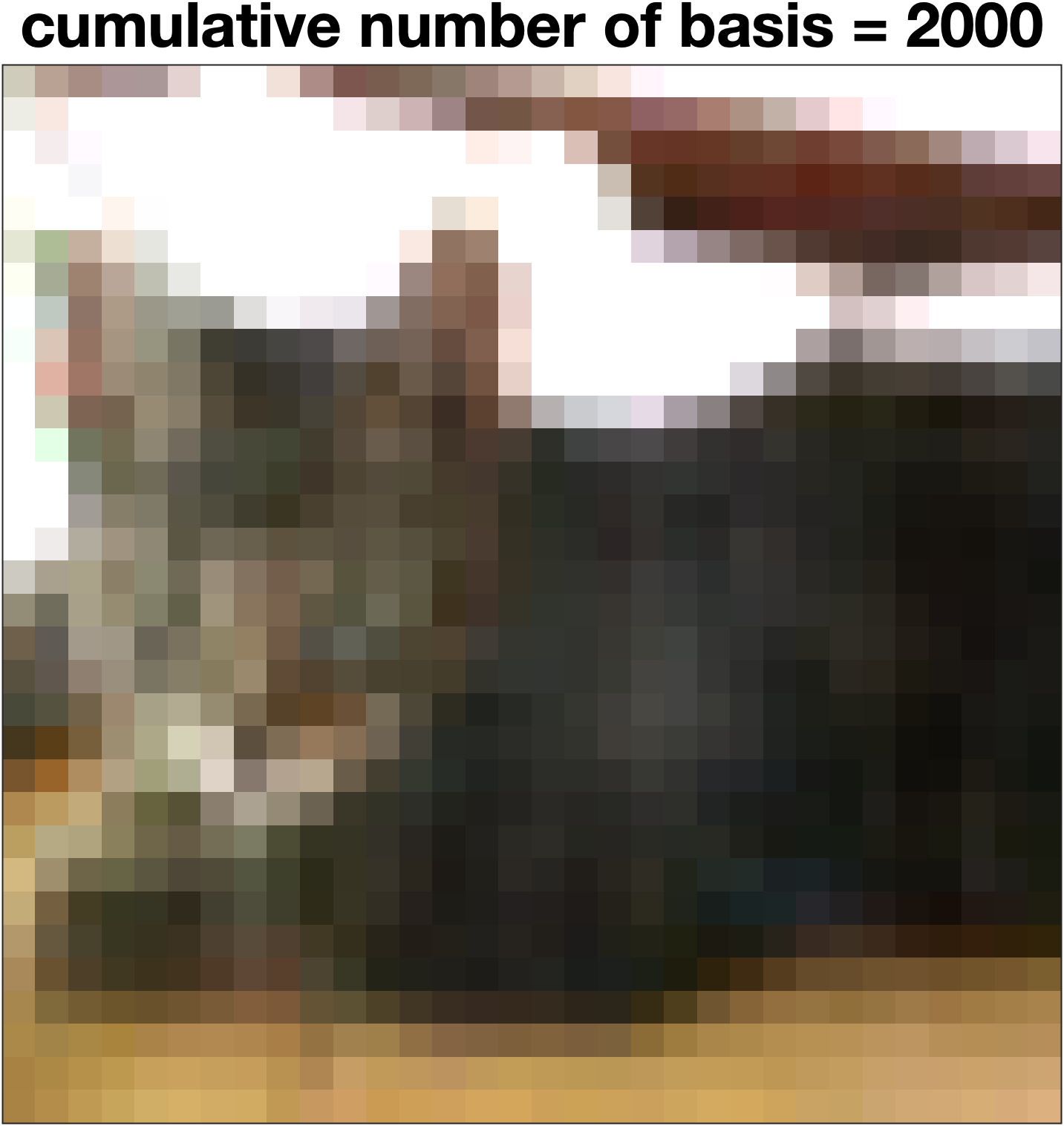}
   \includegraphics[width=0.1\linewidth]{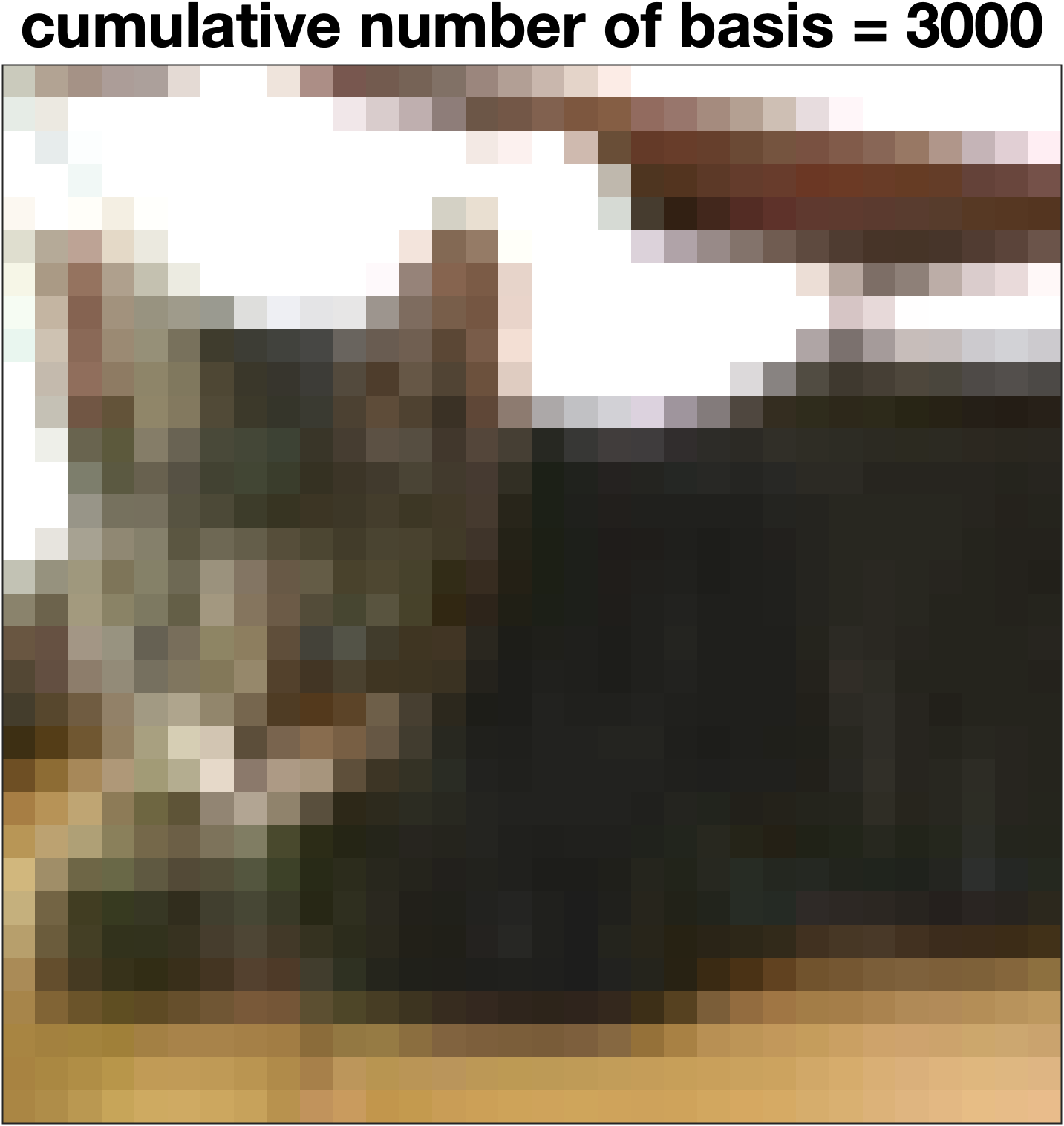}
  \caption{Images reconstructed by cumulatively adding rank-1 patterns in $V$. We see that for many images, the main structure and its distinctive features are obtained by adding a relatively small portion of patterns in the wavelet space. This is the first training image of Cat class.} %A pre-trained model on CIFAR-10 can correctly classify the four images at the right. Hence, we have captured the classification essence of the image by using only a subset of patterns.}
  \label{fig_reconstruction_tr1}
\end{figure}

%\fh{Maybe use this title? `` Low rank approximation constructs image major structure''? or something similar?}\ry{Sure.}
{\bf Low rank approximation constructs image major structure, recognizable by deep neural nets.} Each image is the summation of $m$ patterns, and each of those patterns has rank-1 in the wavelet space. Therefore, we can reconstruct low rank approximation of each image, using a relatively small number of those patterns. Figure~\ref{fig_reconstruction_tr1} shows the evolution of one cat image as $j$ moves from 1 to $m$. For example, the image at the far left of this Figure, is the result of $\sum_{j=1}^{100} \sigma_j U_i(:,j) V_i(j,:)$ and the image at the far right is the result of $\sum_{j=1}^{3000} \sigma_j U_i(:,j) V_i(j,:)$.

\begin{figure}[H]
  \centering
   \includegraphics[width=0.5\linewidth]{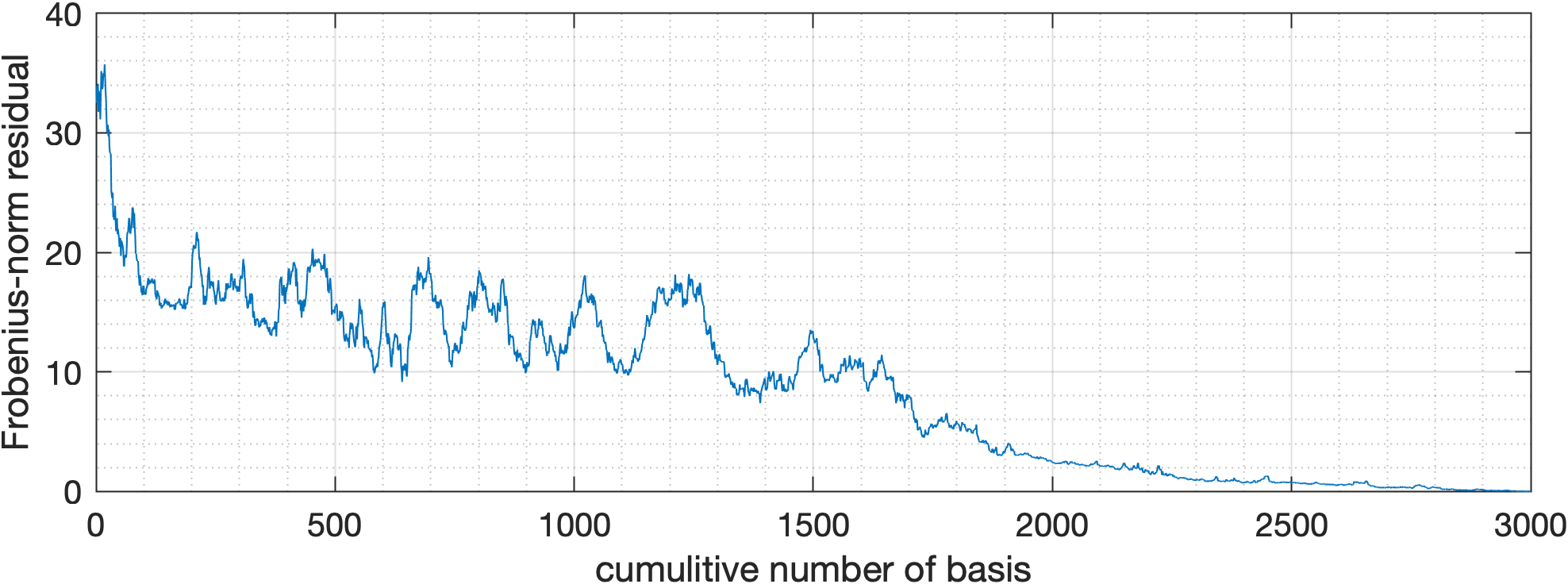}
  \caption{Residual of reconstructed image from the wavelet coefficients as we add more rank-1 patterns to it. The residual usually decreases during the process, implying that most rank-1 patterns are individually a useful contribution to the original image.
}
  \label{fig_residual_tr1}
\end{figure}

Figure~\ref{fig_residual_tr1} shows the change in the residual, where the residual is the Frobenius norm between the original image and the reconstructed image from the wavelet space, during the process of adding rank-1 images. We present similar results for more images in Appendix~\ref{sec:appx_rank1}.

\begin{figure}[H]
  \centering
  \begin{subfigure}[b]{0.495\textwidth}
  \centering
  \fbox{
   \includegraphics[width=0.15\linewidth]{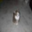}
   \includegraphics[width=0.15\linewidth]{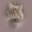}
   \includegraphics[width=0.15\linewidth]{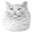}
   \includegraphics[width=0.15\linewidth]{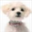}
   \includegraphics[width=0.15\linewidth]{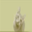}
   \includegraphics[width=0.15\linewidth]{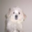}
   }
  \caption{Images with smaller norm in left basis}
  \label{fig_simple}
  \end{subfigure}
   \begin{subfigure}[b]{0.495\textwidth}
   \centering
   \fbox{
   \includegraphics[width=0.15\linewidth]{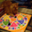}
   \includegraphics[width=0.15\linewidth]{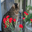}
   \includegraphics[width=0.15\linewidth]{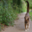}
   \includegraphics[width=0.15\linewidth]{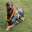}
   \includegraphics[width=0.15\linewidth]{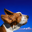}
   \includegraphics[width=0.15\linewidth]{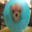}
   }
   \caption{Images with higher norm in left basis}
  \label{fig_notsimple}
  \end{subfigure}
  \caption{Complexity of images reflected through the left basis. \textbf{(a)} Images with smaller norm in left basis look simpler and have fewer distinctive features; \textbf{(b)} images with higher norm in left basis do not look simple and have several distinctive features.}
\end{figure}

{\bf Complexity of images interpreted via the left basis.}
%\fh{Complexity of images interpreted via the left basis? to make the message more explicit?}\ry{Sure!}
The left basis $U_i$ determines how the right basis vectors of $V$ should be put together in order to obtain the image. Each row of $U_i$ corresponds to one image in $D_i$. The rows of $U$ that have smaller norms correspond to images that are simpler, i.e., made of fewer components, e.g., Figure~\ref{fig_simple}. The rows of $U$ with larger norm correspond to images with more components, e.g., Figure~\ref{fig_notsimple}. Additionally, one can investigate the sparsity of rows of the left basis, which we leave in favor of brevity.

%\begin{figure}[H]
%  \centering
%   \includegraphics[width=0.1\linewidth]{cd_train_simple_9115.png}
%   \includegraphics[width=0.1\linewidth]{cd_train_simple_7879.png}
%   \includegraphics[width=0.1\linewidth]{cd_train_simple_6754.png}
%   \includegraphics[width=0.1\linewidth]{cd_train_simple_3579.png}
%   \includegraphics[width=0.1\linewidth]{cd_train_simple_3135.png}
%   \includegraphics[width=0.1\linewidth]{cd_train_simple_7502.png}
%   \caption{Images with higher norm in left basis do not look simple and have several distinctive features.}
%  \label{fig_notsimple}
%\end{figure}
%

%\textbf{Research idea: investigate whether influential right basis for Dogs can be used as adversarial attacks on Cat images, and so on. Relate them to flip points and see if they correspond to flip directions.}

%\textbf{Research Idea: Classify each of these images with a standard ResNet model and see if all of them are classified correctly. If all images are classified correctly, we can try to analyze whether ResNet has learns some of the features that HO-GSVD learns.}

%\textbf{Research idea: investigate whether any of the features in the above figure are similar to the deep features of ResNet reported in the literature.}

%{\bf Canonical Polyadic Decomposition}

{\bf Distinctive Patterns Discovered for All classes of CIFAR-10.} 
%\fh{We might not need a subsection for this?}\ry{I changed it to just bold.}
As we saw previously, one of the important aspects of our interpretation is to associate the patterns to specific classes, which is done by comparing the singular values. When we repeat this analysis for all 10 classes of this dataset, we see that patterns emerge that are associated with more than 1 class.

\begin{figure}[H]
     \centering
     \begin{subfigure}[b]{0.42\textwidth}
        \centering
        \includegraphics[width=.8\linewidth]{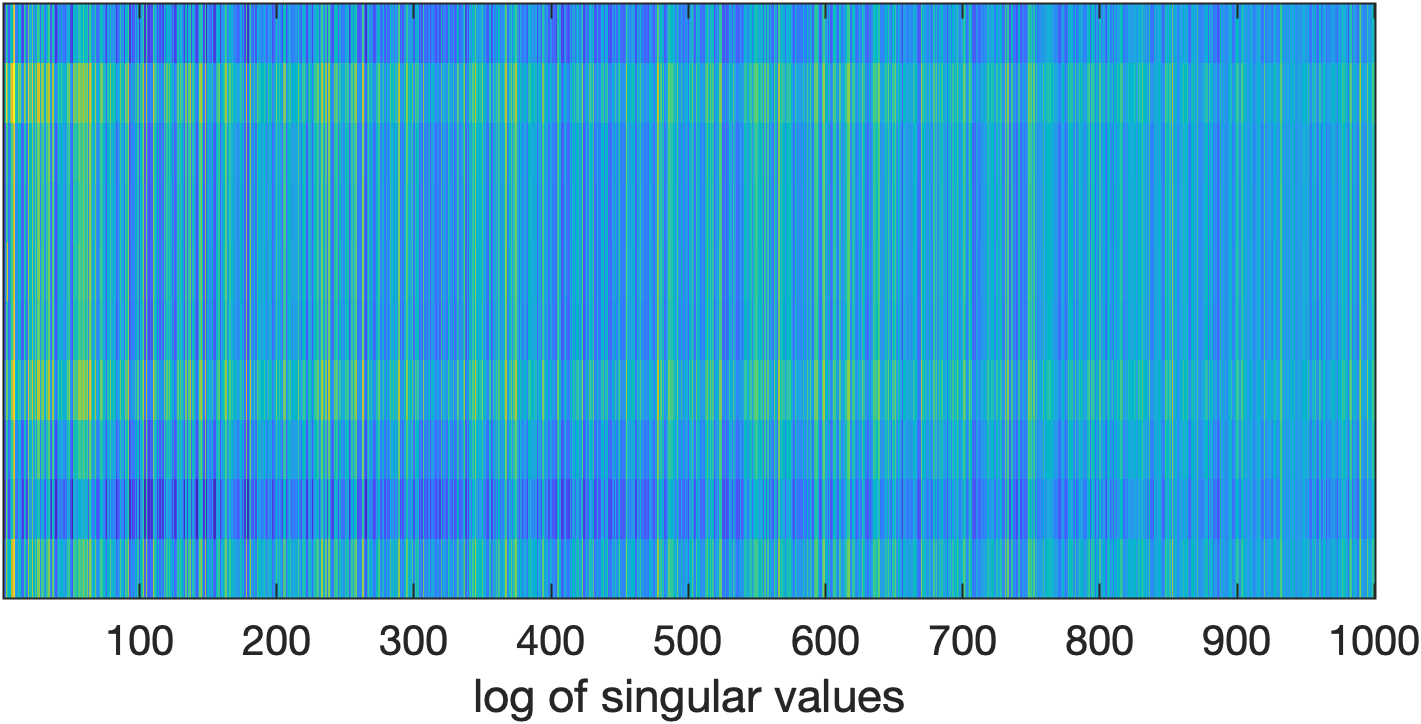}
         \caption{Correct labels}
         \label{fig:cifar_all_singulars_correct}
     \end{subfigure}
     \hspace{-1.5cm}
     \begin{subfigure}[b]{0.52\textwidth}
        \centering
       \includegraphics[width=.8\linewidth]{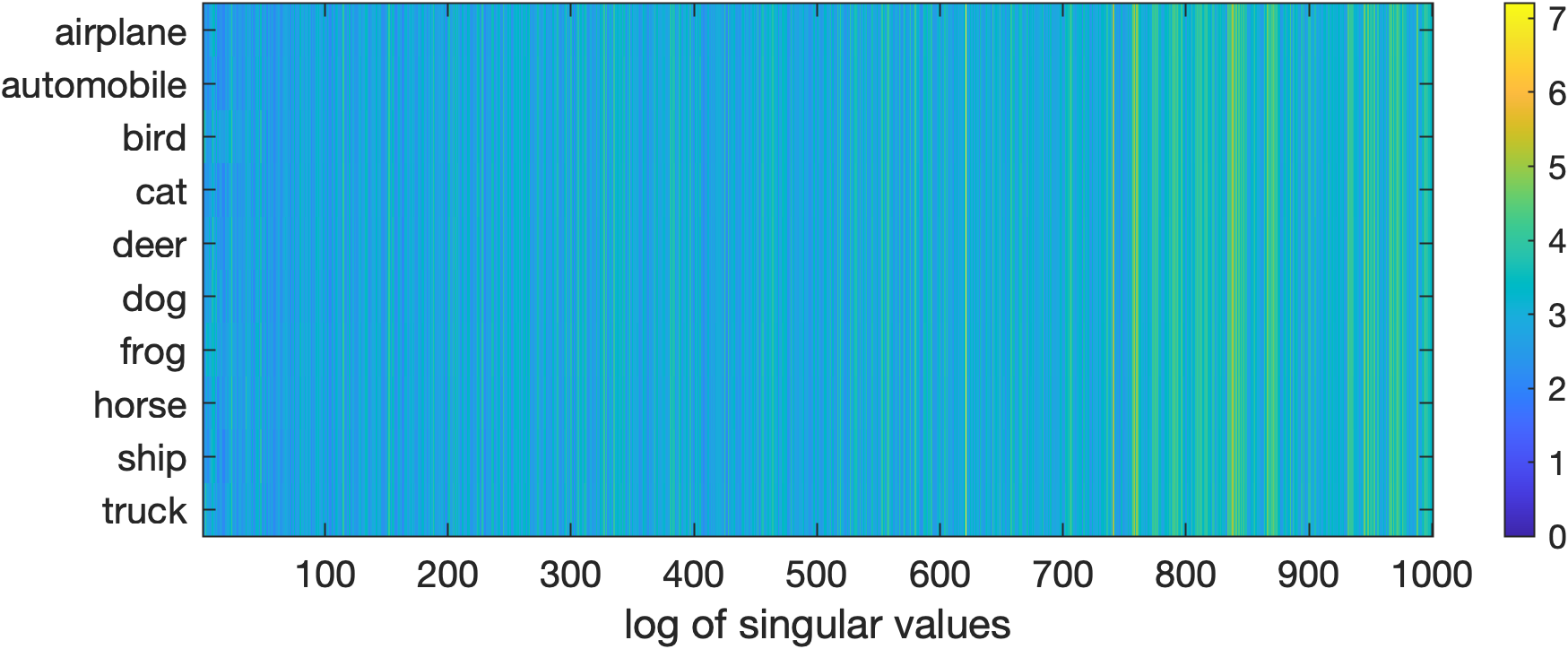}
         \caption{Random labels}
         \label{fig:cifar_all_singulars_random}
     \end{subfigure}
        \caption{The first 1,000 singular values of $\Sigma_i$'s for all 10 classes of CIFAR-10 dataset. Note that we have used logarithmic transformation in this figure. Any association between patterns and classes fades away, when we make the labels random.}
        \label{fig:cifar_all_singulars}
\end{figure}

Figure~\ref{fig:cifar_all_singulars} shows the log of 1,000 singular values obtained for this dataset. Clearly, classes of Truck and Automobile have the highest singular values for many of the patterns, which makes them distinguishable from the others, but not so helpful to distinguish each class from all other classes. This suggests that our approach could be useful in a hierarchical setting where classes are grouped and then analyzed further. Alternatively, one could consider using our method for pairwise comparison of classes as in Siamese networks \citep{bertinetto2016fully}. We next make the labels random to see how it affects the learnability of dataset.

%\fh{We could also merge Figure~\ref{fig:cifar_all_singulars} and Figure~\ref{fig:cifar_all_singulars_rand}. clean one on the left and random one on right. The colormap legend should be shared,  the class labels could be shared as well and put in the middle.}\ry{Done!}

{\bf Labeling the images randomly to diminish learn-ability.}
We repeat the spectral analysis to see if patterns will be associated with classes or not, when labels are random. We know that deep learning models can achieve perfect accuracy on this training set, even when all the images are labeled randomly, and one would not be able to detect the randomness of labels just by training a model. We show that our method reveals whether there is {\bf learnable classification information} present in the training set, useful in practice, and also useful for studying the concept of {\bf memorization vs learning}. 
%Analyzing the learning of randomly labeled data has been the subject of recent study to gain insights about generalization of models and to understand the distinction between memorization vs learning of training sets \citep{zhang2016understanding,belkin2018understand}.
When we make the labels random for all images, the decomposition we obtain is very different and ambiguous. Specifically, the singular values become uniform across classes, as shown in Figure~\ref{fig:cifar_all_singulars_random}.% Although patterns are extracted in $V$, there is no longer any association between patterns and classes. %Clearly, if we receive the randomly labeled dataset, our analysis tells us that there is no classification information to be learned from it.

\begin{figure}[H]
  \centering
   \includegraphics[width=0.6\linewidth]{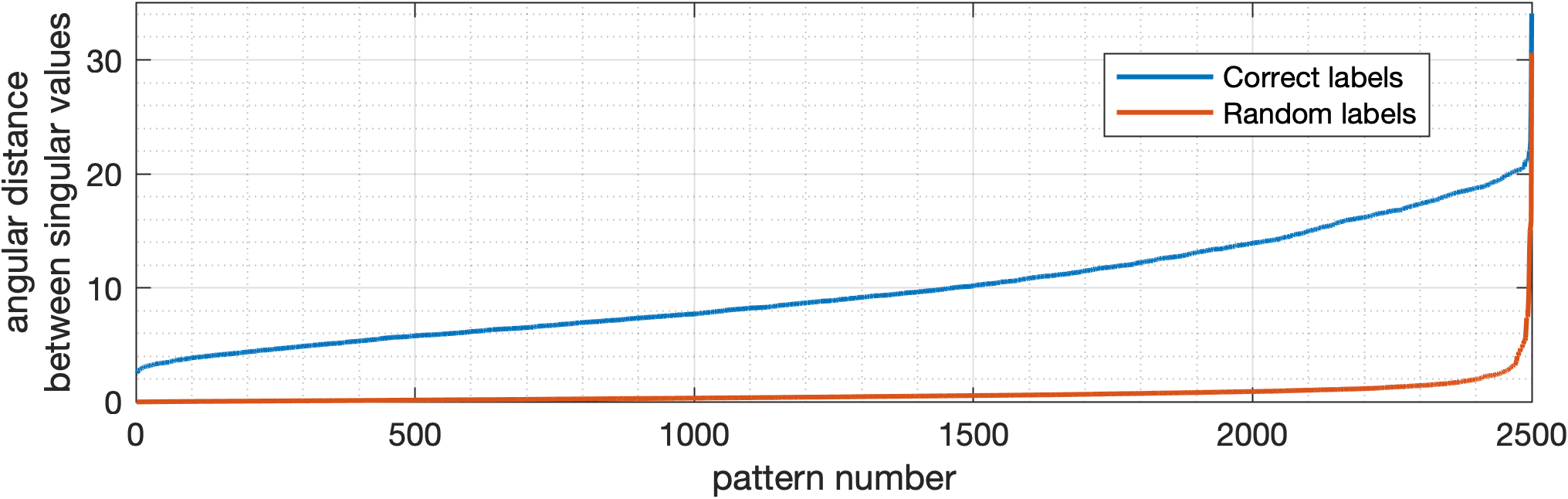}
  \caption{Angular distance (absolute value) between singular values of classes Ship and Truck, when labels are correct (\textcolor{blue}{blue}) and when labels are random (\textcolor{red}{red}). Randomizing the labels leads to collapse of angular distances and makes the patterns disassociated from the classes.}%Let's consider the 2,000th pattern in this figure. The angular distance between singular values of trucks and ships is around 15 degrees, when labels are correct, implying that this pattern is only important for one class. However, when labels are random, the angular distance is almost zero, implying that the 2000th pattern is equally important for both classes and it is not useful to distinguish one class from the other. Clearly, randomizing the labels removes the association of almost all the patterns to any class and diminishes the learnability of images for classification. That is why when a model achieves zero training loss on random labels, we call it memorization, not learning, and we expect it perform poorly on testing sets.}
  \label{fig:cifar_angular}
\end{figure}

To demonstrate this effect more clearly, we measure the angular distance between the singular values of two classes of Ship and Truck, with the correct labels (Figure~\ref{fig:cifar_all_singulars_correct}) and with random labels (Figure~\ref{fig:cifar_all_singulars_random}). As shown in Figure~\ref{fig:cifar_angular}, in the correct label case, singular values are discriminitive between the two classes, i.e., for many patterns, the angular distance is noticeable (\textcolor{blue}{blue} line). However, in the random label case, the angular distance between singular values are close to zero for almost all the patterns (\textcolor{red}{red} line).

If we limit the portion of randomly labeled data, for example, to 20\% and keep the correct labels for the rest of images, we see that the obtained results are not very different from the results we obtained for the correctly labeled data. Repeating this with different portions of random labels shows that the disassociation between patterns and labels is directly related to the portion of random labels.

\section{Conclusions and future work} \label{sec:conclusion}

Here, we showed that spectral decomposition of image classification datasets in the wavelet space can extract the patterns that distinguish each class from the others. We decomposed each image as the summation of finite number of rank-1 images in the wavelet space and showed that low rank approximation to images can capture the classification information to classify them. Our method can also be used to verify existence of learnable classification information in datasets, useful for studies on memorization vs learning of deep models, and also useful in practice for analyzing unfamiliar datasets.

Future directions of research can be to study the adversarial robustness, generalization, and functional behavior of deep classifiers with respect to rank-1 patterns extracted from datasets, and also to study the patterns in relation to deep features.

%In our experiments, {\bf we extract and analyze the patterns (rank-1 components of images in the wavelet space)} and identify which ones are discriminative for each class. We then derive {\bf low rank approximation to each image and show that the distinctive features in images can be captured by a relatively small number of those patterns}. We also explore the effect of random labels on our results. We know that deep learning models can achieve perfect accuracy on this training set, even when all the images are labeled randomly, and one would not be able to detect the randomness of labels just by training a model. We show that our method reveals whether there is {\bf learnable classification information} present in the training set, useful in practice, and also useful for studying the concept of {\bf memorization vs learning}.

%\newpage
%\section*{Broader Impact}
%\input{s7-bi.tex}

\small

\bibliographystyle{plainnat}
\bibliography{refs}

\clearpage

\renewcommand{\thefigure}{A\arabic{figure}}
\renewcommand{\thetable}{A\arabic{table}}
\setcounter{table}{0}
\setcounter{figure}{0}

\appendix

\setcounter{figure}{0}

\section{The choice of wavelet basis and the data extracted} \label{sec:appx_data}

\subsection{CIFAR-10 dataset in the wavelet space} 

About the choice for wavelet basis, it seems natural to choose a 2D basis for the images typically used for classification. A 1D wavelet basis will not be as capable to extract all the information we need from images. A 3D wavelets did not show an advantage over 2D wavelets in our numerical experiments, but it might be advantageous in certain datasets.

For the CIFAR-10 dataset, we experimented with Haar and also the first 5 Daubechies wavelets in 2D. We observed that the data extracted with Daubechies-2 was slightly more informative (higher rank), compared to Haar and Daubechies-1. But, we did not find the information extracted with Daubechies-3 to 5 to be more informative. So, our experiments on this dataset are with Daubechies-2 wavelets. This wavelet basis extracts 3,468 features from each image in this dataset, but not all of these features are influential in our pattern analysis for classification. Hence, we chose 3,000 of those coefficients using RR-QR algorithm.

We note that the end result of our analysis, e.g., the patterns extracted in the tensor decomposition were not noticeably sensitive to the choice of wavelet basis.

\begin{figure}[H]
  \centering
   \includegraphics[width=0.45\linewidth]{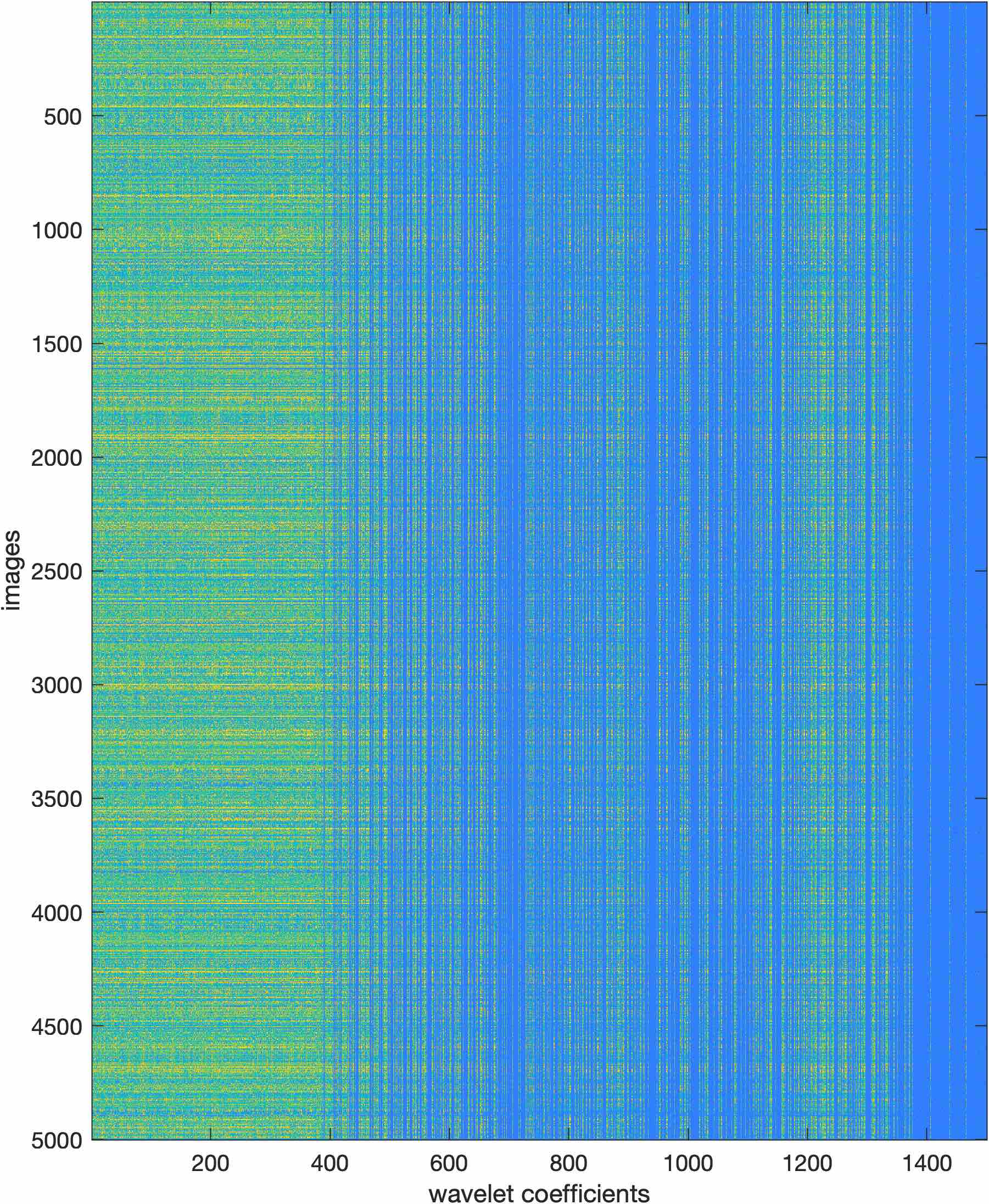}
   \hspace{.2cm}
   \includegraphics[width=0.45\linewidth]{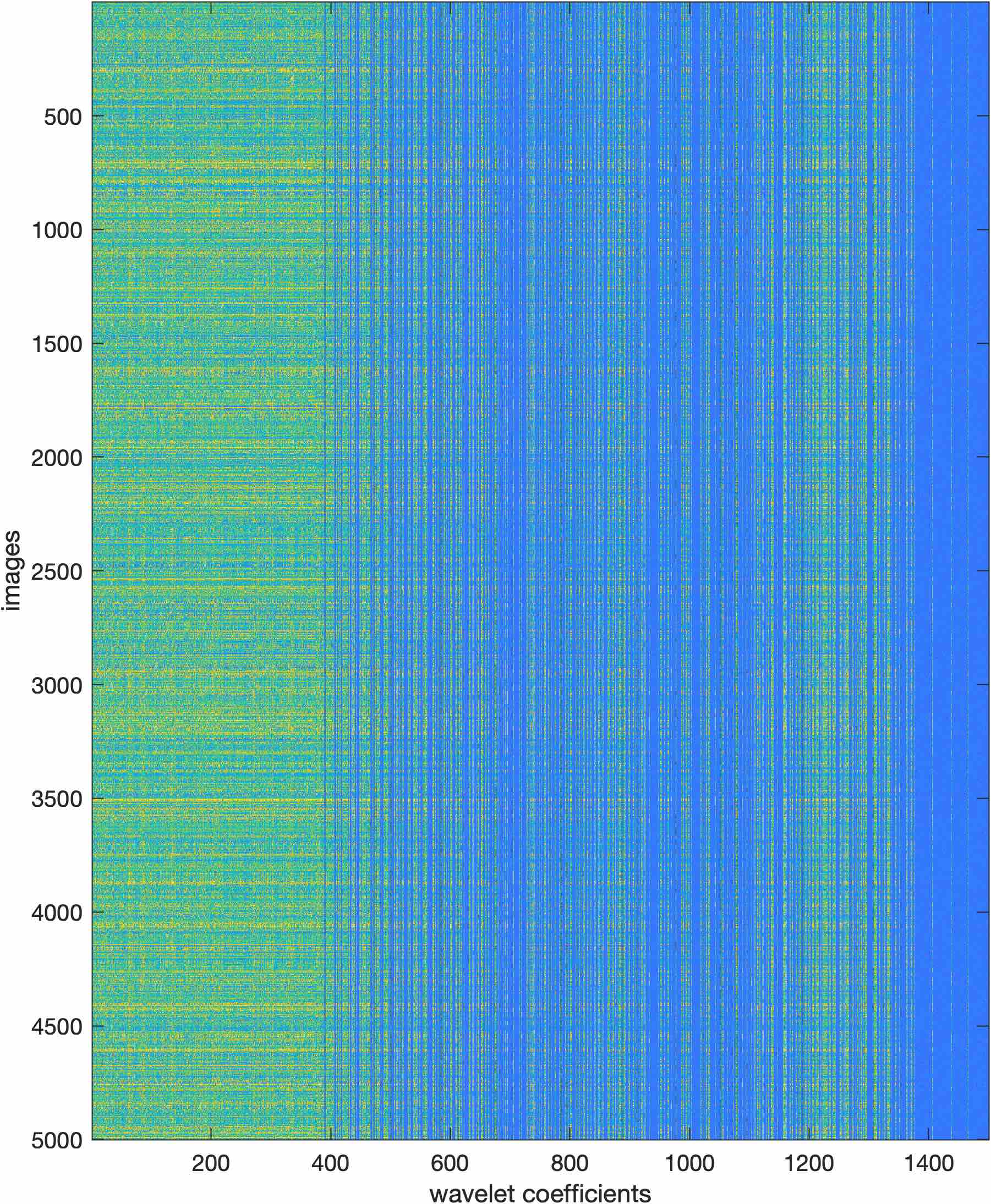}
  \caption{Matrix of wavelet coefficients for training images for classes of Cats (left) and Dogs (right) in CIFAR-10 dataset. This figure shows the first 1,500 most influential wavelet coefficients of images, chosen by the RR-QR algorithm.}
  \label{fig:cifar10_cd_matrix}
\end{figure}

Figure~\ref{fig:cifar10_cd_matrix} shows only the first 1,500 columns for each $\mathcal{D}_i$, to show how the data generally looks like. But, note that we actually used 3,000 wavelet coefficients in our experiments.
 
\subsection{MNIST dataset in the wavelet space}

For MNIST, again we did not find the choice of wavelet basis to be noticeable in the patterns extracted. We found the Daubechies-1 to transform the images similar to higher order Daubechies wavelets. We note that the data extracted from this dataset has significant redundancies and linear dependencies compared to the data extracted from CIFAR-10. This is to be expected because the images of MNIST are very simple compared to images of CIFAR-10.

\begin{figure}[H]
  \centering
   \includegraphics[width=.25\linewidth]{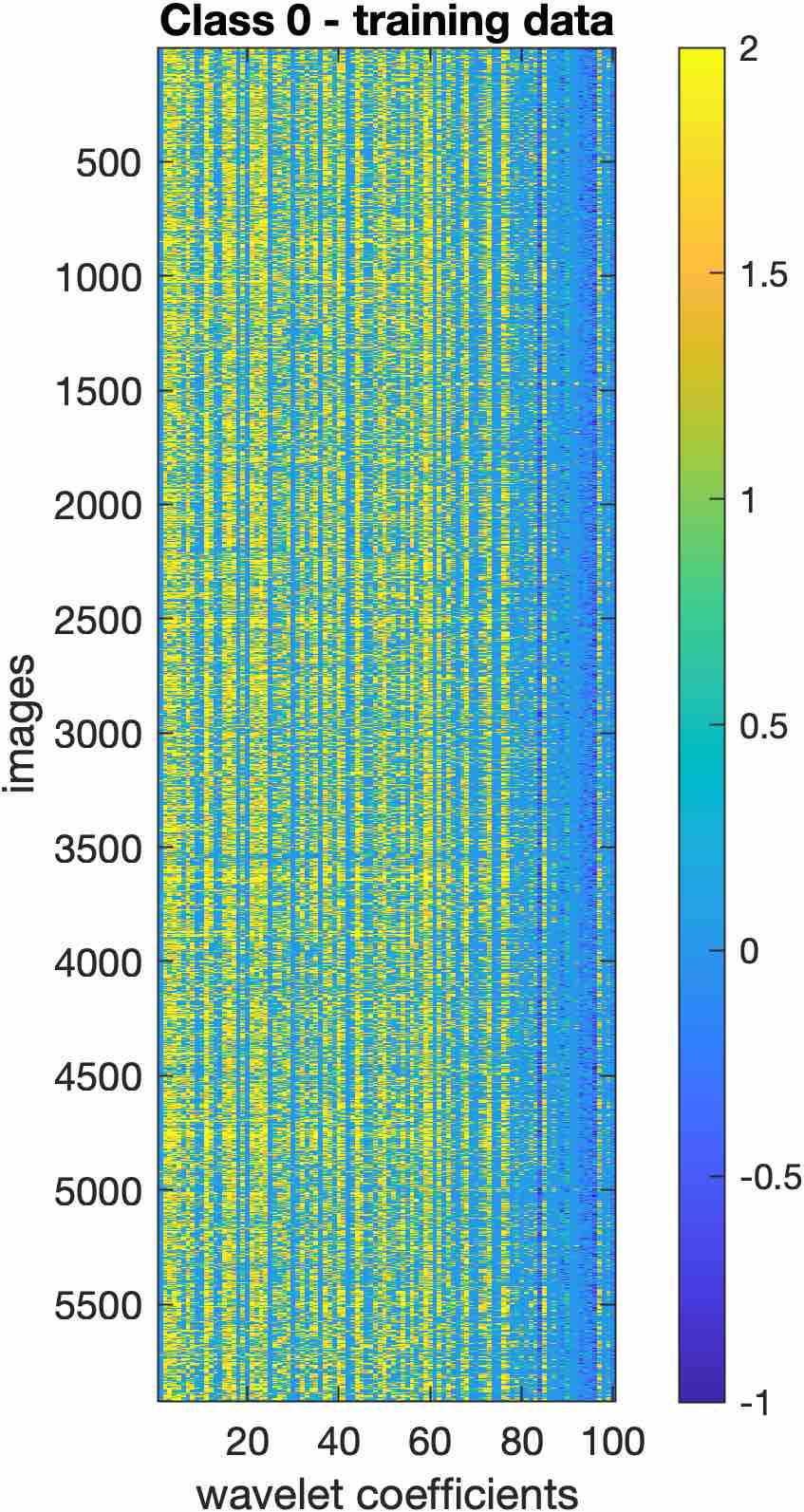}
   \includegraphics[width=0.22\linewidth]{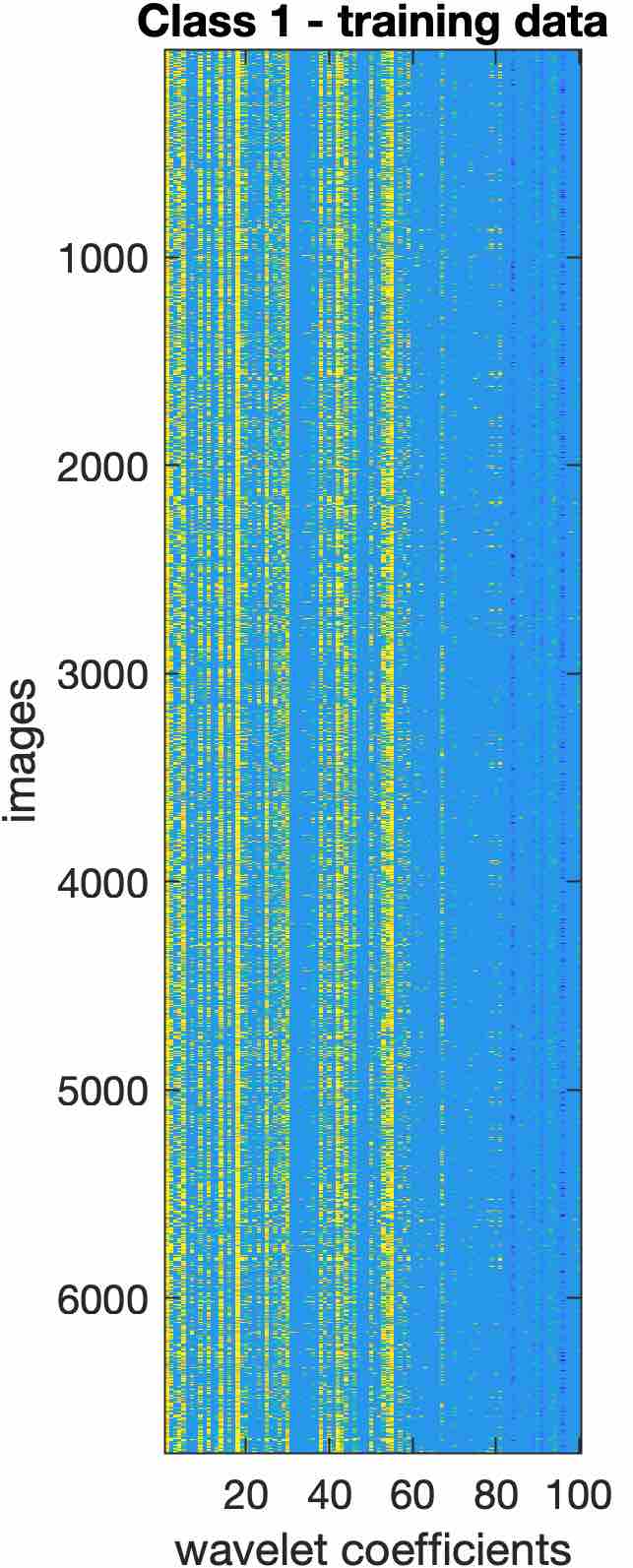}
   \includegraphics[width=0.22\linewidth]{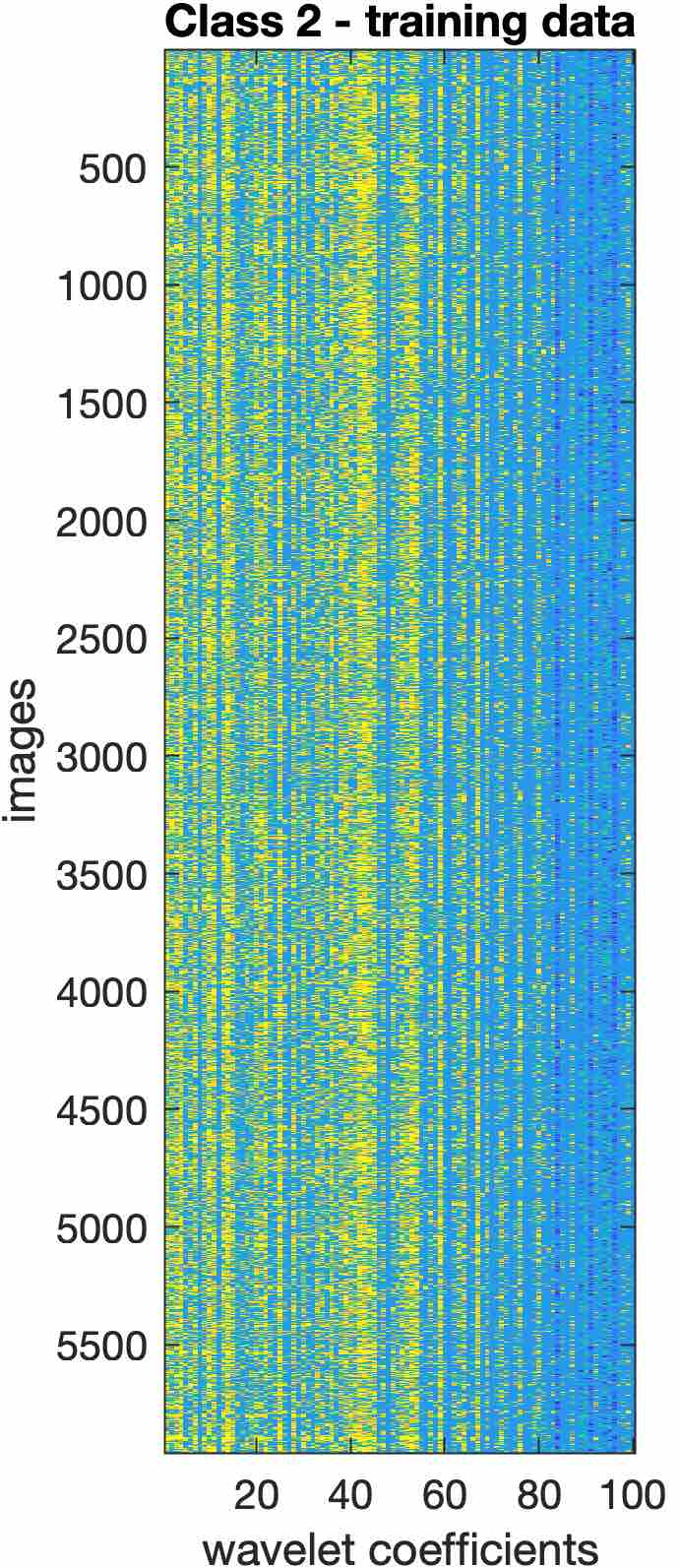}
   \includegraphics[width=0.22\linewidth]{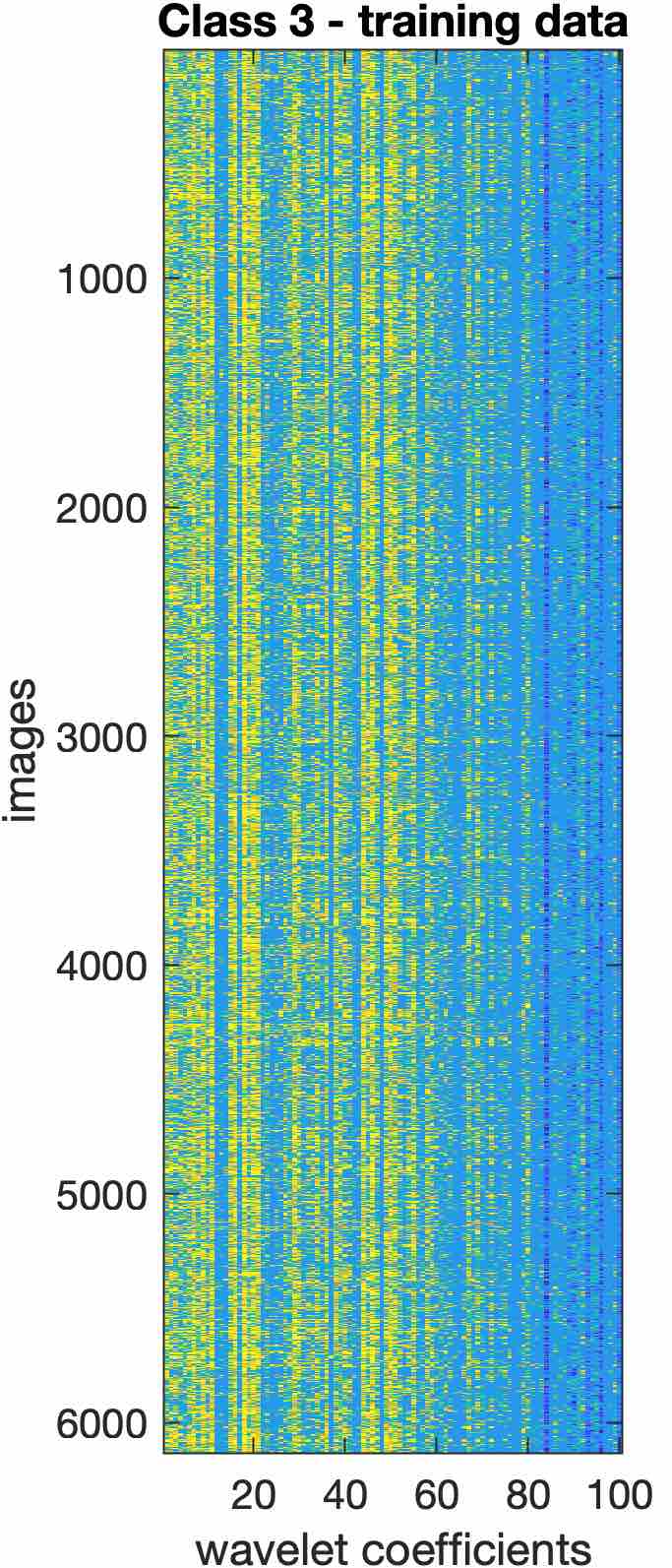}\\
   \vspace{.2cm}
   \includegraphics[width=0.22\linewidth]{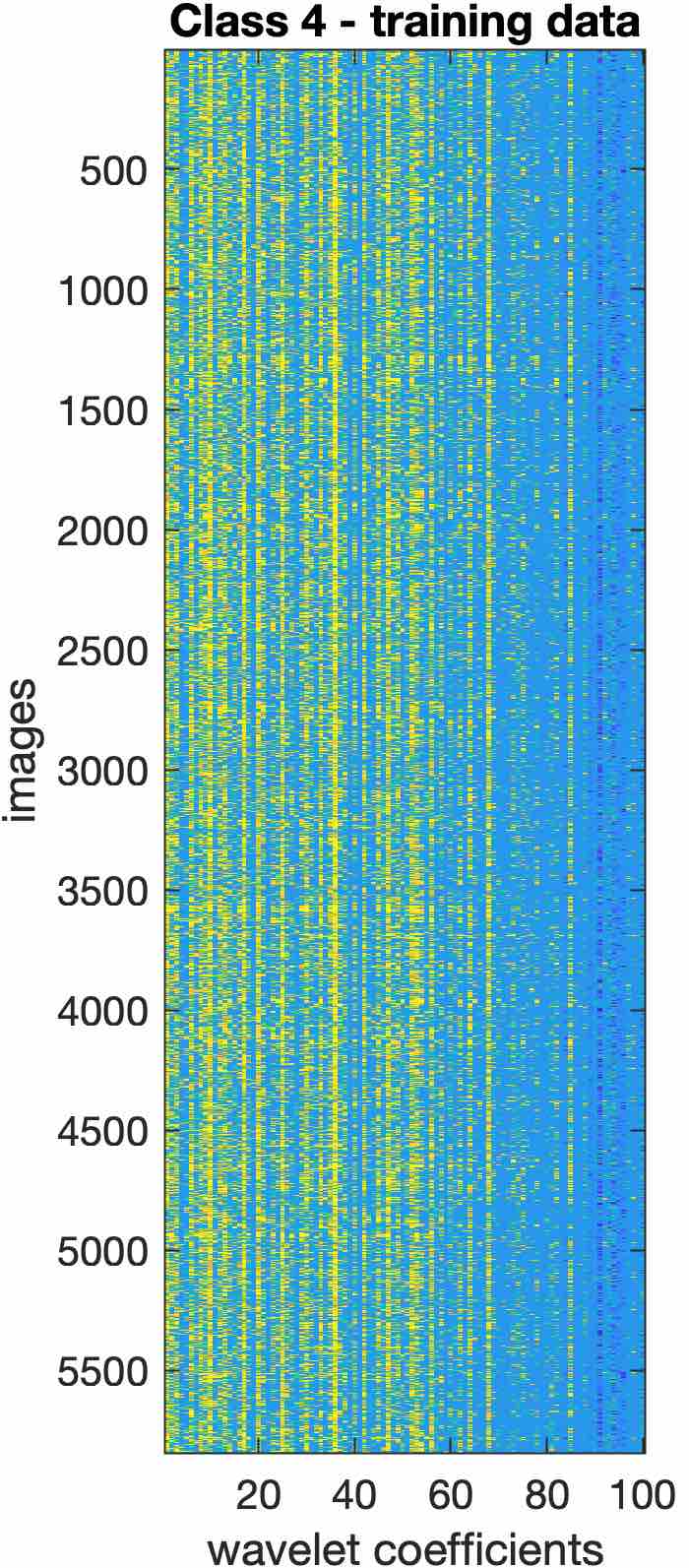}
   \includegraphics[width=0.22\linewidth]{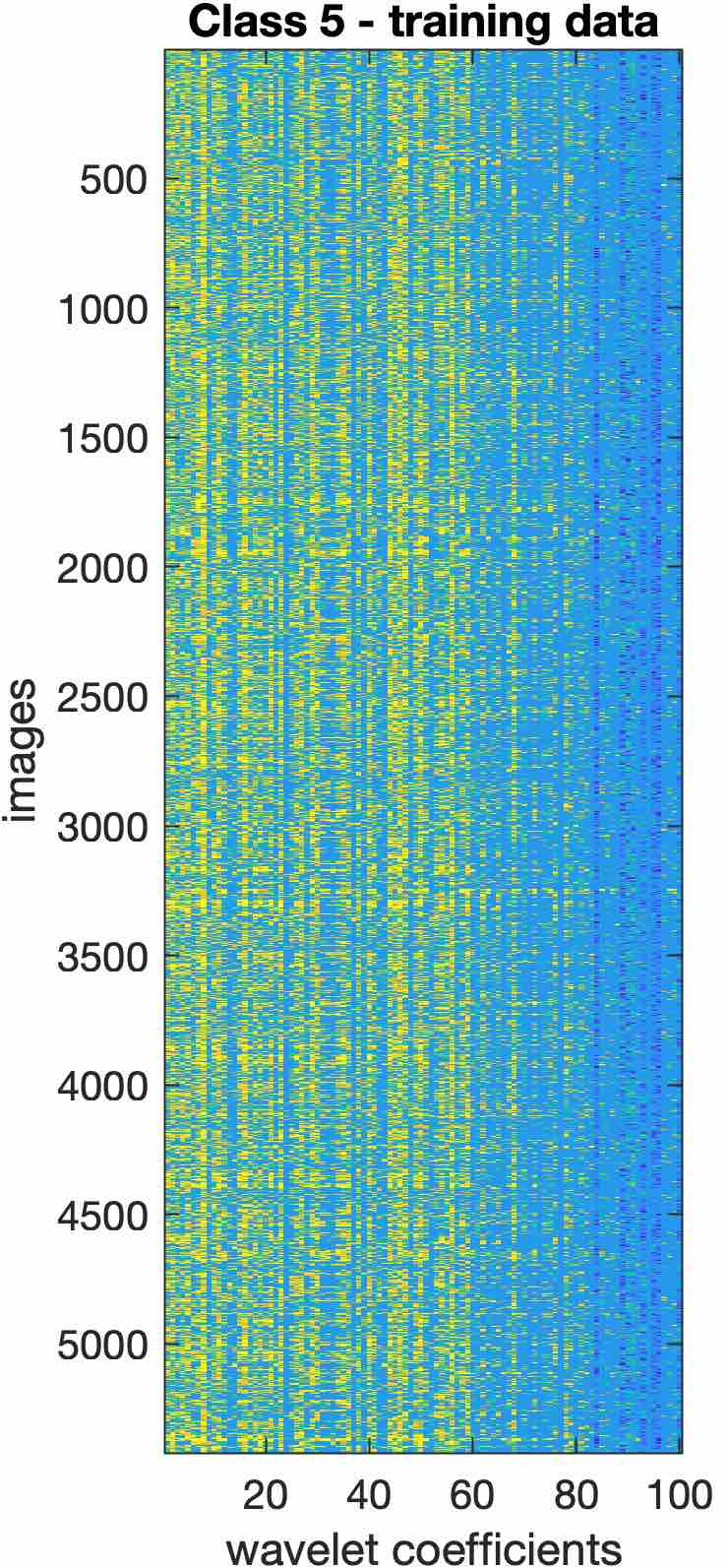}
   \includegraphics[width=0.22\linewidth]{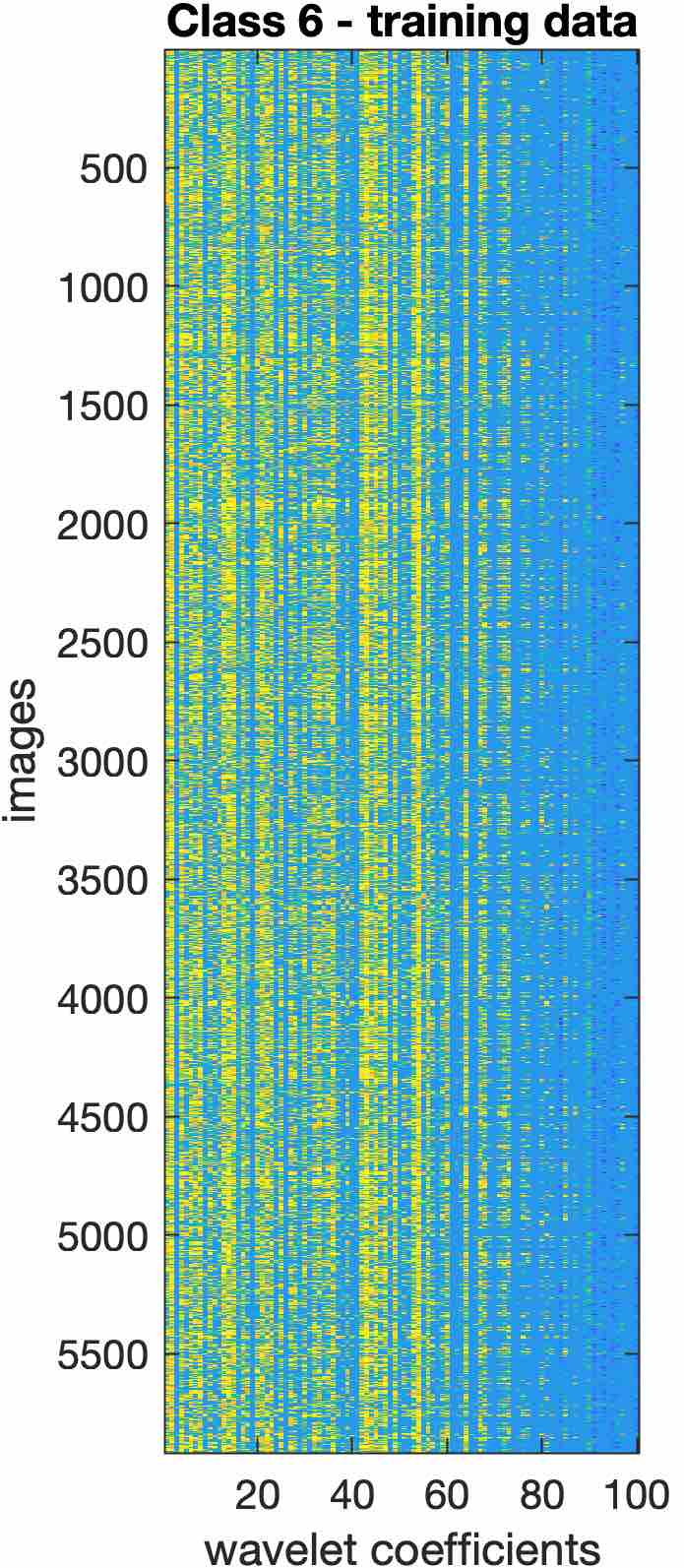}
   \includegraphics[width=0.22\linewidth]{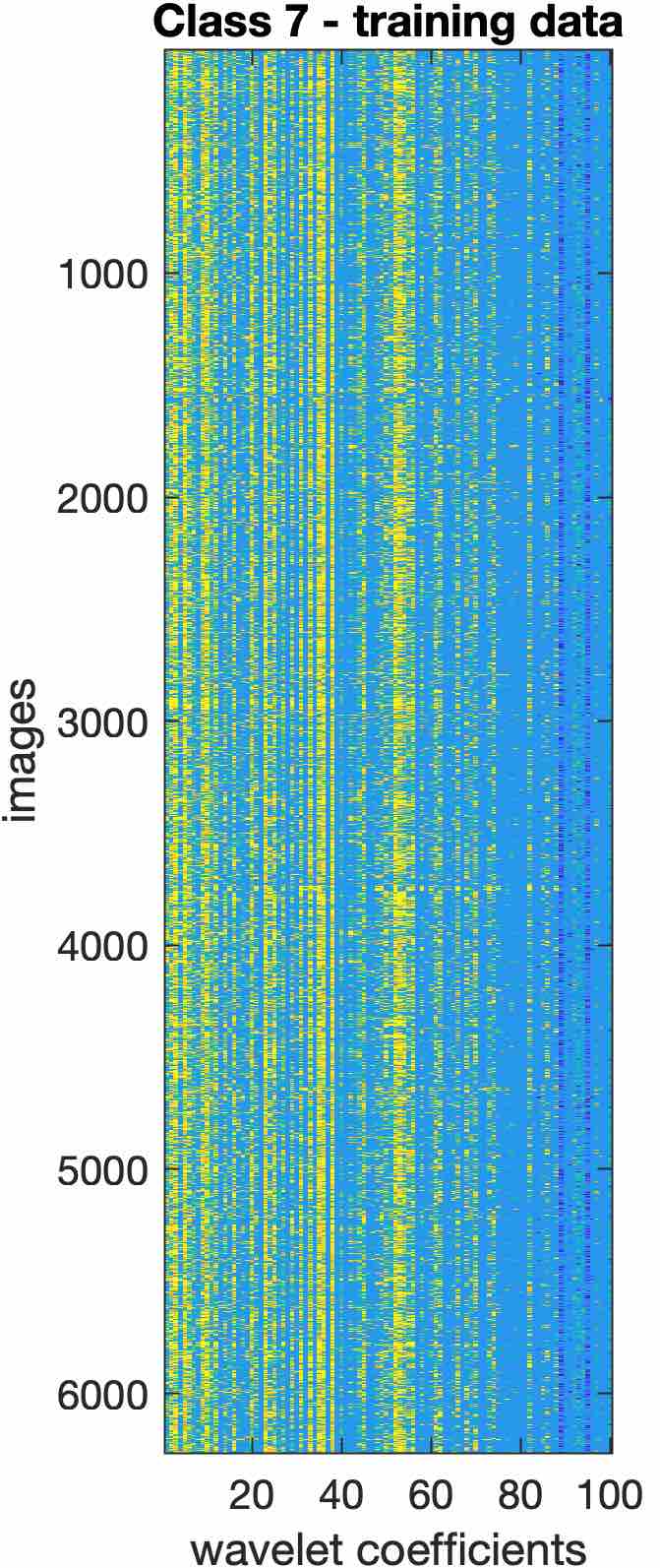}\\
   \vspace{.2cm}
   \includegraphics[width=0.22\linewidth]{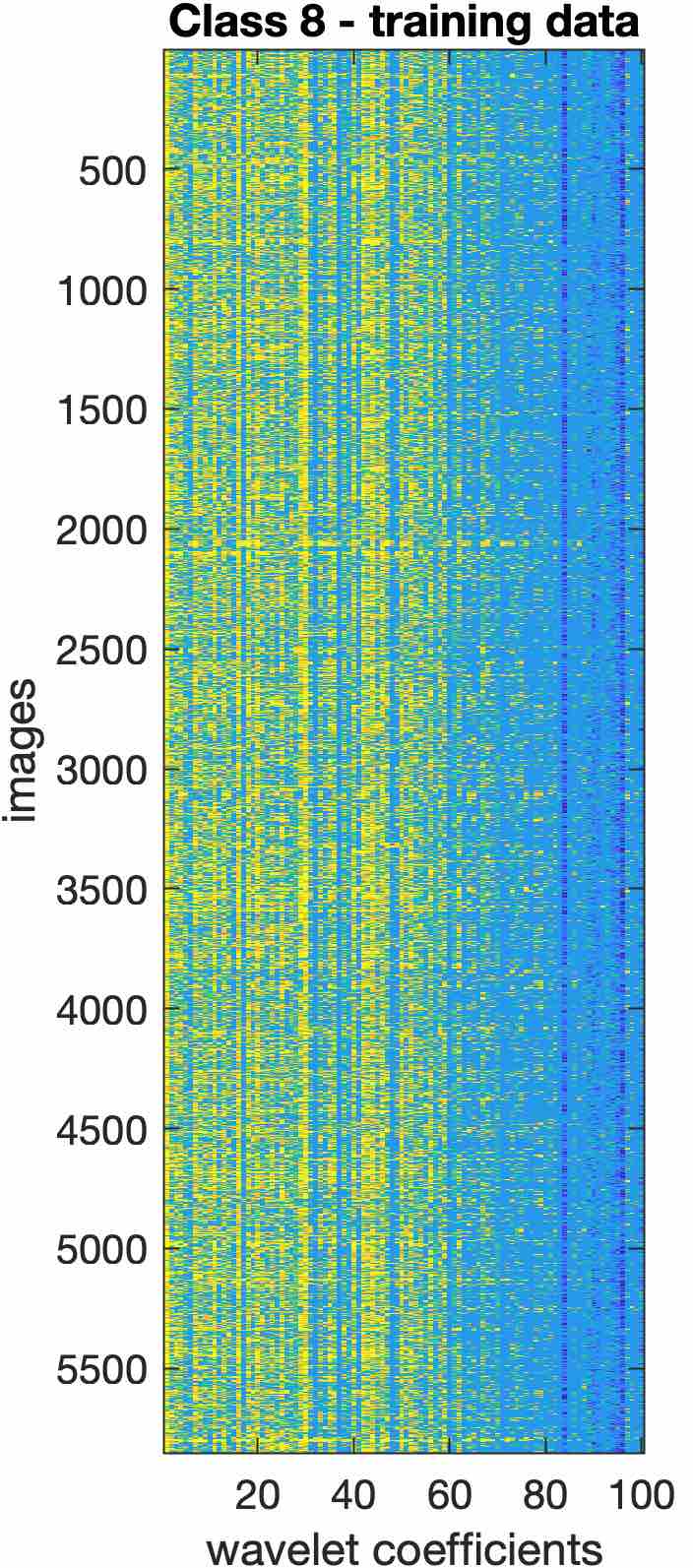}
   \includegraphics[width=0.22\linewidth]{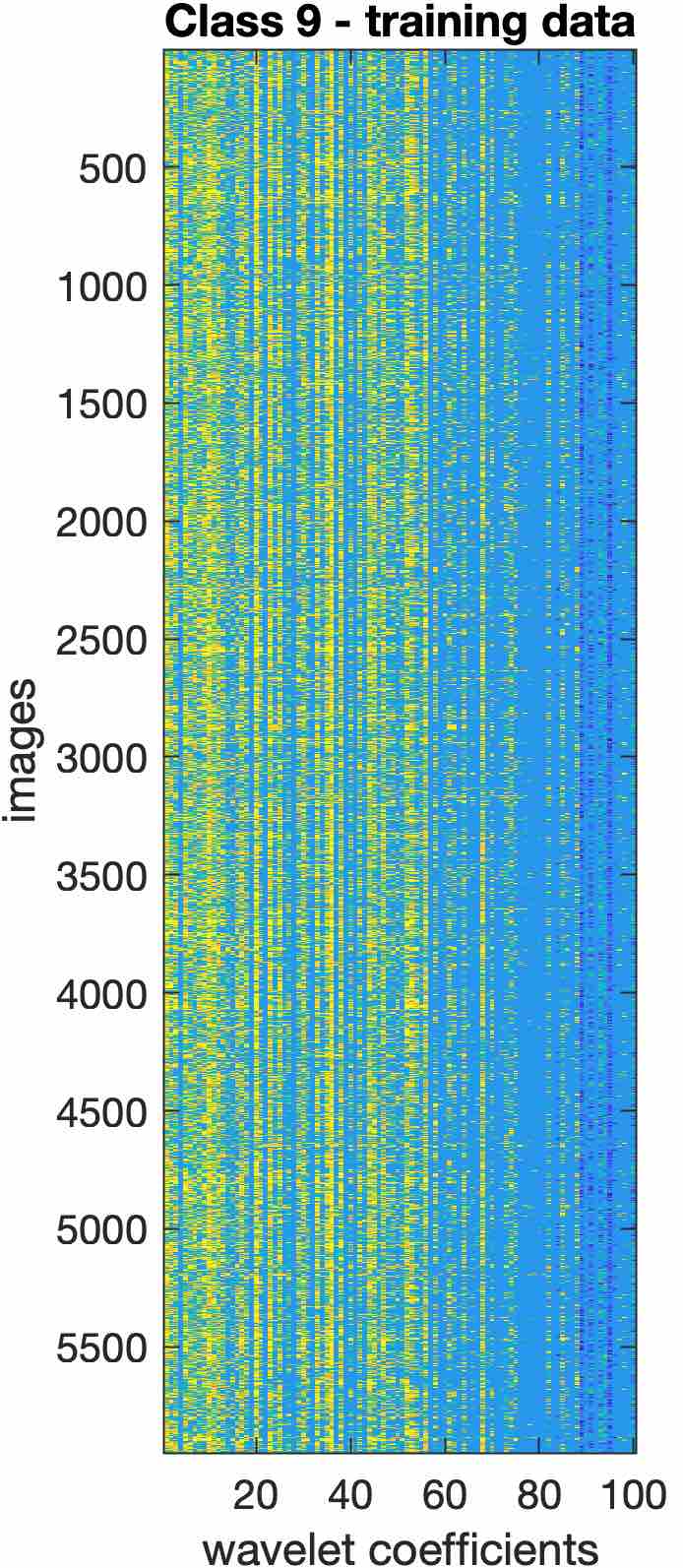}
  \caption{Matrix of wavelet coefficients for training images of 10 classes of MNIST dataset. Columns represent 100 of the most influential wavelet coefficients of images, chosen by the RR-QR algorithm. The same color scale is used for all figures.}
  \label{fig:mnist_wmatrix}
\end{figure}

Figure~\ref{fig:mnist_wmatrix} shows each of the $\mathcal{D}_i$'s for the 10 classes of MNIST.

\begin{figure}[H]
  \centering
   \includegraphics[width=0.4\linewidth]{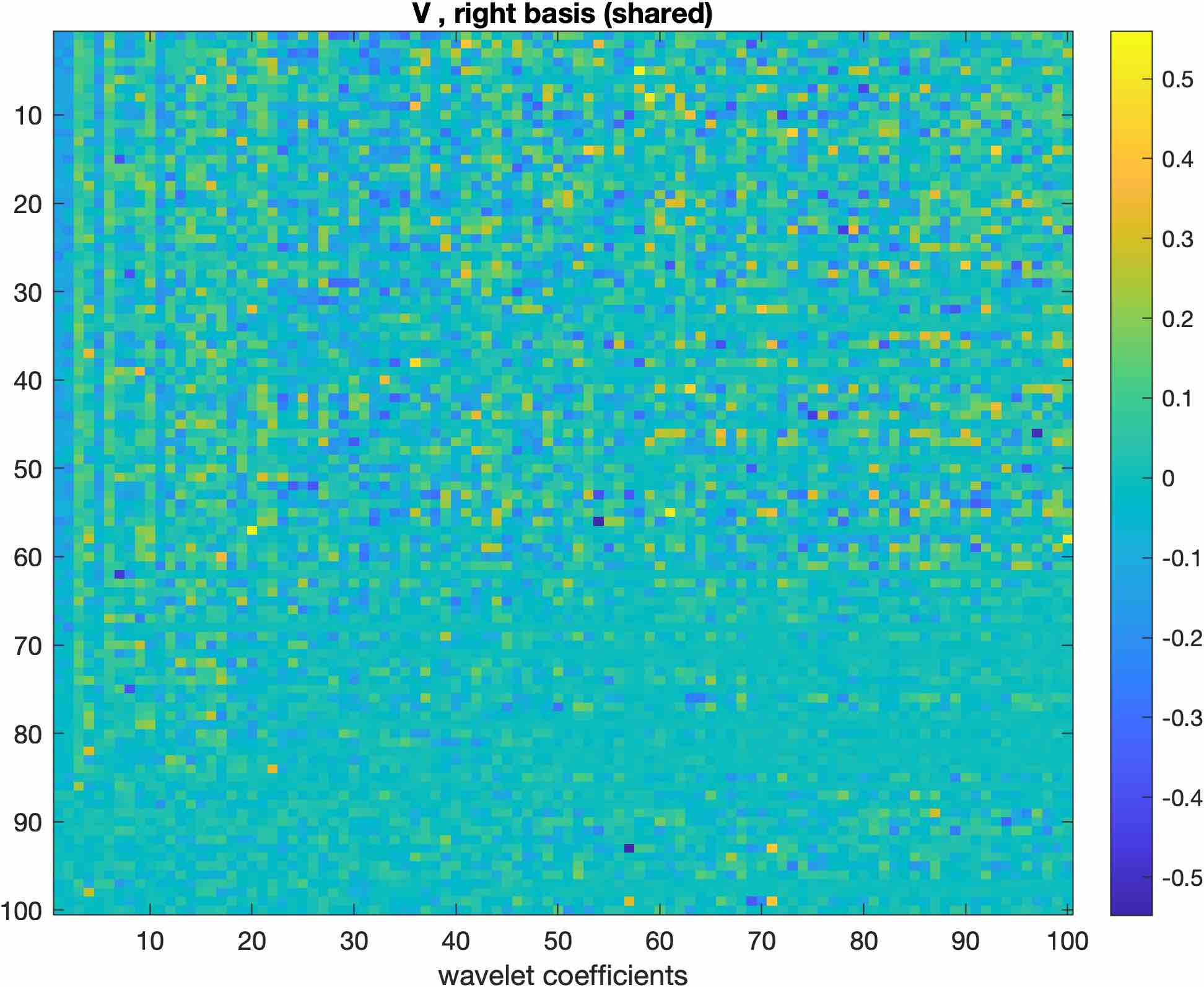}
  \caption{The right basis $V$ obtained from Higher Order GSVD, which is common among all classes of MNIST. Images in Figure~\ref{fig:mnist_rightbasis} are reconstruction of specific columns of this matrix, back into the pixel space.}
  \label{fig:mnist_right}
\end{figure}

Figure~\ref{fig:mnist_right} shows the right basis $V$ which is common among all classes.

\setcounter{figure}{0}
\renewcommand{\thefigure}{B\arabic{figure}}

\section{Additional examples for reconstructing images as summation of rank-1 images} \label{sec:appx_rank1}

Here, we provide an example of writing a testing image as summation of rank-1 images obtained from the training set (in the wavelet space). Hence the same common basis obtained from all classes of training set, works for efficient decomposition of testing images, as well.

\begin{figure}[H]
  \centering
   \includegraphics[width=0.11\linewidth]{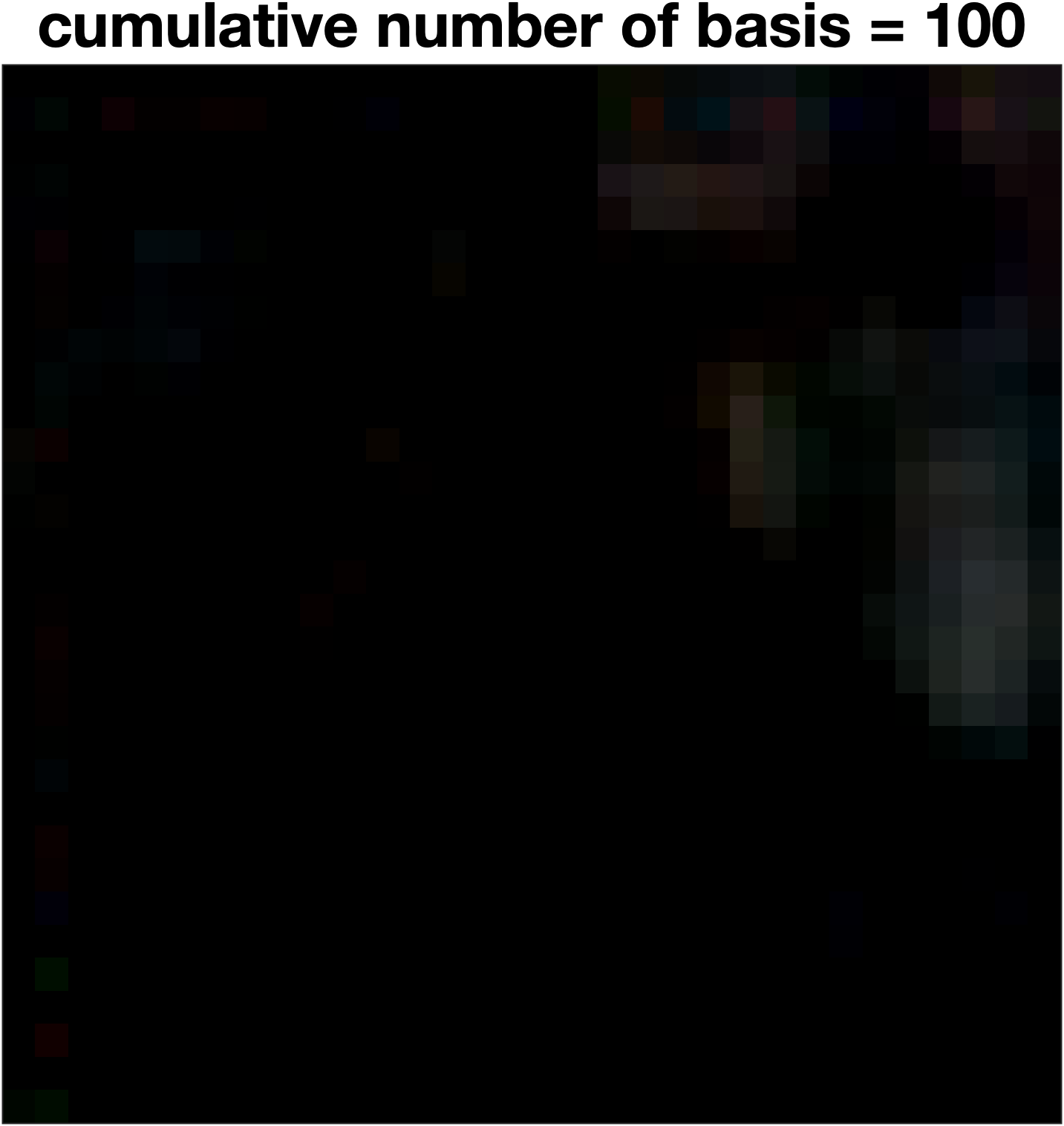}
   \includegraphics[width=0.11\linewidth]{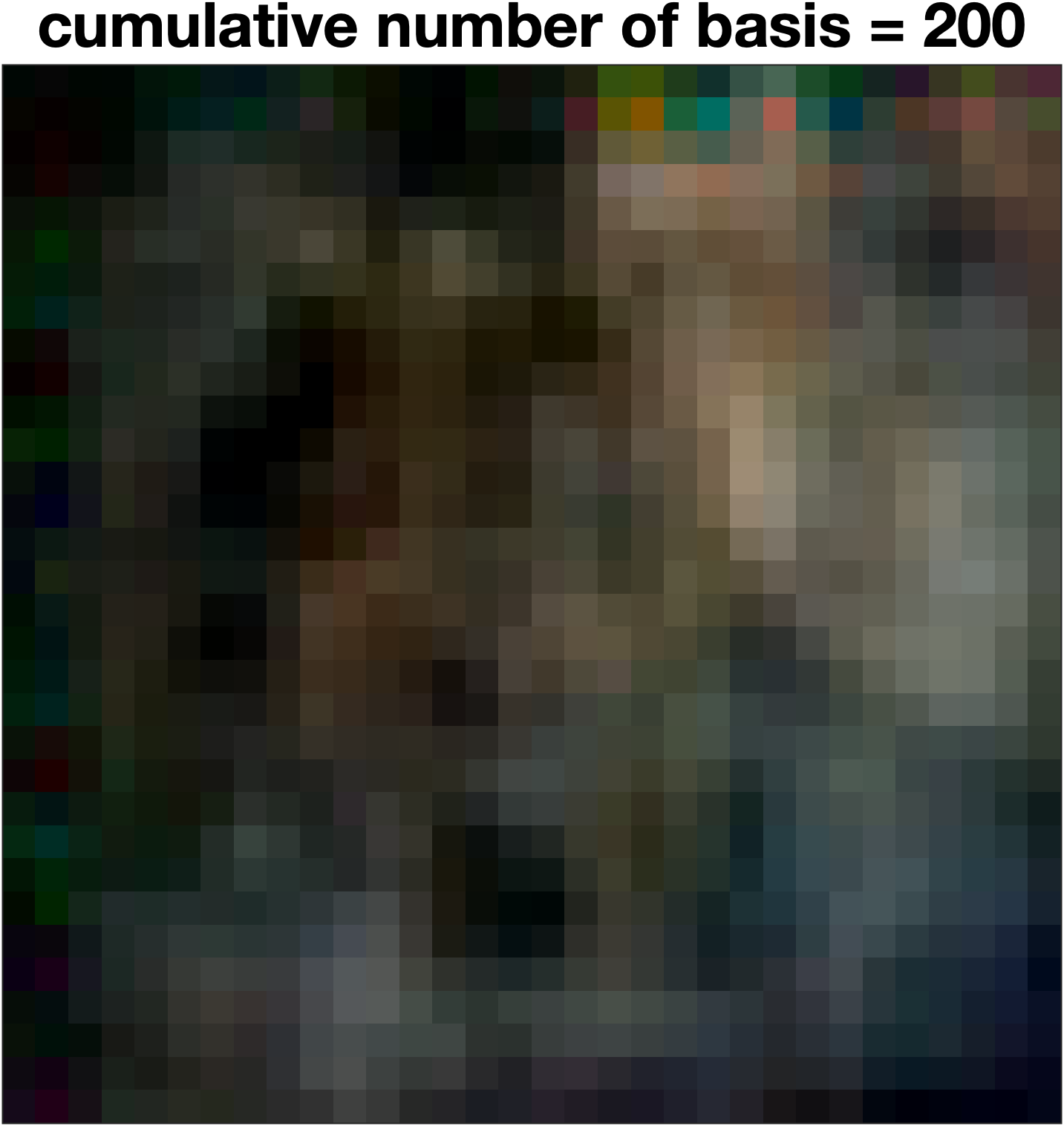}
   \includegraphics[width=0.11\linewidth]{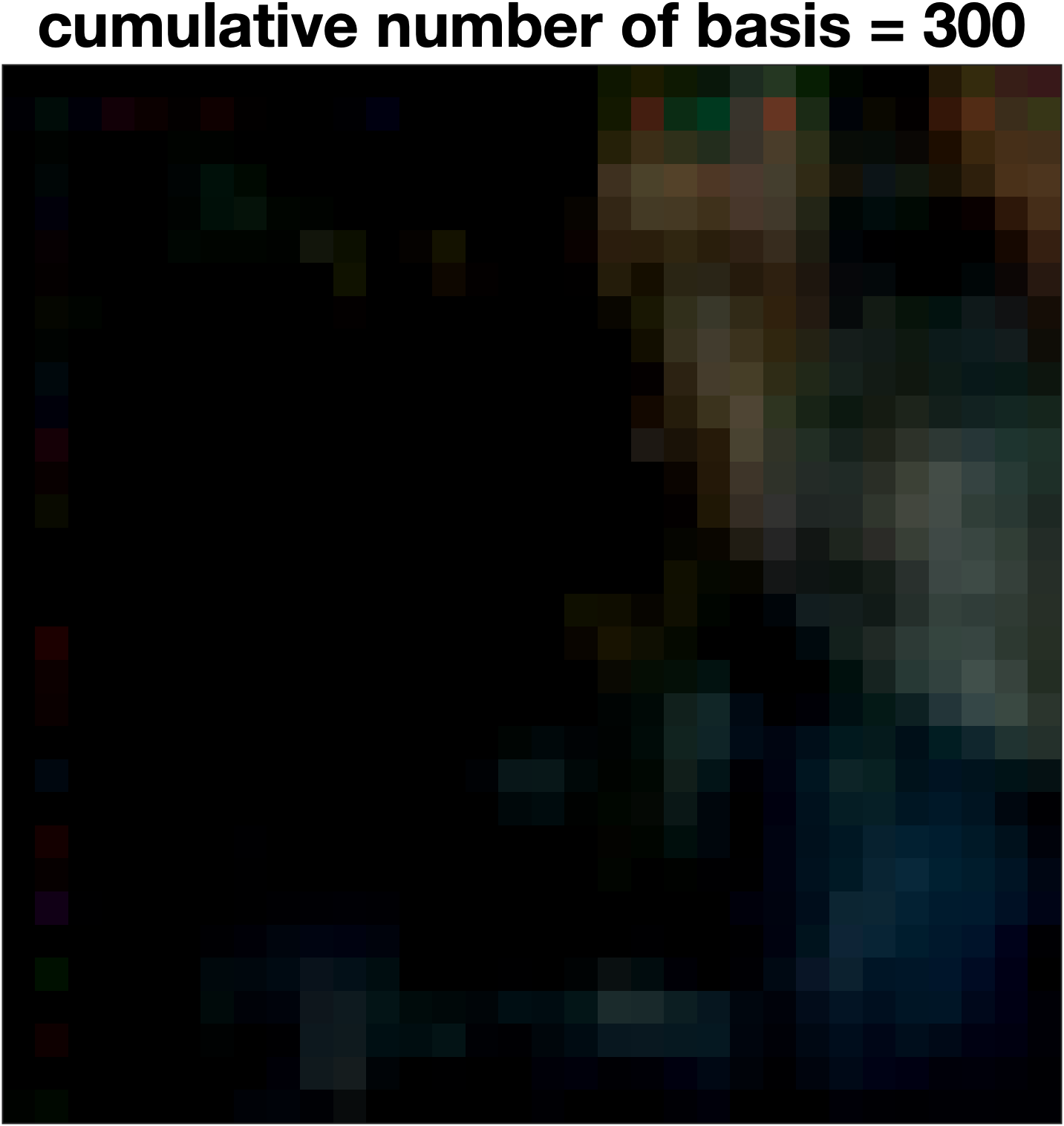}
   \includegraphics[width=0.11\linewidth]{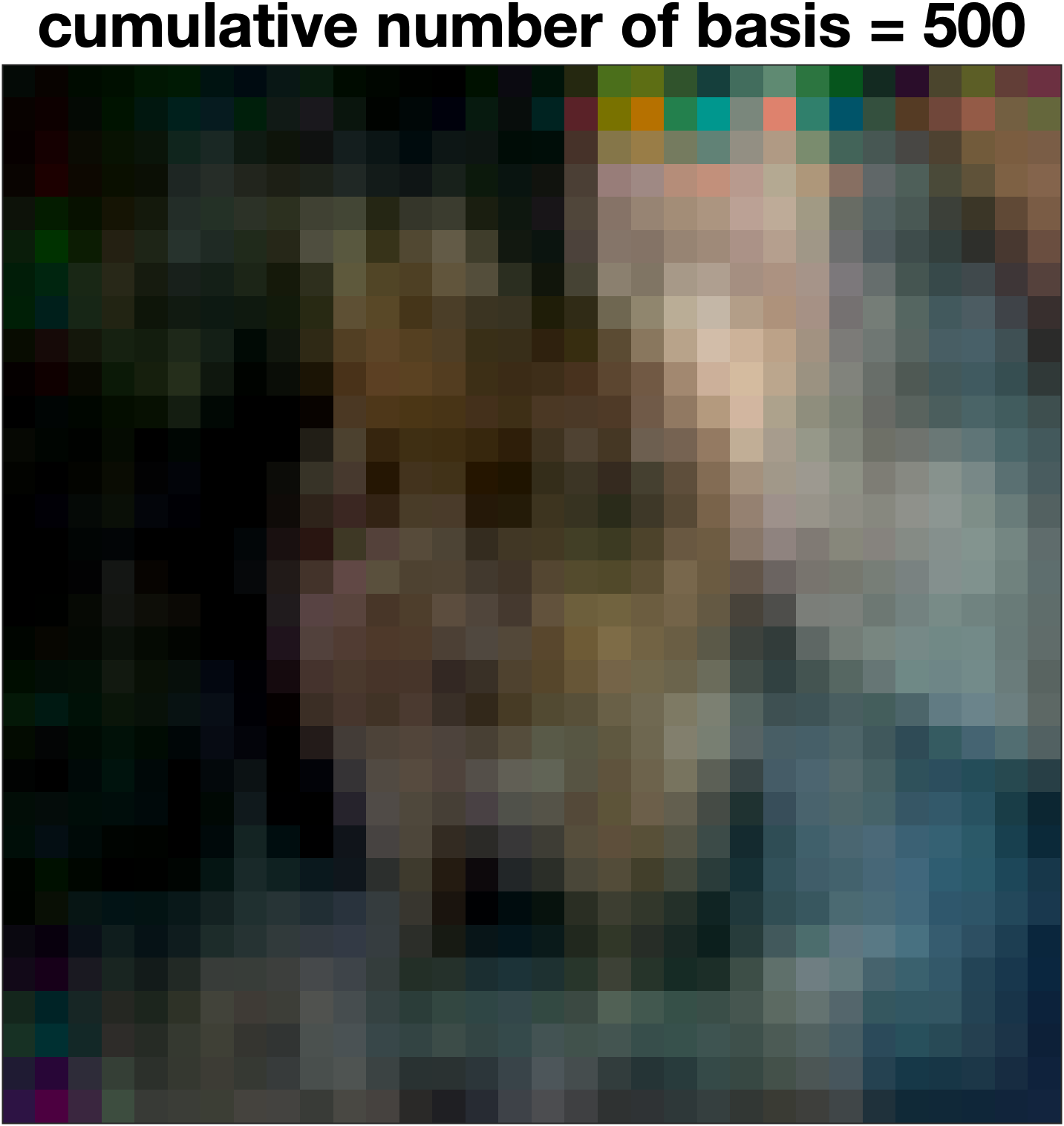}
   \includegraphics[width=0.11\linewidth]{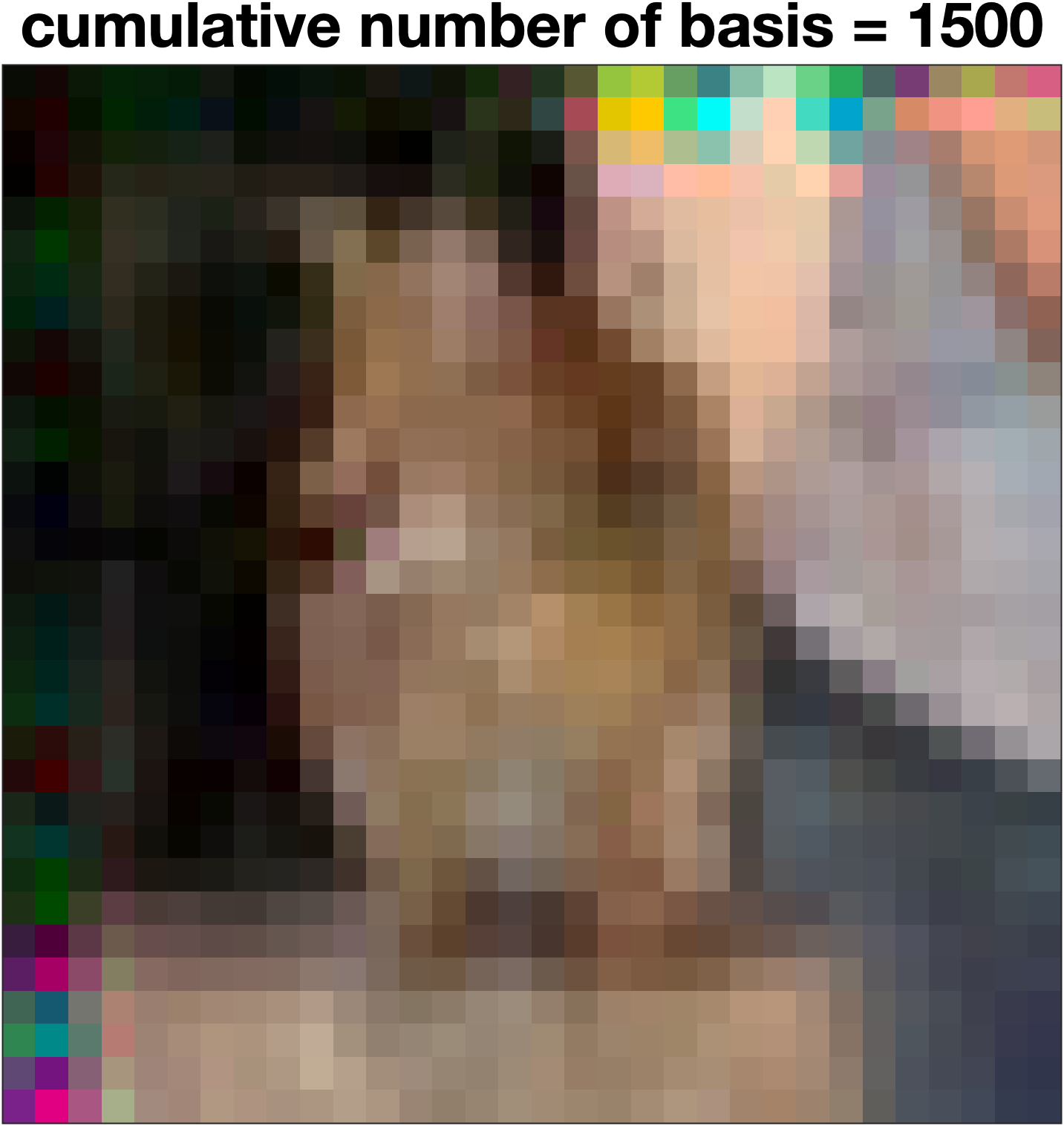}
   \includegraphics[width=0.11\linewidth]{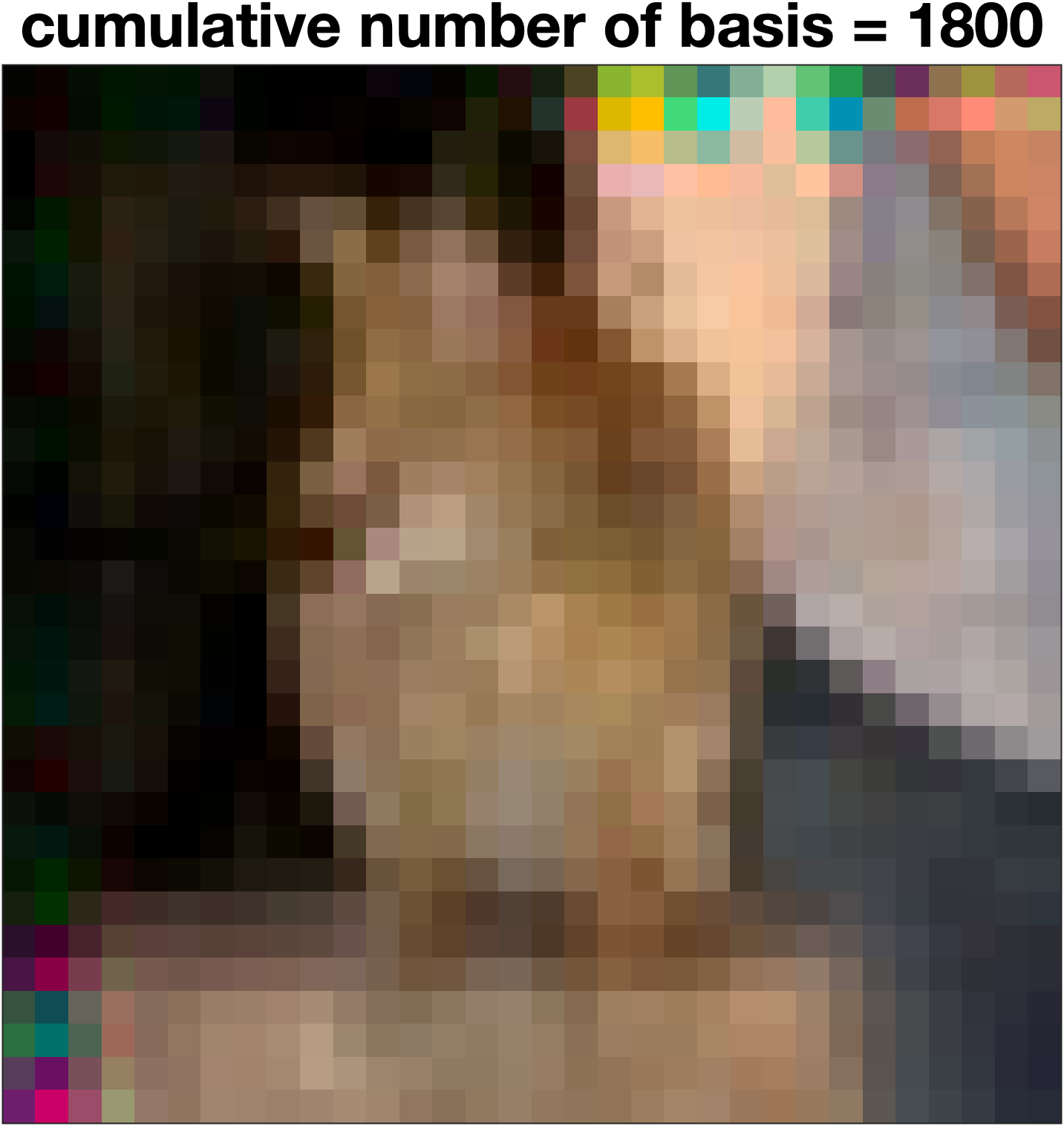}
   \includegraphics[width=0.11\linewidth]{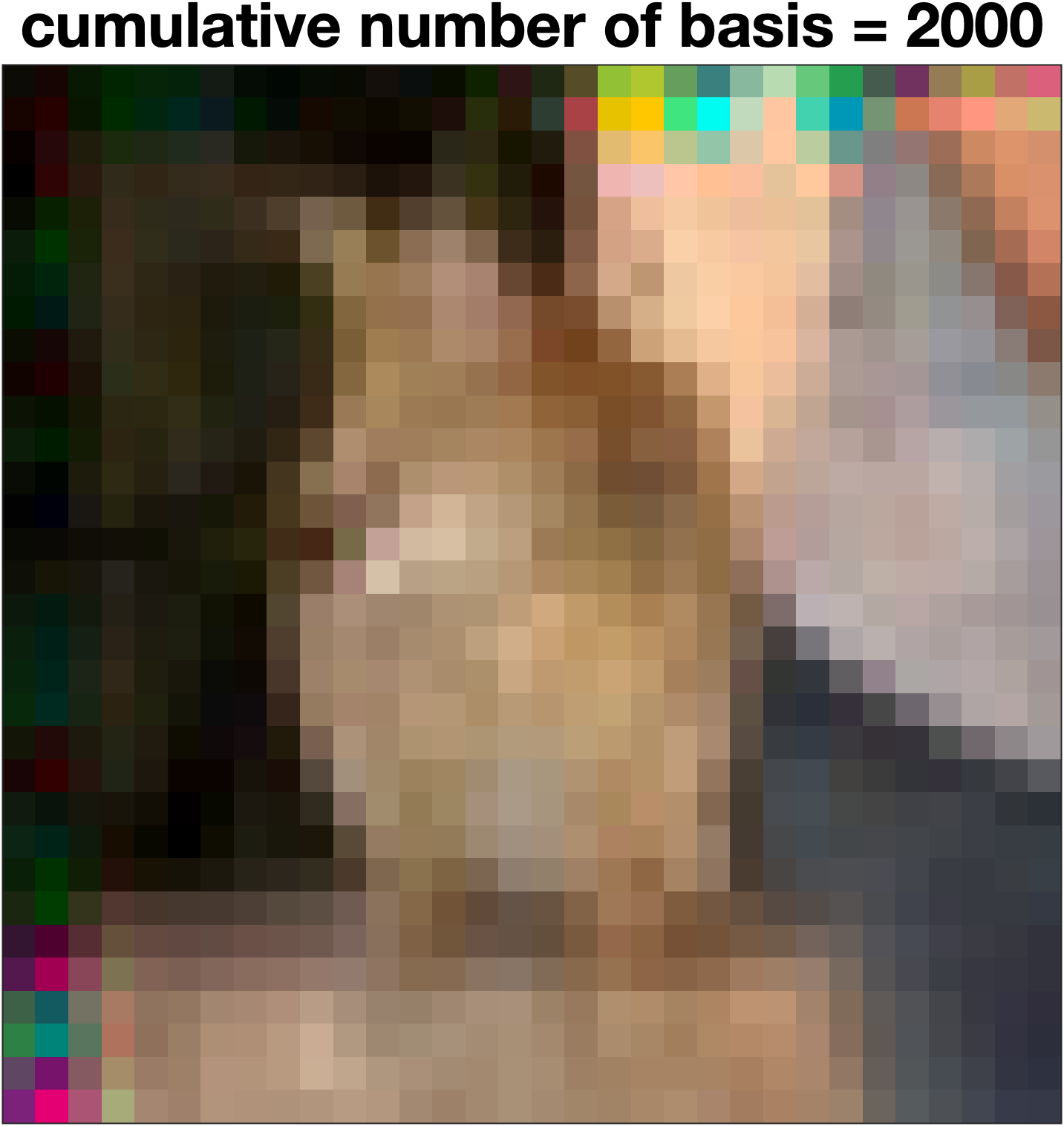}
   \includegraphics[width=0.11\linewidth]{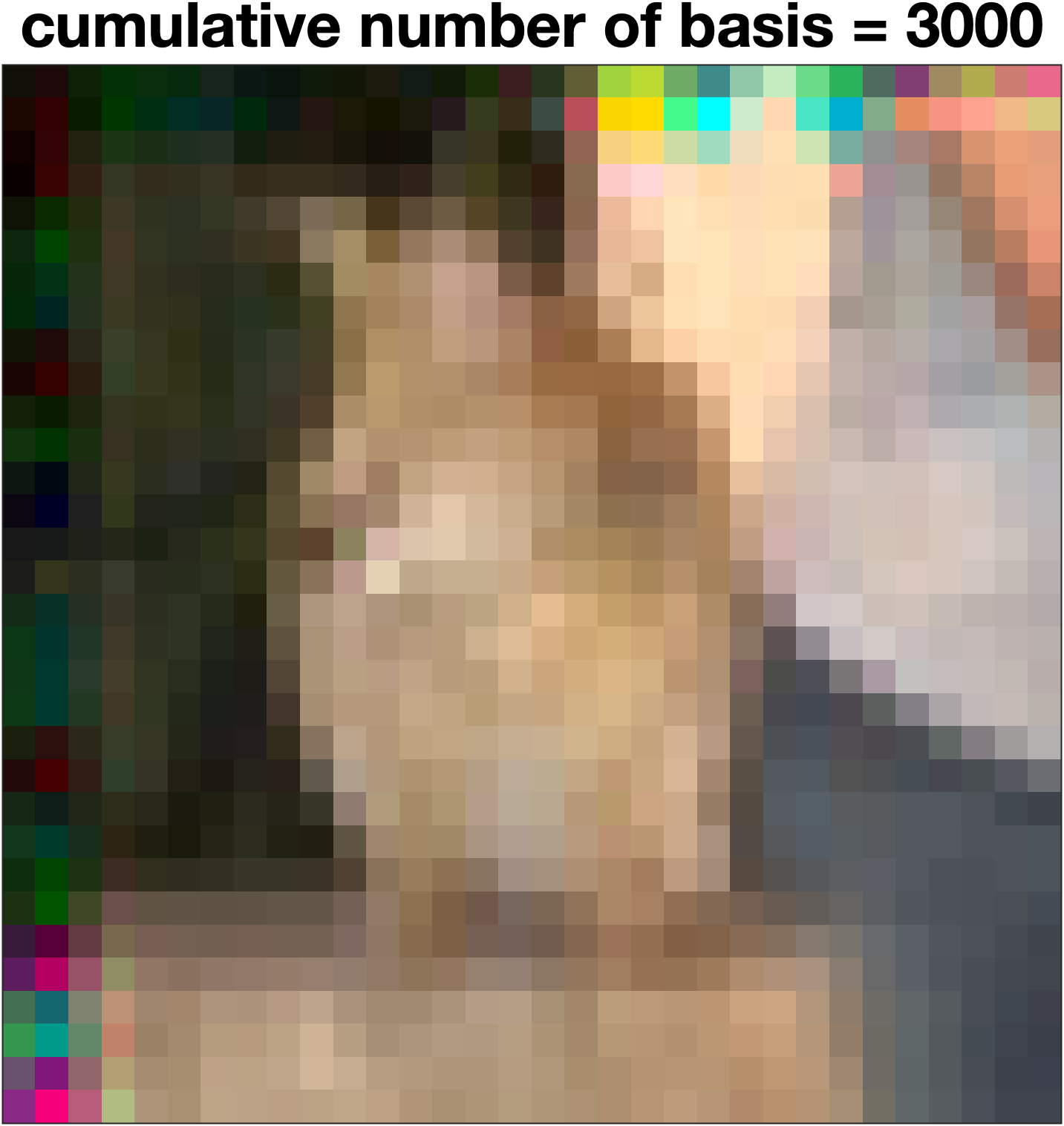}\\
       \vspace{0.5cm}
   \includegraphics[width=0.7\linewidth]{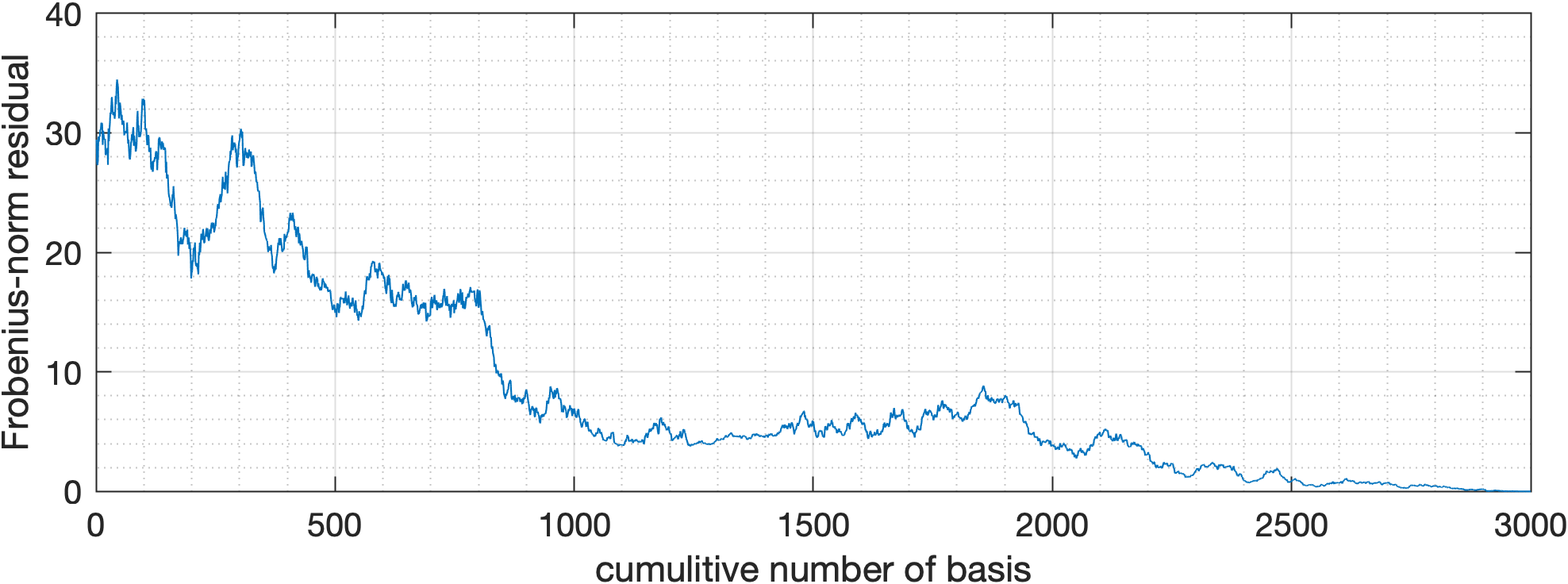}
  \caption{3rd testing image of the Cat class in CIFAR-10. We have used the common right basis obtained from training set to decompose this image into rank-1 images.}
  \label{fig_reconstruction_te2}
\end{figure}

Figure~\ref{fig_reconstruction_te2} shows the reconstruction of the 3rd testing image of the Cat class in CIFAR-10.

\begin{figure}[H]
  \centering
   \includegraphics[width=0.11\linewidth]{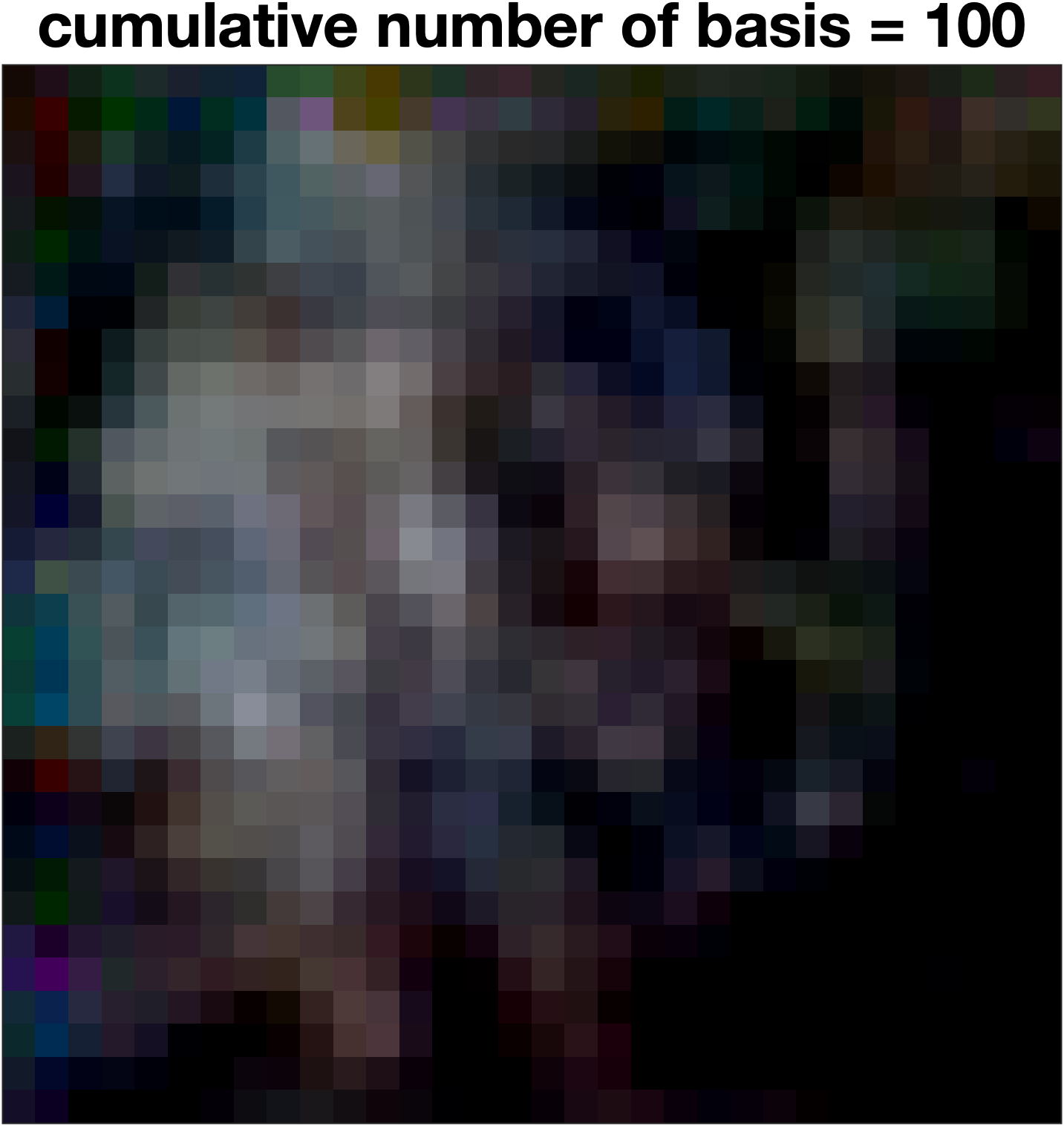}
   \includegraphics[width=0.11\linewidth]{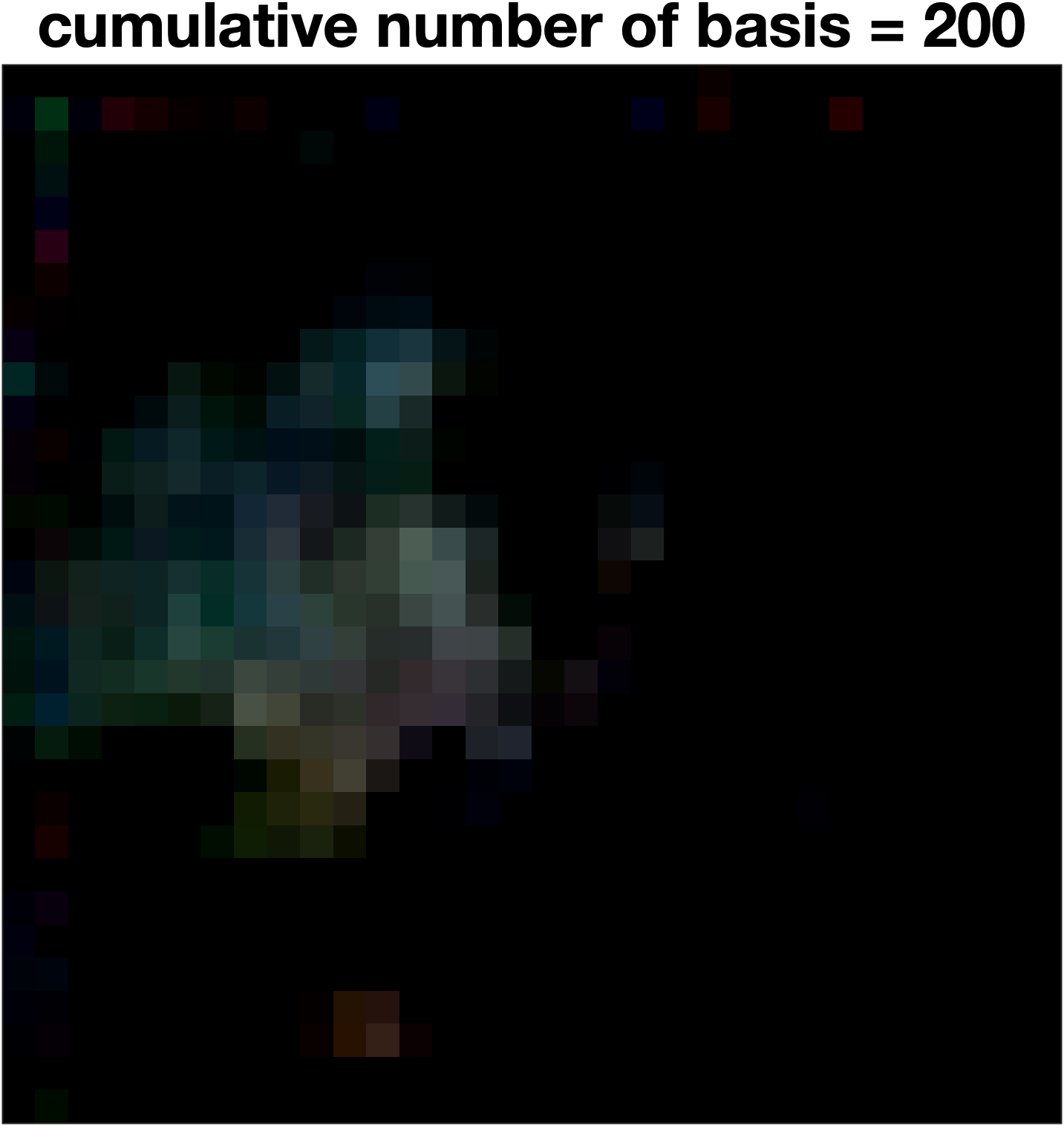}
   \includegraphics[width=0.11\linewidth]{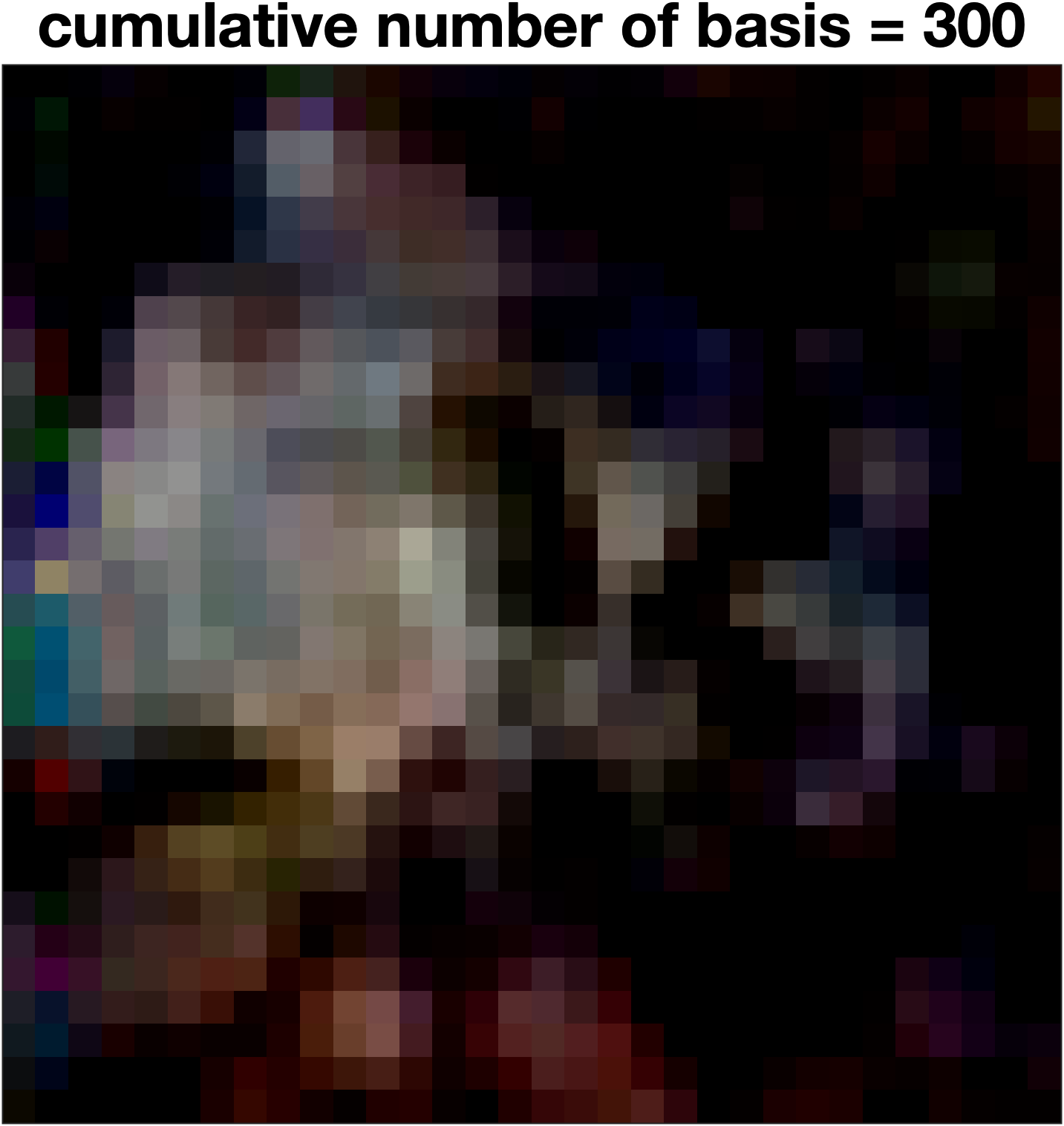}
   \includegraphics[width=0.11\linewidth]{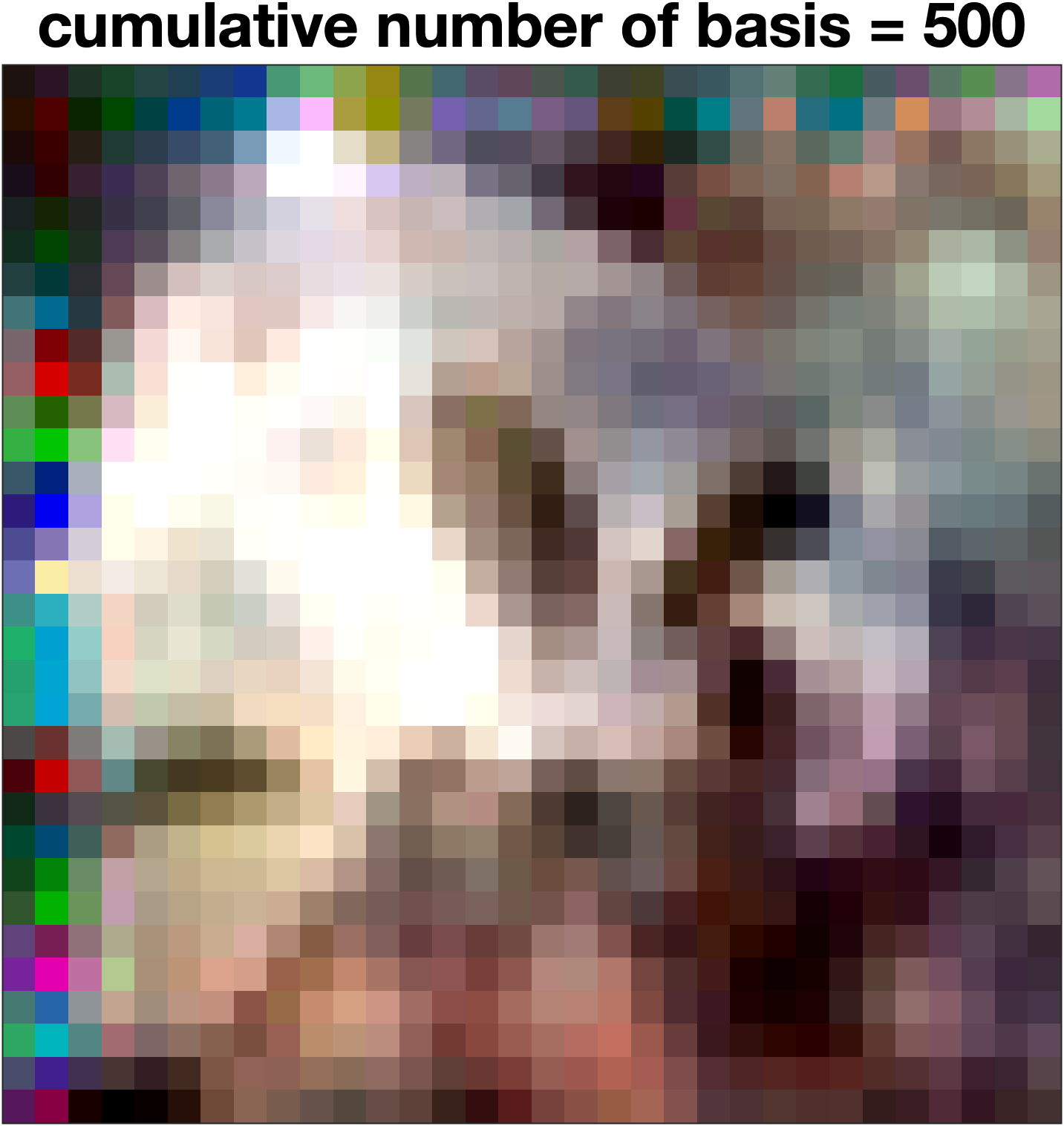}
   \includegraphics[width=0.11\linewidth]{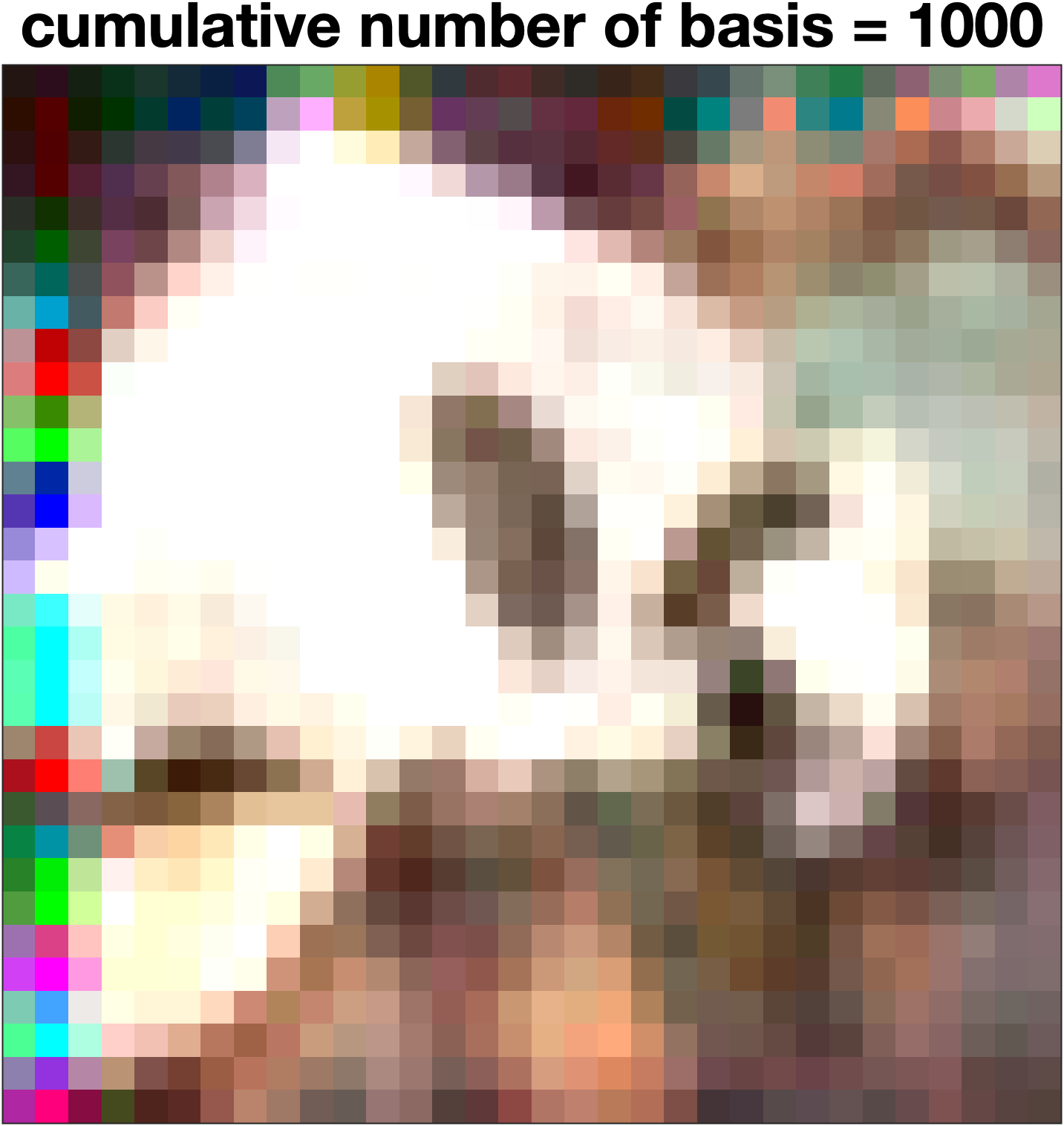}
   \includegraphics[width=0.11\linewidth]{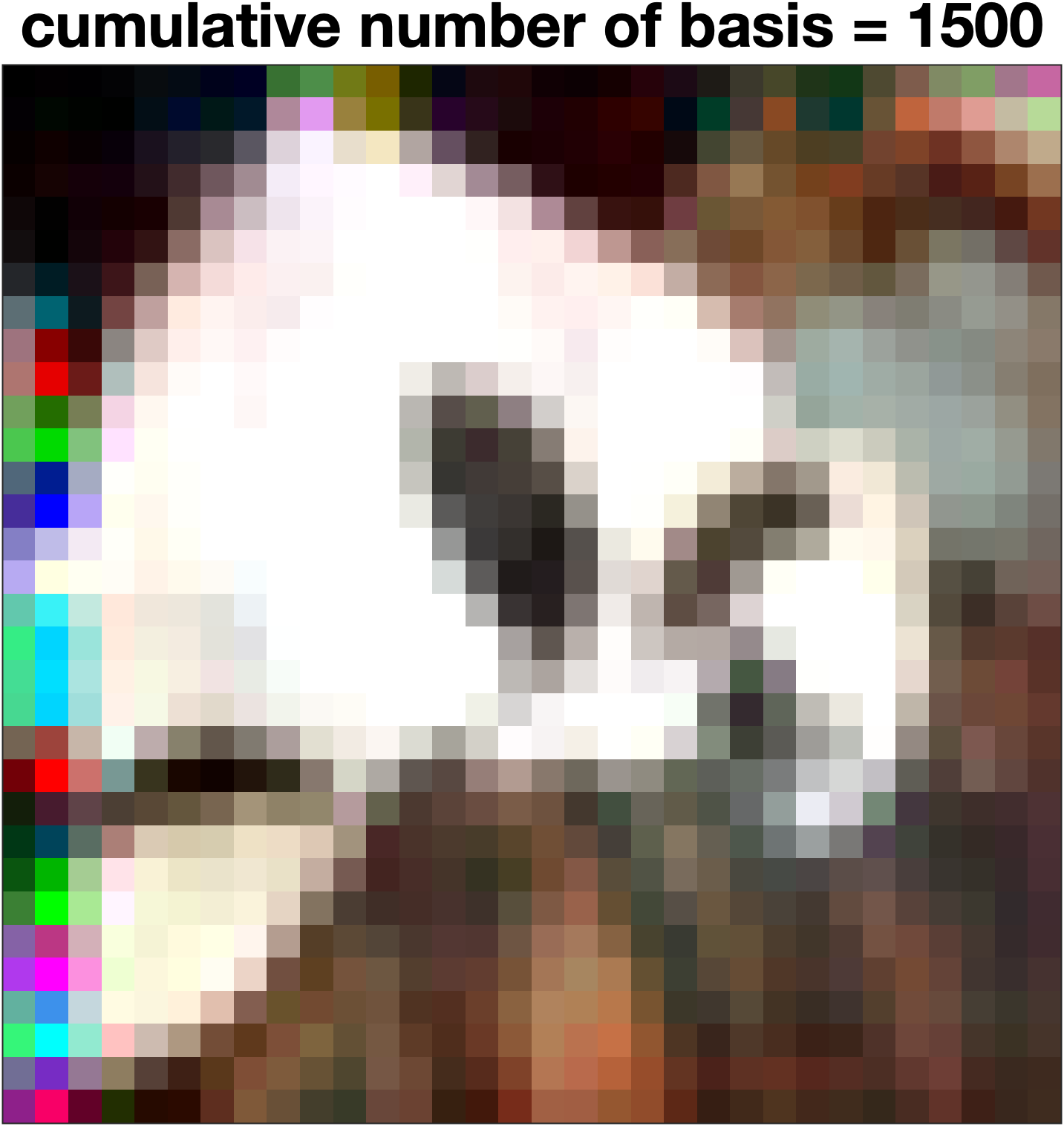}
   \includegraphics[width=0.11\linewidth]{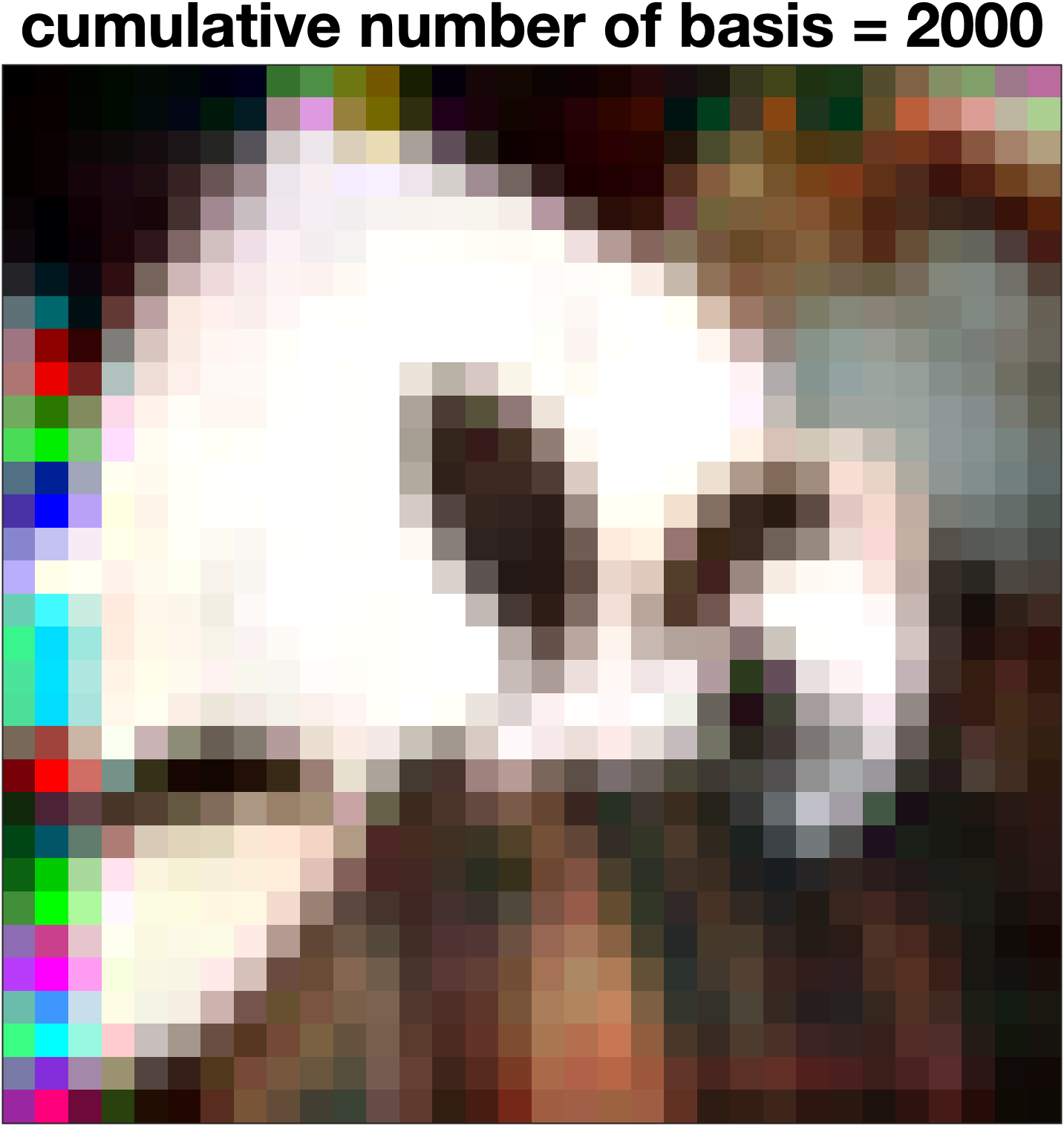}
   \includegraphics[width=0.11\linewidth]{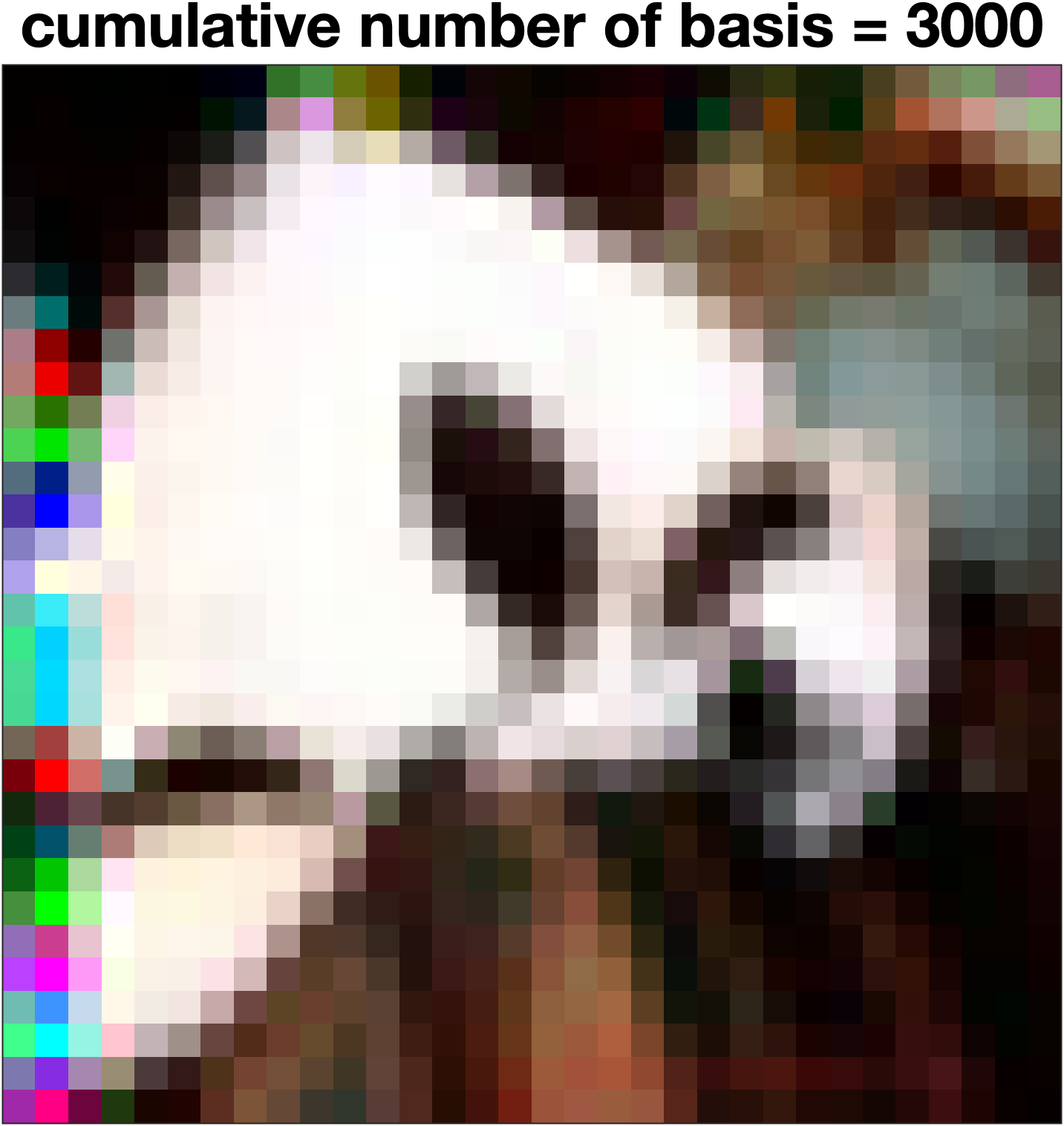}\\
       \vspace{0.5cm}
   \includegraphics[width=0.7\linewidth]{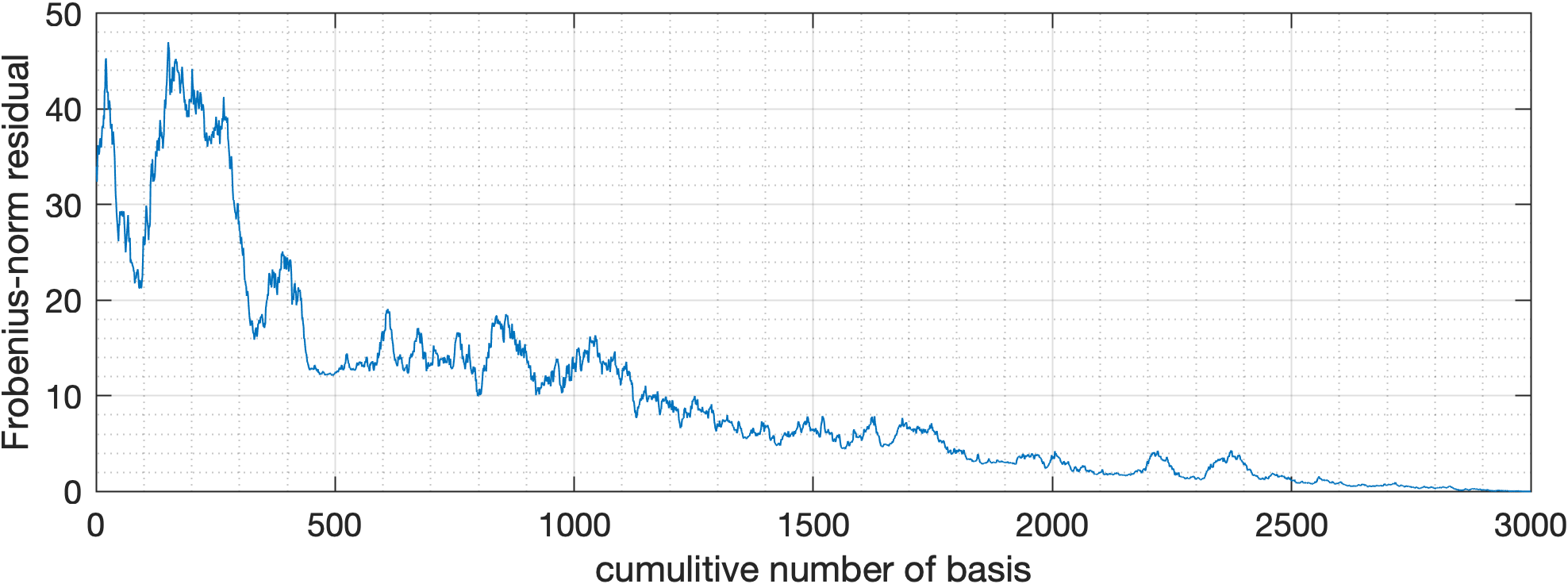}
  \caption{331st testing image of the Cat class which is misclassified by state of art model on CIFAR-10. For this decomposition,we have used the same $V$ obtained from training set.}
  \label{fig_reconstruction_te331}
\end{figure}

Figure~\ref{fig_reconstruction_te331} shows the reconstruction of a testing image of Cat class that is misclassified by state of art model on CIFAR-10.

%*****

% Figures~\ref{fig_reconstruction_tr2} and \ref{fig_residual_tr2} show the reconstruction of second training image of Cat class in CIFAR-10 dataset.

% \begin{figure}[H]
%   \centering
%   \includegraphics[width=0.11\linewidth]{fig_cd_tr2_100.png}
%   \includegraphics[width=0.11\linewidth]{fig_cd_tr2_200.png}
%   \includegraphics[width=0.11\linewidth]{fig_cd_tr2_300.png}
%   \includegraphics[width=0.11\linewidth]{fig_cd_tr2_500.png}
%   \includegraphics[width=0.11\linewidth]{fig_cd_tr2_1000.png}
%   \includegraphics[width=0.11\linewidth]{fig_cd_tr2_1500.png}
%   \includegraphics[width=0.11\linewidth]{fig_cd_tr2_2000.png}
%   \includegraphics[width=0.11\linewidth]{fig_cd_tr2_3000.png}
%   \caption{2nd training image of Cat class, reconstructed as summation of rank-1 basis of Higher Order GSVD. We see that for many images, the main structure of image and its distinctive features are obtained by adding a relatively small portion of rank-1 matrices in the wavelet space. This is the .}
%   \label{fig_reconstruction_tr2}
% \end{figure}

% \begin{figure}[H]
%   \centering
%   \includegraphics[width=0.7\linewidth]{fig_cd_tr2_residual.png}
%   \caption{Residual of reconstructed image from the wavelet coefficients as we add more rank-1 images from the Higher Order-GSVD. Image is the second training image of the Cat class.{\bf make this figure wider.}}
%   \label{fig_residual_tr2}
% \end{figure}

\setcounter{figure}{0}

\section{Practical notes and other extensions} \label{sec:appx_algorithm}

\subsection{Practical notes about our algorithm}

Our Algorithm~\ref{alg:main}, uses the parameter $\tau$
to impose an upper bound on the condition numbers of the slices of data. Since we choose the wavelet coefficients using the RR-QR algorithm, imposing the upper bound on condition numbers of the slices of the data leads to discarding the possible redundancies and rank deficiencies in the data and ensures that the remaining data is not rank deficient. As a practical value based on our numerical experiments, we recommend $\tau$ to be chosen between $10^{5}$ to $10^{7}$. Choosing much larger values might lead to rank deficiency.

Generally, choosing a smaller value for $\tau$ would lead to a smaller $m$, which means fewer wavelet coefficients are involved in our analysis, which might lead to extracting vague patterns at the end. It is important to note that when we decompose an image with a wavelet basis, we can reconstruct it perfectly (to some computational precision), if we use {\em all} the wavelet coefficients of the decomposition. If we choose only a subset of those coefficients, the reconstructed image may have a lower quality, some patterns might become blurry and vague, and some pixel values might become completely lost. Therefore, we would not want to discard too many of the wavelet coefficients by choosing the $m$ very small. 

Finally, for datasets with high resolution images, it would be most efficient to first reduce the resolution of images and then perform the analysis. Some images might have pixels in the order of millions. Seeing a high-resolution image with millions of pixels is visually nice for humans, but we would not need the fine level resolution to identify patterns that distinguish one class from the others. Additionally, decomposing extremely large tensors might be computationally intractable.%\fh{This could be moved to experiments or appendix}

\subsection{Shearlets}

We note that shearlets are also a multi-scale framework, similar to wavelets, which can have certain advantages in capturing edges and other an-isotropic features in images \citep{labate2005sparse,kutyniok2012shearlets}. For example, shearlets are used to extract features from images for edge detection \citep{andrade2019shearlets,schug2015wavelet} and image interpolation \citep{lakshman2015image}. Investigating whether shearlets can extract more useful information and patterns from the image-classification datasets can be a future direction of research.

\setcounter{figure}{0}
\renewcommand{\thefigure}{D\arabic{figure}}

\section{Our results on MNIST} \label{sec:appx_mnist}

We consider all 10 classes of MNIST. To decompose the images, we choose Daubechies-1 wavelet basis. More information about the data is presented in Appendix~\ref{sec:appx_data}. We also use $m=100$ which satisfies the full rank requirement for all classes.

\begin{figure}[H]
  \centering
   \includegraphics[width=0.08\linewidth]{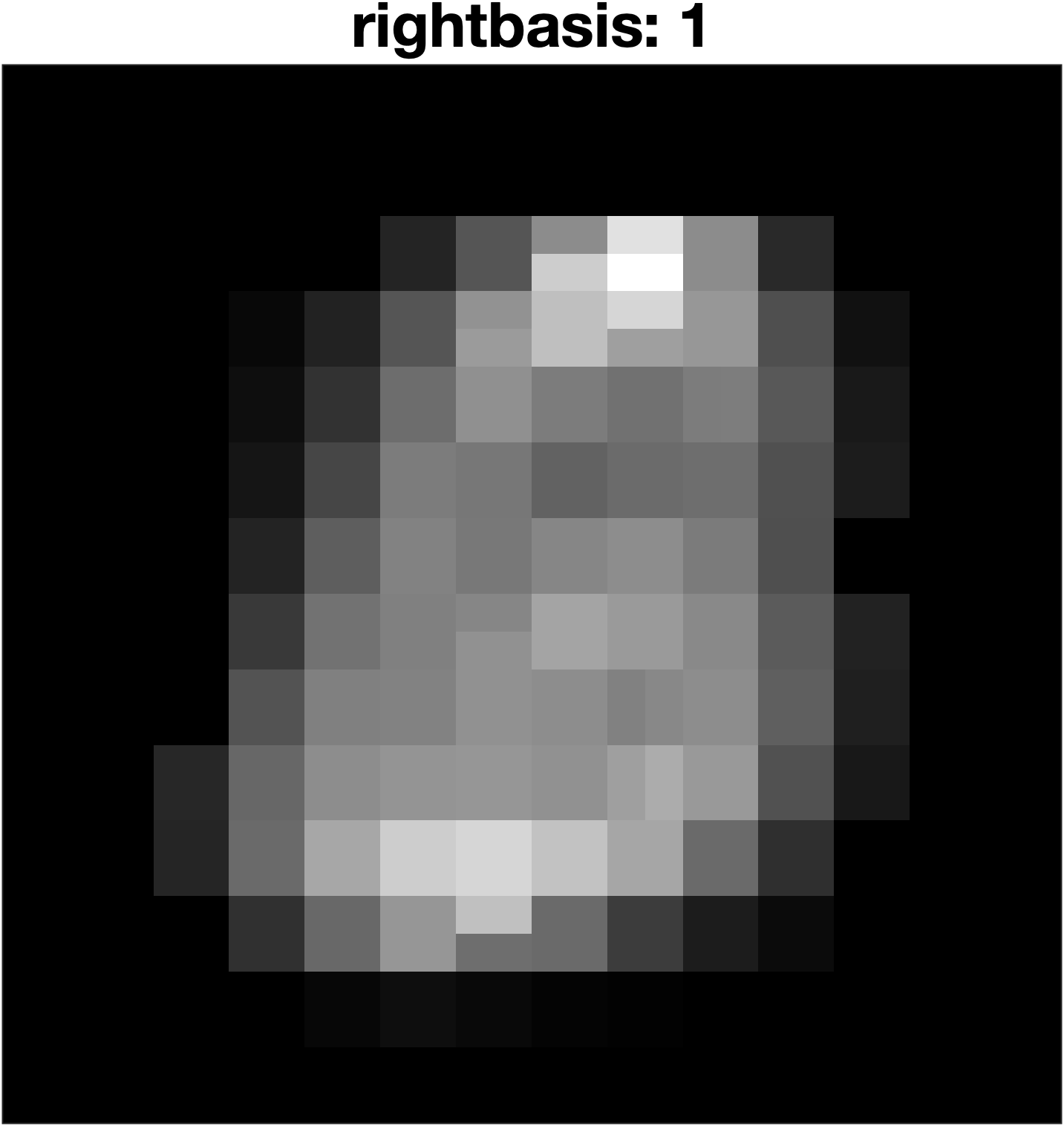}
   \includegraphics[width=0.08\linewidth]{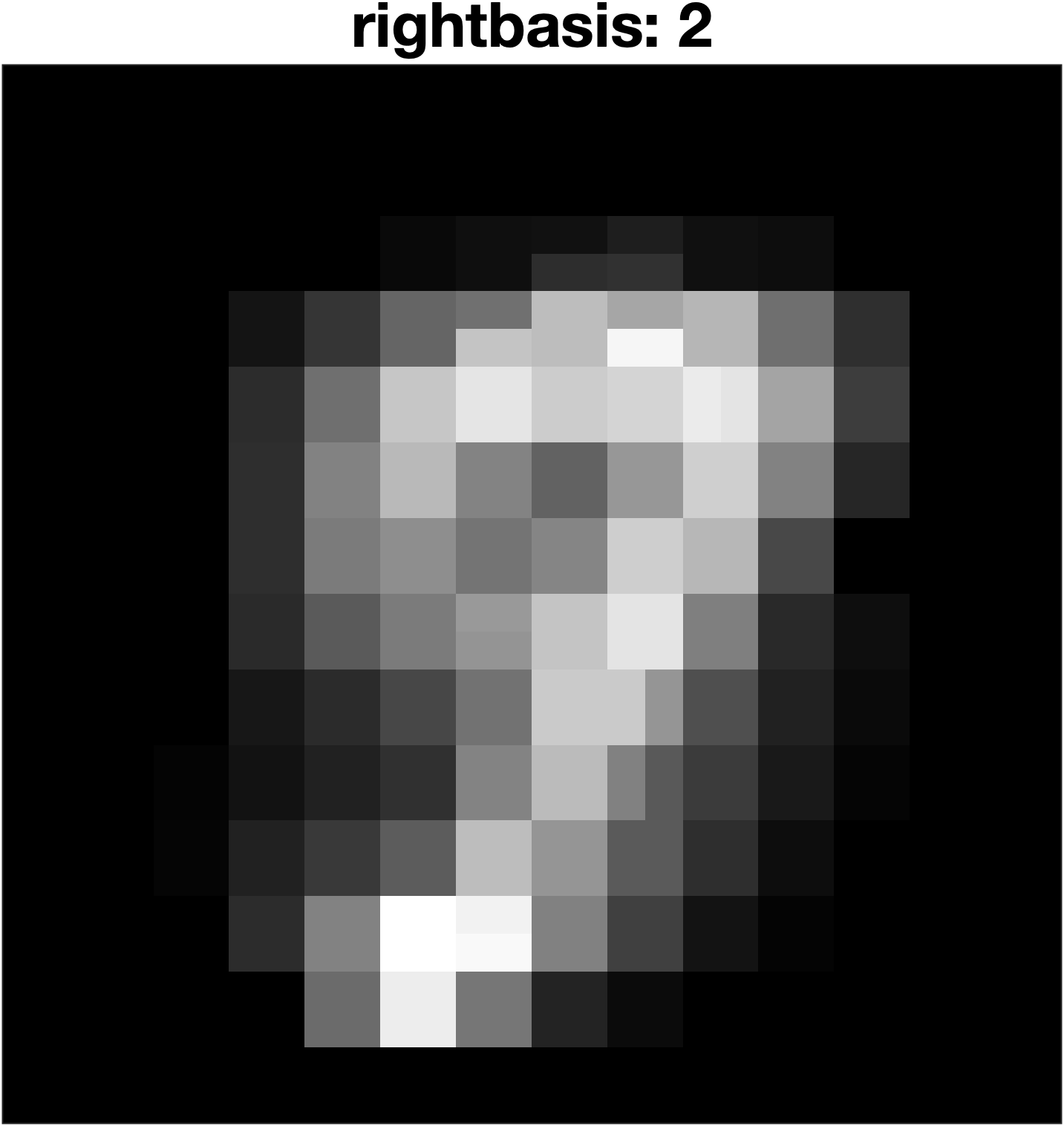}
   \includegraphics[width=0.08\linewidth]{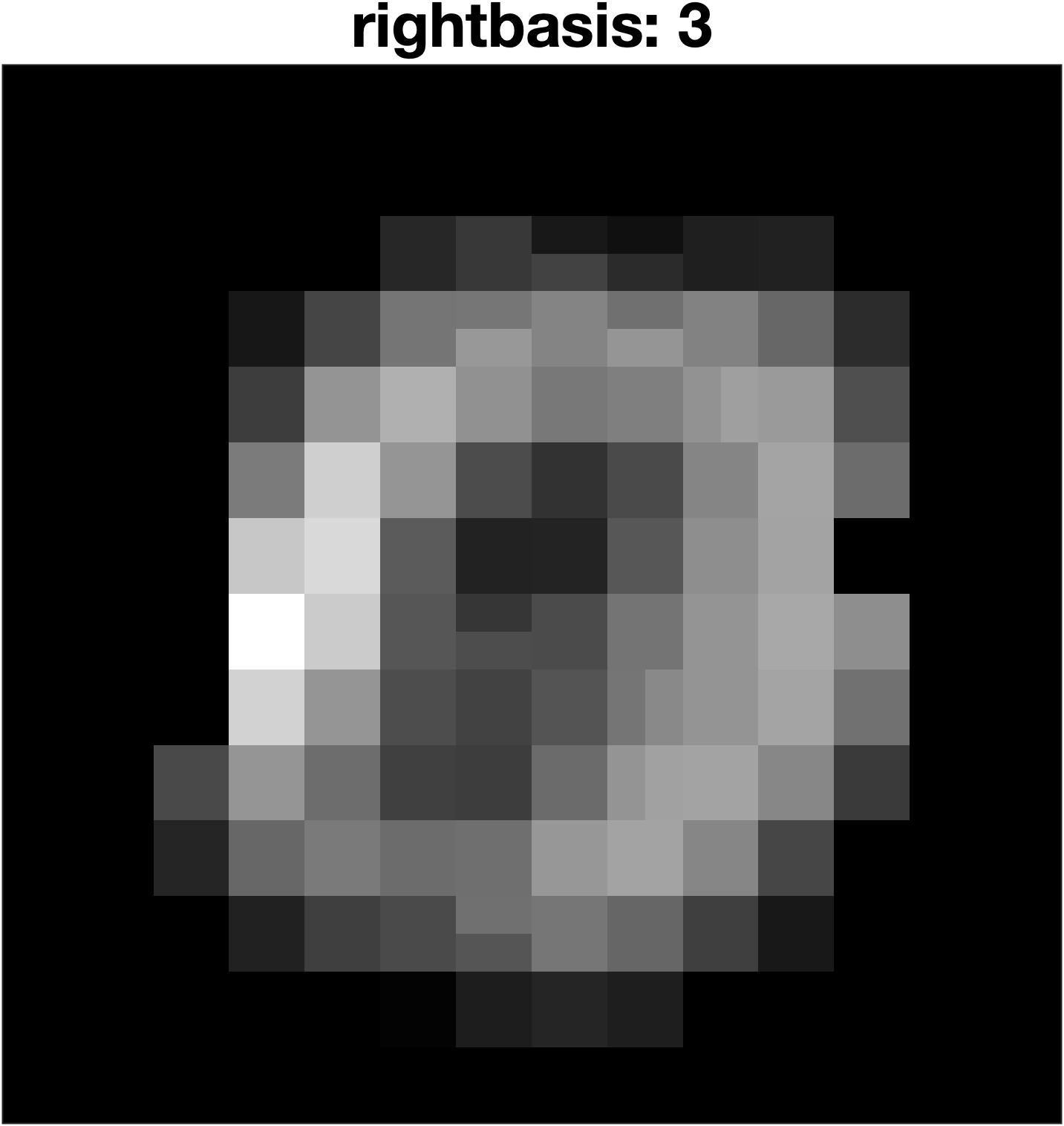}
   \includegraphics[width=0.08\linewidth]{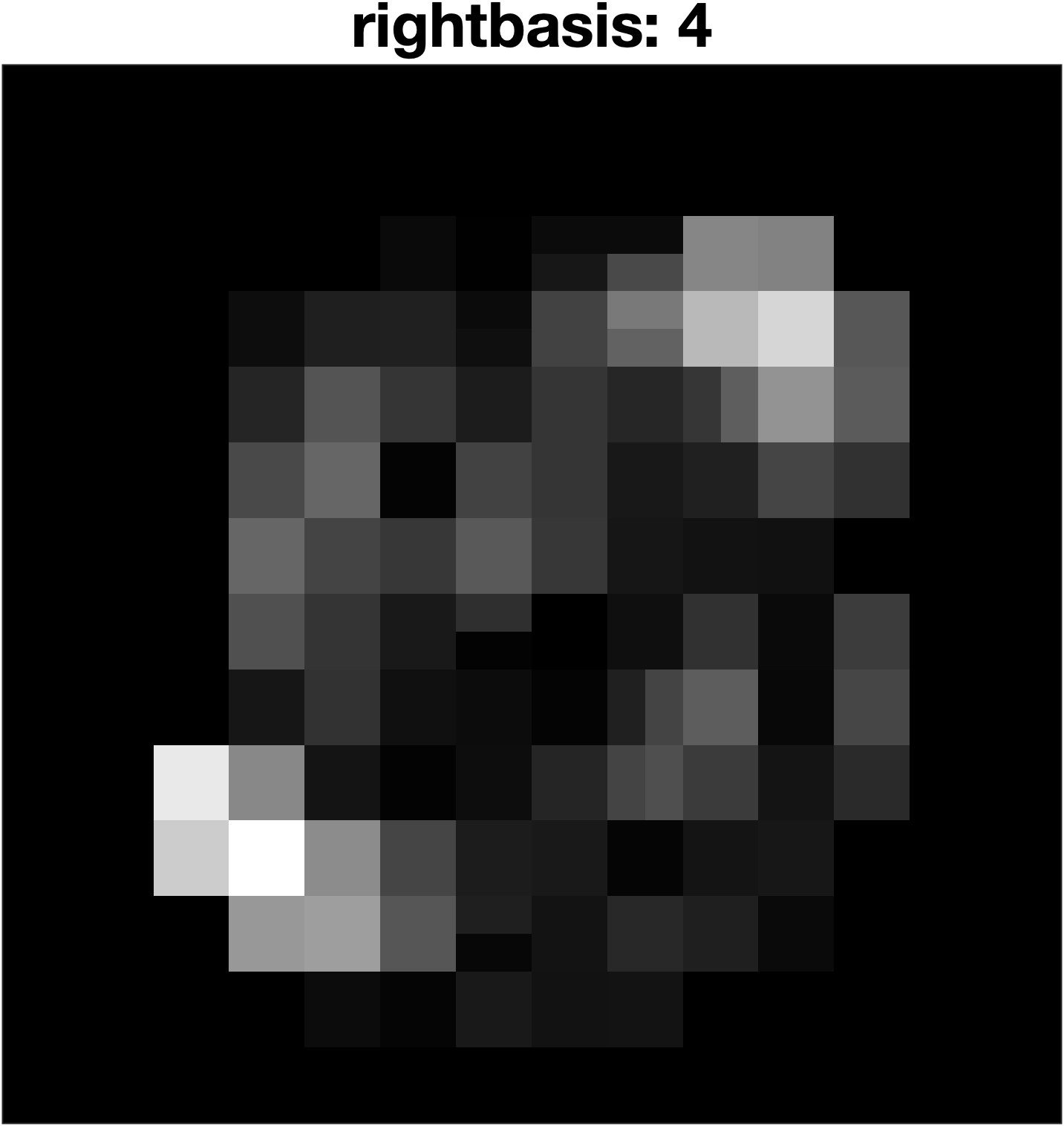}
   \includegraphics[width=0.08\linewidth]{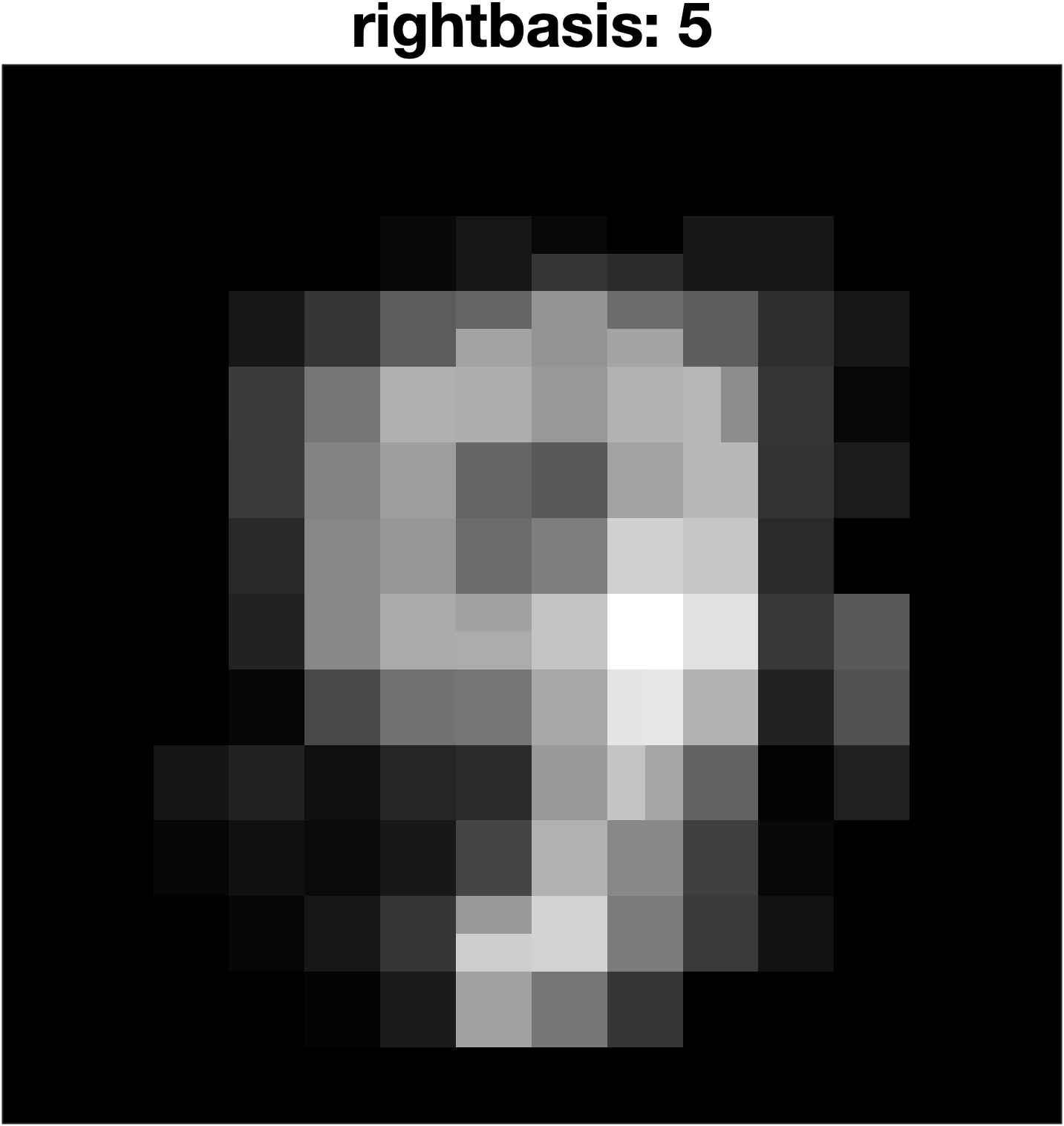}
   \includegraphics[width=0.08\linewidth]{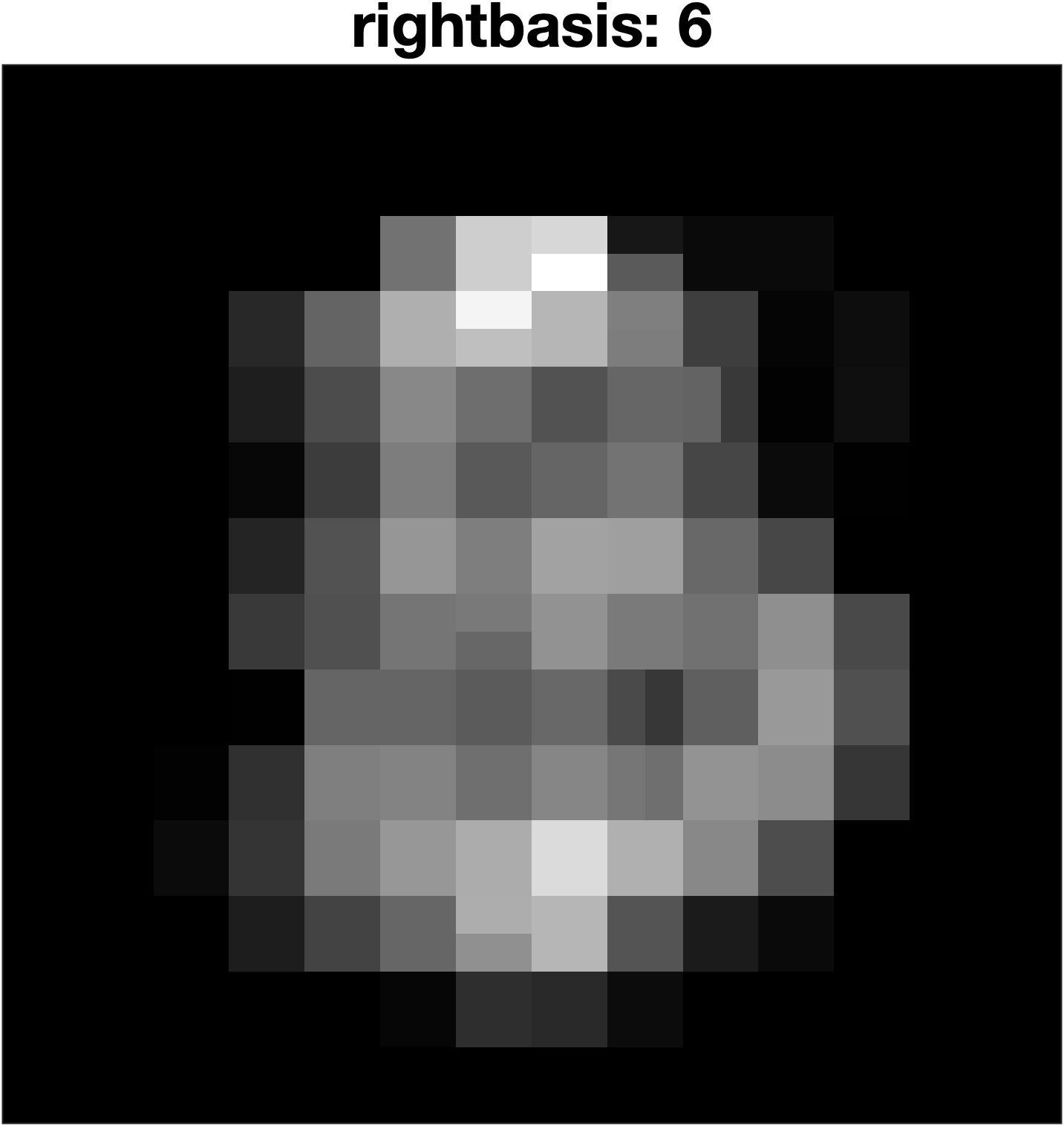}
   \includegraphics[width=0.08\linewidth]{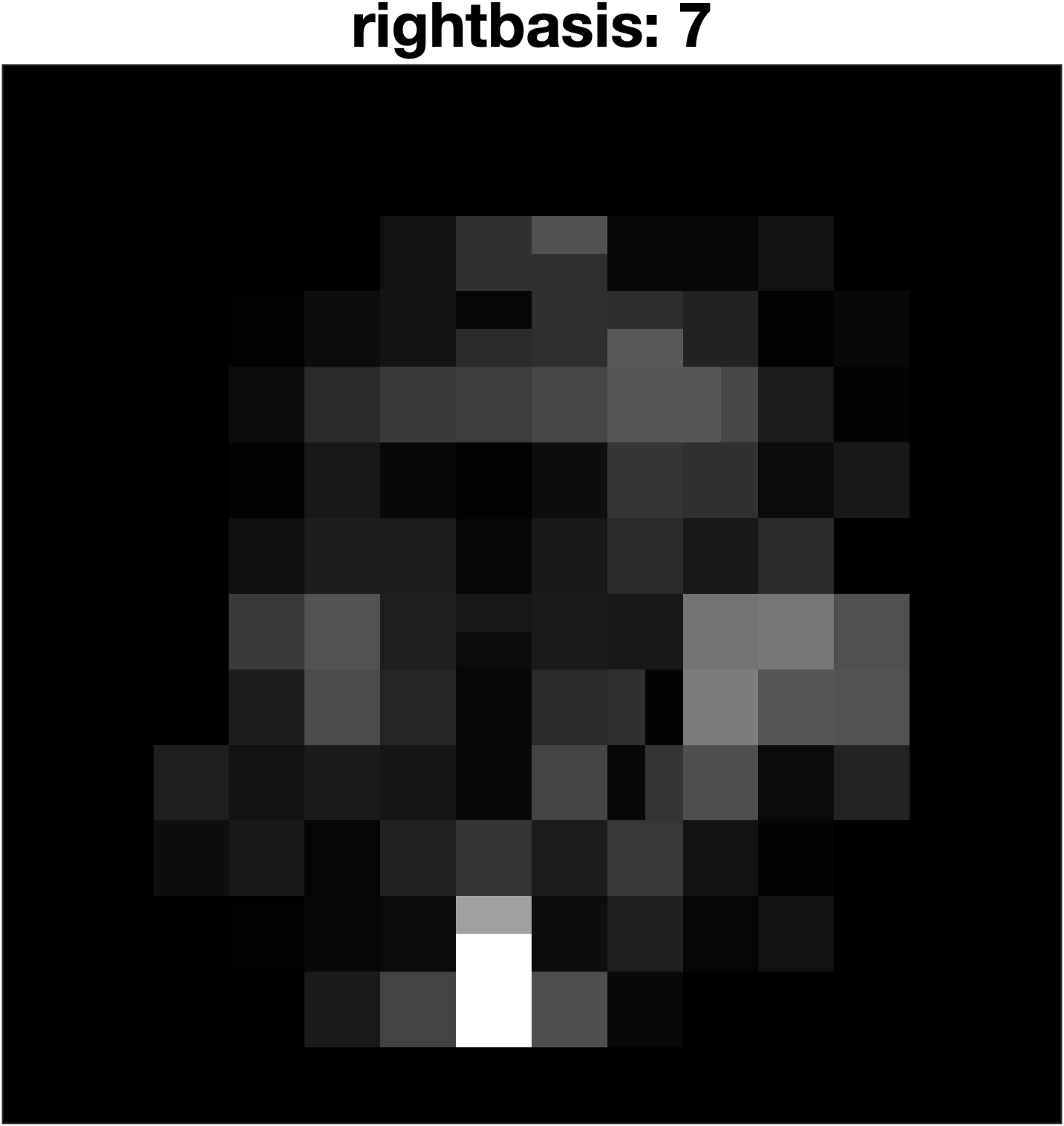}
   \includegraphics[width=0.08\linewidth]{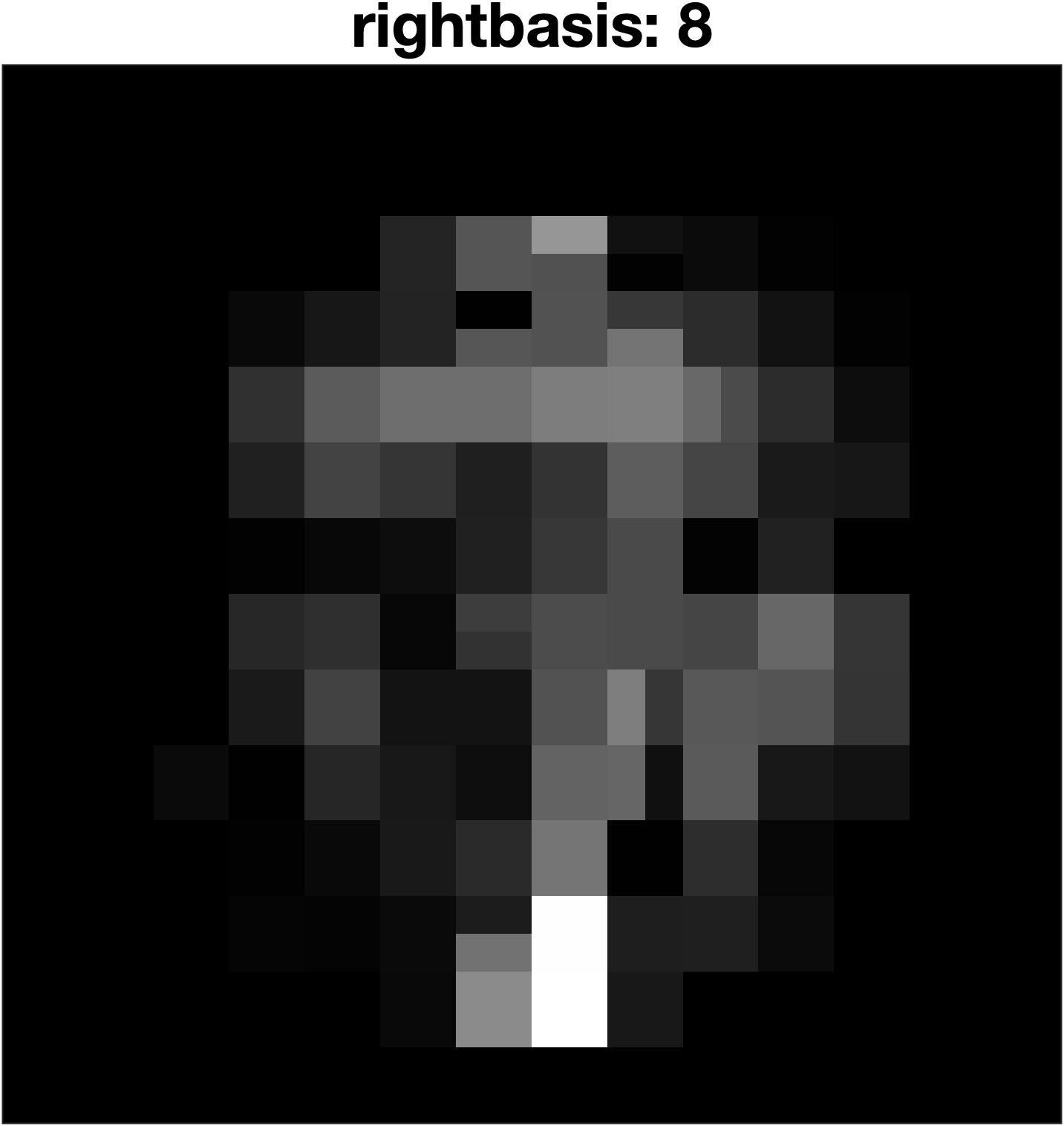}
   \includegraphics[width=0.08\linewidth]{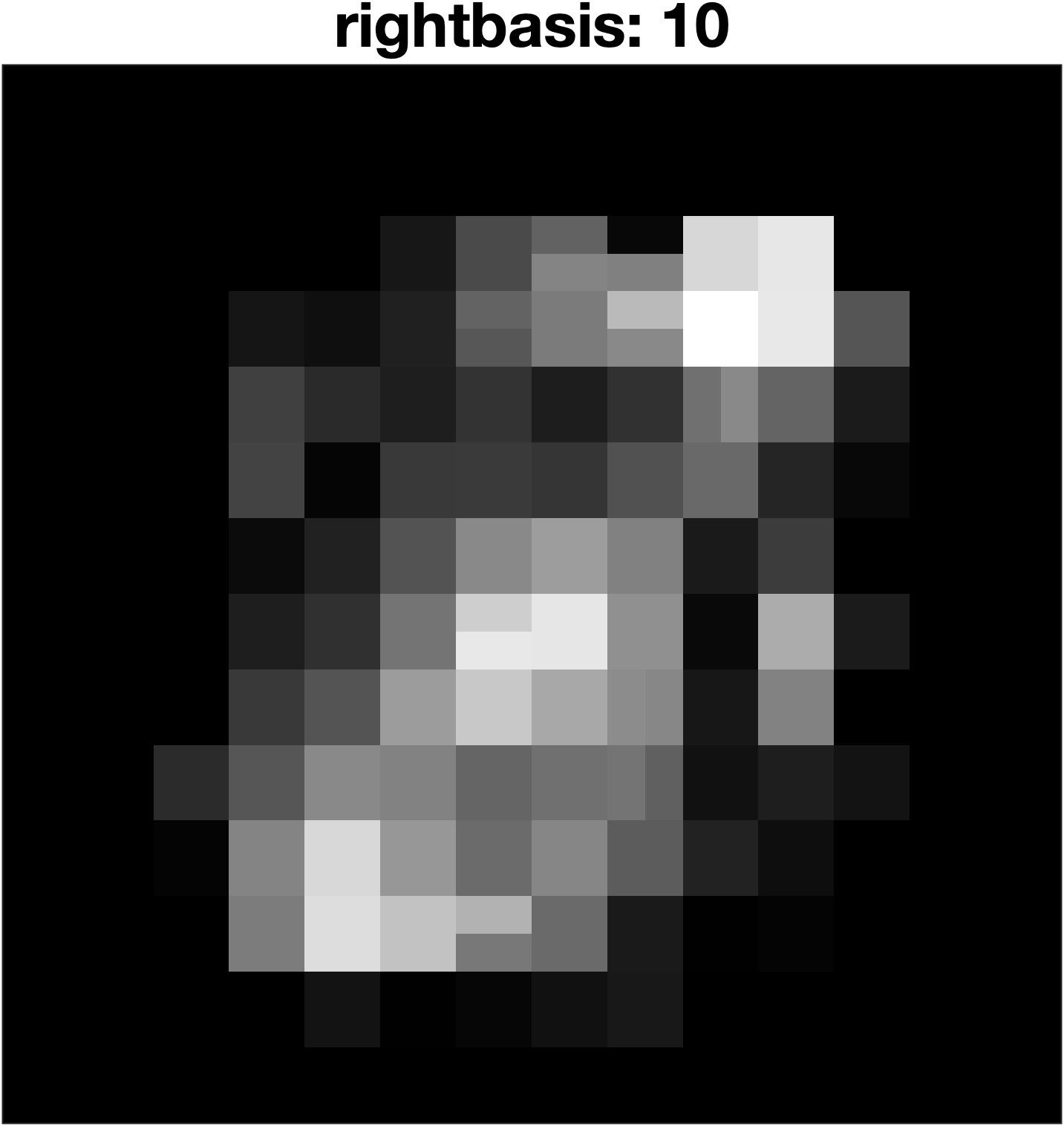}
  \caption{Images reconstructed from individual columns of the right basis $V$.}
  \label{fig:mnist_rightbasis}
\end{figure}

{\bf Patterns in the dataset.}
Again, we use the Higher Order GSVD as our spectral decomposition method. Figure~\ref{fig:mnist_rightbasis} shows the images reconstructed from the first 9 columns of the right basis, $V$. 

\begin{figure}[H]
     \centering
     \begin{subfigure}[b]{0.44\textwidth}
         \centering
         \includegraphics[width=.7\linewidth]{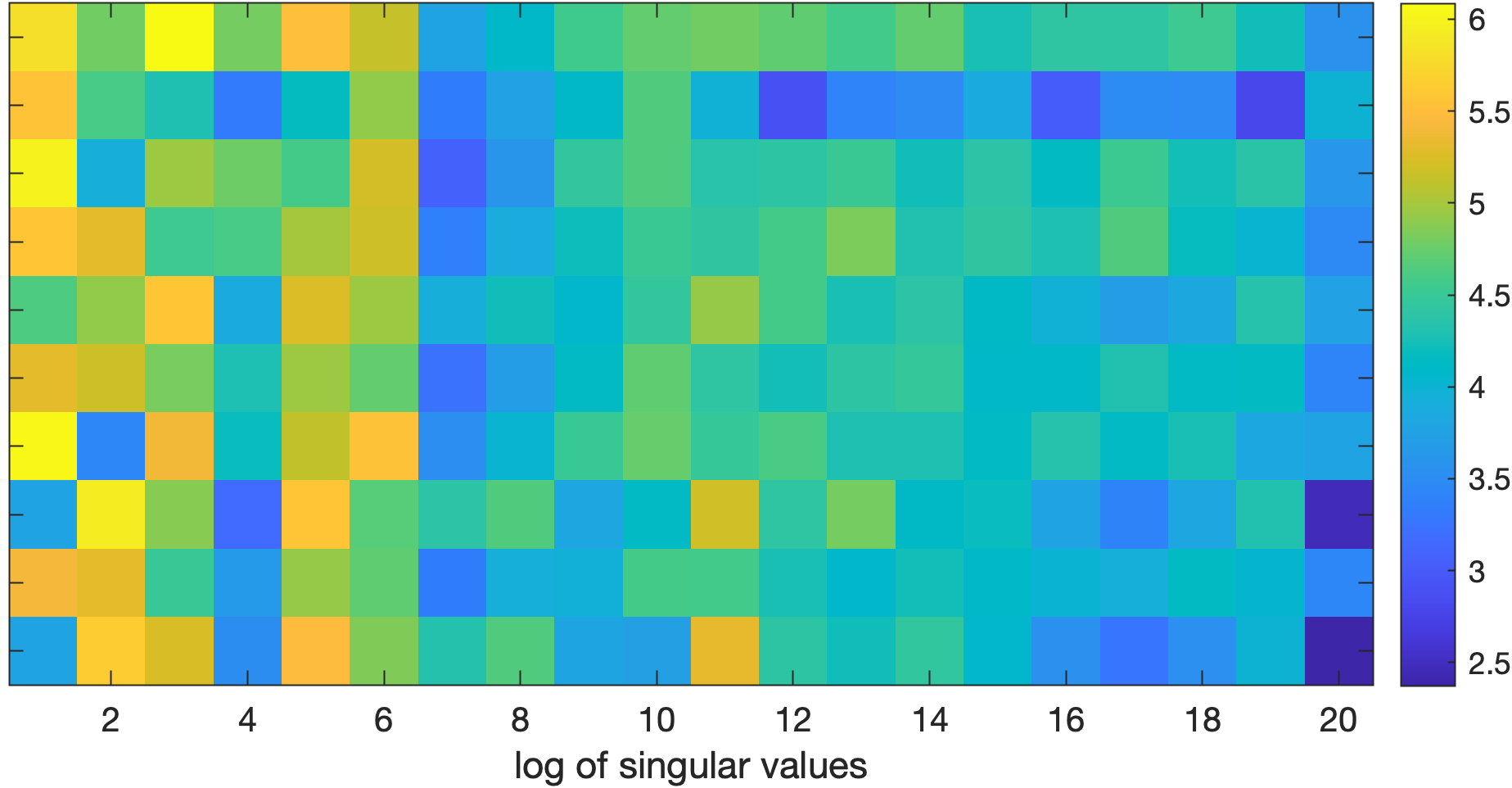}
         \caption{Singular values 1 through 20}
         \label{fig:mnist_singulars1}
     \end{subfigure}
     \hspace{-1.8cm}
     \begin{subfigure}[b]{0.48\textwidth}
        \centering
        \includegraphics[width=.7\linewidth]{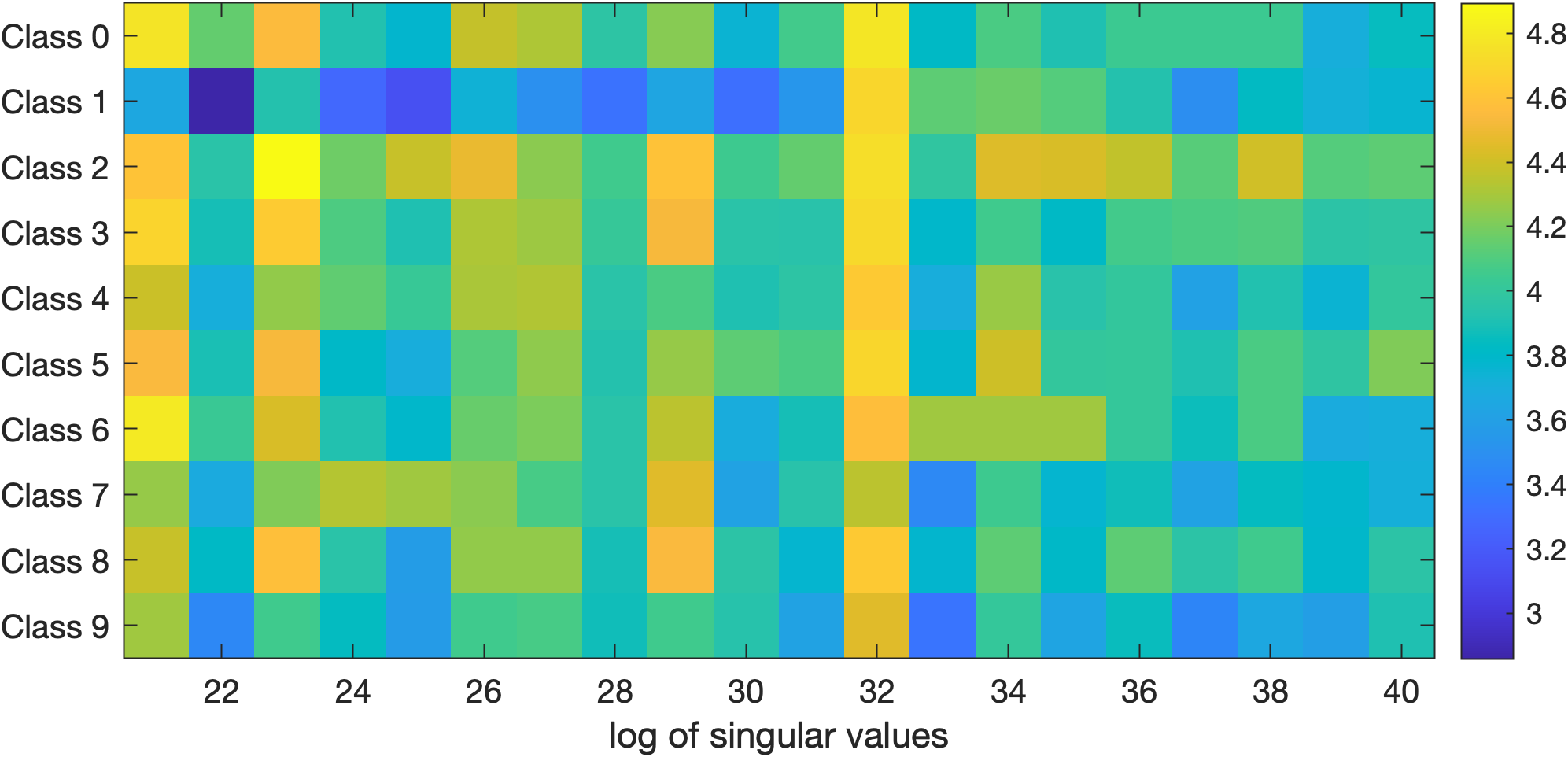}
         \caption{Singular values 21 through 40}
         \label{fig:mnist_singulars2}
     \end{subfigure}
        \caption{Log of singular values for all classes of MNIST. Some singular values are dominant for a particular class and some of them are dominant for more than one class.}
        \label{fig:mnist_singulars}
\end{figure}

{\bf Singular values and association of patterns to classes.}
Figure~\ref{fig:mnist_singulars1} shows the first 20 singular values for all classes, and Figure~\ref{fig:mnist_singulars2} shows the second 20 singular values, i.e., singular values 21 through 40.

{\bf Interpreting the patterns.}
Consider the images for columns 7 and 8 of right basis, reconstructed in Figure~\ref{fig:mnist_rightbasis}. One could rationalize that they represent distinctive features for classes of 4,7, and 9, because those patterns have a vertical stroke at the bottom of image and they all have some vague content at the center. Looking at the singular values (columns 7 and 8 in Figure~\ref{fig:mnist_singulars2}), reveals that the dominant singular values are exactly those digits (4, 7, and 9).

\begin{figure}[H]
  \centering
   \includegraphics[width=0.1\linewidth]{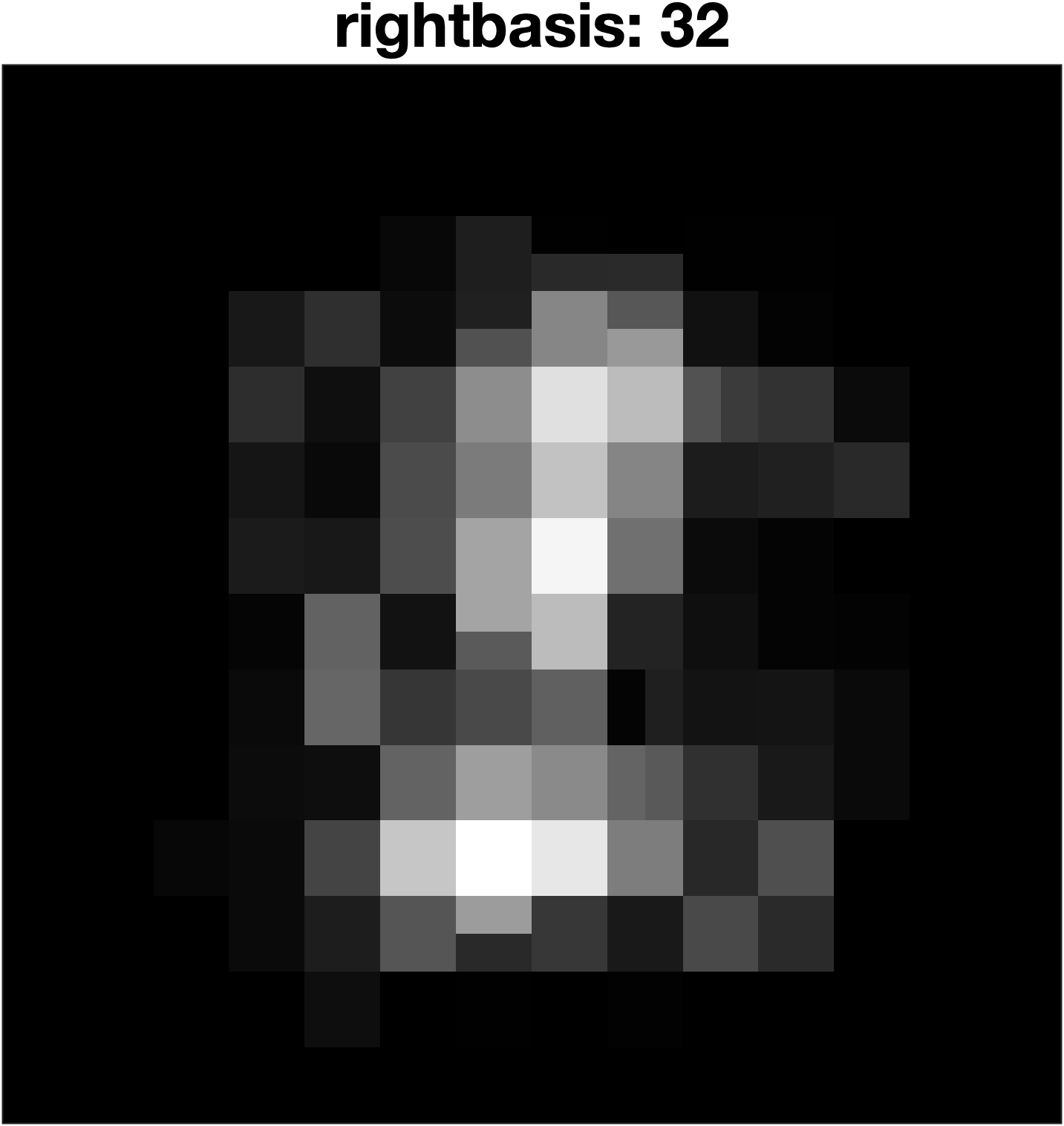}
   \includegraphics[width=0.1\linewidth]{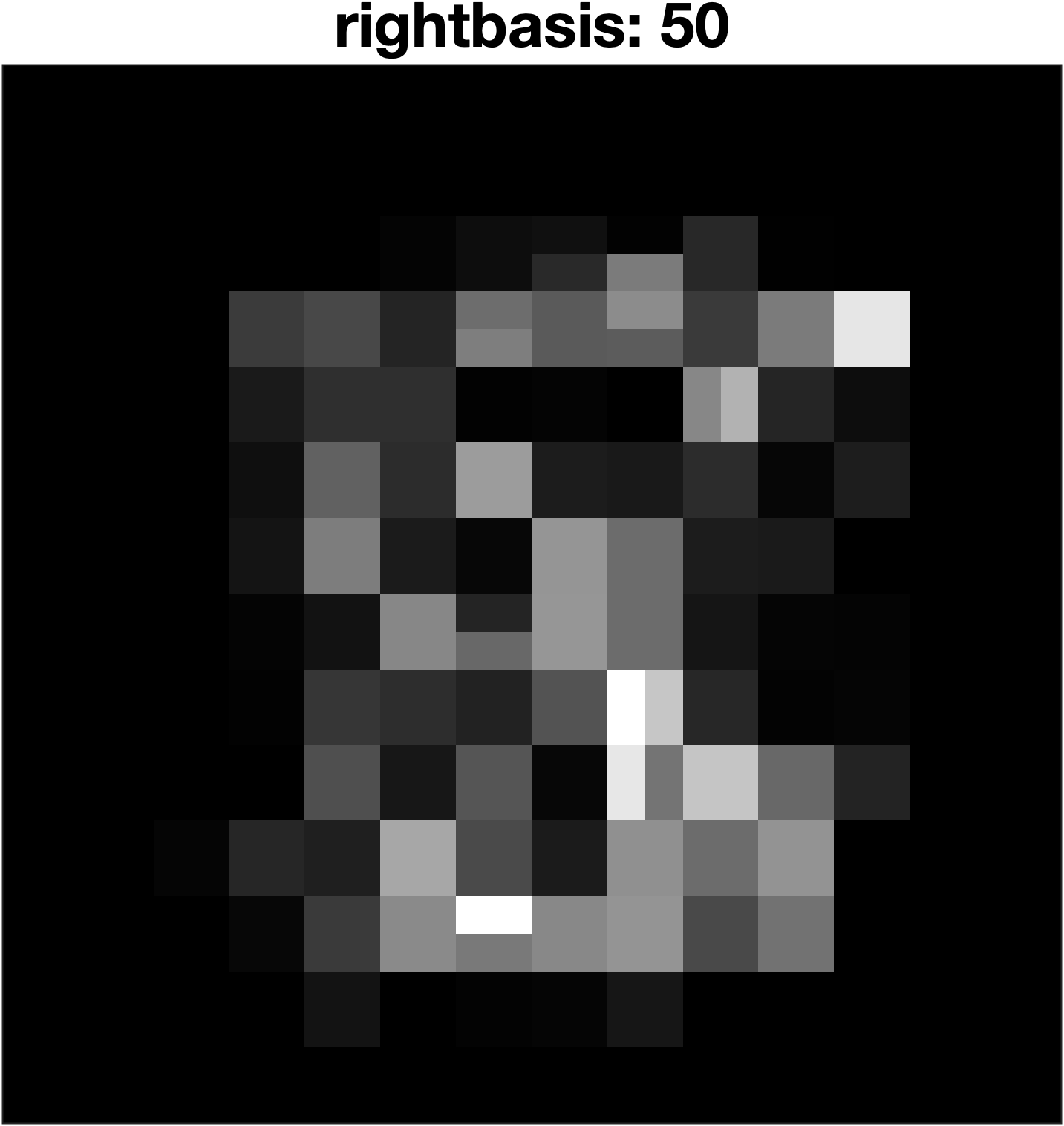}
  \caption{Images reconstructed from individual columns of the right basis $V$.}
  \label{fig:mnist_rightbasis2}
\end{figure}

Let's consider the 32nd singular value which seems to be relatively similar for all classes, i.e., the column 32 in Figure~\ref{fig:mnist_singulars2}. The image reconstructed from the 32nd column of right basis, $V$, is shown in Figure~\ref{fig:mnist_rightbasis2} (left), which seems to be a general basis, not distinctive for any particular digit. Now, lets' consider the 50th column in $V$ where its reconstruction is shown in Figure~\ref{fig:mnist_rightbasis2} (right). The dominant singular values for this class are classes 5 and 8, which seems to correspond to the distinctive patterns one would expect for digits 5 and 8.

\begin{figure}[H]
     \centering
     \begin{subfigure}[b]{0.49\textwidth}
      \centering
       \includegraphics[width=0.18\linewidth]{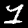}
       \includegraphics[width=0.18\linewidth]{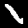}
       \includegraphics[width=0.18\linewidth]{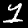}
      \caption{}
      \label{fig:mnist_notsimple}
     \end{subfigure}
     %\hfill{.1cm}
     \begin{subfigure}[b]{0.49\textwidth}
      \centering
       \includegraphics[width=0.18\linewidth]{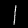}
       \includegraphics[width=0.18\linewidth]{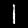}
       \includegraphics[width=0.18\linewidth]{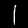}
      \caption{}
      \label{fig:mnist_simple}
     \end{subfigure}
        \caption{\textbf{(a)} Images that have the largest norm in left basis are made from many patterns and seem atypical. \textbf{(b)} Images that have a small norm in left basis are simple to reconstruct from the patterns.}
        \label{fig:mnist_simple_complex}
\end{figure}

{\bf Left basis interpretation.}
Now, let's consider the left basis for digit 1. The three images with the largest norm in the left basis are shown in Figure~\ref{fig:mnist_notsimple}. On the other hand, the images with the smallest norm in the left basis are shown in Figure~\ref{fig:mnist_simple}. Clearly, the simple images that one would consider straightforward to reconstruct or classify are the ones with small norm in left basis, and vice versa.

% {\bf Canonical Polyadic Decomposition}
% Here, we have space to briefly present the

\setcounter{figure}{0}
\renewcommand{\thefigure}{E\arabic{figure}}

\section{Analyzing the similarities using the left basis} \label{sec:appx_similarities}

%\ry{I modified this paragraph a bit. I think the relationship to out-of distribution detection methods is interesting, especially because of the type of images we have identified in Figure~\ref{fig:cifar_dog_anomalies}. But, I hesitate to call our method, an out-of-distribution detector, yet. Maybe that is a direction we can focus on.}
Here, we take a more unsupervised approach to analyze the similarities of images and to identify isolated images present in a dataset.

Using HO-GSVD, we explained how the vectors in $V$ represent the orthogonal patterns extracted from the entire dataset. We also know that each image in the dataset can be reconstructed by a specific combination of patterns in $V$, defined by the singular values and the left basis vectors. In fact, $V$ is common among all the classes in the dataset $V$ and what distinguishes each image from other images, is its corresponding vector in $U_i \Sigma_i$. The number of rows in $U \Sigma$ correspond to the number of images in the class $i$. Let's consider the $j^{th}$ image in class $i$. We can reconstruct this image using the coefficients in the $j^{th}$ row of $U_i \Sigma_i$ to combine the vectors of $V$.\footnote{Clearly, all of these computations will be in  the wavelet space, and at the end we have to reconstruct them back in the pixel space.} Mathematically, if we have two identical images in the dataset, we will get identical coefficients for them. 

Overall, we can say that the HO-GSVD extracts a set of common patterns in the entire dataset, and gives us a set of coefficients in $U_i \Sigma_i$, determining how these patterns should be combined in order to obtain the images. To analyze the similarity of images, we can analyze the similarity of their coefficients, because patterns are common for all of them. In fact, patterns can be considered images in the pixel space, but coefficients are just a vector of real numbers. This way, analyzing the similarity of two images is reduced to basically comparing two vectors.

\begin{figure}[H]
  \centering
   \includegraphics[width=0.4\linewidth]{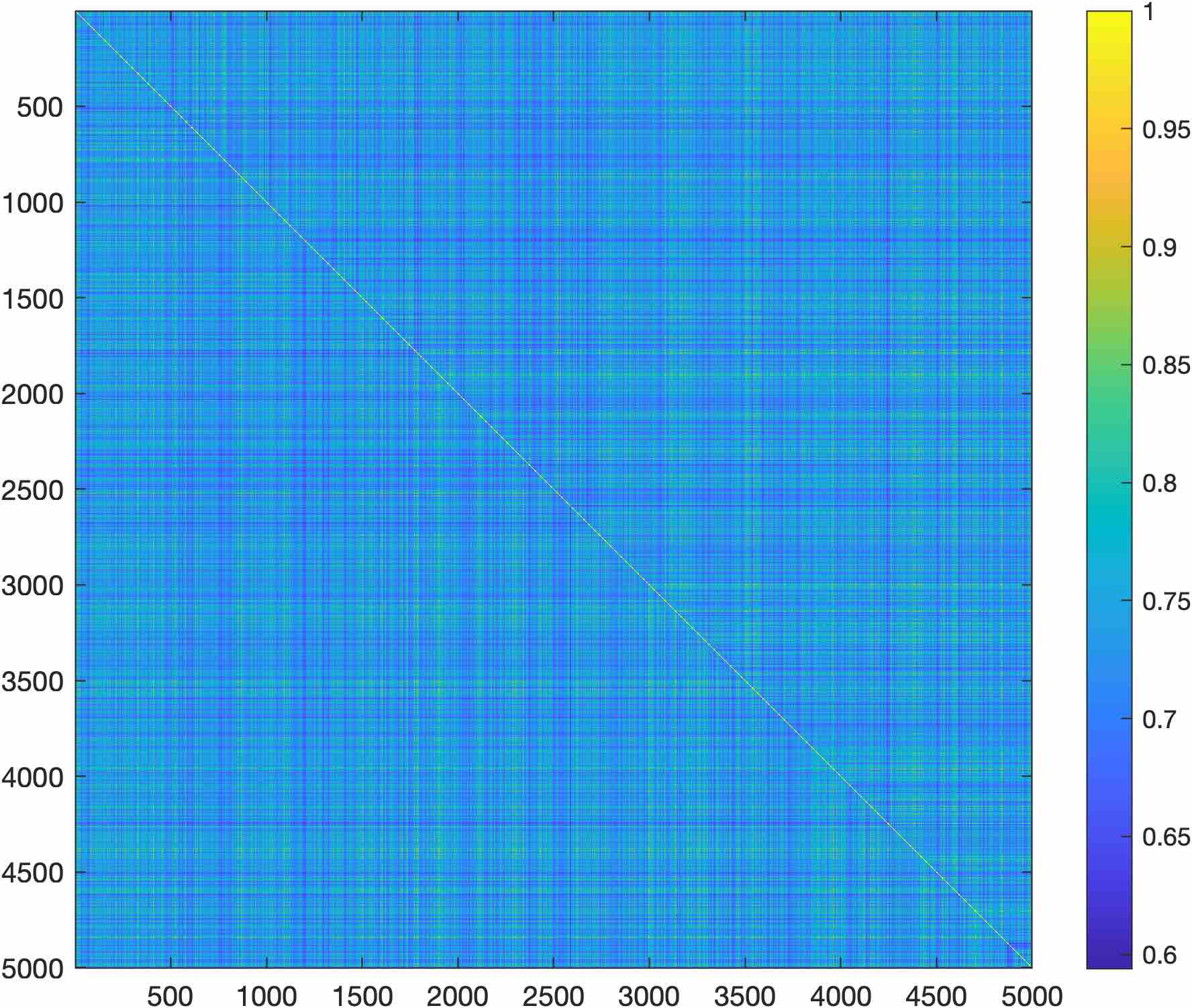}
  \caption{Similarity matrix of images in the Cat class based on the coefficients in $U \Sigma$.}
  \label{fig:cifar_cat_similarity}
\end{figure}

Figure~\ref{fig:cifar_cat_similarity} shows the similarity matrix of all images in the Cat class. To derive this similarity matrix, we followed these steps:

\begin{enumerate}
    \item Compute the matrix $U_i \Sigma_i$ which has 5,000 rows and 3,000 columns for the Cat class.
    \item Compute the 2-norm distance between each row in this matrix to obtain a distance matrix of 5,000 rows and columns.
    \item Convolve the distance matrix with a Gaussian kernel to obtain the similarity matrix.
\end{enumerate}

Once we obtain a similarity matrix, a wide array of analysis can be performed. For example, one can perform clustering to identify communities of similar images. By analyzing the eigenvalues of a graph Laplacian built from the similarity matrix, we identified that there are not any large communities inside this dataset, as previously reported by \cite{yousefzadeh2020using}. However, coarse graining (i.e., organizing the data in large number of clusters) can easily identify small groups of images (usually 2 to 5) that are very similar, and sometimes nearly identical.

\begin{figure}[H]
  \centering
   \includegraphics[width=0.1\linewidth]{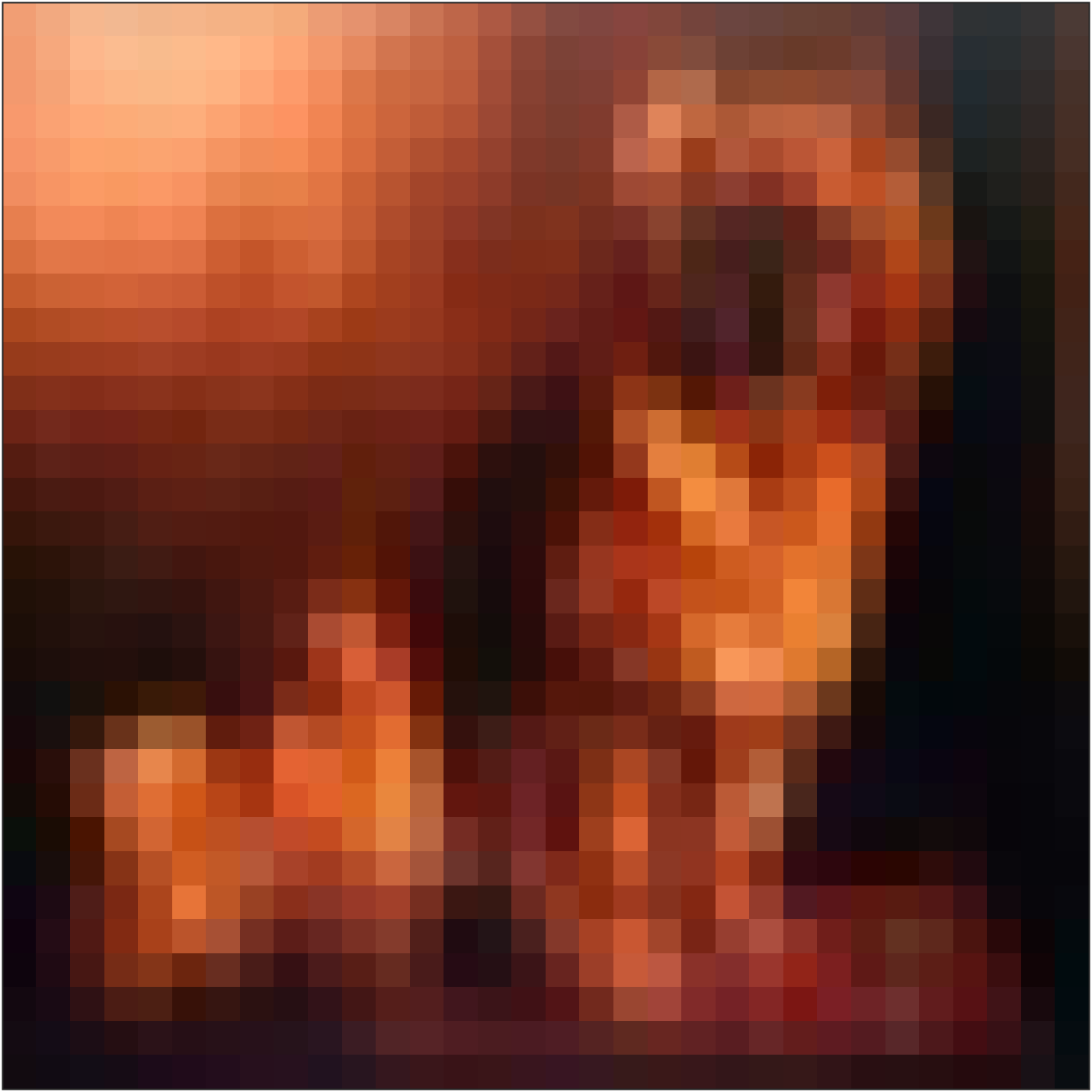}
   \includegraphics[width=0.1\linewidth]{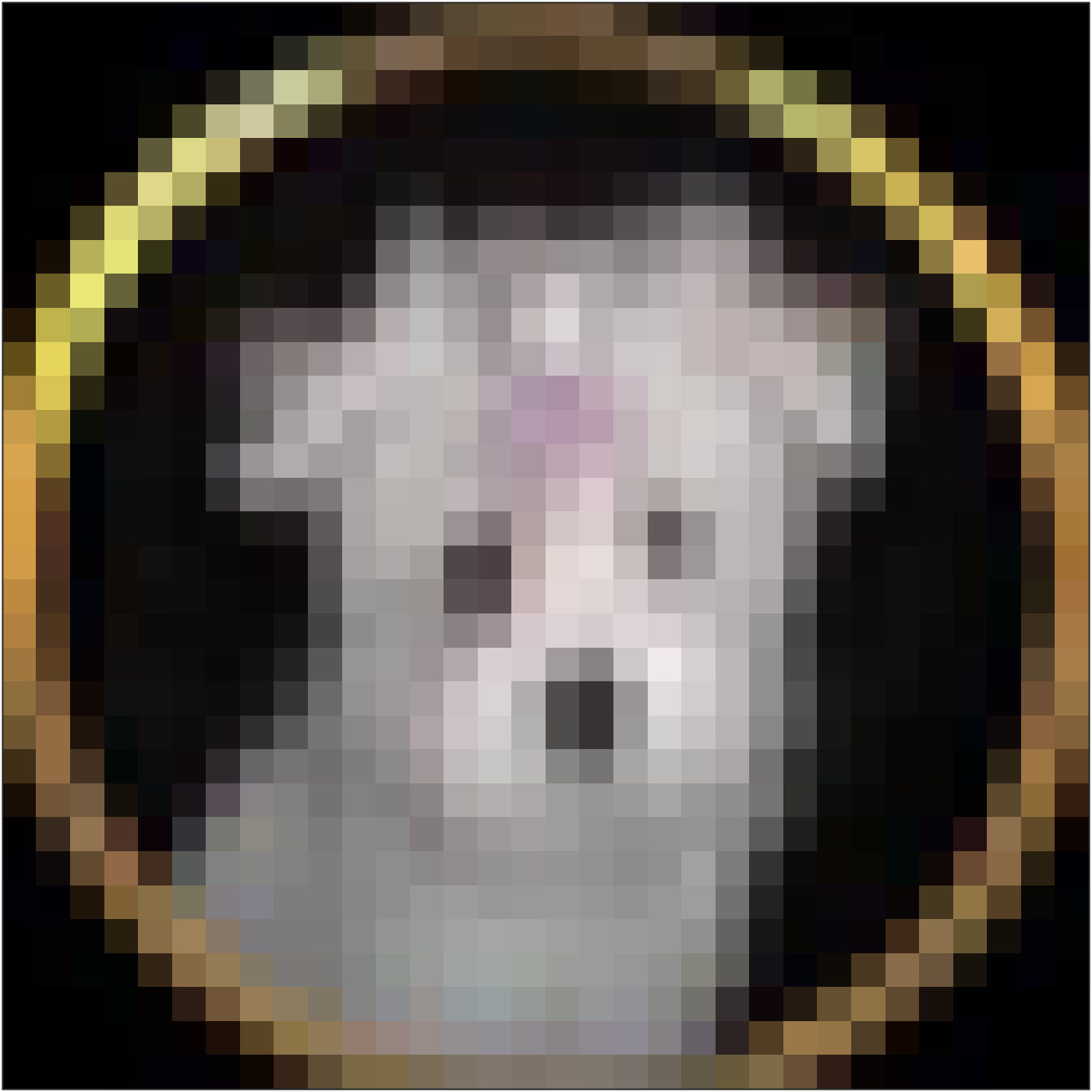}
   \includegraphics[width=0.1\linewidth]{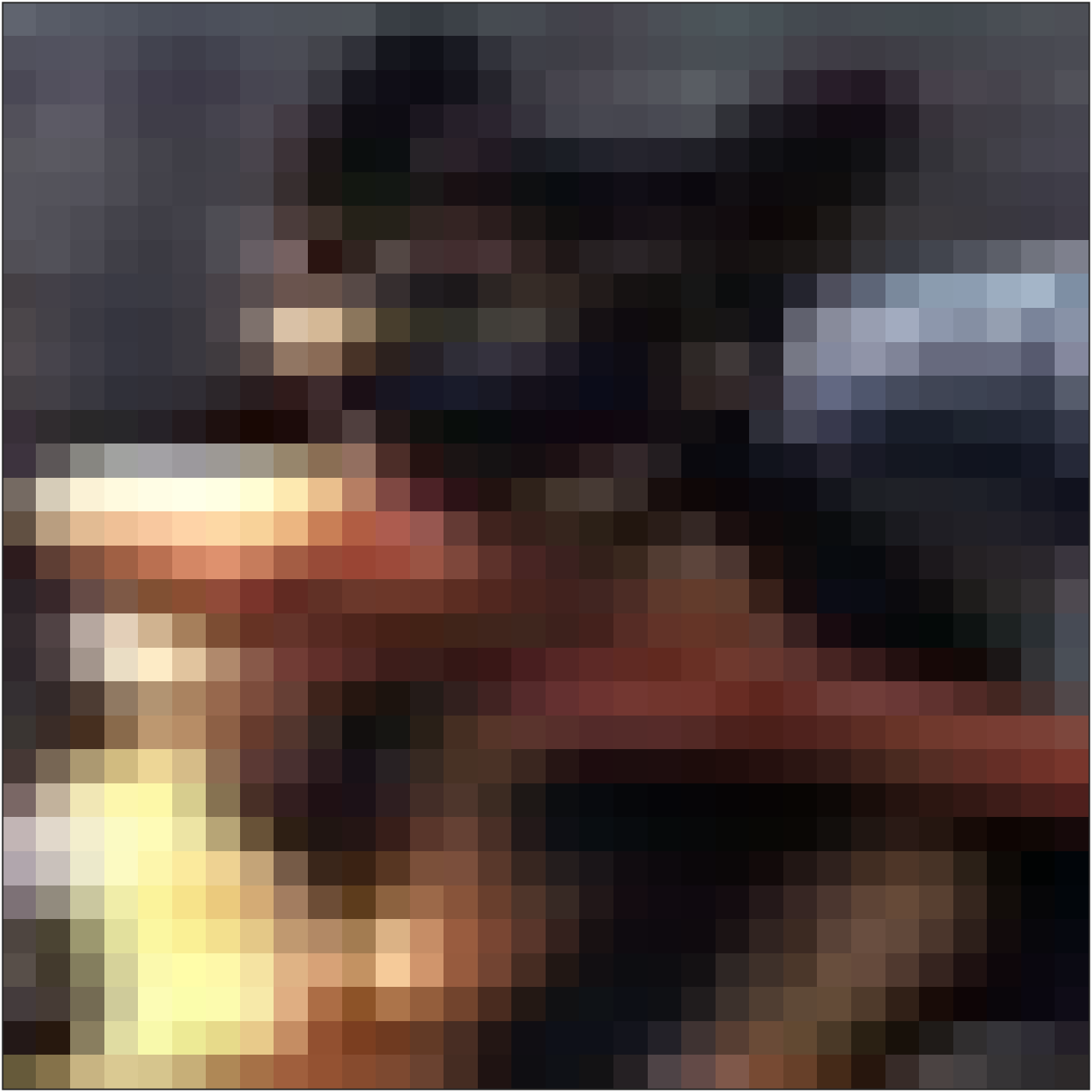}
   \includegraphics[width=0.1\linewidth]{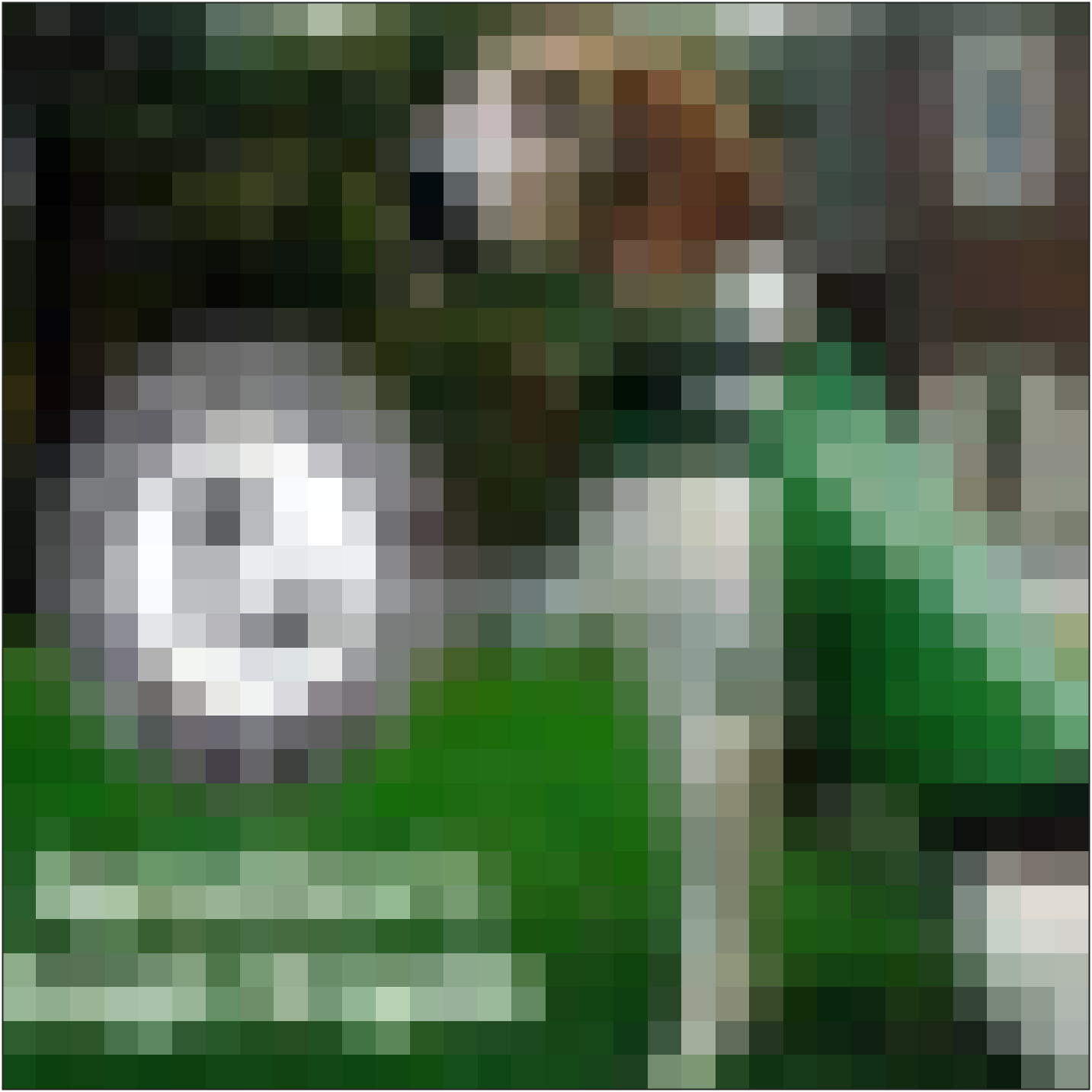}
   \includegraphics[width=0.1\linewidth]{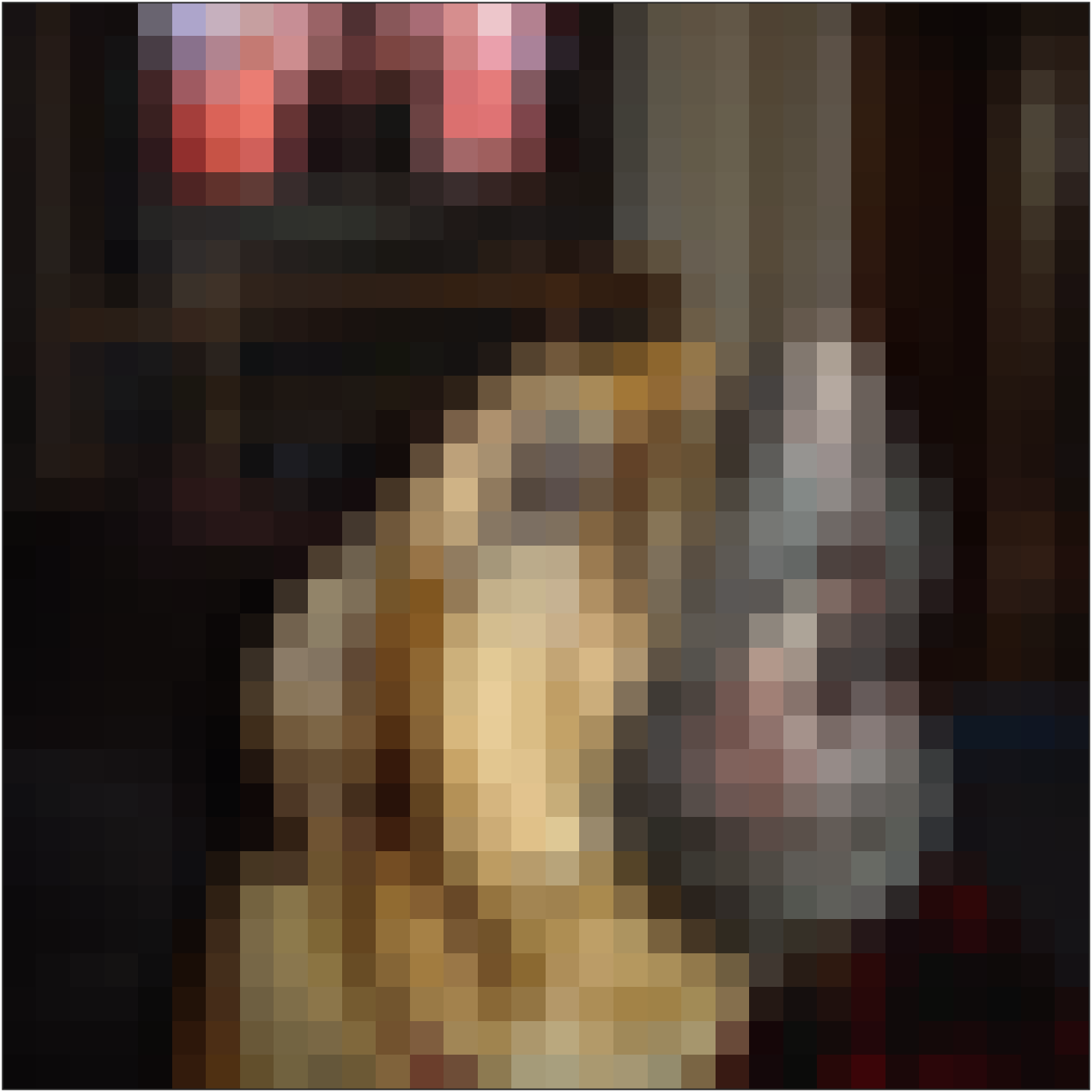}
   \includegraphics[width=0.1\linewidth]{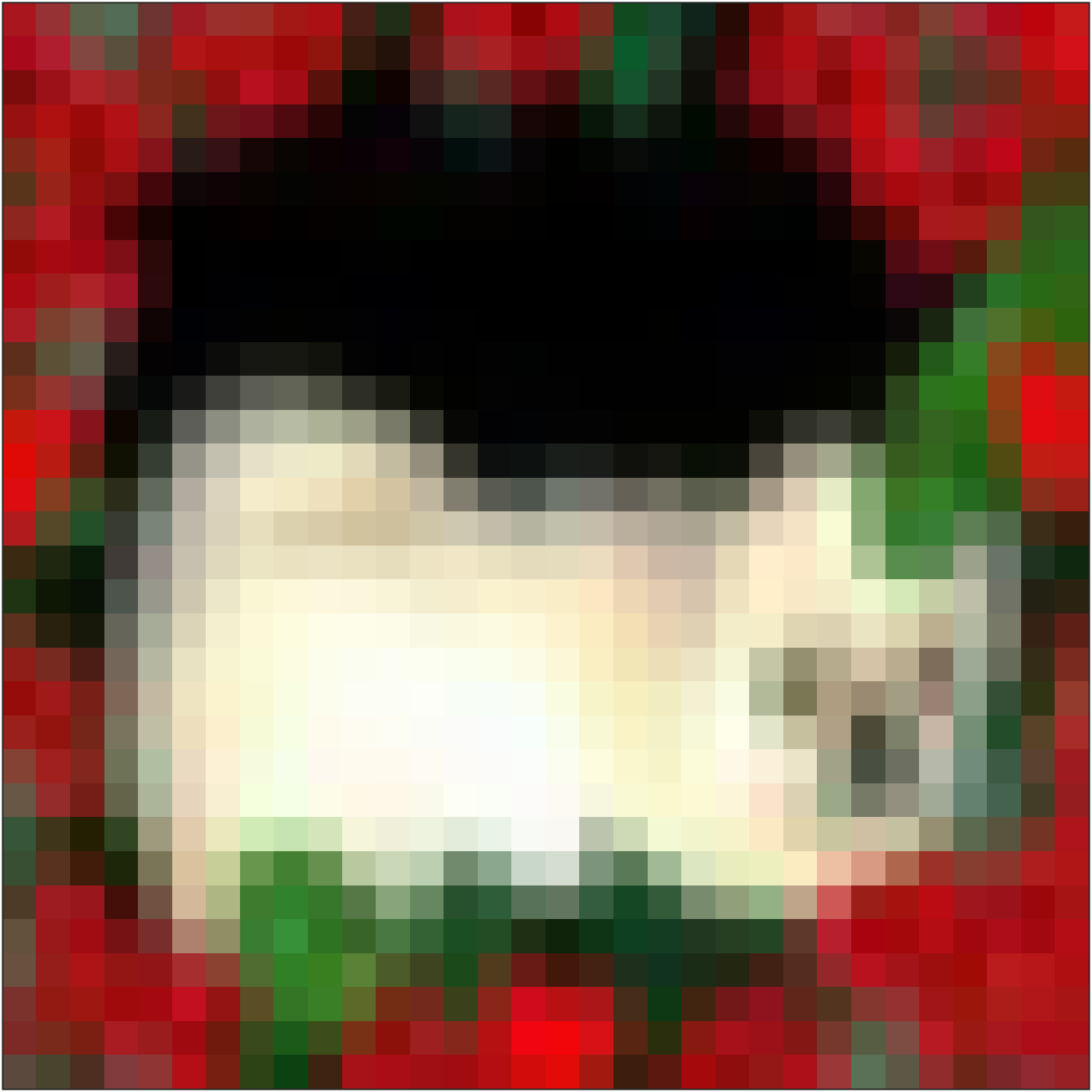}
   \includegraphics[width=0.1\linewidth]{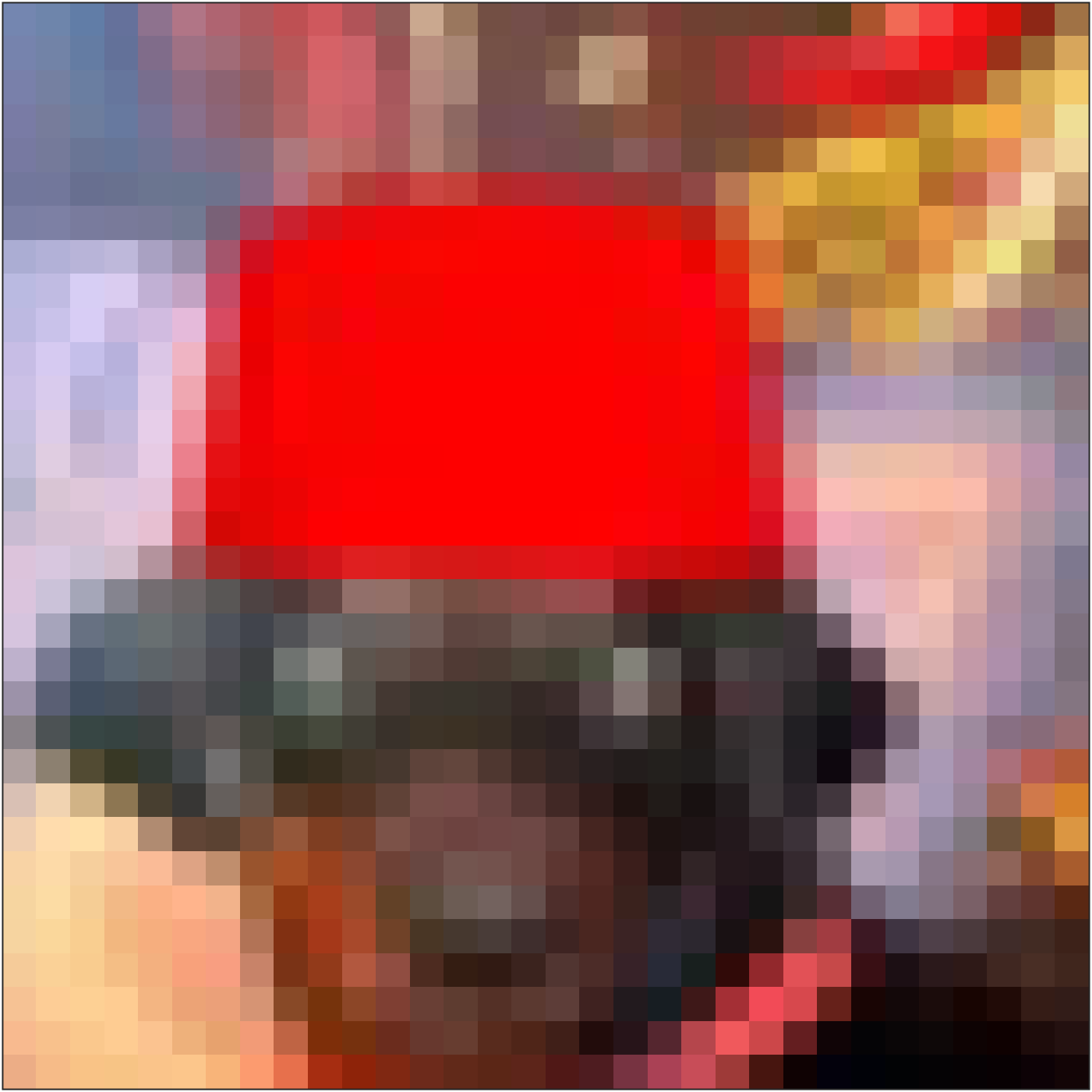}
  \caption{Dog images that are most isolated within the dog class, identified by the similarity matrix. We can consider these out-of-distribution images in the context of CIFAR-10 training set.}
  \label{fig:cifar_dog_anomalies}
\end{figure}

Figure~\ref{fig:cifar_dog_anomalies} shows dog images that are most isolated within the dog class. Clearly, we may consider them anomalies or out-of distribution images in the context of this dataset, because picture of a dog with a red bucket on its head, or picture of a dog with red flowers around the perimeter of image are not common. This relates to the broad literature on detecting out-of-distribution images and anomalies in image classification datasets, e.g., \cite{liang2018enhancing,huang2019out,ren2019likelihood}. This is another possible extension of our method.

\begin{figure}[H]
  \centering
  \includegraphics[width=0.9\linewidth]{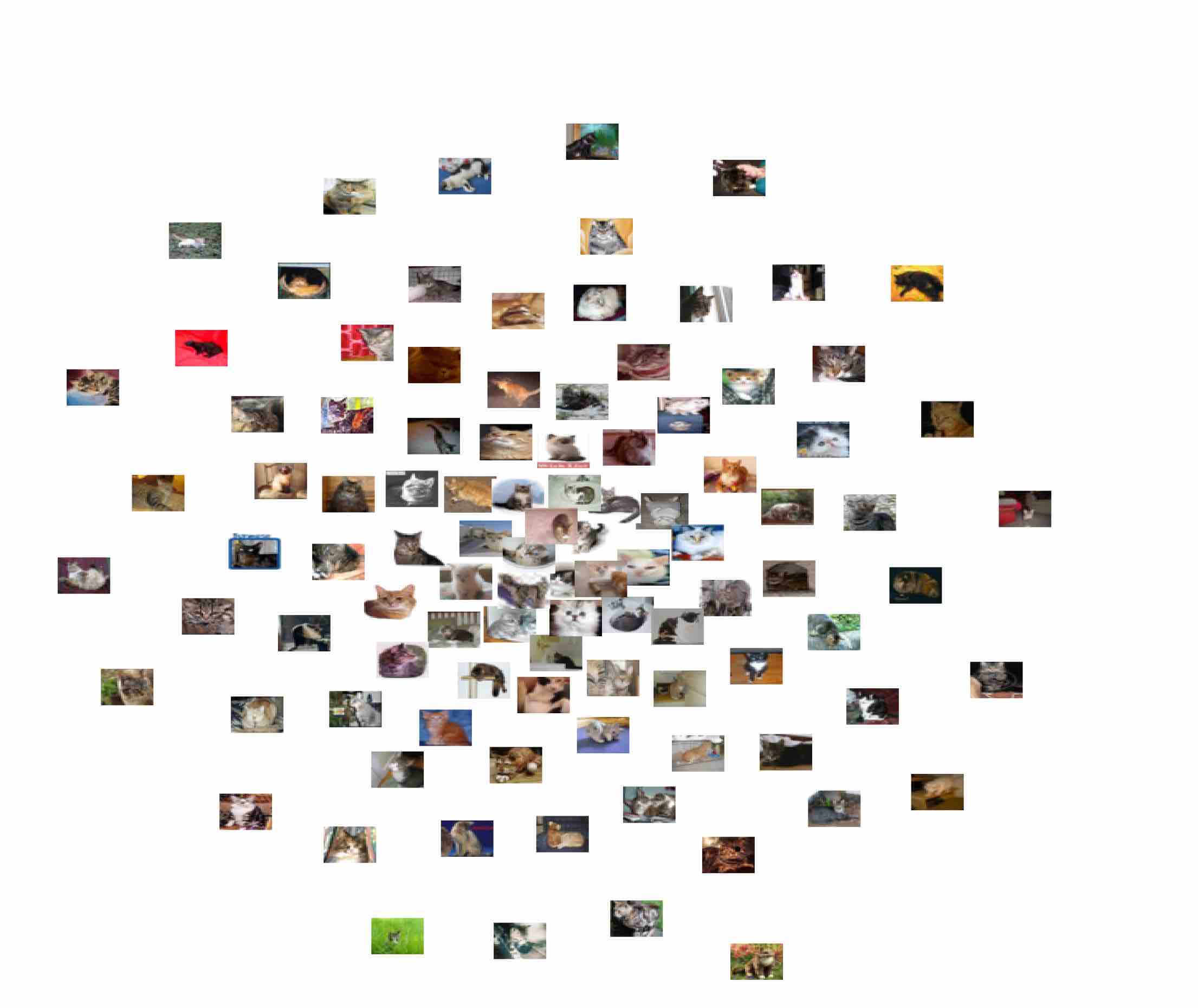}
  \caption{t-SNE plot obtained from the $U_i \Sigma_i$ matrix of cat images.}
  \label{fig:cifar_cat_tsne}
\end{figure}

We can also use the coefficients in $U_i \Sigma_i$ to visualize the dataset in form of an embedding. Figure~\ref{fig:cifar_cat_tsne} shows the embedding obtained by the t-SNE (t-Distributed Stochastic Neighbor Embedding) algorithm \citep{maaten2008visualizing}. t-SNE is a dimensionality reduction algorithm, specifically designed for visualizing high dimensional data.

This approach of visualizing a set of images in form of an embedding relates to a separate line of research in the literature. Those methods usually suffer from very high computational costs, because they rely on pairwise comparison of images in the pixel space, e.g., \cite{vo2019unsupervised}. We also note that t-SNE is used before for visualizing image datasets, for example by \cite{parde2017face}, but such approaches use a pre-trained VGG model for comparison of images. Our approach does not require a pre-trained model, however, we note that its usefulness requires further experiments and comparison with other methods.

\setcounter{figure}{0}
\renewcommand{\thefigure}{F\arabic{figure}}

\section{COVID-19 CT-Scan Images} \label{sec:appx_covid}

\begin{figure}[H]
  \centering
  \fbox{
   \includegraphics[width=0.22\linewidth]{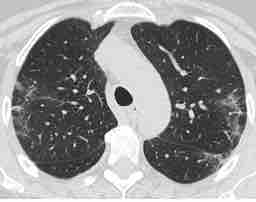}
   \includegraphics[width=0.22\linewidth]{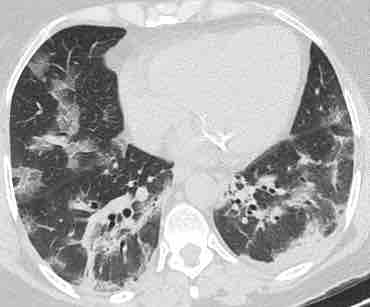}}
   \hfil
  \fbox{
   \includegraphics[width=0.2\linewidth]{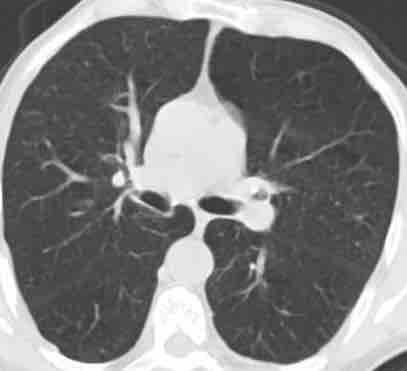}
   \includegraphics[width=0.2\linewidth]{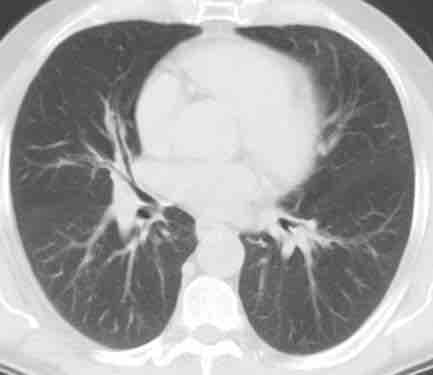}}
   \caption{Sample of images in the SARS-COV-2 CT-Scan dataset. Two images in the left box are from {\em infected} patients and the ones in the right box are from {\em non-infected} patients.}
  \label{fig:covid_samples}
\end{figure}

Here, we analyze the contents of the SARS-COV-2 CT-Scan Dataset \citep{angelov2020explainable}. This dataset contains 2,482 images of CT-Scan, 1,230 of which belong to infected patients and 1,252 belong to non-infected patients. Figure~\ref{fig:covid_samples} shows some samples from this dataset.

\begin{figure}[H]
  \centering
   \includegraphics[width=0.25\linewidth]{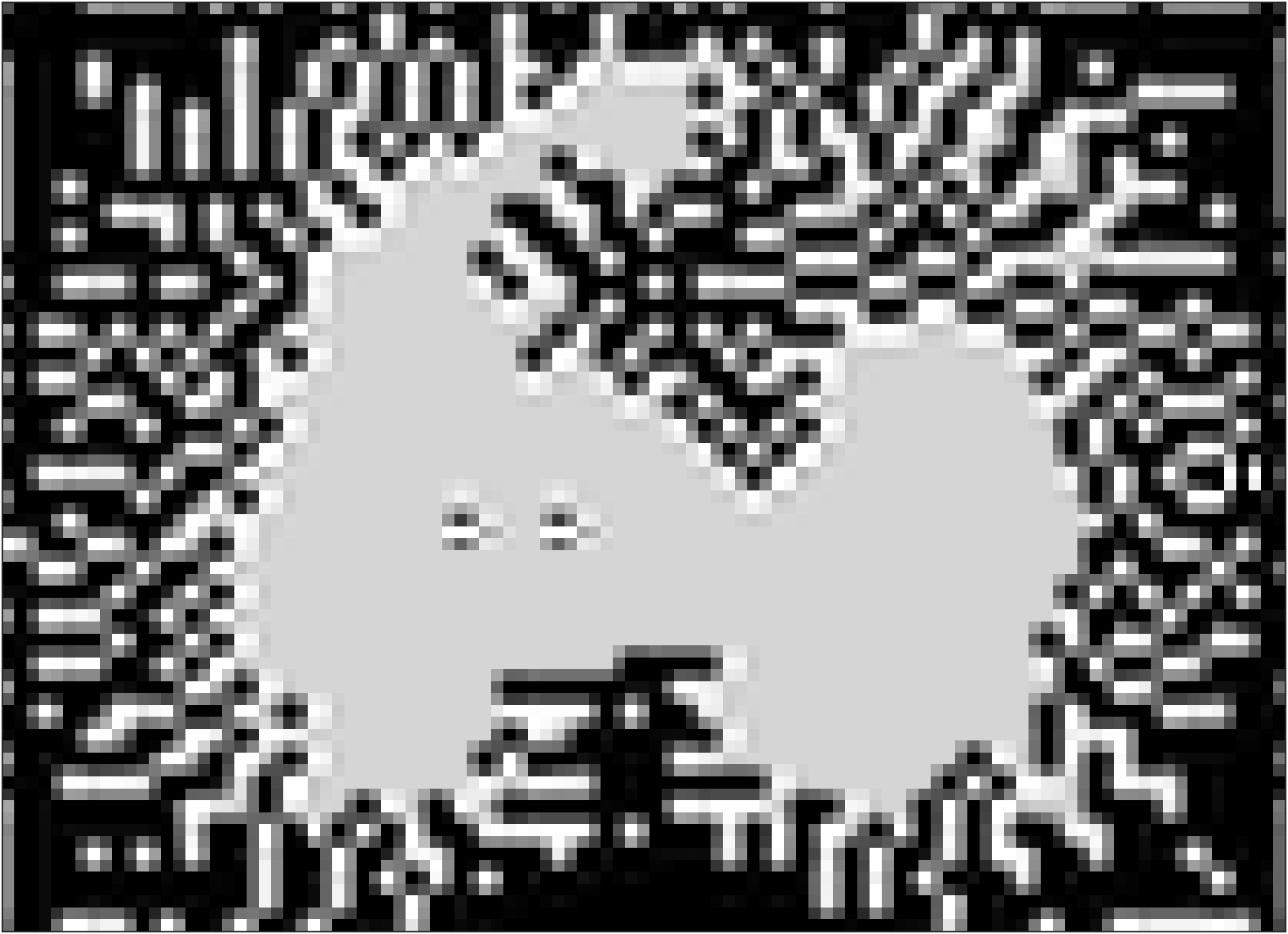}
   \caption{Pixels corresponding to the most influential coefficients chosen by the RR-QR algorithm. A radiologist could verify whether the patterns corresponding to infections are expected to be in those regions. RR-QR detects that most of the variations among images are in the white regions, and changes in the black regions are either small or linearly dependent.}
  \label{fig:covid_stencil}
\end{figure}

We form the data into two matrices, $\mathcal{D}_1$ for COVID samples and $\mathcal{D}_2$ for non-COVID samples. For wavelet transformation, we use Daubechies-3 wavelets. We use the RR-QR algorithm to choose a subset of $m=1200$ most influential wavelet coefficients. Figure~\ref{fig:covid_stencil} highlights the pixels corresponding to those wavelet coefficients.

\begin{figure}[H]
  \centering
   \includegraphics[width=0.95\linewidth]{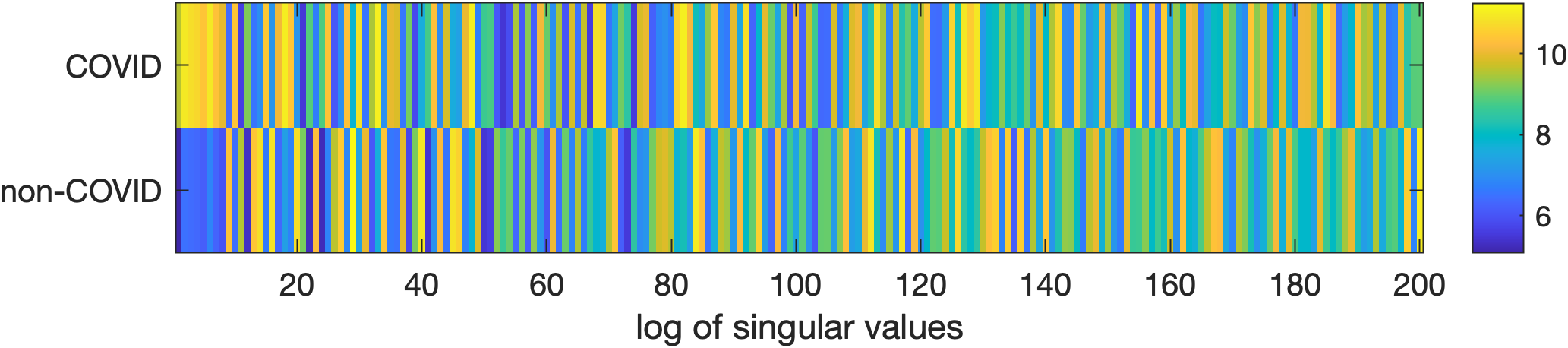}
   \caption{First 200 singular values (in logarithmic scale) show a clear separation between many of the patterns in COVID and non-COVID patients. Nitoce that when singular values are large for the COVID class, singular values are usually small for the non-COVID class, and vice versa.}
  \label{fig:covid_singulars}
\end{figure}

Using Higher Order GSVD, we decompose the tensor of wavelet coefficients $\mathcal{D}$. The resulting singular values, shown in Figure~\ref{fig:covid_singulars} show a clear separation between the Covid and non-Covid patients.

\begin{figure}[H]
  \centering
   \includegraphics[width=0.25\linewidth]{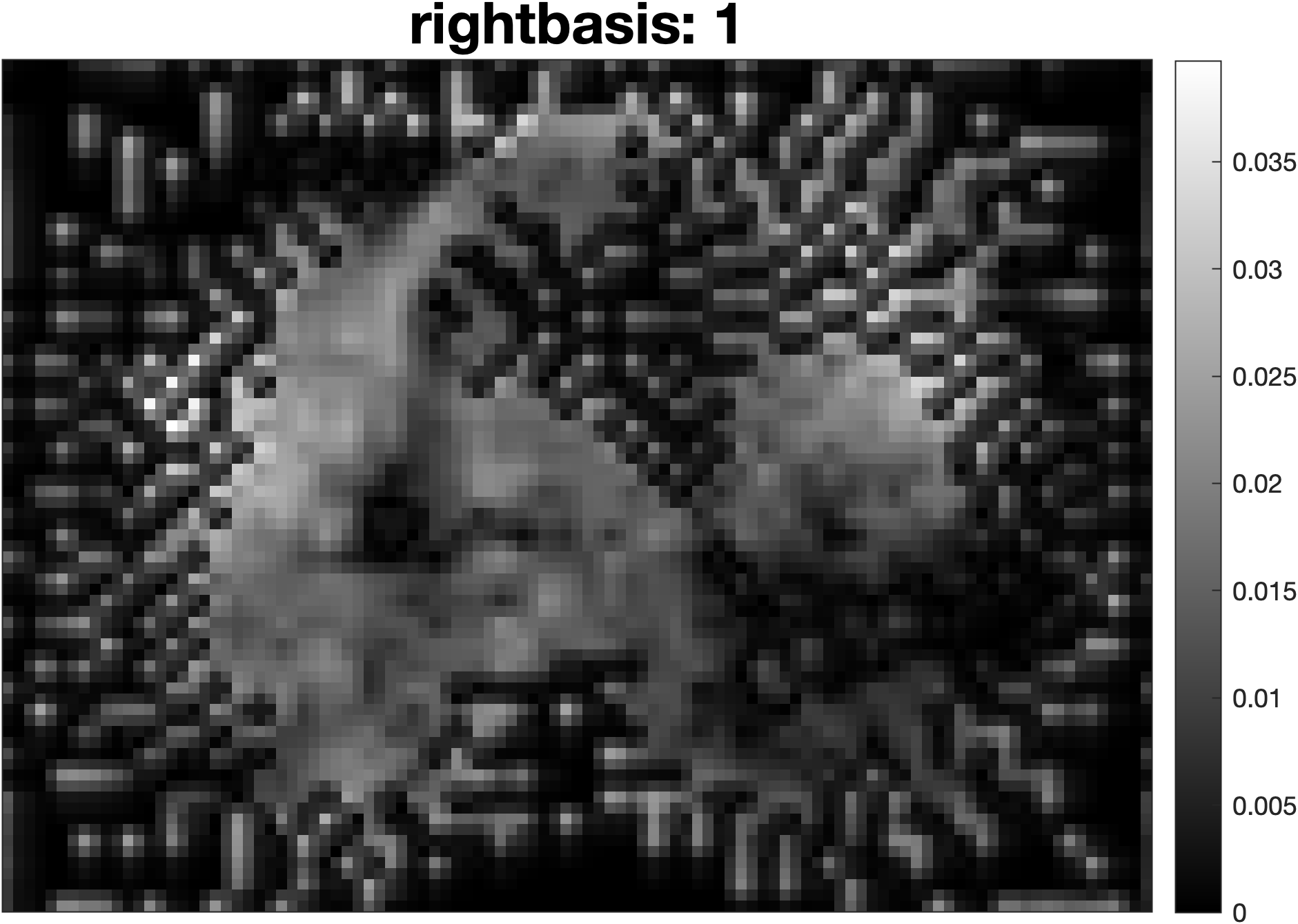}
   \includegraphics[width=0.25\linewidth]{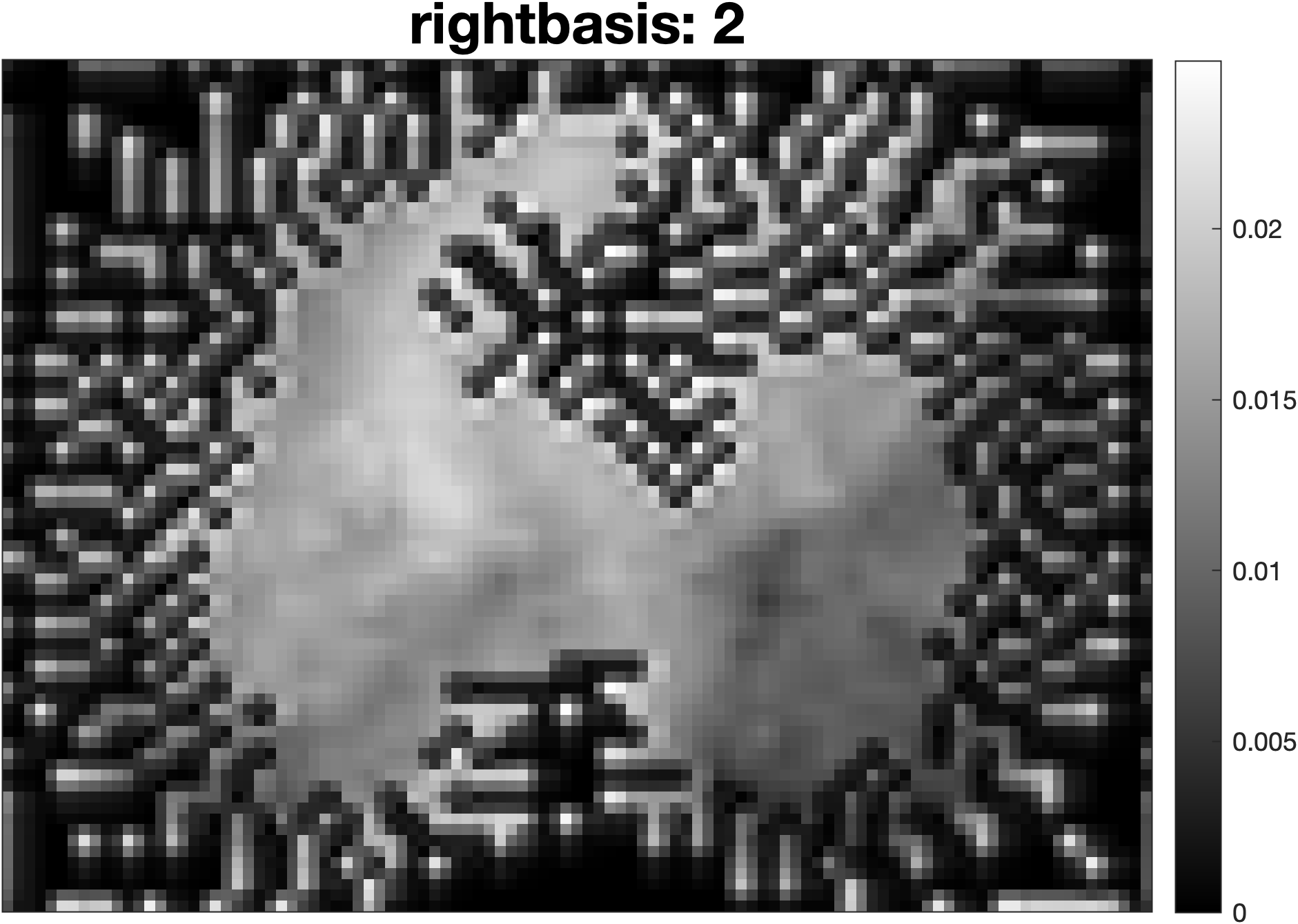}
   \includegraphics[width=0.25\linewidth]{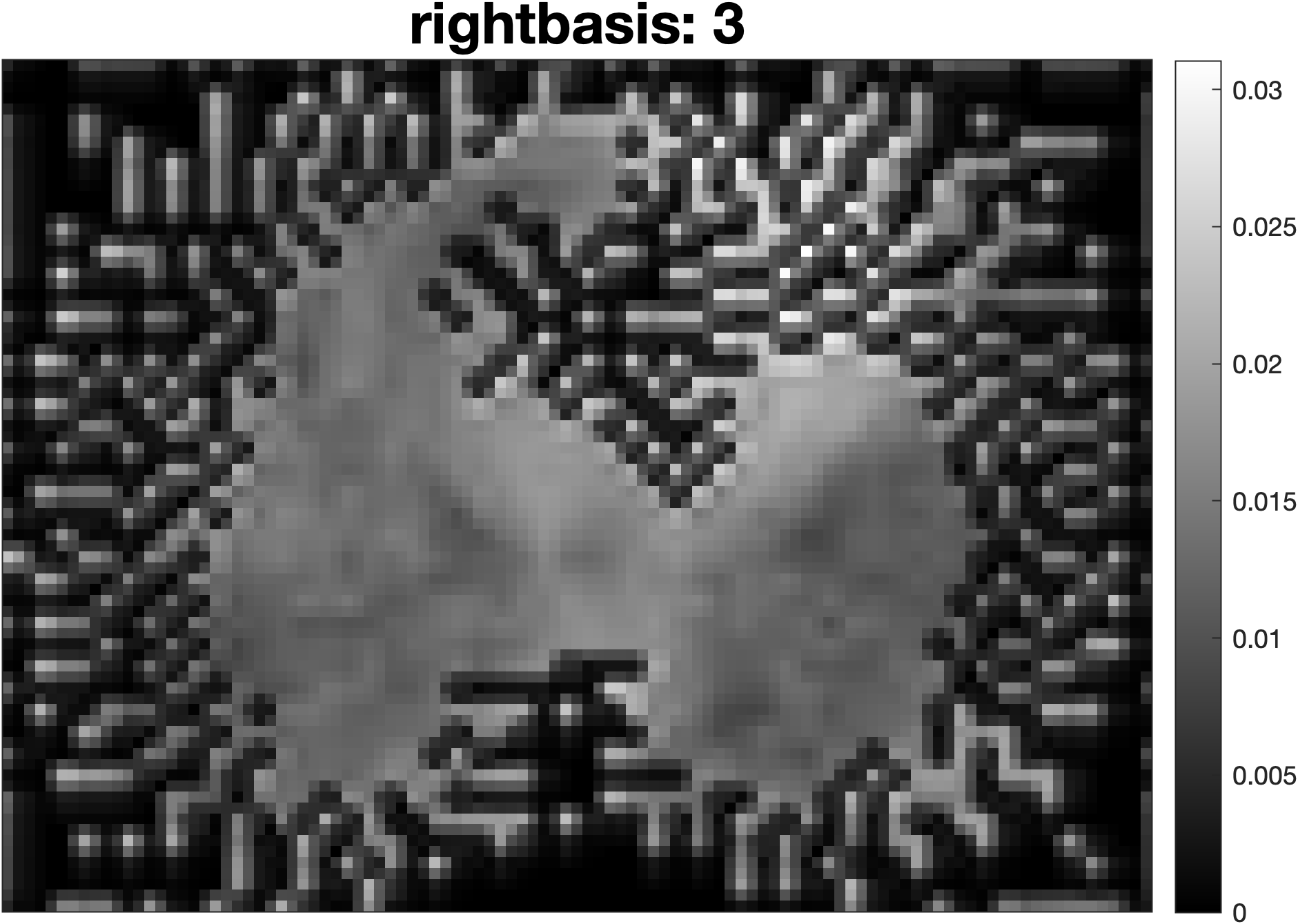}
   \includegraphics[width=0.25\linewidth]{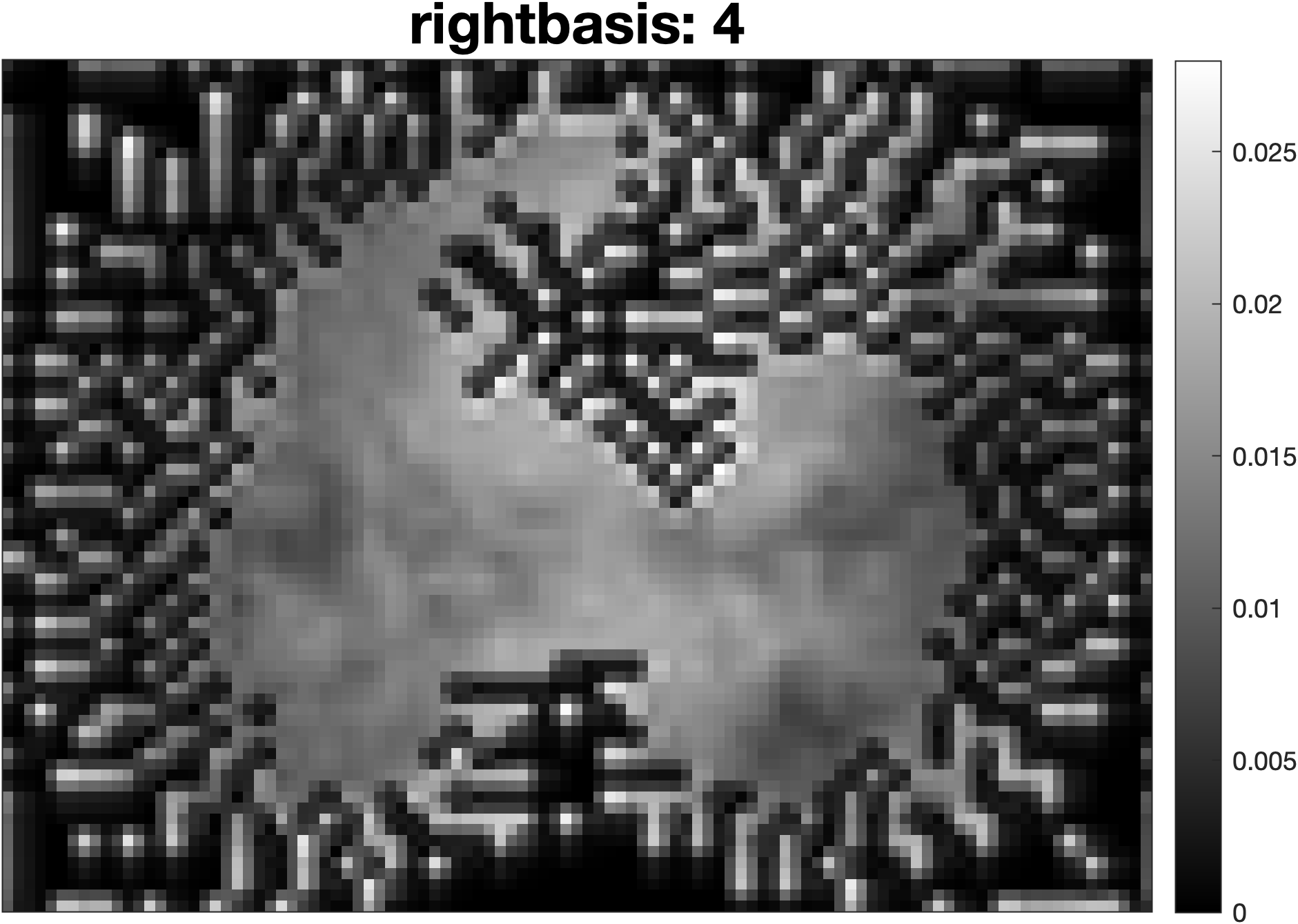}
   \includegraphics[width=0.25\linewidth]{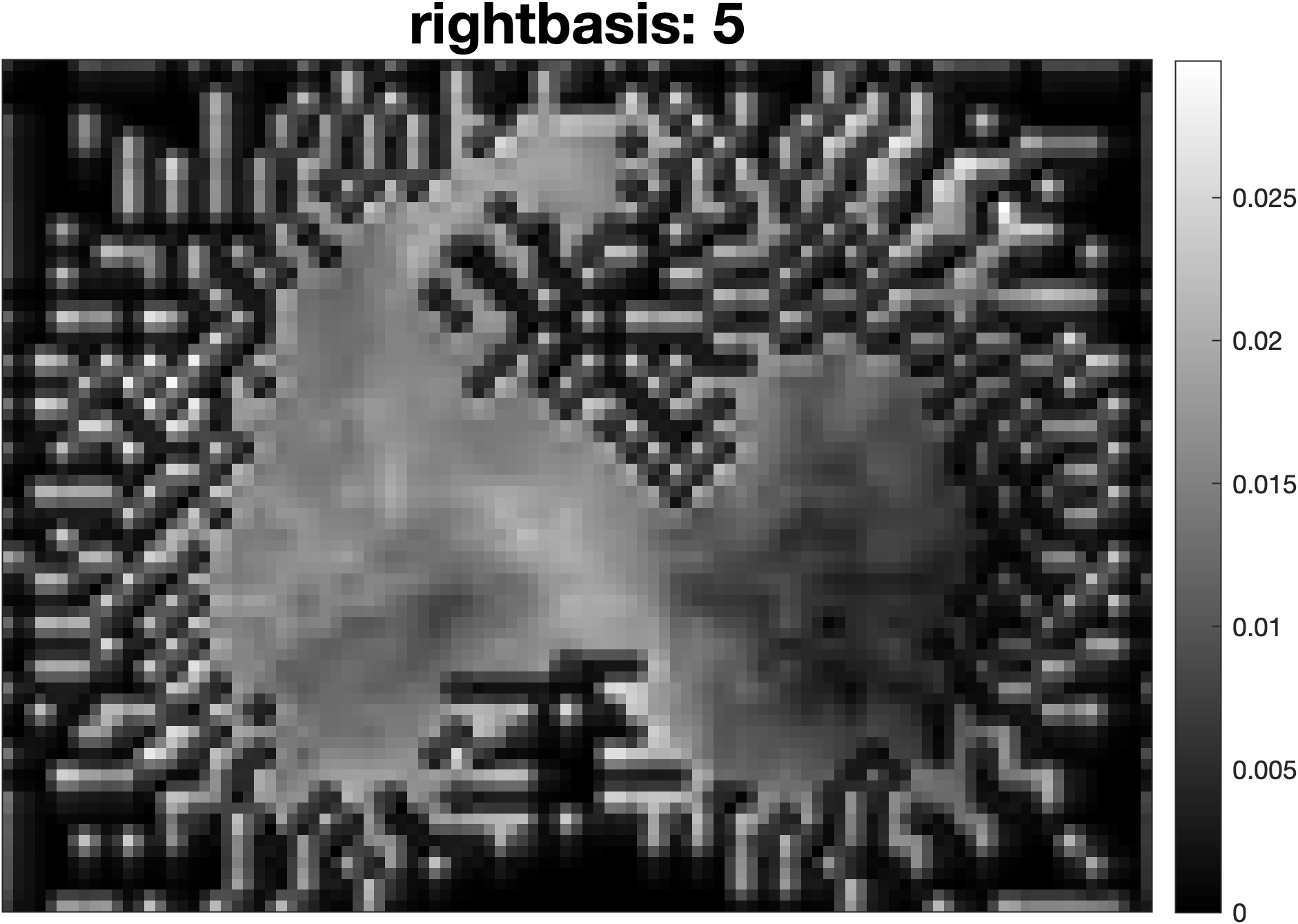}
   \caption{The most dominant patterns for COVID infected patients. Notice the color variations within the boundaries defined in Figure~\ref{fig:covid_stencil}.}
  \label{fig:covid_patterns}
\end{figure}

\begin{figure}[H]
  \centering
   \includegraphics[width=0.25\linewidth]{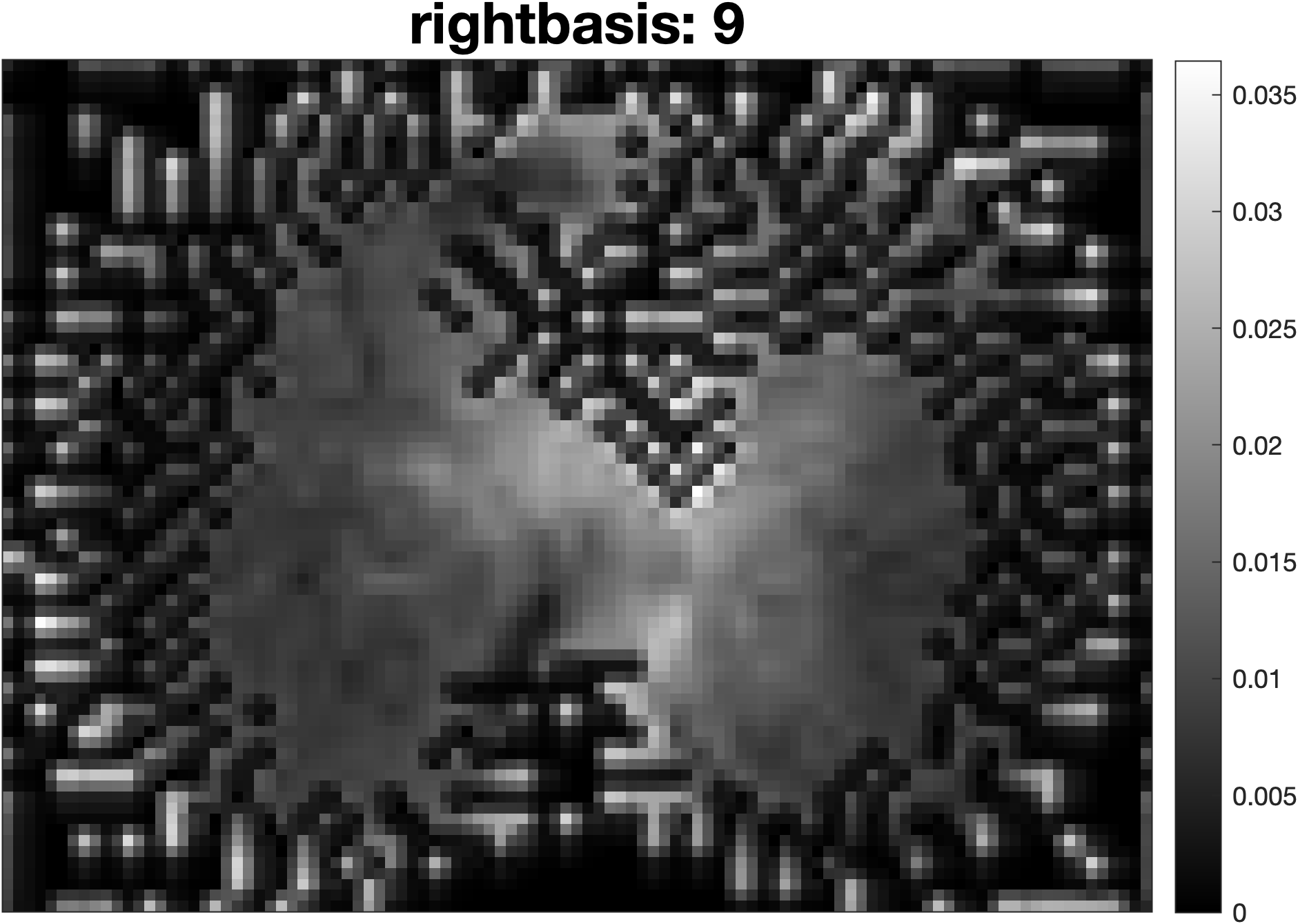}
   \includegraphics[width=0.25\linewidth]{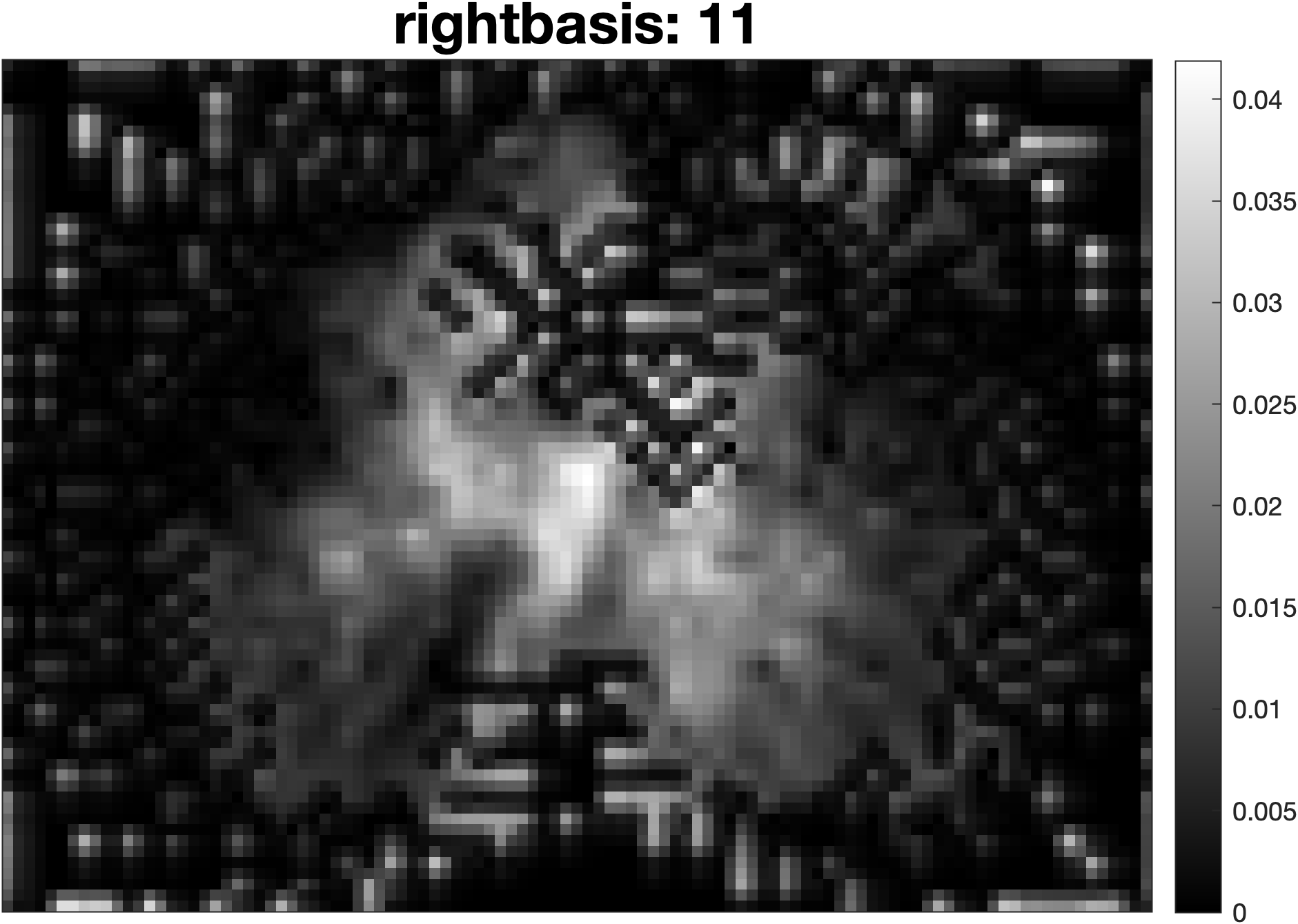}
   \includegraphics[width=0.25\linewidth]{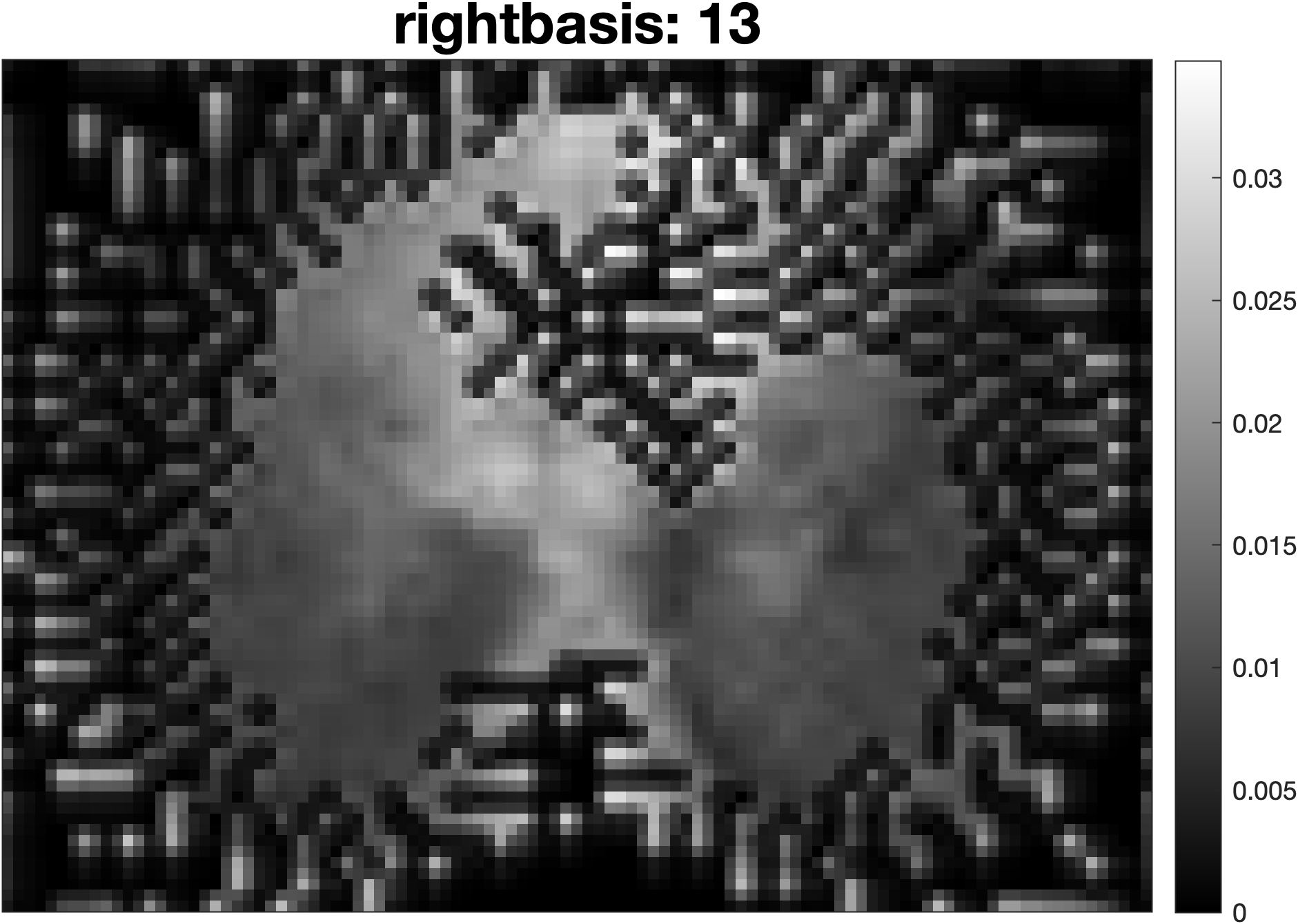}
   \includegraphics[width=0.25\linewidth]{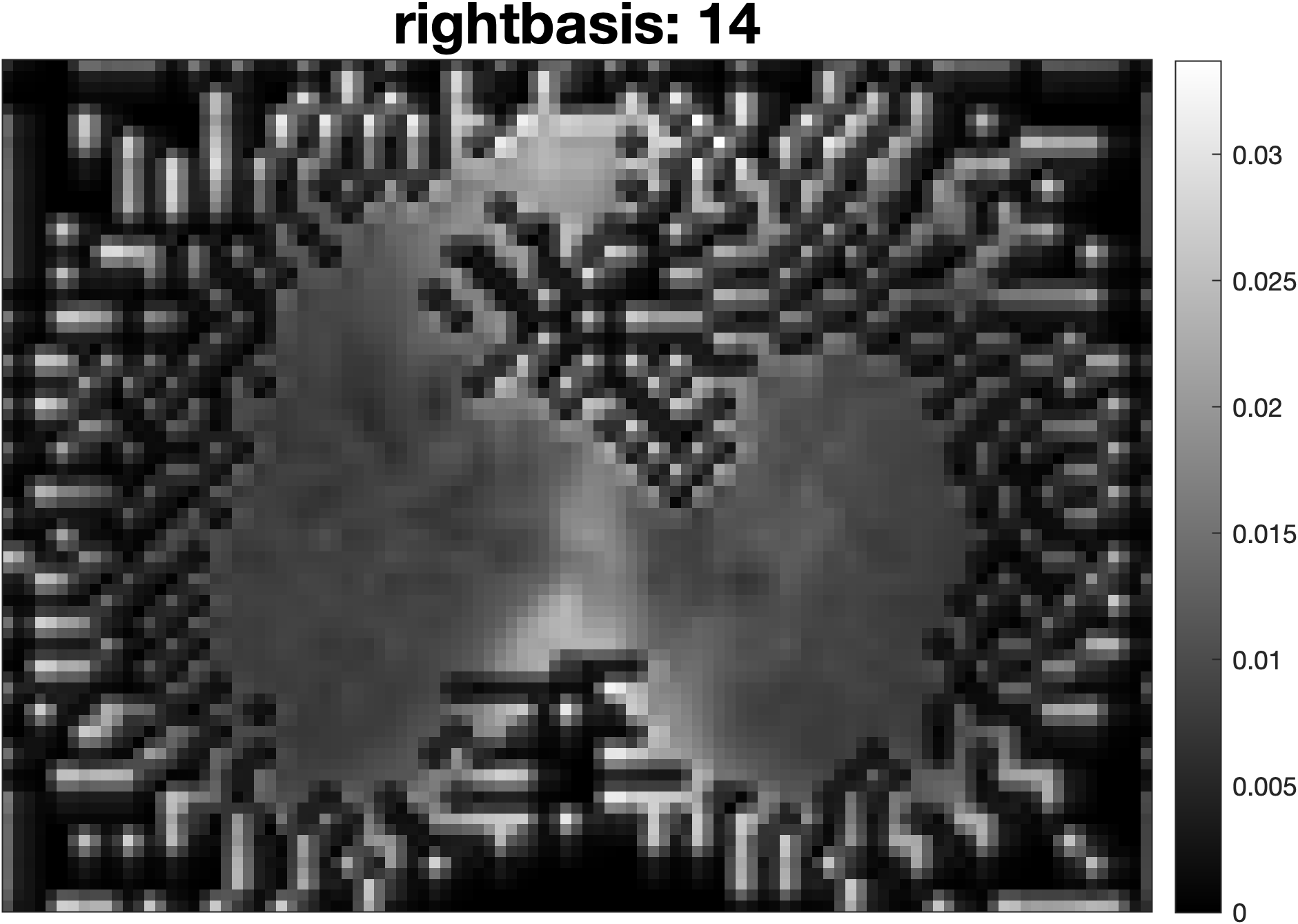}
   \includegraphics[width=0.25\linewidth]{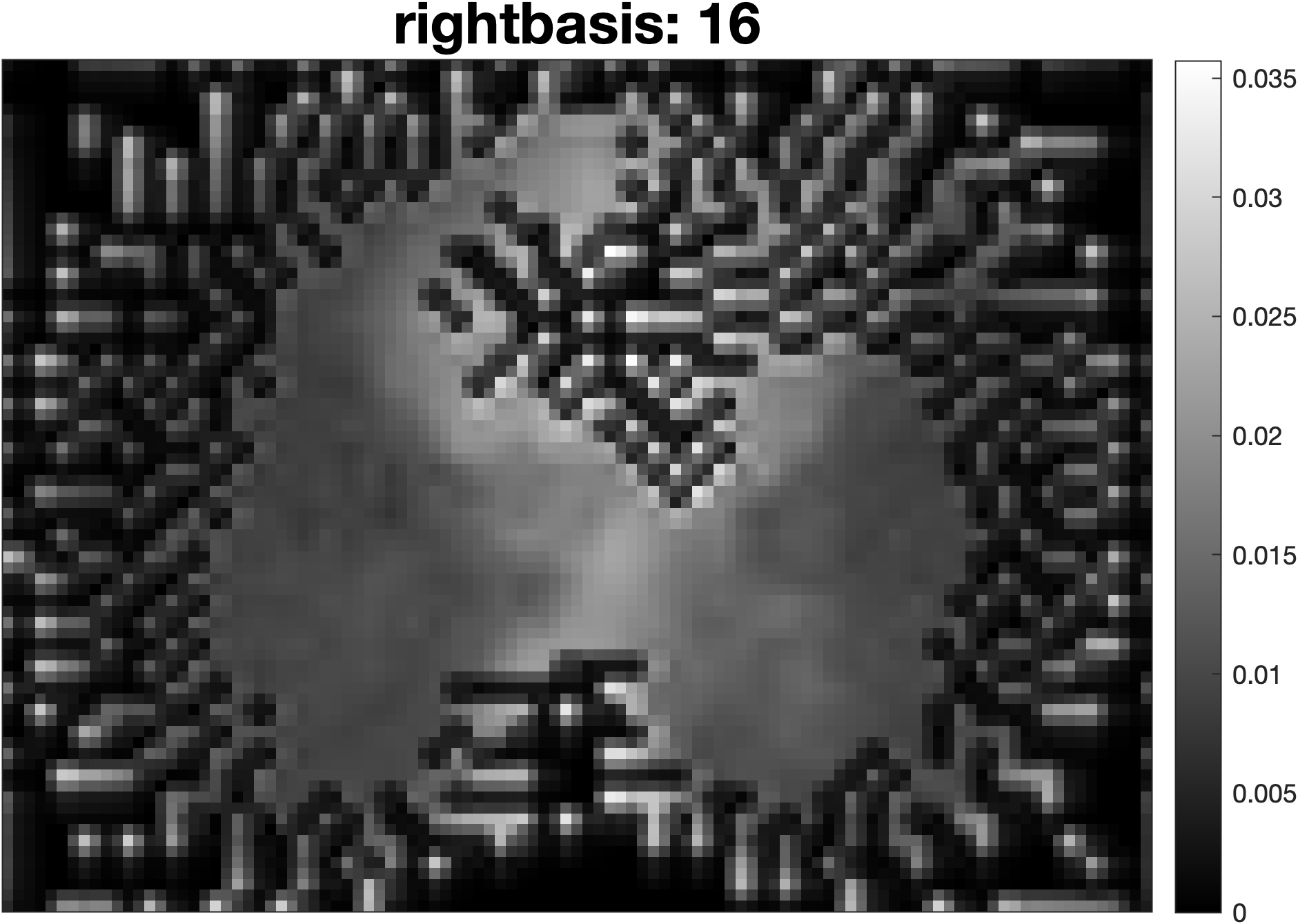}
   \caption{The most dominant patterns for non-COVID patients. Notice the color variations within the boundaries defined in Figure~\ref{fig:covid_stencil}. Also, notice how the color variations are different from the COVID-19 infected patients in Figure~\ref{fig:covid_stencil}.}
  \label{fig:noncovid_patterns}
\end{figure}

Figure~\ref{fig:covid_patterns} shows the most dominant patterns obtained for the COVID patients and Figure~\ref{fig:noncovid_patterns} shows the most dominant patterns obtained for non-COVID patients. Clearly patterns seem very different for non-COVID and COVID patients.

%Also, Figure~\ref{fig:covid_patterns2} shows some of the general pattern in $V$ that are not specifically dominant for any class.

% \begin{figure}[H]
%   \centering
%   \includegraphics[width=0.9\linewidth]{images-covid/covid_singulars.png}
%   \caption{First 200 singular values (in logarithmic scale) show a clear separation between many of the patterns in COVID and non-COVID patients.}
%   \label{fig:covid_patterns2}
% \end{figure}

\begin{figure}[H]
  \centering
   \includegraphics[width=0.24\linewidth]{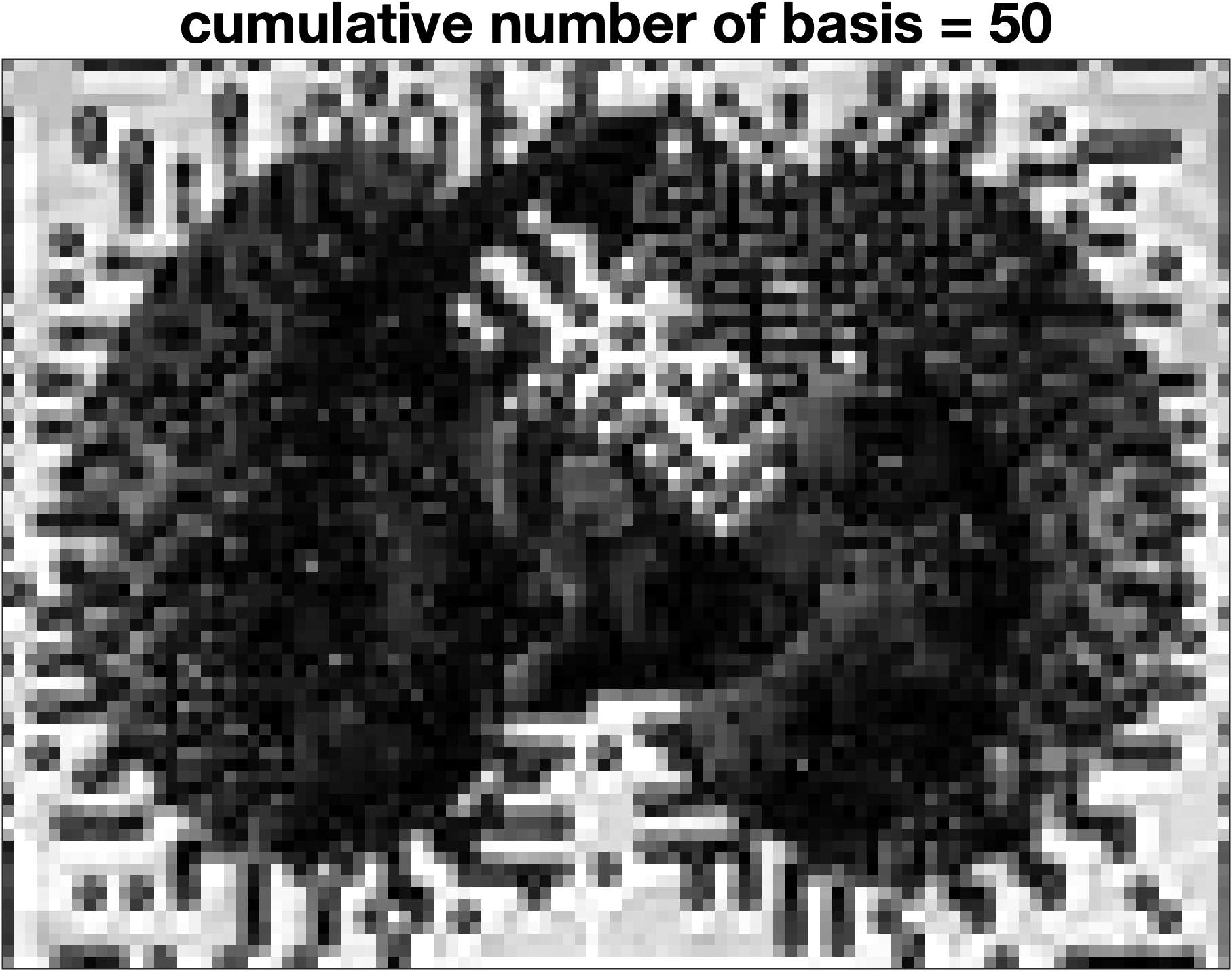}
   \includegraphics[width=0.24\linewidth]{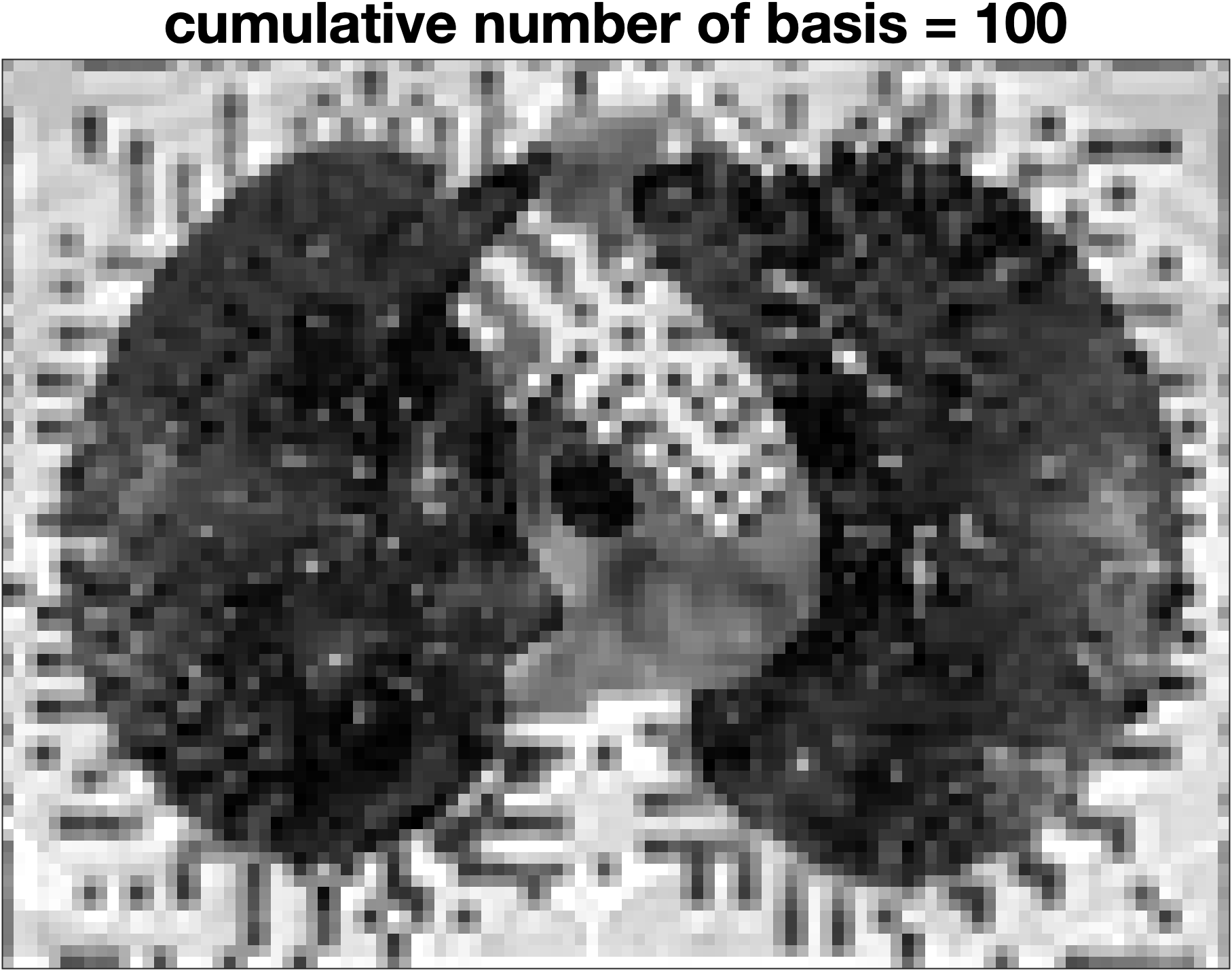}
   \includegraphics[width=0.24\linewidth]{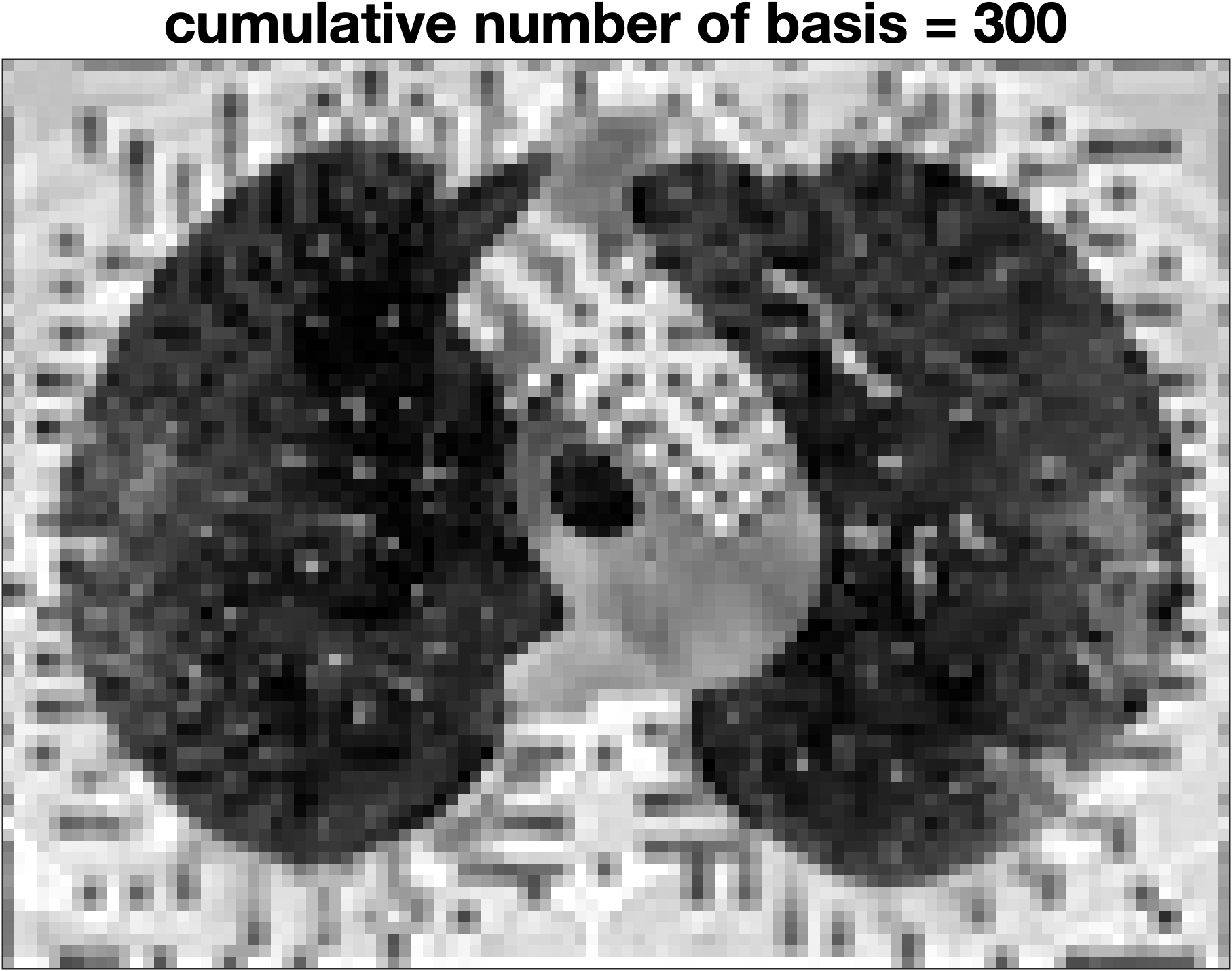}
   \includegraphics[width=0.24\linewidth]{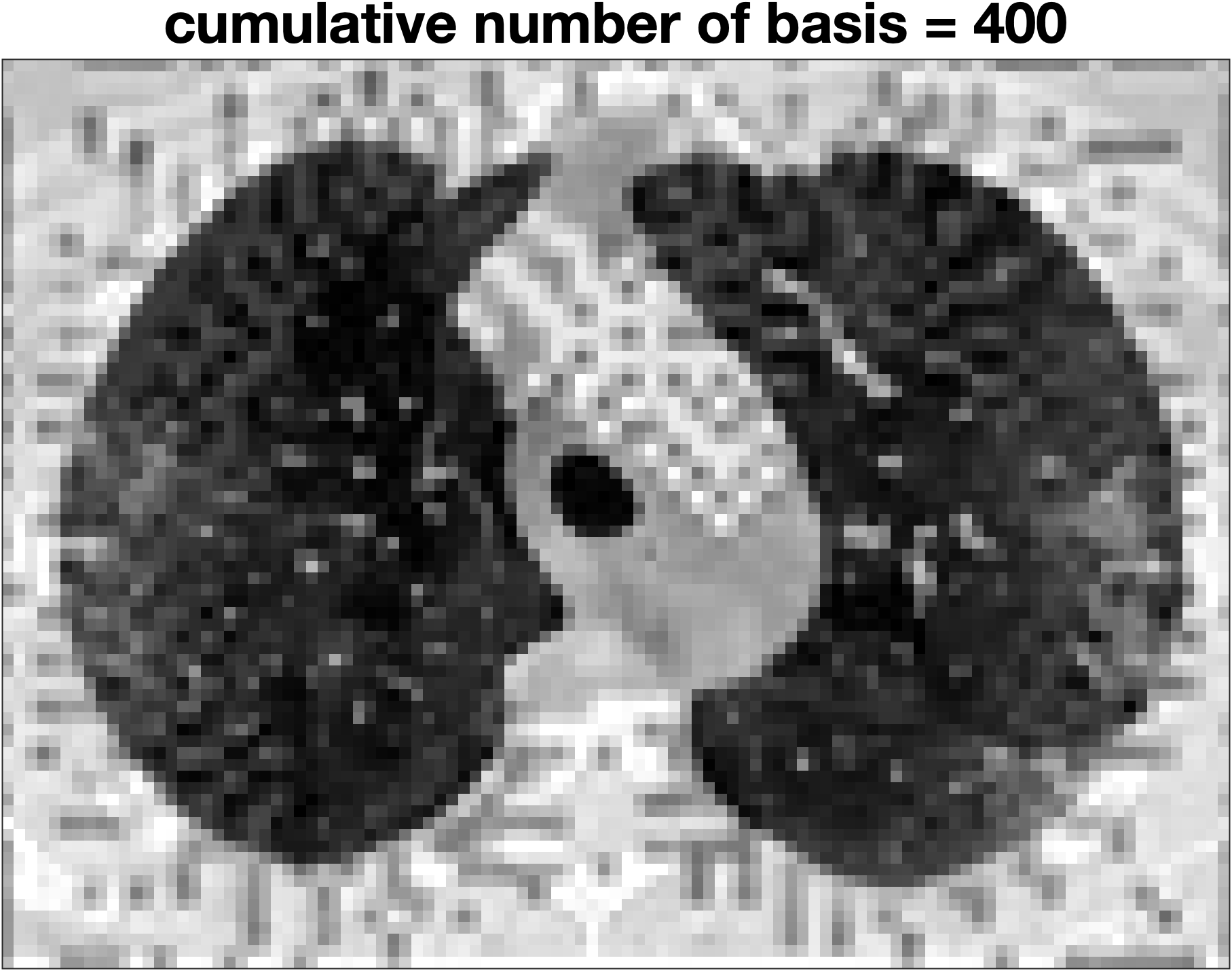}
   \includegraphics[width=0.24\linewidth]{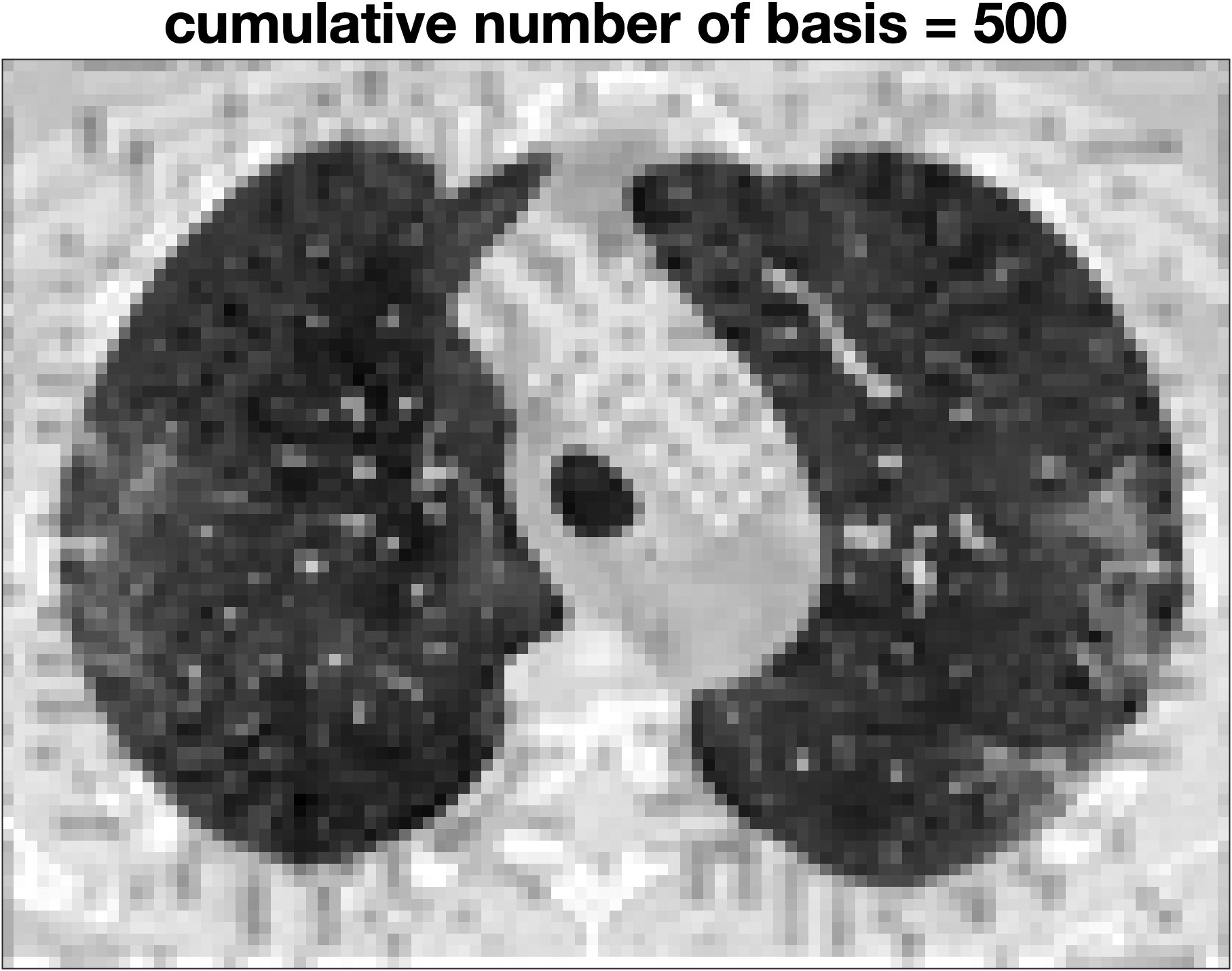}
   \includegraphics[width=0.24\linewidth]{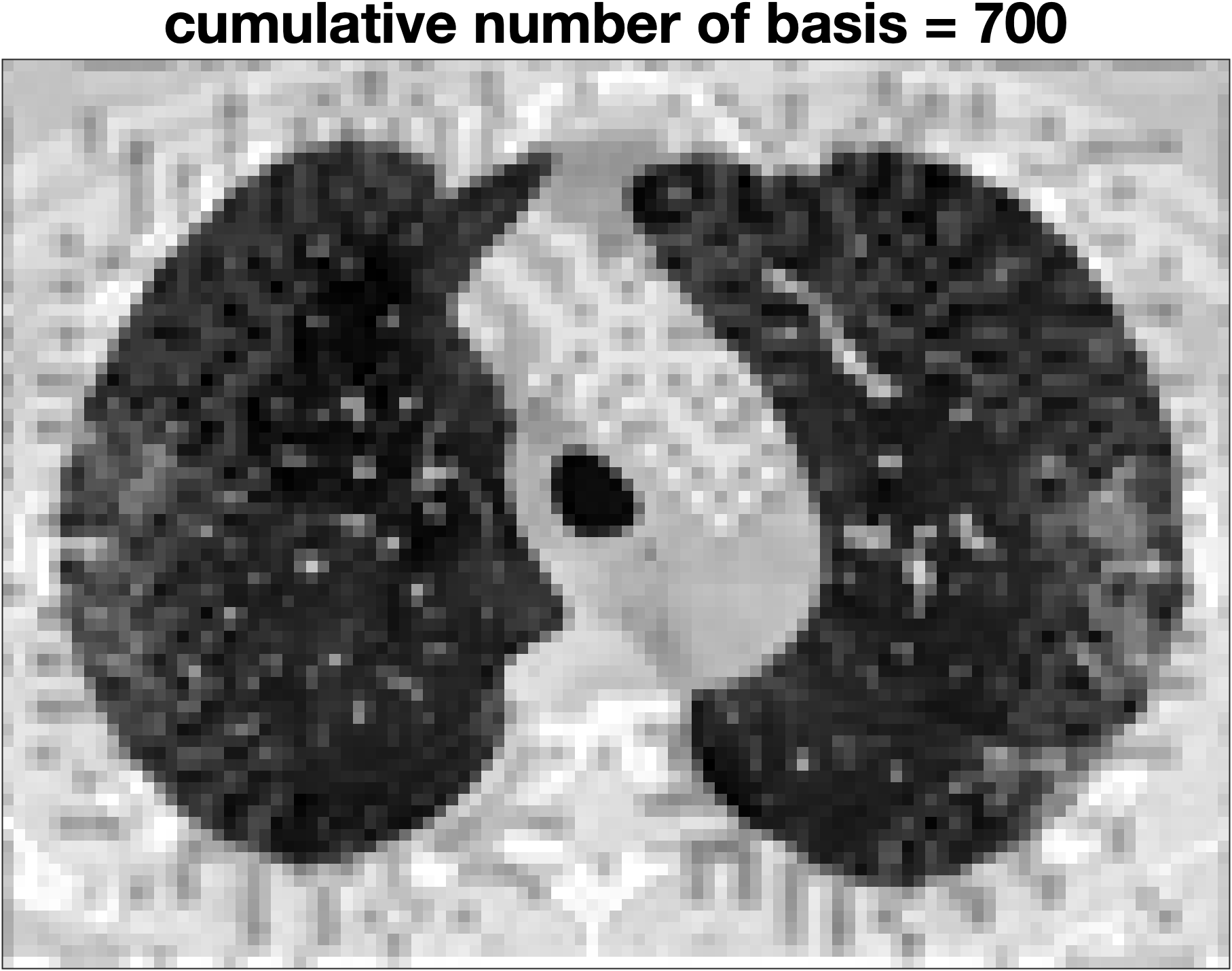}
   \includegraphics[width=0.24\linewidth]{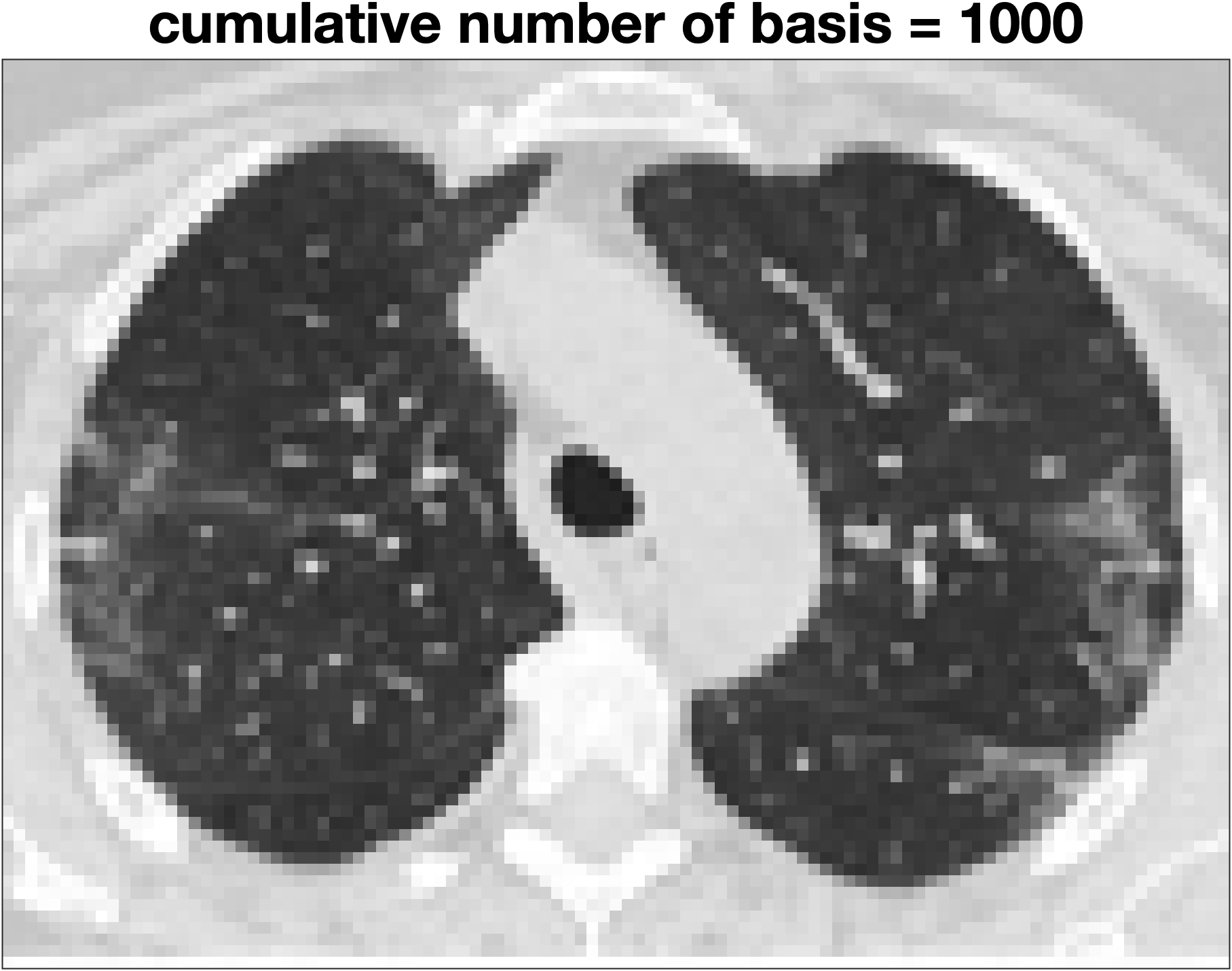}
   \includegraphics[width=0.24\linewidth]{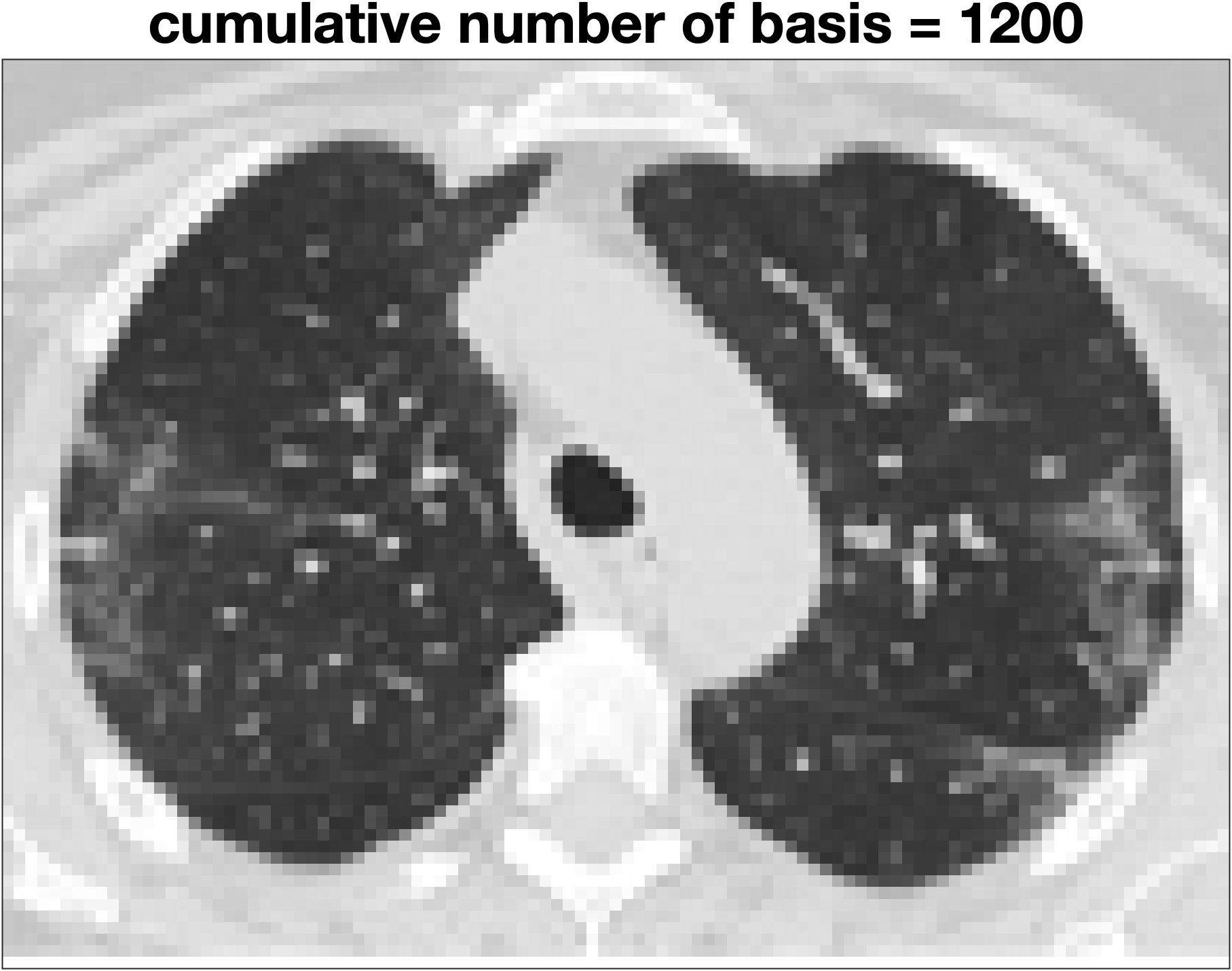}
   \includegraphics[width=0.7\linewidth]{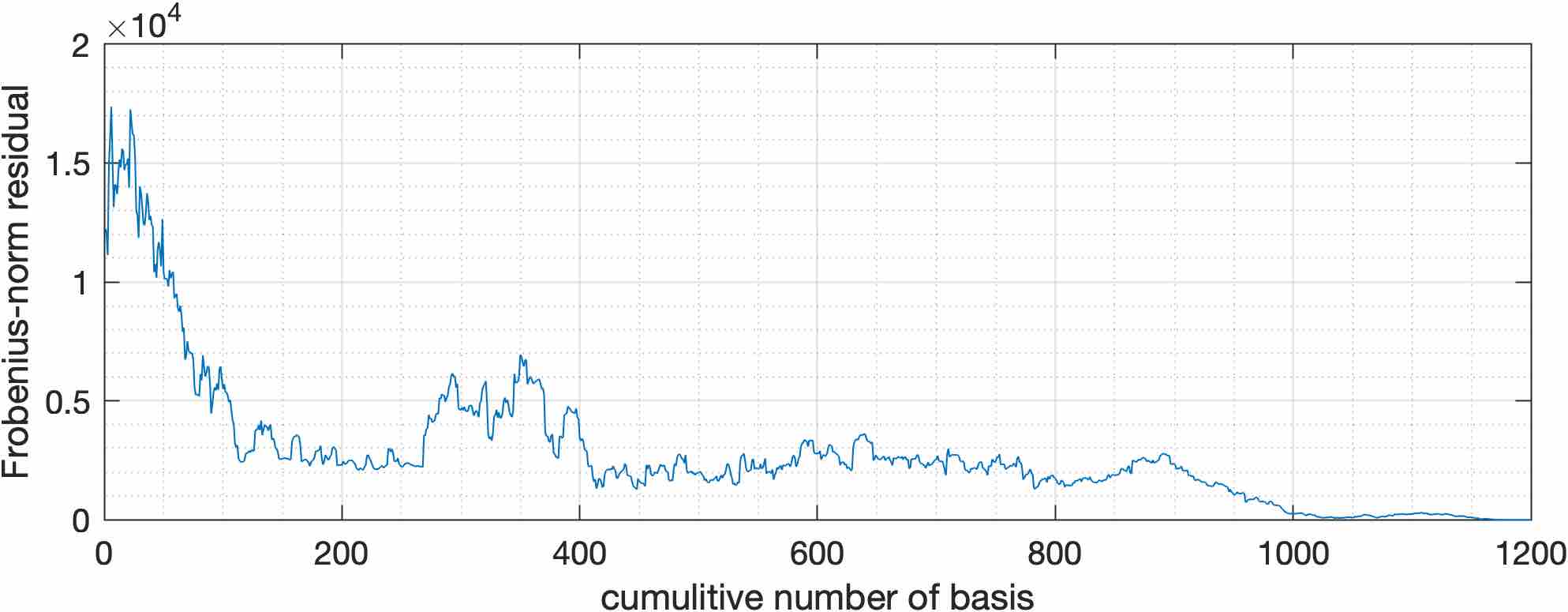}
   \caption{Reconstruction of an image using the patterns extracted in $V$. We can see the evolution of image as more patterns are added to it. The residual also drops relatively quickly. We have captured most distinctive features of the image by using about 400 patterns. Using 1,000 patterns captures the image almost perfectly.}
  \label{fig:covid_reconstruct}
\end{figure}

We can use these patterns to reconstruct an image. Figure~\ref{fig:covid_reconstruct} shows reconstruction of an image using low-rank approximation defined by Equation~\ref{eq:sum_rank1}.

\begin{figure}[H]
  \centering
   \includegraphics[width=0.24\linewidth]{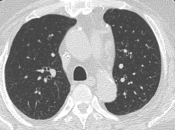}
   \includegraphics[width=0.24\linewidth]{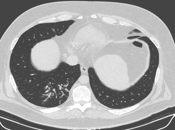}
   \includegraphics[width=0.24\linewidth]{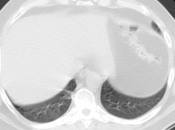}
   \includegraphics[width=0.24\linewidth]{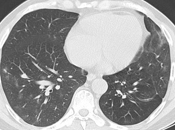}
   \caption{Images with small norm in the left basis appear are made from fewer patterns and appear to have less distinctive features.}
  \label{fig:covid_simple}
\end{figure}

\begin{figure}[H]
  \centering
   \includegraphics[width=0.24\linewidth]{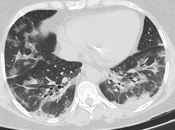}
   \includegraphics[width=0.24\linewidth]{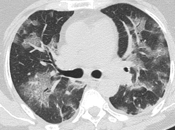}
   \includegraphics[width=0.24\linewidth]{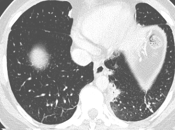}
   \includegraphics[width=0.24\linewidth]{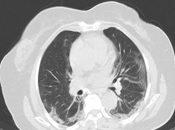}
   \caption{Images with large norm in the left basis appear relatively complex, have many distinctive features and many patterns contribute to them.}
  \label{fig:covid_complex}
\end{figure}

The left basis defines how the patterns should be merged together. Images with lower norm in the left basis appear to have fewer distinctive features in them, as some of them are shown in Figure~\ref{fig:covid_simple}. In other word, they are made from fewer patterns in $V$. On the other hand, images with larger norm in the left basis have more distinctive features and many patterns contribute to them, as some of them are shown in Figure~\ref{fig:covid_complex}.

We note that evaluating the usefulness of our patterns, from the practical point of view, requires domain expertise. A radiology scientist, for example, could verify whether the patterns in Figure~\ref{fig:covid_patterns} are sensible and useful in practice.

\setcounter{figure}{0}
\renewcommand{\thefigure}{G\arabic{figure}}

\section{Demonstrating the effectiveness of wavelet transformation in our analysis} \label{sec:appx_similarities}

In all the results we have presented so far, we have relied on wavelet transformation of images. In this last appendix, we repeat some of our experiments without using the wavelet transformation, i.e., directly applying the Higher Order GSVD on the pixel information of images. %The results we obtain clearly demonstrate the necessity and effectiveness of using wavelets.
In this experiment we use the COVID data in Appendix E.

\begin{figure}[H]
  \centering
   \includegraphics[width=0.25\linewidth]{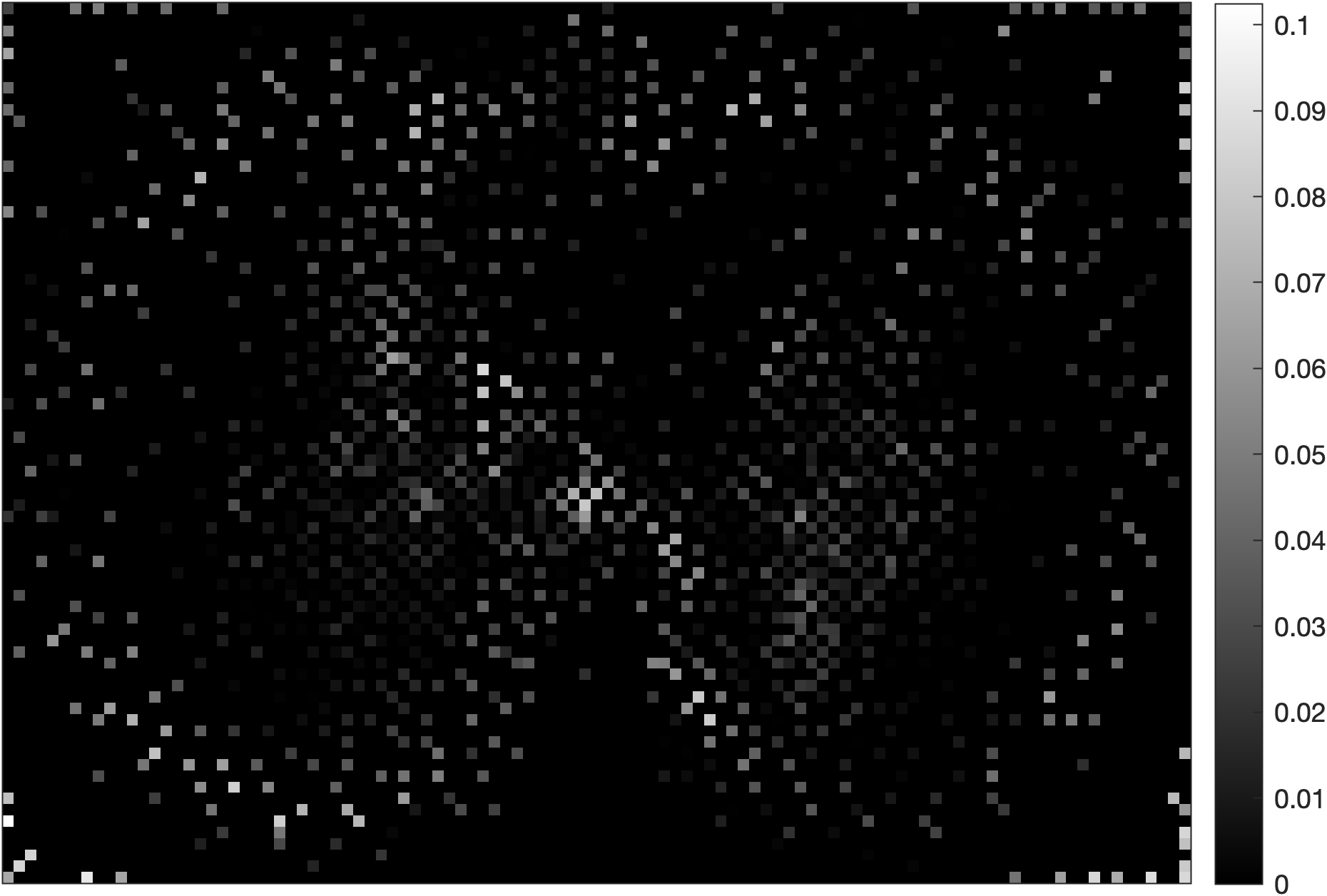}
   \includegraphics[width=0.25\linewidth]{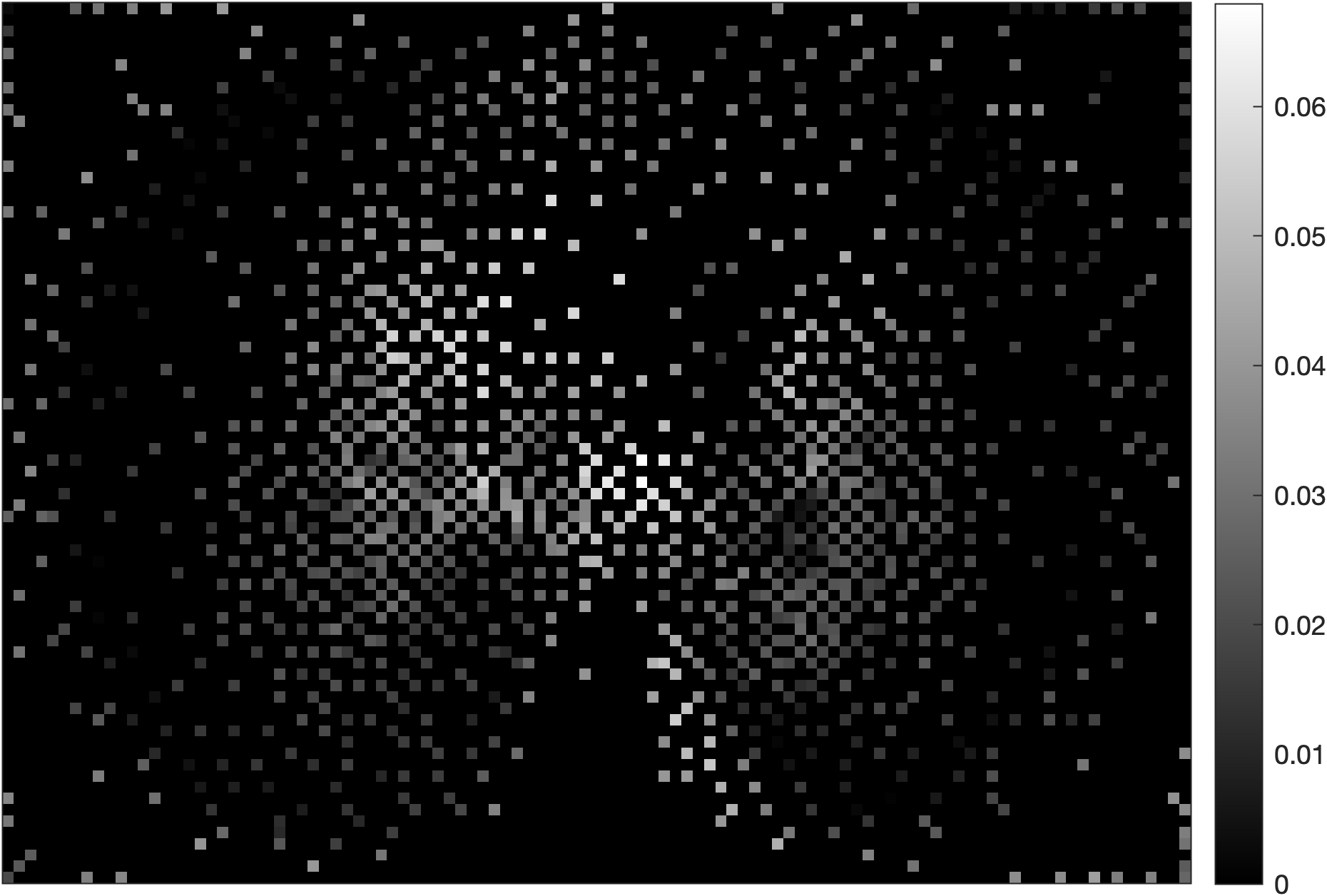}
   \includegraphics[width=0.25\linewidth]{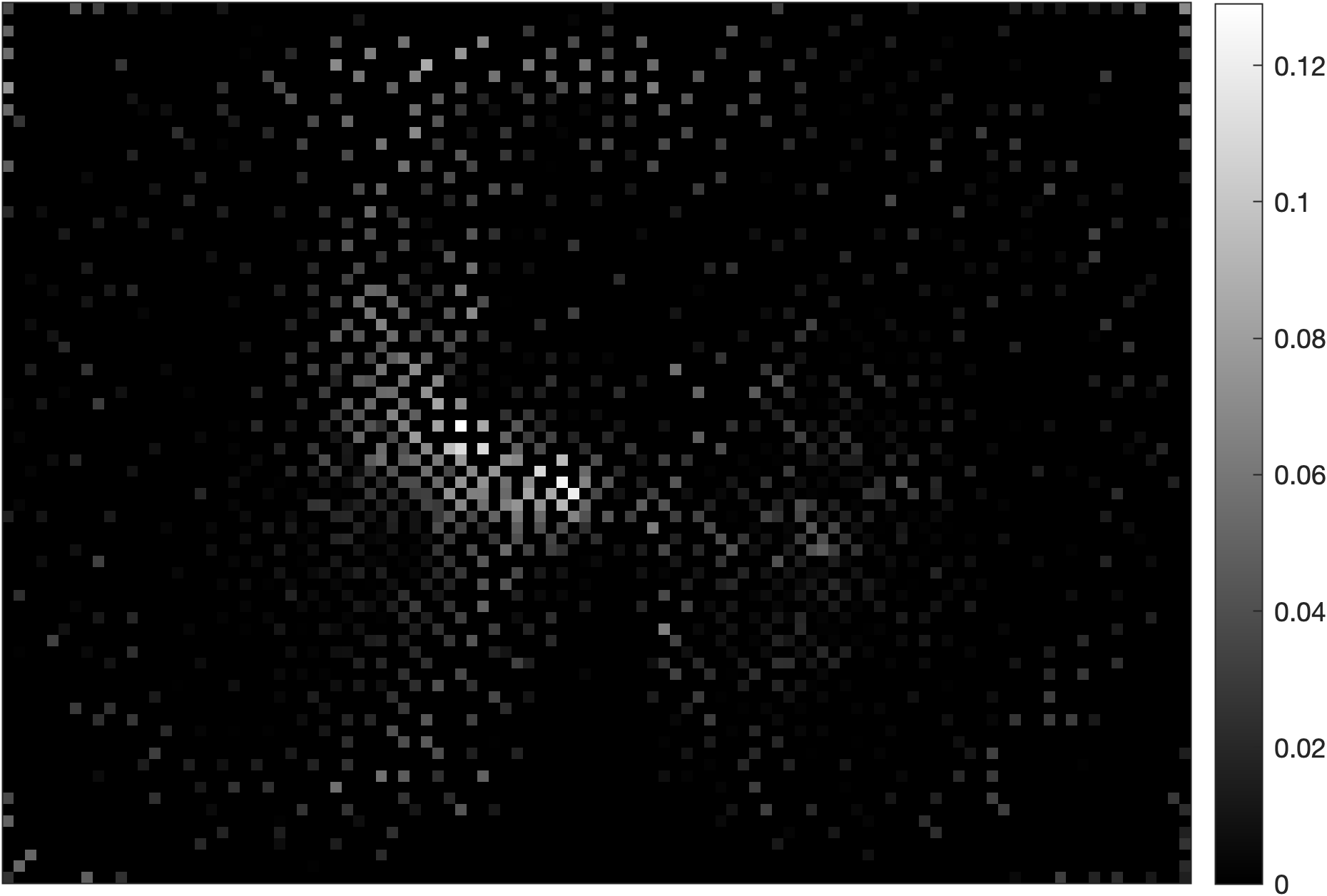}
   \includegraphics[width=0.25\linewidth]{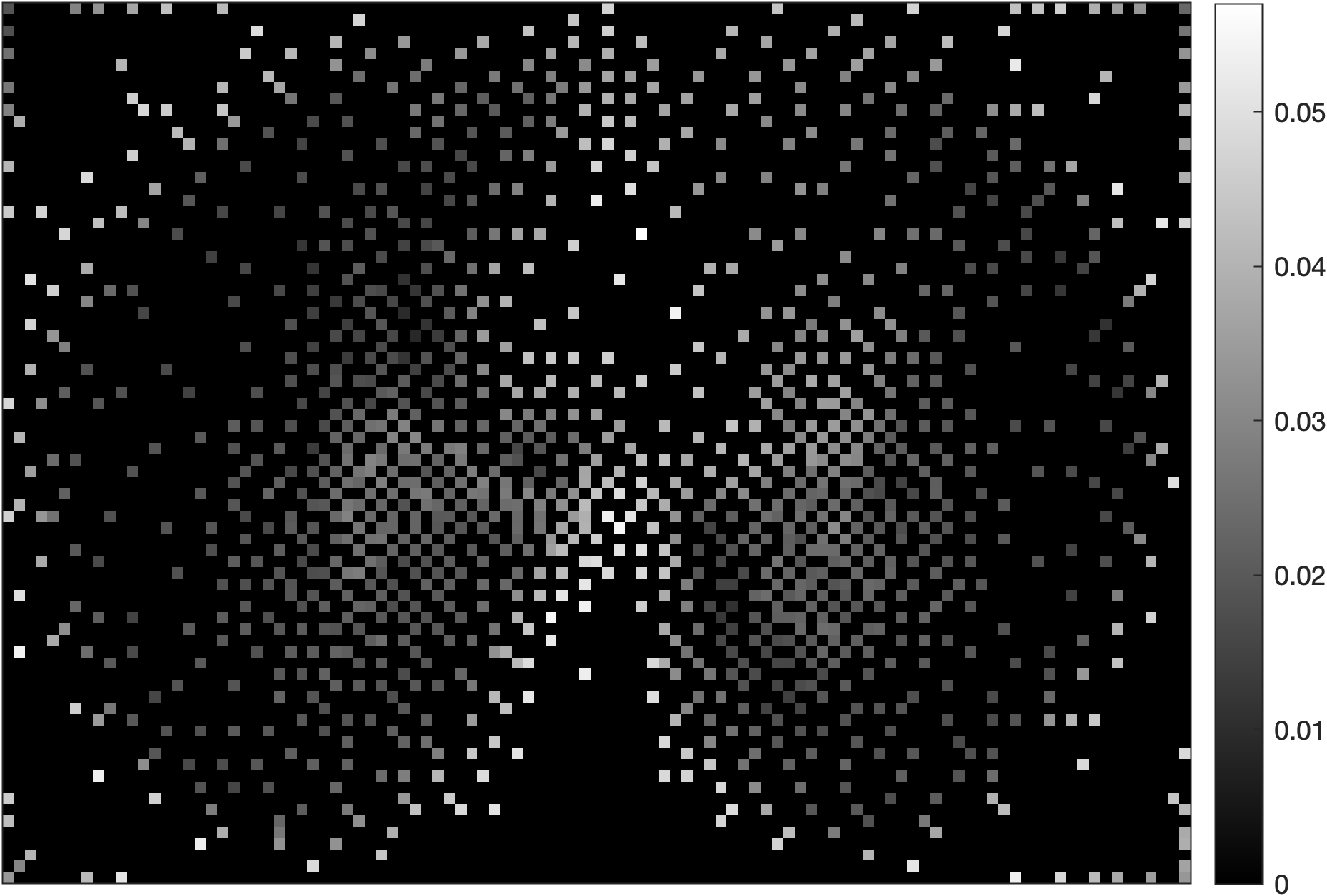}
   \includegraphics[width=0.25\linewidth]{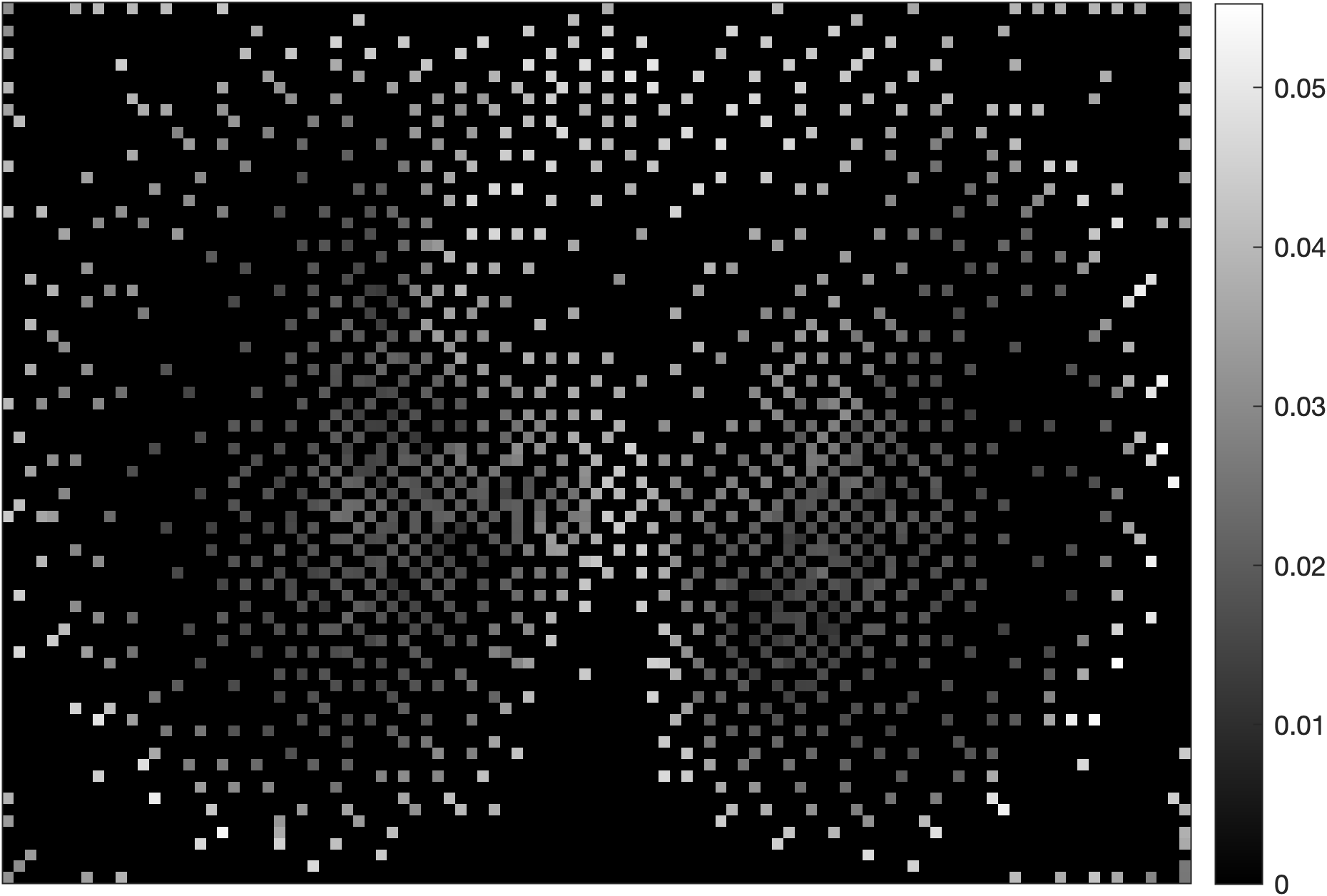}
   \caption{The most dominant patterns for non-COVID patients.}
  \label{fig:noncovid_pixel_patterns}
\end{figure}

For example, Figures~\ref{fig:covid_pixel_patterns} and~\ref{fig:noncovid_pixel_patterns} show the dominant patterns we obtain in $V$ for COVID and non-COVID patients. Compare these to the patterns we obtained in Figures~\ref{fig:noncovid_patterns} and~\ref{fig:noncovid_patterns}, when we applied the HO-GSVD on the wavelet coefficients, instead of the pixels. Although the results obtained from the pixels seem to provide some information, they are relatively vague and scattered.

\begin{figure}[H]
  \centering
   \includegraphics[width=0.25\linewidth]{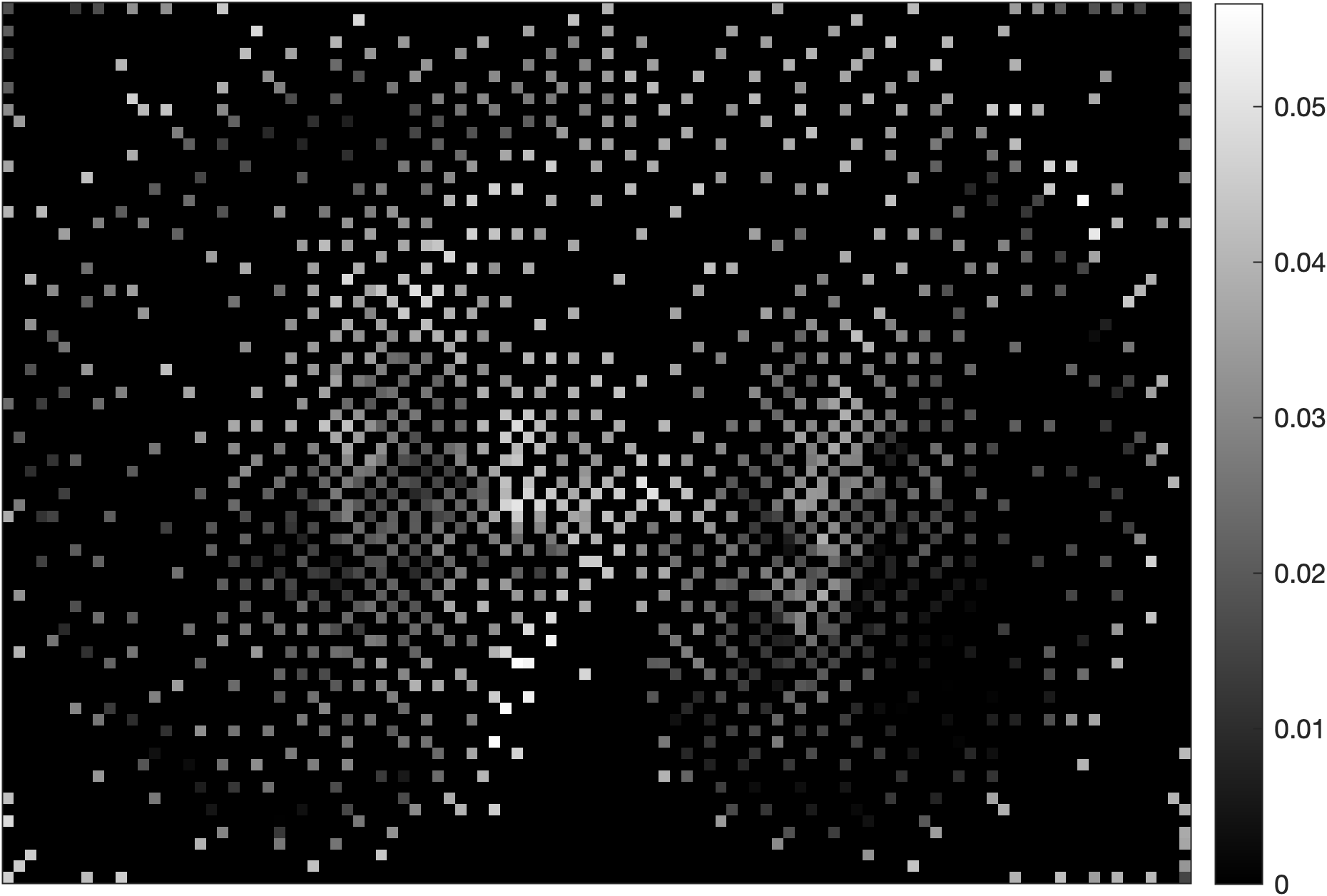}
   \includegraphics[width=0.25\linewidth]{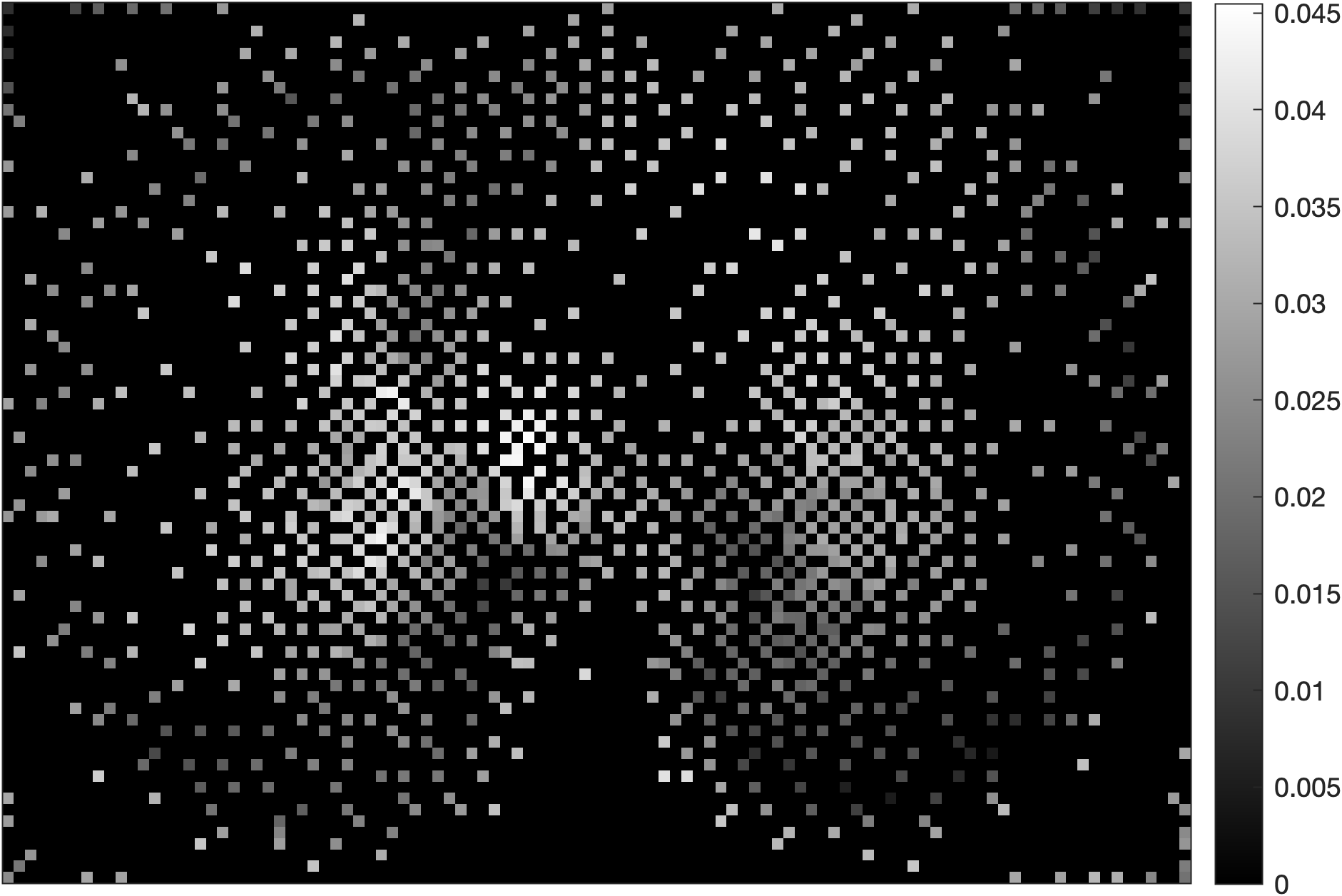}
   \includegraphics[width=0.25\linewidth]{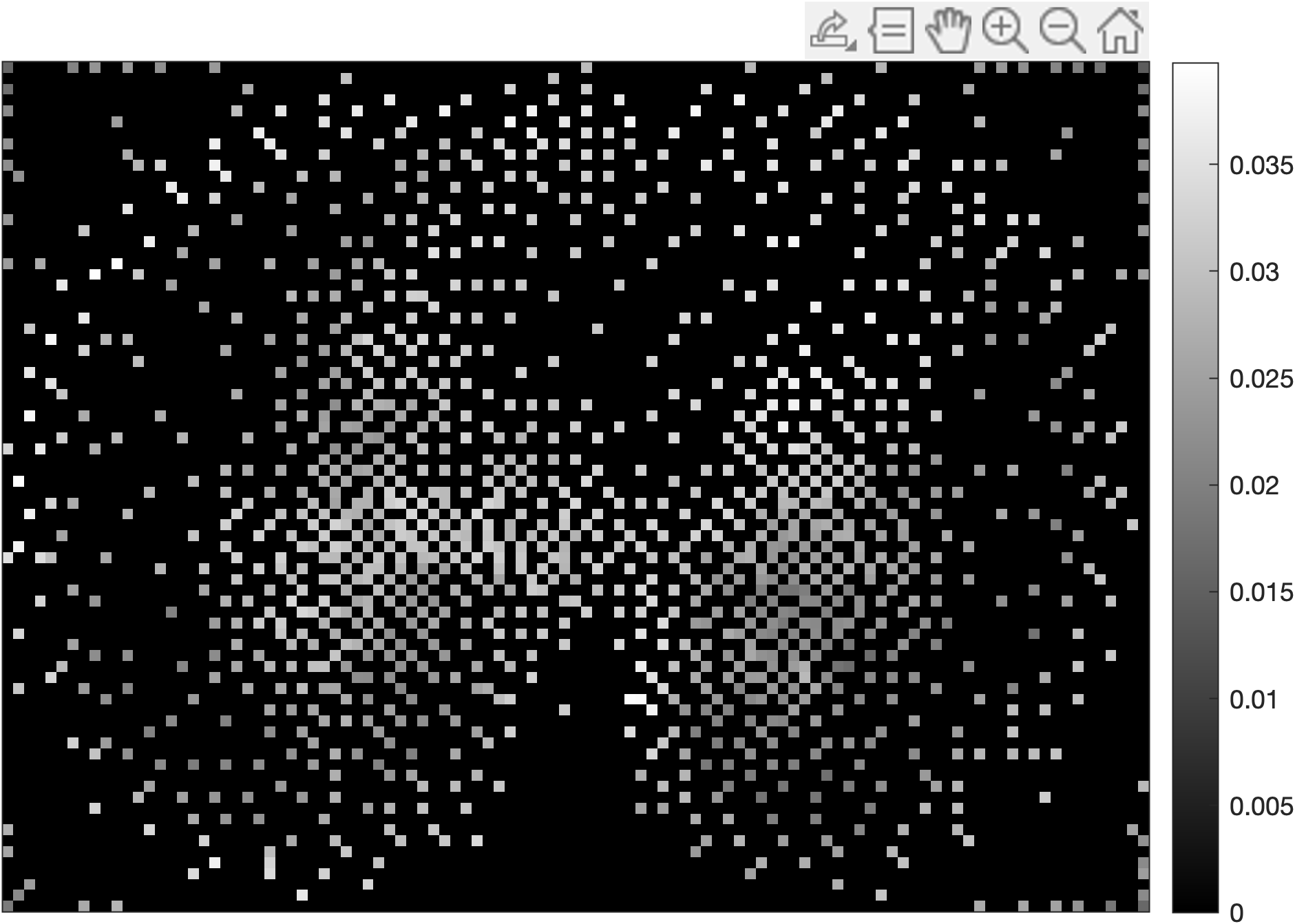}
   \includegraphics[width=0.25\linewidth]{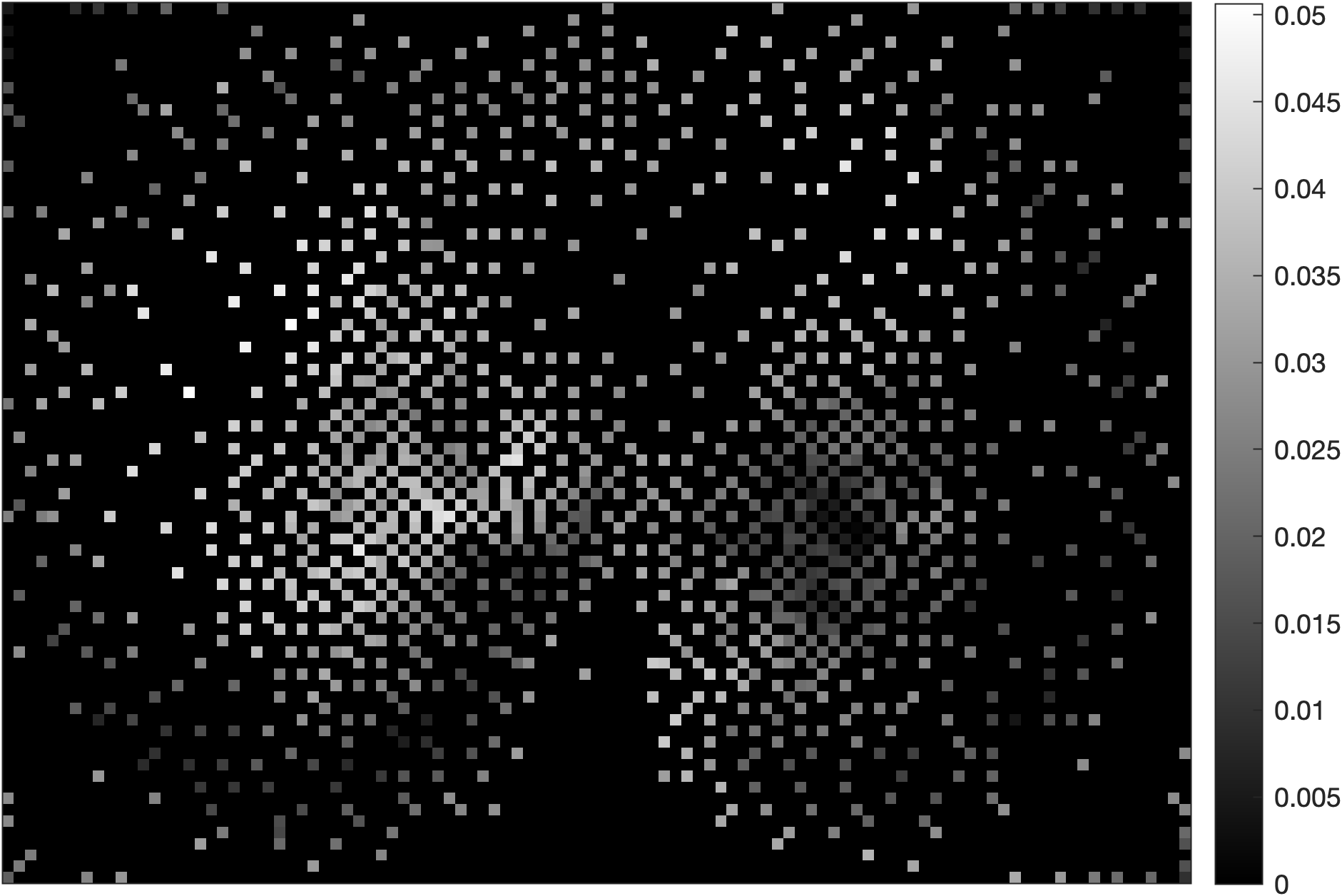}
   \includegraphics[width=0.25\linewidth]{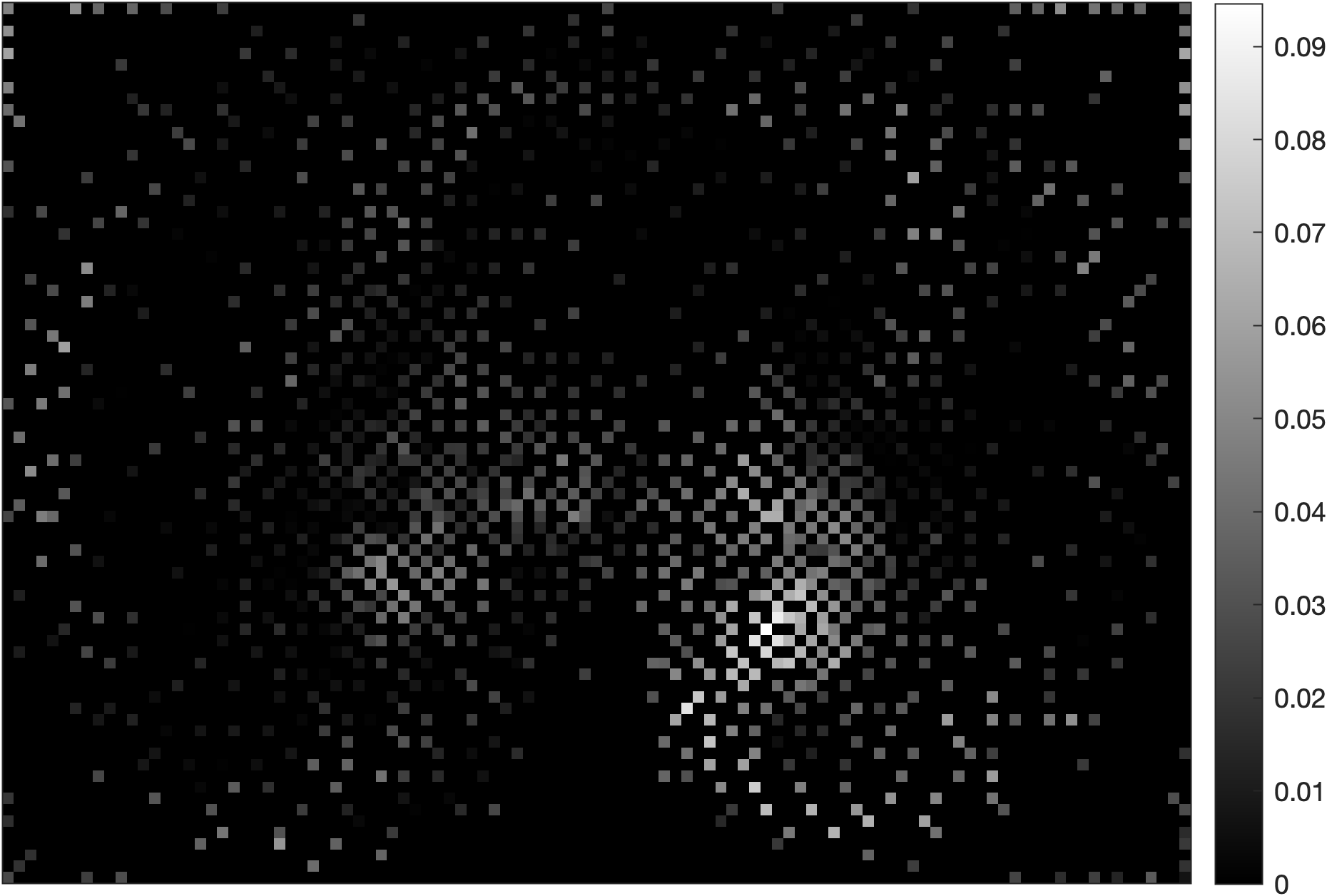}
   \caption{The most dominant patterns for COVID infected patients. This corresponds to Figure~\ref{fig:covid_patterns} which was obtained using wavelet transformation. The patterns obtained without the wavelets are relatively vague and scattered.}
  \label{fig:covid_pixel_patterns}
\end{figure}

\begin{figure}[H]
  \centering
   \includegraphics[width=0.24\linewidth]{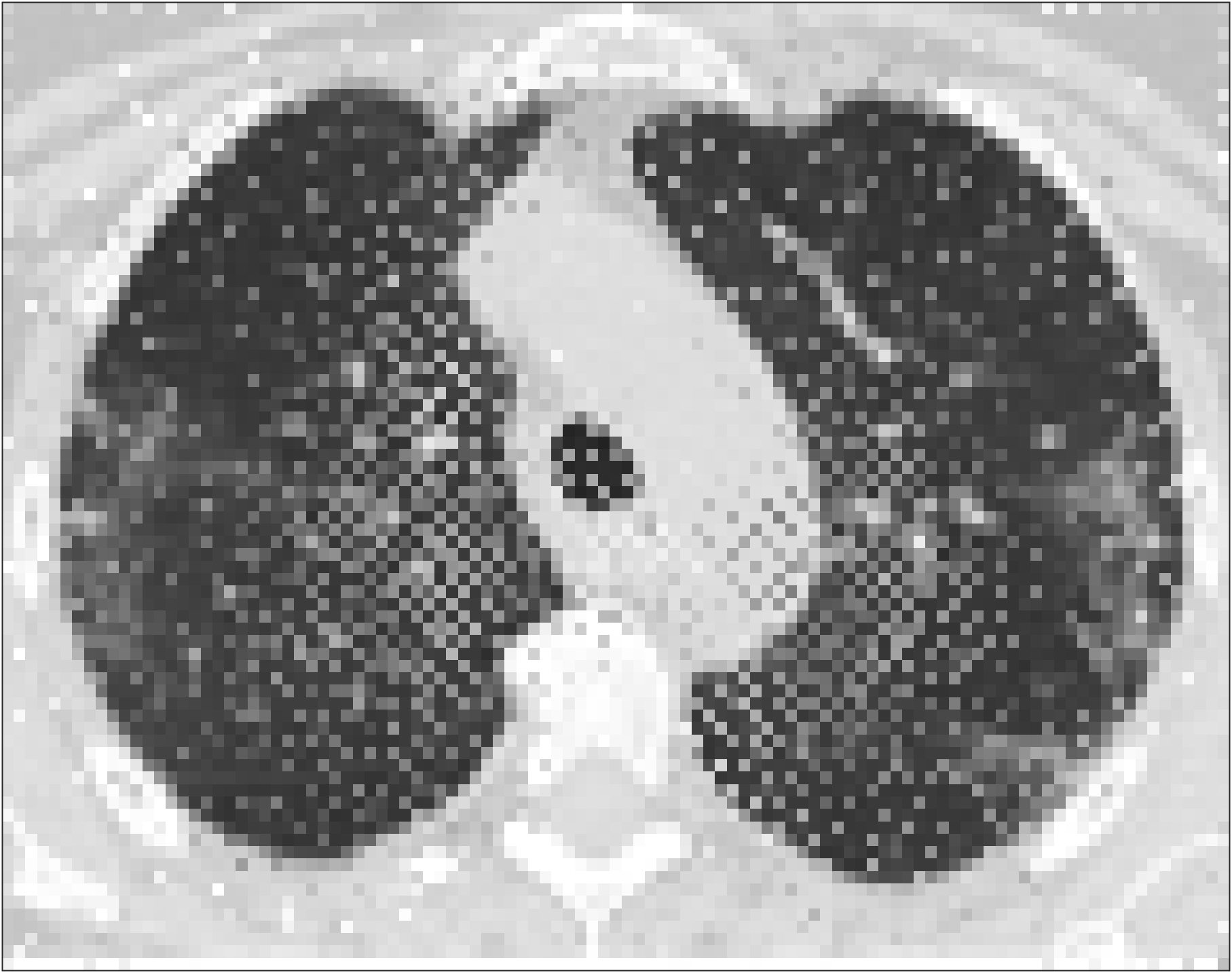}
   \includegraphics[width=0.24\linewidth]{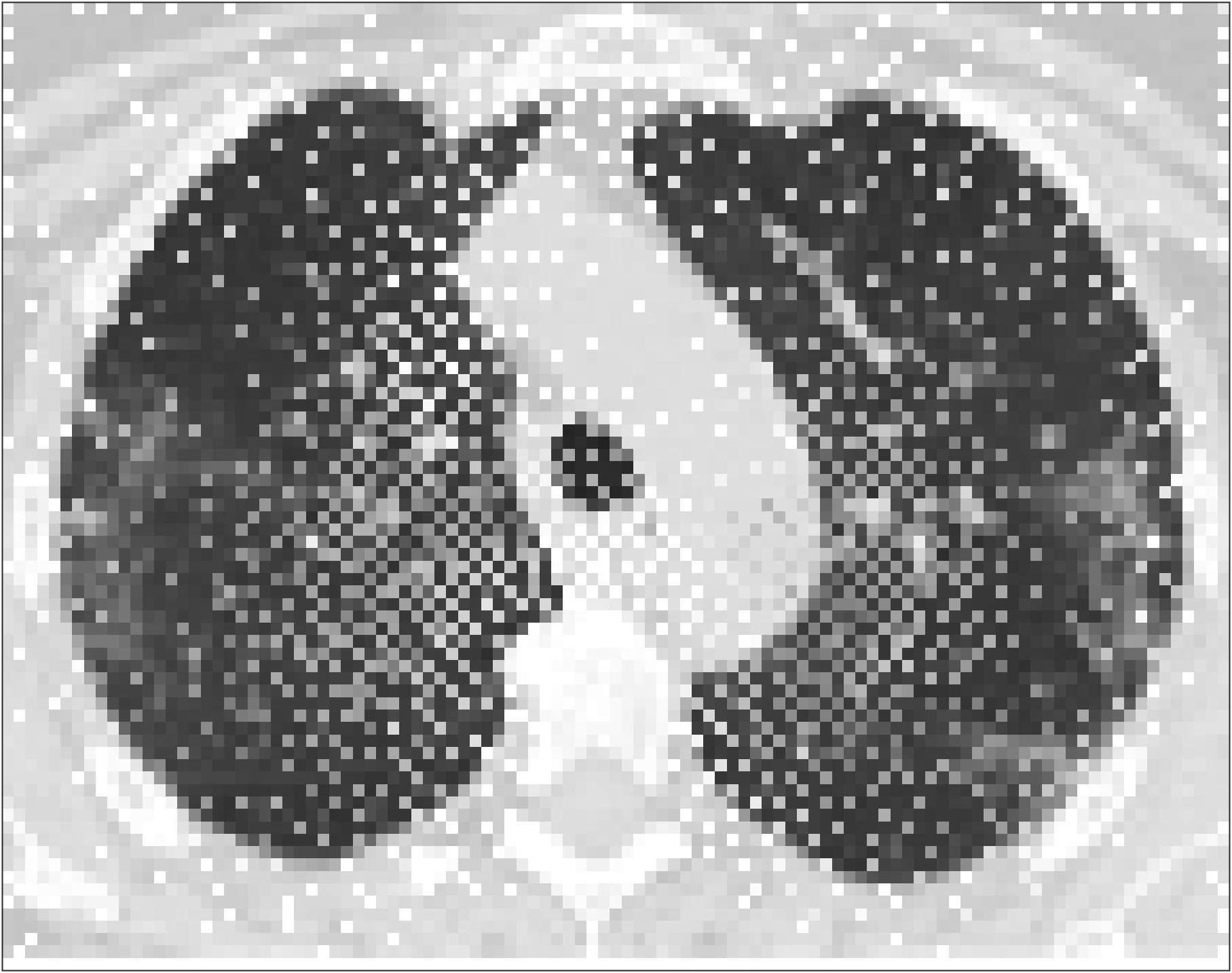}
   \includegraphics[width=0.24\linewidth]{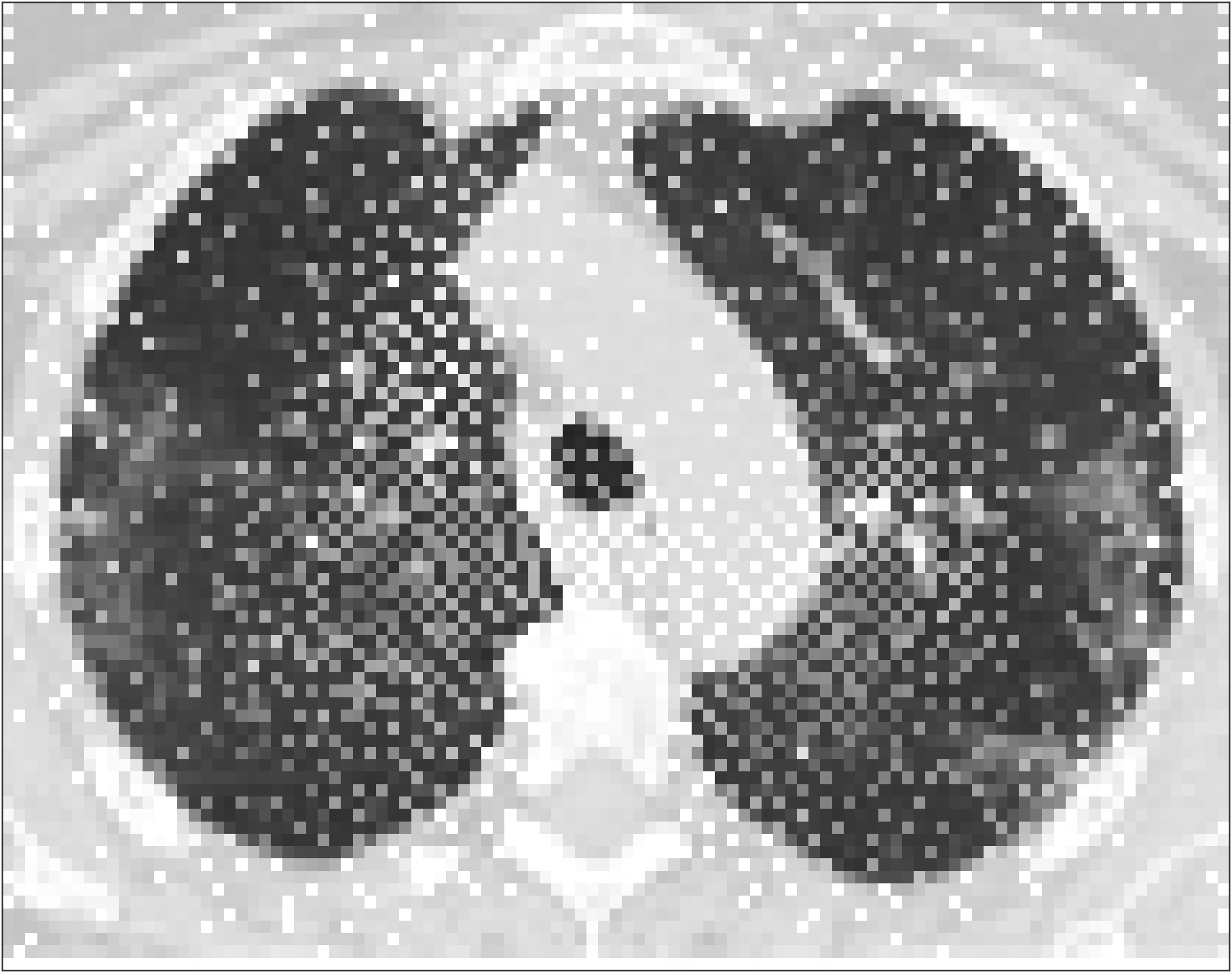}
   \includegraphics[width=0.24\linewidth]{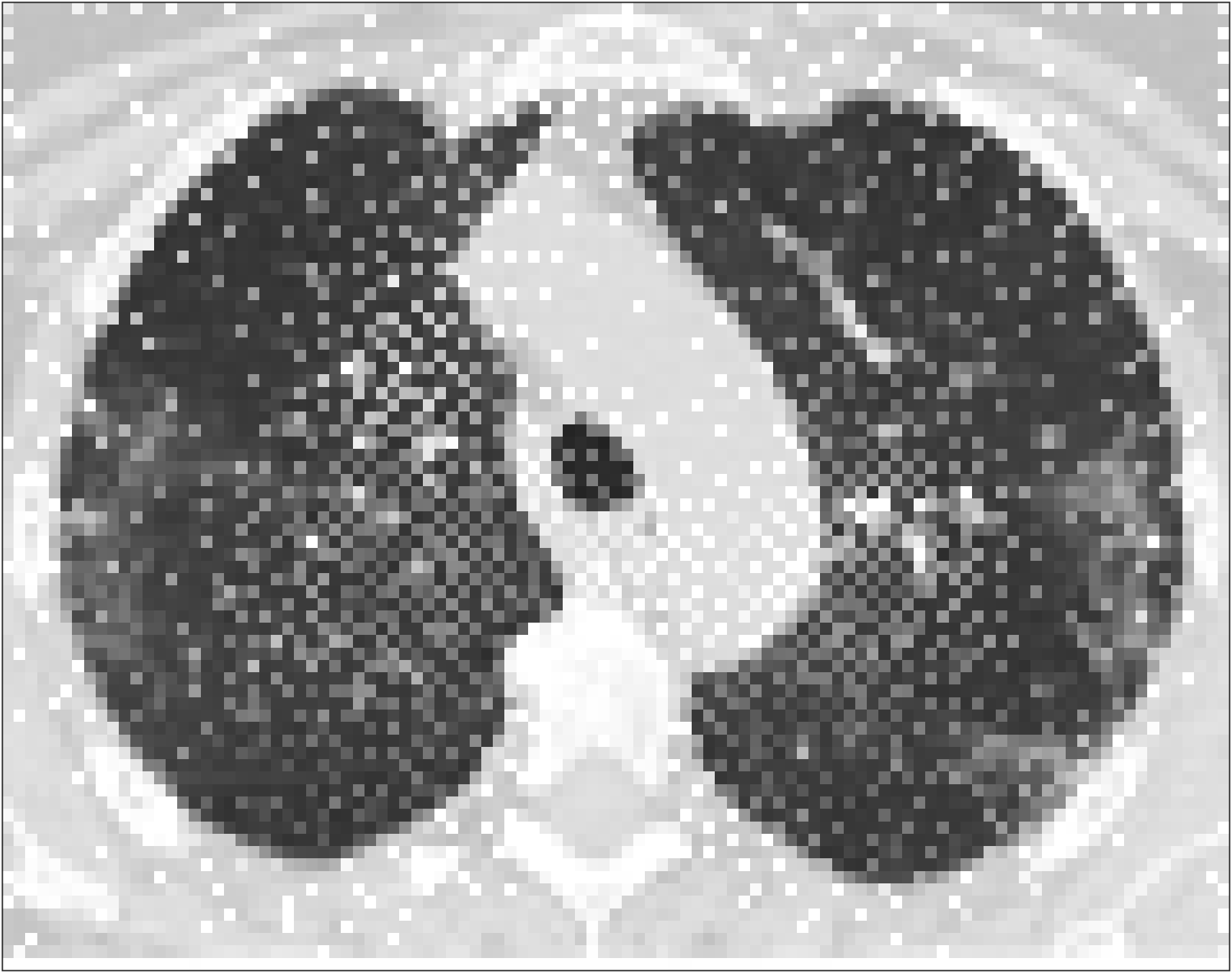}
   \includegraphics[width=0.24\linewidth]{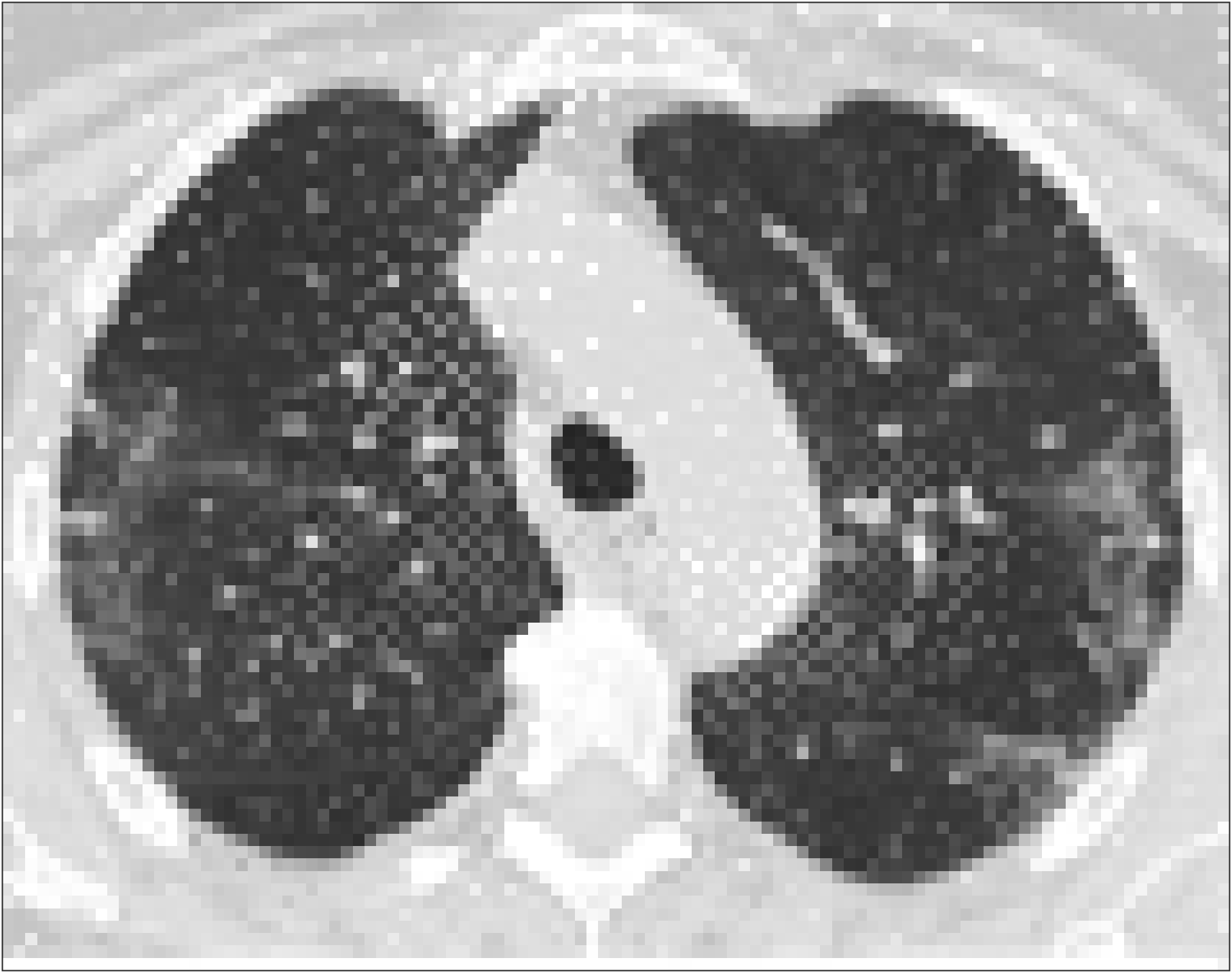}
   \includegraphics[width=0.24\linewidth]{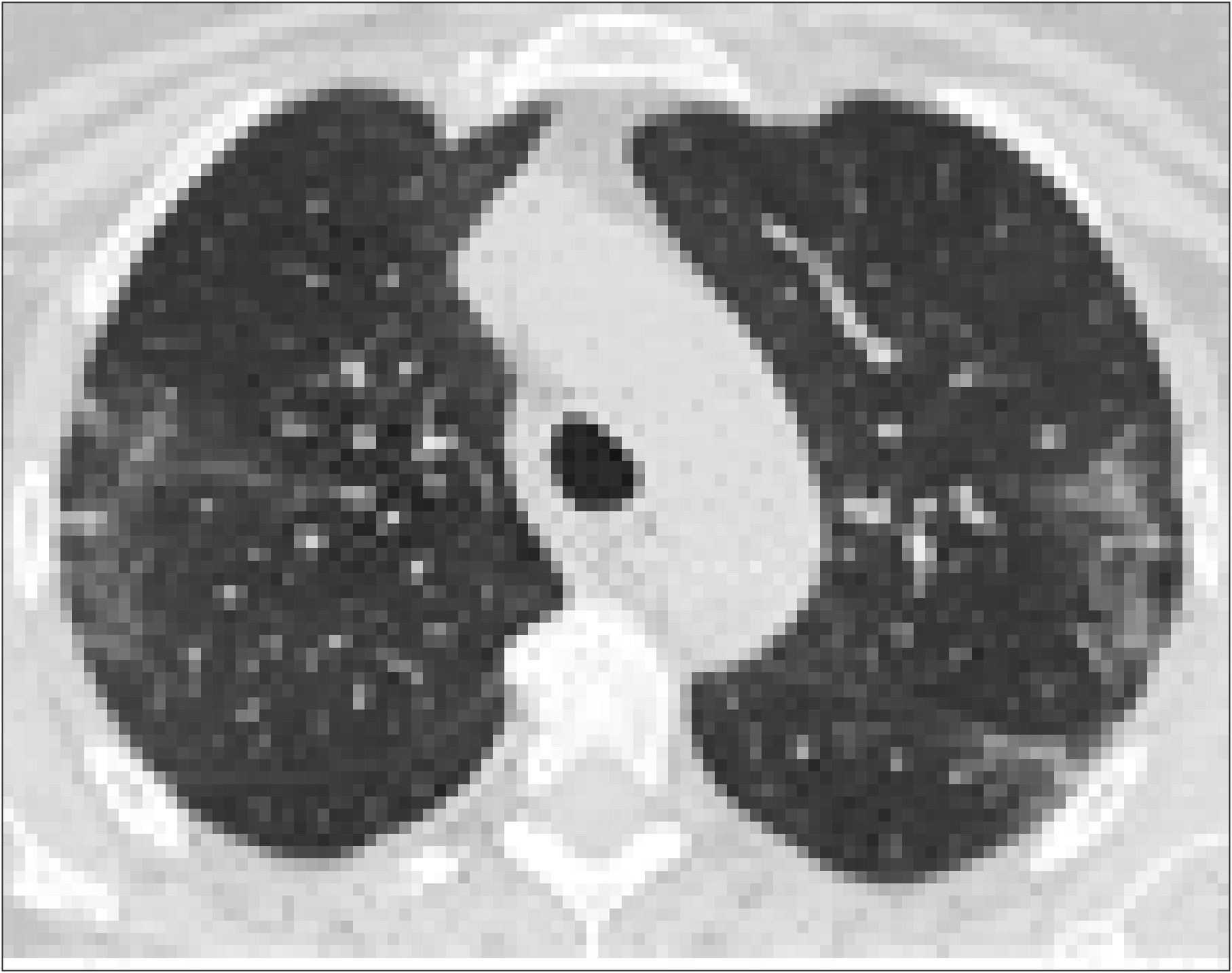}
   \includegraphics[width=0.24\linewidth]{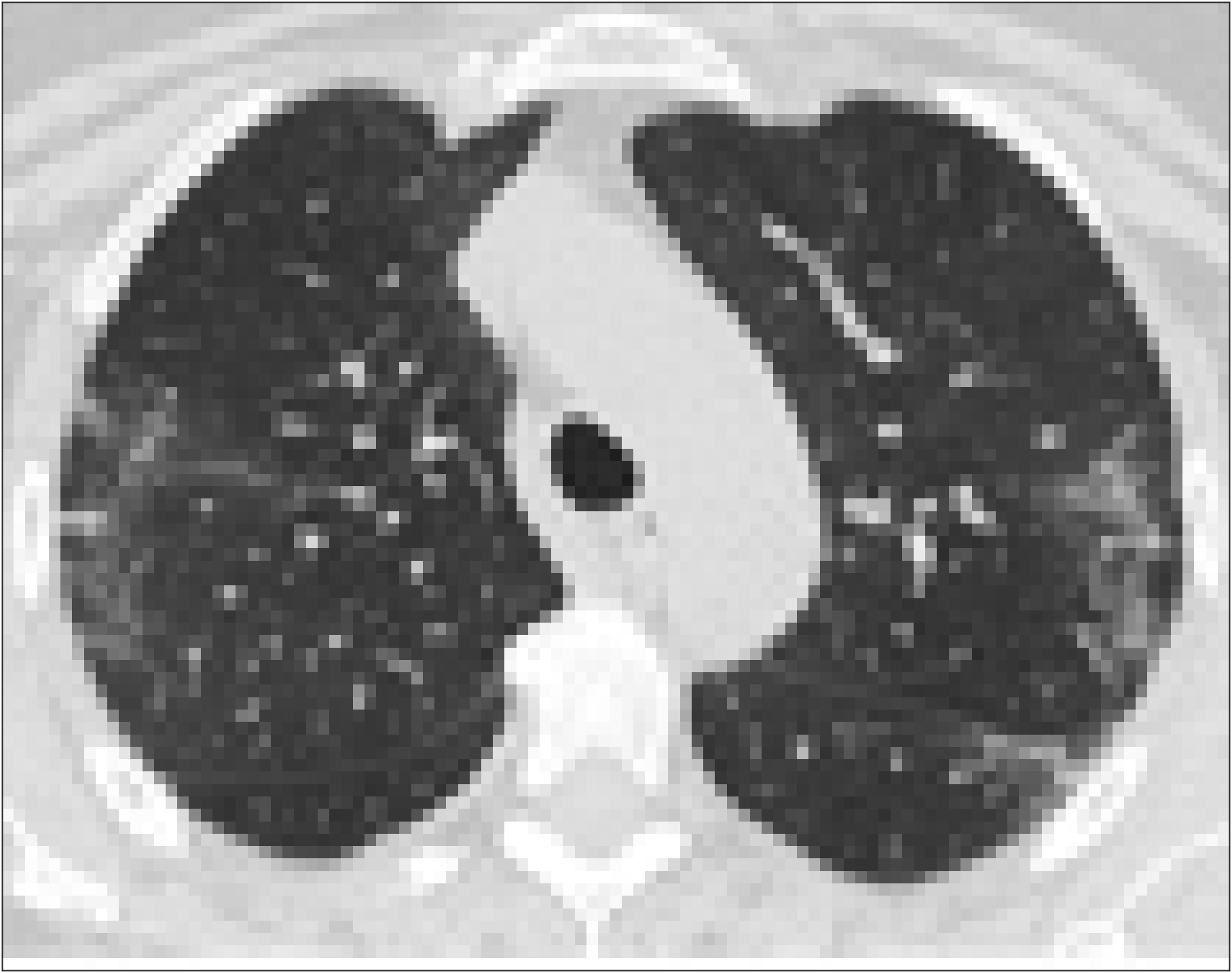}
   \includegraphics[width=0.24\linewidth]{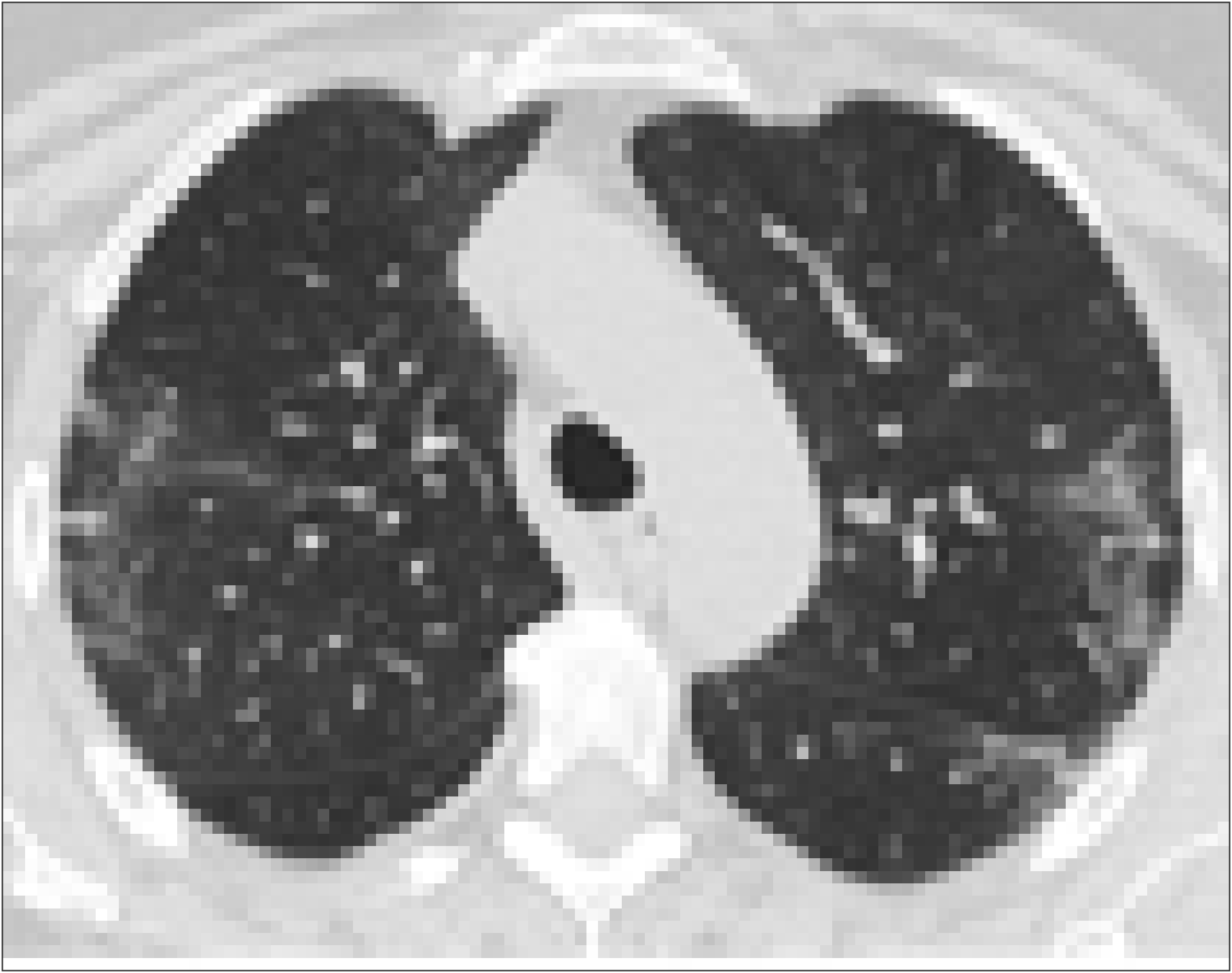}
   \includegraphics[width=0.7\linewidth]{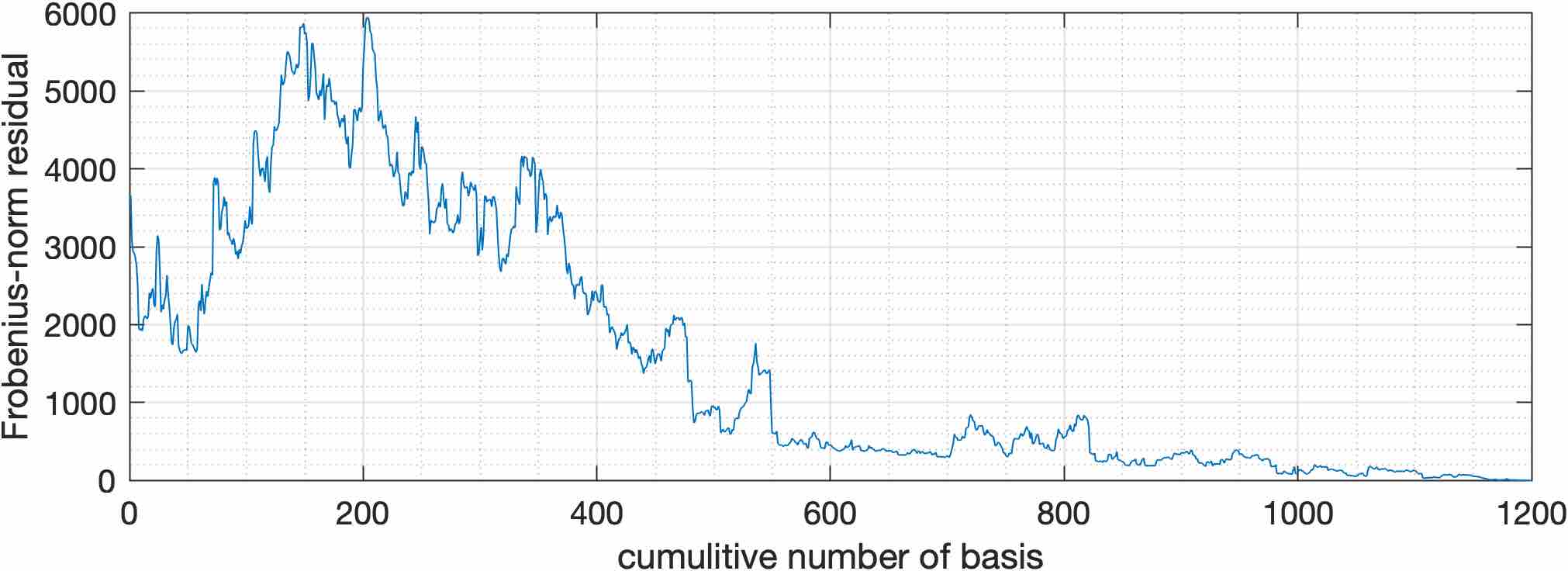}
   \caption{Reconstruction of an image using the patterns extracted in $V$, by applying HO-GSVD directly on the pixel information of dataset. This is the same image as in Figure~\ref{fig:covid_reconstruct}. This time, many of the patterns are adding noise to the image, leading to significant increase in the residual at the early stages of reconstruction.}
  \label{fig:covid_pixel_reconstruct}
\end{figure}

Moreover, Figure~\ref{fig:covid_pixel_reconstruct} shows the reconstruction of the same image as in Figure~\ref{fig:covid_reconstruct}, this time with rank-1 patterns obtained without wavelet transformation.

\begin{figure}[H]
  \centering
   \includegraphics[width=0.25\linewidth]{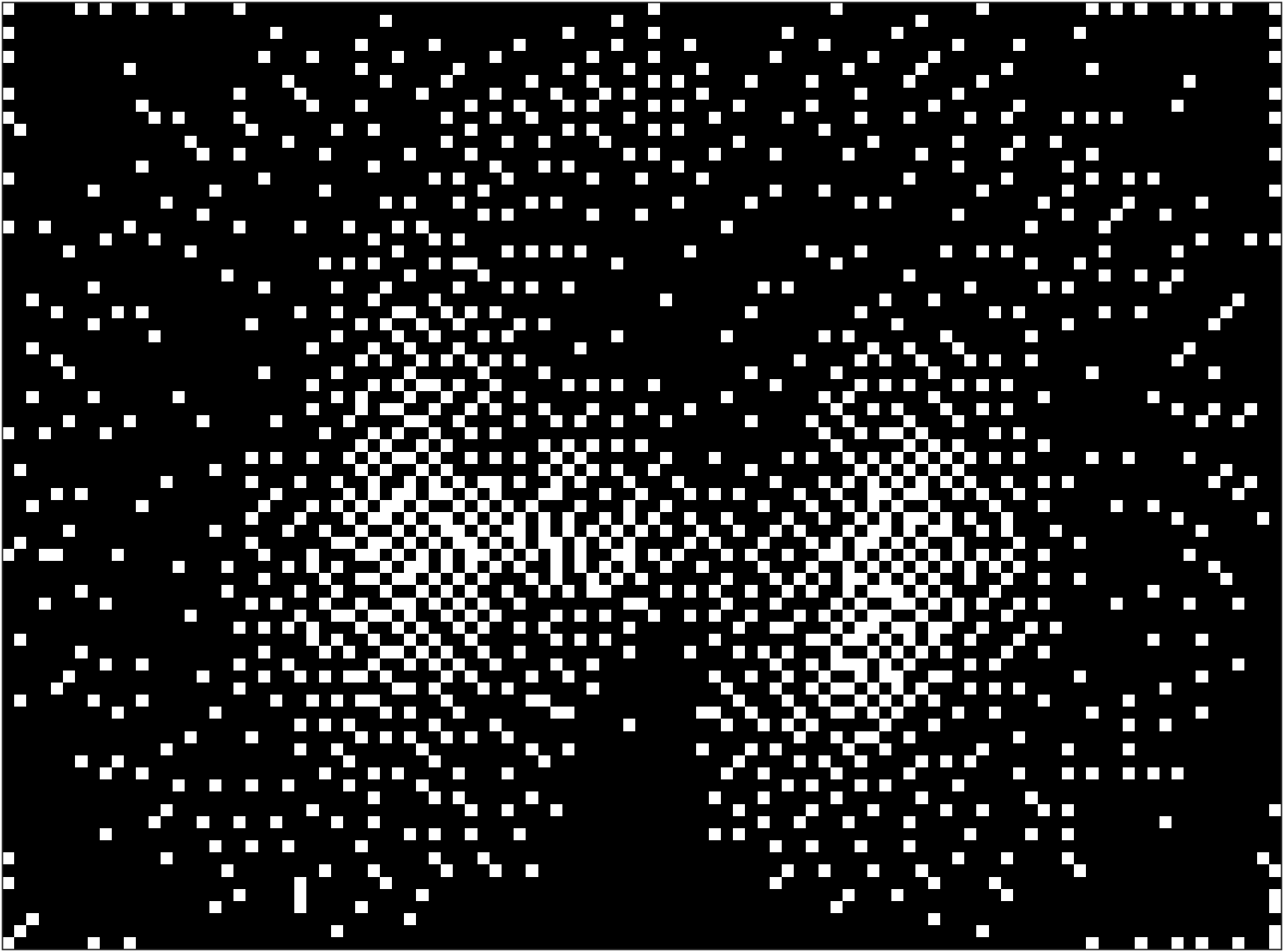}
   \caption{1,200 of the pixels with linearly independent variations among images. Some of these pixels are at the perimeter of image which clearly do not relate to COVID infection. Compare these pixels to the pixels in Figure~\ref{fig:covid_stencil}, and how the 1,200 wavelet coefficients have engaged large and relevant regions of the image.}
  \label{fig:covid_pixel_stencil}
\end{figure}

Notice that in this reconstruction only 1,200 pixels (out of 8,190) are involved, as shown in Figure~\ref{fig:covid_pixel_stencil}, and the rest of pixels are the same as the original image.\footnote{Using more than 1,220 pixels would violate the requirement of full column rank.} Hence, the reconstruction without using the wavelets needs to start from an image mostly similar to the original image and cannot consider many of the pixels. This seems to hollow the point of performing low rank approximation.

However, when using the 1,200 wavelet coefficients, the majority of pixels are involved in our analysis, as the stencil in Figure~\ref{fig:covid_stencil} demonstrates. Also, wavelets make the low rank approximations of images informative.

% \section{Omniglot dataset} \label{sec:appx_omniglot}

% Here, we perform similar analysis for 30 alphabets of the Omniglot dataset \citep{lake2015human}.

\end{document}